\documentclass{article} 
\usepackage{iclr2025_conference,times}
\usepackage{xspace}
\usepackage{mathpazo}
\usepackage{float}

\usepackage{amsmath,amsfonts,bm}

\def\eqref#1{equation~\ref{#1}}

\def\1{\bm{1}}

\DeclareMathAlphabet{\mathsfit}{\encodingdefault}{\sfdefault}{m}{sl}
\SetMathAlphabet{\mathsfit}{bold}{\encodingdefault}{\sfdefault}{bx}{n}

\usepackage{booktabs}  
\usepackage{multirow}  
\usepackage{siunitx}   
\usepackage{makecell}  
\usepackage{algorithm}
\usepackage{algpseudocode}
\usepackage{url}
\usepackage{graphicx}
\usepackage{listings}  
\usepackage{upquote}
\usepackage{xcolor}    
\usepackage{soul} 
\usepackage{pifont}
\usepackage{subcaption}
\usepackage{fvextra}
\usepackage[utf8]{inputenc}
\usepackage[export]{adjustbox} 
\usepackage{svg}

\usepackage[most]{tcolorbox}

\newtcolorbox[list inside=prompt,auto counter,number within=section]{prompt}[1][]{
    colbacktitle=black!60,
    fonttitle=\small,
    coltitle=white,
    fontupper=\footnotesize,
    boxsep=3pt,
    left=0pt,
    right=0pt,
    top=0pt,
    bottom=0pt,
    boxrule=1pt,
    #1,
    breakable, 
}

\usepackage{nicematrix}
\definecolor{beige}{rgb}{0.9254901960784314, 0.9254901960784314, 0.9058823529411765}
\definecolor{purple}{rgb}{0.5098039215686274, 0.4235294117647059, 0.4980392156862745}

\definecolor{skyblue}{rgb}{0.5294117647058824, 0.807843137254902, 0.9215686274509803}   
\definecolor{steelblue}{rgb}{0.27450980392156865, 0.5098039215686274, 0.7058823529411765} 
\definecolor{royalblue}{rgb}{0.2549019607843137, 0.4117647058823529, 0.8823529411764706} 
\definecolor{mintgreen}{rgb}{0.596078431372549, 0.984313725490196, 0.596078431372549}   
\definecolor{seagreen}{rgb}{0.1803921568627451, 0.5450980392156862, 0.3411764705882353} 
\definecolor{tealgreen}{rgb}{0.0, 0.5019607843137255, 0.5019607843137255}              
\definecolor{brickred}{rgb}{0.796078431372549, 0.2549019607843137, 0.32941176470588235} 

\definecolor{c-navy}{rgb}{0,0.08,0.45}
\definecolor{c-blue}{RGB}{29,59,105}
\usepackage{hyperref}
\hypersetup{ %
colorlinks=true,
linkcolor=c-navy,
citecolor=c-navy,
filecolor=c-navy,
urlcolor=c-navy,
}
\usepackage{url}

\newcommand{\githublogo}[1]{\hspace{2pt}\includegraphics[width=13pt, valign=c]{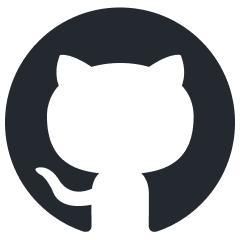}\hspace{2pt} \texttt{#1}}
\newcommand{\hflogo}[1]{\includegraphics[width=15pt, valign=c]{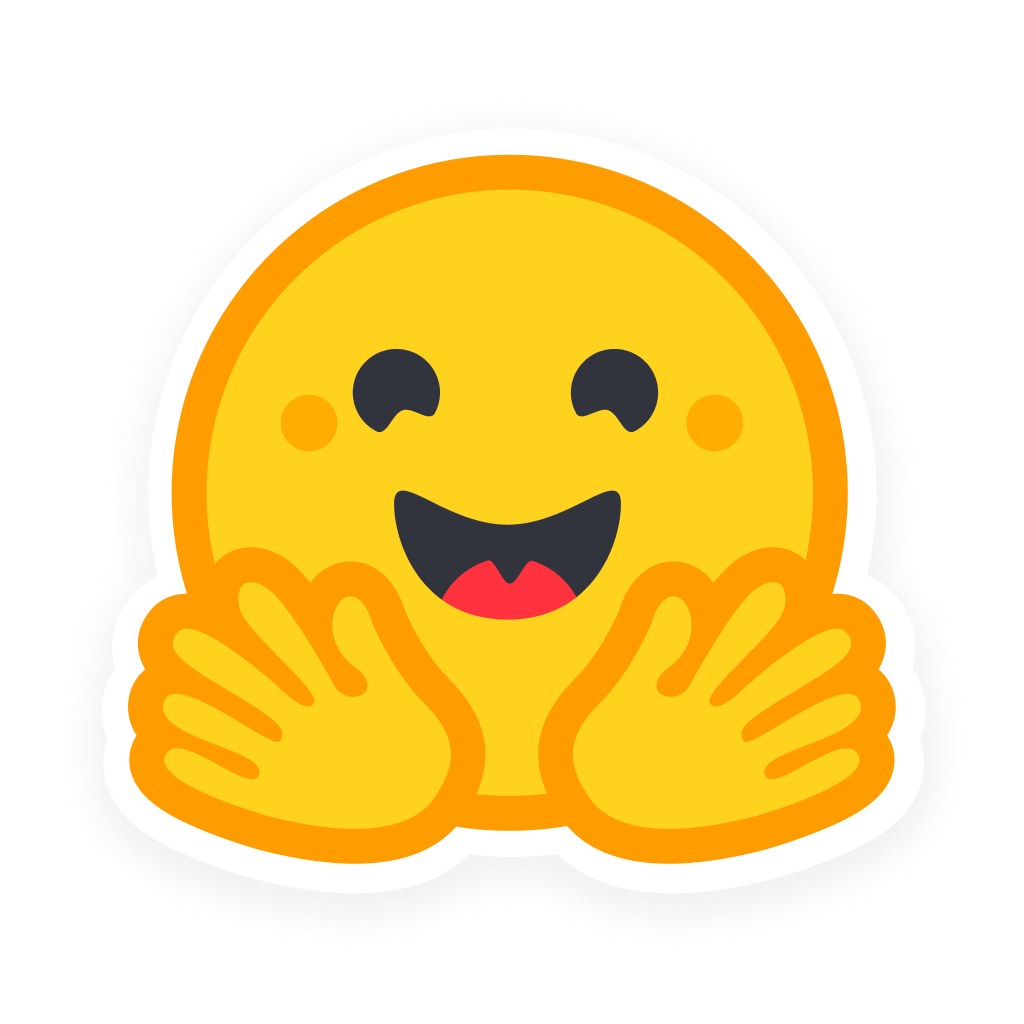} \texttt{#1}}
\newcommand{\datalogo}[1]{\hspace{1pt}\includegraphics[width=15pt, valign=c]{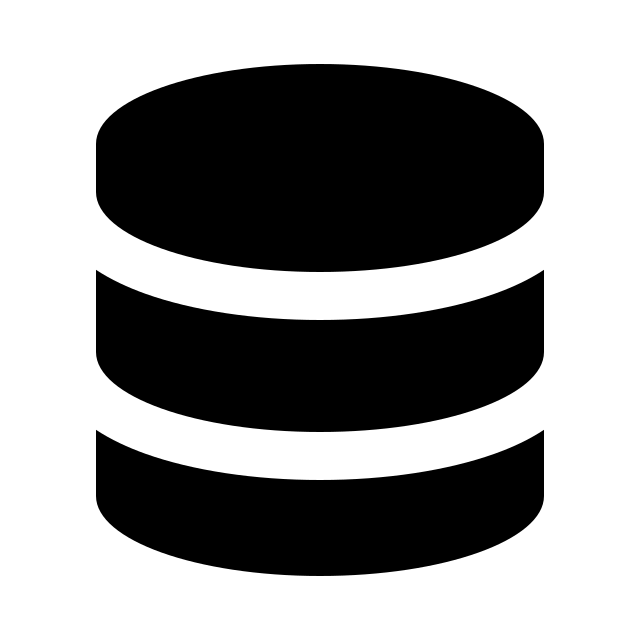}\hspace{1pt} \texttt{#1}}
\newcommand{\benchmarklogo}[1]{\hspace{1pt}\includegraphics[width=15pt, valign=c]{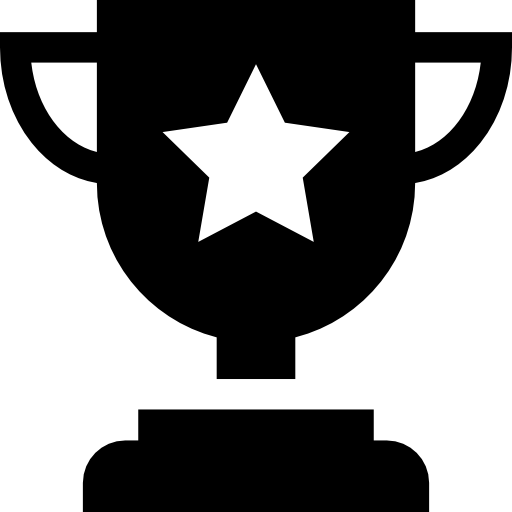}\hspace{1pt} \texttt{#1}}

\title{\tucanotitle \\[1em]
\centering
Better Open Source LLMs for Portuguese\\[2mm]
\tucanologo
}

\author{Project Polyglot\thanks{Polyglot is a project supported by Germany’s Federal Ministry of Education and Research (BMBF) and North Rhine-Westphalia’s Ministry of Culture and Science (MWK) under the TRA Sustainable Futures initiative at the University of Bonn.} \\ \\ 
\textbf{Leads:} Nicholas Kluge Corrêa\textsuperscript{1,2,3}, Aniket Sen\textsuperscript{4} \\ \\
\textbf{Core Team:}
Nicholas Kluge Corrêa\textsuperscript{1,2,3}, 
Shiza Fatimah\textsuperscript{1,2},
Sophia Falk\textsuperscript{5},
Aniket Sen\textsuperscript{4} \\ \\ 
\textbf{Other Contributors:} 
Lucie Flek\textsuperscript{1,2},
Julia Kastner\textsuperscript{3},
Lennard Landgraf\textsuperscript{3} \\ \\ \\ \\ \\
\textbf{Affiliations:}\\
  \textsuperscript{1}Bonn-Aachen International Center for Information Technology (b-it) / CAISA Lab\\
  \textsuperscript{2}Lamarr Institute for Machine Learning and Artificial Intelligence \\
  \textsuperscript{3}Center for Science and Thought \\
  \textsuperscript{4}Helmholtz-Institut für Strahlen- und Kernphysik\\
  \textsuperscript{5}Bonn Sustainable AI Lab\\
}

\newcommand{\tucanotitle}{\resizebox{\linewidth}{!}{\includegraphics{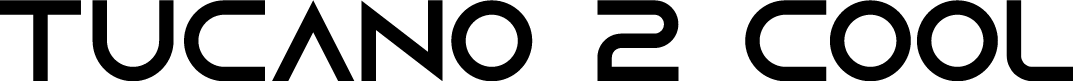}}}
\newcommand{\tucanologo}{\resizebox{10cm}{!}{\includegraphics{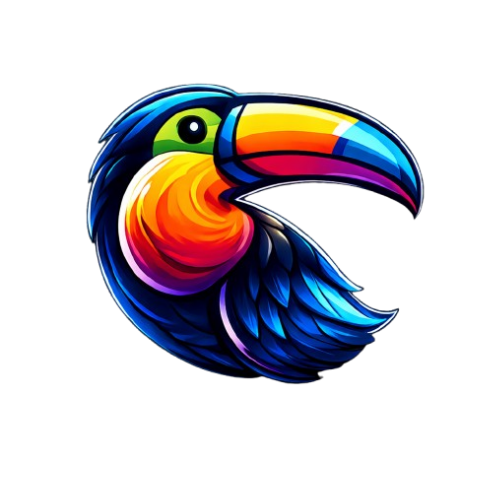}}}  

\newcommand{\gigaverbo}{\textcolor{seagreen}{GigaVerbo-v2}\xspace}
\newcommand{\gigaverbosynth}{\textcolor{seagreen}{GigaVerbo-v2 Synth}\xspace}
\newcommand{\gigaverbosft}{\textcolor{seagreen}{GigaVerbo-v2 SFT}\xspace}
\newcommand{\gigaverbopreferences}{\textcolor{seagreen}{GigaVerbo-v2 Preferences}\xspace}

\newcommand{\tucano}{\textcolor{seagreen}{\textbf{Tucano 2}}\xspace}

\newcommand{\tucanoptsmall}{\textcolor{seagreen}{\textbf{Tucano2-0.6B-Base}}\xspace}
\newcommand{\tucanocptsmall}{\textcolor{seagreen}{\textbf{Tucano2-qwen-0.5B-Base}}\xspace}
\newcommand{\tucanocptsmallinstruct}{\textcolor{seagreen}{\textbf{Tucano2-qwen-0.5B-Instruct}}\xspace}
\newcommand{\tucanocptsmallthink}{\textcolor{seagreen}{\textbf{Tucano2-qwen-0.5B-Think}}\xspace}
\newcommand{\tucanocptmedium}{\textcolor{seagreen}{\textbf{Tucano2-qwen-1.5B-Base}}\xspace}
\newcommand{\tucanocptmediuminstruct}{\textcolor{seagreen}{\textbf{Tucano2-qwen-1.5B-Instruct}}\xspace}
\newcommand{\tucanocptmediumthink}{\textcolor{seagreen}{\textbf{Tucano2-qwen-1.5B-Think}}\xspace}
\newcommand{\tucanocptlarge}{\textcolor{seagreen}{\textbf{Tucano2-qwen-3.7B-Base}}\xspace}
\newcommand{\tucanocptlargeinstruct}{\textcolor{seagreen}{\textbf{Tucano2-qwen-3.7B-Instruct}}\xspace}
\newcommand{\tucanocptlargethink}{\textcolor{seagreen}{\textbf{Tucano2-qwen-3.7B-Think}}\xspace}

\begin{document}

\maketitle

\newpage

\begin{abstract}
We present \tucano, a fully open suite of large language models (LLMs) with 0.5-3.7 billion parameters, designed to address certain gaps in open-source development for Portuguese LLMs. Following our previous works, we now extend our dataset, \gigaverbo, to a new degree of quality and scale, while also introducing a new synthetic dataset, \gigaverbosynth, aimed at filling missing gaps in \gigaverbo, and two post-training datasets, \gigaverbosft and \gigaverbopreferences, that allow Portuguese LLMs to be trained in domains like retrieval augmented generation, coding, tool use, chain-of-thought reasoning, and many other domains of interest. Through extensive ablation studies, we design both pretraining and continual pretraining recipes for the \tucano suite (\textit{Base, Instruct, and Think}), which achieve state-of-the-art performance on several Portuguese-language modeling benchmarks. We also extend and refine the evaluation harness introduced in our earlier work, yielding a comprehensive evaluation suite that provides strong signals across different pretraining, continual pretraining, and post-training regimes. All artifacts associated with \tucano are openly released, including training recipes, logs, and source code, ensuring that our work is reproducible, accessible, and extendable by the broader Portuguese NLP community. 

\end{abstract}

\newpage
\tableofcontents

\newpage
\section{Overview}
\label{sec:introduction}

Large language models (LLMs) have radically transformed NLP, but the fruits of this progress have been unevenly distributed across languages \citep{cohere2024gap}. High-resource languages such as English have vast amounts of data and state-of-the-art models, whereas many low-resource languages lag far behind \citep{hedderich-etal-2021-survey, ranathunga2021neuralmachinetranslationlowresource, joshi-etal-2020-state}. Multilingual models still exhibit large performance gaps across underrepresented languages \citep{virtanen2019multilingualenoughbertfinnish, Martin_2020, armengolestape2021multilingualmodelsbestchoice, Corr_a_2025}, and reliance on closed, compute-intensive foundations limits adoption in resource-poor settings. It is worth further stressing the (un)reproducibility aspect. In Portuguese, for example, although recent years have seen growing interest in language-specific modeling \citep{silva2025evaluationaiethicstools, Cruz-Castaneda_Amadeus_2025}, only a small number of models have been developed and released in a fully open manner. Most high-performing systems remain closed-source or provide limited transparency regarding data, training procedures, and evaluation \citep{abonizio2025sabia3technicalreport, gaia-gemma-3-4b-2025, jurema_7b_2025}. This lack of openly available models, datasets, and reproducible recipes constrains independent research, hinders community-driven improvements, and potentially slows the creation of downstream applications. Consequently, there remains substantial space for the open-source community to contribute foundational LLM resources that are openly accessible, reproducible, and tailored to low-resource scenarios.

Meanwhile, the frontier of open-weight models is rapidly advancing. A new generation of publicly released LLMs is increasingly rivaling (or even surpassing) leading proprietary systems across a range of benchmarks. For example, recent models from DeepSeek AI \citep{deepseekai2025deepseekv3technicalreport} have been reported to match or exceed OpenAI’s GPT-5 and Google’s Gemini 3.0 Pro across several reasoning and coding evaluations. Similarly, Moonshot AI’s trillion-parameter Kimi K2 \citep{kimiteam2026kimik2openagentic} has demonstrated competitive or superior performance relative to Claude Sonnet 4.5 on demanding reasoning benchmarks. These trends, alongside strong results from the GLM series \citep{5team2025glm45agenticreasoningcoding}, demonstrate that open-weight models can approach the capabilities of state-of-the-art proprietary systems.

However, coming back to the (un)reproducibility aspect, it is important to distinguish between \textit{open-weight} and \textit{open-source} models. While many recent releases provide access to trained parameters, only a small number of projects disclose the complete stack required for reproducible LLM development, including training data, preprocessing pipelines, hyperparameters, and evaluation protocols. Notable exceptions include the SmolLM \citep{allal2025smollm2smolgoesbig, bakouch2025smollm3} and OLMo series \citep{groeneveld2024olmoacceleratingsciencelanguage, olmo2025olmo3}, among other inspiring efforts \citep{liu2023llm360, zhou2025megamath, k2team2025k2v2360openreasoningenhancedllm, apertus2025apertusdemocratizingopencompliant}. These initiatives exemplify a growing movement toward transparent, reproducible LLM research, in which models are accompanied by the artifacts needed for inspection, verification, and retraining. Nevertheless, such fully open efforts remain rare, particularly for low-resource languages, underscoring a critical gap---and an opportunity---for projects that prioritize openness and reproducibility.

In prior work, we sought to ``seed the soil'' for Portuguese NLP by providing foundational resources and models in an open-source fashion. In \cite{correa2024teenytinyllama}, we explored whether monolingual models could outperform multilingual ones when trained to the level recommended by the Chinchilla scaling laws \citep{hoffmann2022training} as compute-optimal. One of our main findings was that the 460-million-parameter version of our native models (the \textit{TeenyTinyLlama} pair) outperformed some multilingual models (e.g., Bloom 1b7 \citep{workshop2022bloom}) while approaching others (e.g., Qwen \citep{bai2023qwen}).

In \cite{correa2025tucanoadvancingneuraltext}, we started the \textit{Tucano} series. While the TTL pair was trained up to the theoretical optimal-compute allocation set by \cite{hoffmann2022training}---20 tokens $\times$ parameter---this series, we opted to follow certain results from the literature that showed that models past a certain size ($\sim$ 400 million parameters) tend to endure saturation for longer \citep{biderman2023pythia, touvron2023llama}, hence, leading us to experiment with more prolonged training runs. Something enabled by our data curation pipeline, resulting in the (at the time of release) largest Portuguese text dataset ever assembled: \textit{GigaVerbo}, a 200B-token Portuguese corpus we used to train the first Tucanos.

The first iteration of the Tucano series demonstrated highly promising results, with natively pretrained models outperforming---or matching---the performance of multilingual models of comparable size that were trained with substantially larger computational budgets and far broader multilingual corpora. However, the open-weight multilingual landscape has evolved dramatically, and achieving performance on par with current state-of-the-art multilingual LLMs---such as Qwen3 \citep{yang2025qwen3}---has become increasingly challenging, prompting us to rethink and redesign our approaches to developing language models tailored to specific linguistic domains.

In this work, we introduce \tucano, a new generation of open Portuguese LLMs that substantially extends our earlier efforts. We construct a large-scale, high-quality Portuguese corpus of approximately 320 billion tokens, enriched with educational and toxicity annotations and complemented by 9.3 billion synthetic tokens to mitigate domain gaps. To support data curation, we develop auxiliary datasets and train specialized annotation models for educational content, toxicity, and instruction quality. We further assemble a diverse supervised fine-tuning (SFT) corpus covering coding, tool use, structured outputs, mathematical problem solving, and chain-of-thought reasoning, and release a novel dual-reasoning preference dataset to further power direct alignment efforts. Leveraging these resources, we train a family of Portuguese LLMs with parameter counts ranging from 0.5B to 3.5B, using pretraining and continual pretraining regimes. Our models achieve strong gains over prior open Portuguese baselines and outperform similarly sized multilingual models across multiple Portuguese benchmarks. Finally, we release a comprehensive evaluation harness for Portuguese that supports assessment at both early and late training stages, including long-context settings.

\textbf{In summary, our main contributions are:}

\begin{itemize}
    \item \textbf{Large-scale Portuguese corpus.} A $\sim$320B-token cleaned corpus with rich educational and toxicity annotations (\gigaverbo), plus 9.3B synthetic tokens for domain coverage (\gigaverbosynth).
    \item \textbf{Annotation and filtering models.} Educational, toxicity, and instruction-quality classifiers trained from new auxiliary datasets and released for community use.
    \item \textbf{Diverse SFT data.} A large supervised fine-tuning collection spanning coding, tool calling, structured output, reasoning, and mathematics (\gigaverbosft).
    \item \textbf{Dual-reasoning preference dataset.} A new resource for preference optimization that captures complementary aspects of reasoning quality (\gigaverbopreferences).
    \item \textbf{Tucano 2 model family.} Open Portuguese foundation models from 0.5B–3.5B parameters that outperform most prior Portuguese models of similar size (\tucano).
    \item \textbf{Portuguese evaluation suite.} Benchmarks covering early- and late-stage training, including long-context evaluation.
    \item \textbf{Fully open release.} All datasets, models, training recipes, and evaluation code are publicly released under permissive licenses.
\end{itemize}

This work is part of the broader Polyglot project. The methodology and findings presented here extend to additional language-specific studies conducted within the same framework, including Hindi (e.g., \href{https://huggingface.co/collections/Polygl0t/lilmoo}{LilMoo}) and Bengali (e.g., \href{https://huggingface.co/collections/Polygl0t/liltii}{LilTii}). For further details on these parallel efforts and associated resources, please refer to the Polyglot project page: \href{https://huggingface.co/Polygl0t}{huggingface.co/Polygl0t}.

\newpage
\section*{Released Assets}
\textbf{Base Models}
\\[1mm]
\begin{small}
\hflogo{\href{https://huggingface.co/Polygl0t/Tucano2-0.6B-Base}{Polygl0t/Tucano2-0.6B-Base}} \\
\hflogo{\href{https://huggingface.co/Polygl0t/Tucano2-qwen-0.5B-Base}{Polygl0t/Tucano2-qwen-0.5B-Base}} \\
\hflogo{\href{https://huggingface.co/Polygl0t/Tucano2-qwen-1.5B-Base}{Polygl0t/Tucano2-qwen-1.5B-Base}} \\
\hflogo{\href{https://huggingface.co/Polygl0t/Tucano2-qwen-3.7B-Base}{Polygl0t/Tucano2-qwen-3.7B-Base}} \\
\end{small}

\textbf{Instruct Models}
\\[1mm]
\begin{small}
\hflogo{\href{https://huggingface.co/Polygl0t/Tucano2-qwen-0.5B-Instruct}{Polygl0t/Tucano2-qwen-0.5B-Instruct}} \\
\hflogo{\href{https://huggingface.co/Polygl0t/Tucano2-qwen-1.5B-Instruct}{Polygl0t/Tucano2-qwen-1.5B-Instruct}} \\
\hflogo{\href{https://huggingface.co/Polygl0t/Tucano2-qwen-3.7B-Instruct}{Polygl0t/Tucano2-qwen-3.7B-Instruct}} \\
\end{small}

\textbf{Reasoning Models}
\\[1mm]
\begin{small}
\hflogo{\href{https://huggingface.co/Polygl0t/Tucano2-qwen-0.5B-Think}{Polygl0t/Tucano2-qwen-0.5B-Think}} \\
\hflogo{\href{https://huggingface.co/Polygl0t/Tucano2-qwen-1.5B-Think}{Polygl0t/Tucano2-qwen-1.5B-Think}} \\
\hflogo{\href{https://huggingface.co/Polygl0t/Tucano2-qwen-3.7B-Think}{Polygl0t/Tucano2-qwen-3.7B-Think}} \\
\end{small}

\textbf{Auxiliary Models}
\\[1mm]
\begin{small}
\hflogo{\href{https://huggingface.co/Polygl0t/portuguese-bertimbau-edu-classifier}{Polygl0t/portuguese-bertimbau-edu-classifier}} \\
\hflogo{\href{https://huggingface.co/Polygl0t/portuguese-bertimbau-large-edu-classifier}{Polygl0t/portuguese-bertimbau-large-edu-classifier}} \\
\hflogo{\href{https://huggingface.co/Polygl0t/portuguese-bertimbau-toxicity-classifier}{Polygl0t/portuguese-bertimbau-toxicity-classifier}} \\
\hflogo{\href{https://huggingface.co/Polygl0t/portuguese-bertabaporu-large-toxicity-classifier}{Polygl0t/portuguese-bertabaporu-large-toxicity-classifier}} \\
\hflogo{\href{https://huggingface.co/Polygl0t/portuguese-qwen3-4b-instruct-quality-classifier}{Polygl0t/portuguese-qwen3-4b-instruct-quality-classifier}} \\
\hflogo{\href{https://huggingface.co/Polygl0t/portuguese-qwen3-4b-instruct-quality-judge}{Polygl0t/portuguese-qwen3-4b-instruct-quality-judge}} \\
\end{small}

\textbf{Datasets}
\\[1mm]
\begin{small}
\datalogo{\href{https://huggingface.co/datasets/Polygl0t/gigaverbo-v2}{Polygl0t/gigaverbo-v2}}\\
\datalogo{\href{https://huggingface.co/datasets/Polygl0t/gigaverbo-v2-synth}{Polygl0t/gigaverbo-v2-synth}}\\
\datalogo{\href{https://huggingface.co/datasets/Polygl0t/gigaverbo-v2-sft}{Polygl0t/gigaverbo-v2-sft}}\\
\datalogo{\href{https://huggingface.co/datasets/Polygl0t/gigaverbo-v2-preferences}{Polygl0t/gigaverbo-v2-preferences}}\\
\datalogo{\href{https://huggingface.co/datasets/Polygl0t/portuguese-edu-qwen-annotations}{Polygl0t/portuguese-edu-qwen-annotations}}\\
\datalogo{\href{https://huggingface.co/datasets/Polygl0t/portuguese-toxicity-qwen-annotations}{Polygl0t/portuguese-toxicity-qwen-annotations}}\\
\datalogo{\href{https://huggingface.co/datasets/Polygl0t/portuguese-instruct-quality-qwen-annotations}{Polygl0t/portuguese-instruct-quality-qwen-annotations}}\\
\end{small}

\textbf{Benchmarks}
\\[1mm]
\begin{small}
\benchmarklogo{\href{https://huggingface.co/datasets/Polygl0t/IFEval-PT}{IFEval-PT}}\\
\benchmarklogo{\href{https://huggingface.co/datasets/Polygl0t/gsm8k-pt}{GSM8K-PT}}\\
\benchmarklogo{\href{https://github.com/Polygl0t/lm-evaluation-harness/blob/ruler_pt/lm_eval/tasks/ruler_pt/README.md}{RULER-PT}}\\
\end{small}

\textbf{Code:}
\\[1mm]
\begin{small}
\githublogo{\href{https://github.com/Polygl0t/llm-foundry}{Polygl0t/llm-foundry}} \\
\githublogo{\href{https://github.com/Polygl0t/lm-evaluation-harness}{Polygl0t/lm-evaluation-harness}} \\
\end{small}

\newpage
\begin{tcolorbox}[colback=gray!5!white, colframe=brickred!75!black, title=Safety Considerations]
\small

Like most large language models trained on web data, \tucano should not be considered a turnkey solution for safety-critical or high-stakes applications. Despite extensive data curation and filtering, the model can generate hallucinated content that appears plausible but is factually incorrect. It may also reflect social and historical biases present in its training data, potentially producing toxic, harmful, or offensive outputs. Given that the model is primarily optimized for Portuguese, performance in other languages may be inconsistent, leading to misinterpretations. Additionally, the model may exhibit undesirable generation behaviors such as excessive verbosity or repetitive loops under certain decoding settings. Consequently, although \tucano is released under a permissive license, we strongly recommend that users conduct appropriate risk assessments, apply additional safeguards, and perform domain-specific evaluation before deploying the model in real-world systems.
\end{tcolorbox}

\newpage
\section{Related Work}
\label{sec:related-work}

\begin{tcolorbox}[colback=gray!5!white, colframe=gray!75!black, title=Summary / TL;DR]
\small

Taken together, the 2025 landscape presents both challenges and opportunities for the development of Portuguese LLMs. On one hand, native Portuguese models (e.g., Curi{\'o}, Amadeus-Verbo) demonstrate that targeted adaptation can yield strong domain performance, particularly when training budgets are constrained. On the other hand, leading multilingual models (Qwen3, Gemma 3, Falcon 3) now allocate substantial capacity to Portuguese within trillion-token multilingual corpora, raising the bar for competitive performance. However, reproducibility remains inconsistent: while models like SmolLM3 and OLMo 3 exemplify full-stack openness, many Portuguese-adapted models provide limited documentation of data composition, training procedures, and evaluation protocols. Our work seeks to address this gap by contributing not only competitive Portuguese foundation models but also the assets---datasets, training recipes, and evaluation harnesses---necessary to sustain open research.
\end{tcolorbox}

In our previous work with Tucano, we presented a detailed timeline of large language model development in Portuguese \citep{correa2025tucanoadvancingneuraltext}, highlighting key milestones and models that have advanced NLP in the language up to 2024.\footnote{A timeline of releases and publications is available in \href{https://github.com/Nkluge-correa/Tucano/blob/main/img/timeline.pdf}{github.com/Nkluge-correa/Tucano/blob/main/img/timeline.pdf}.} In 2025, substantial progress was made in both Portuguese-specific and multilingual language model development. This section surveys key releases that inform the context and baselines for our work, focusing on models that either target Portuguese or include comprehensive Portuguese support within a multilingual framework. For further \textit{expos{\'e}s} on this topic, specifically where it comes to the developments tied to the field in Portuguese, we refer readers to the works of \cite{silva2025evaluationaiethicstools} and \cite{Cruz-Castaneda_Amadeus_2025}.

\subsection{Portuguese-Centric Models}

\paragraph{Curi{\'o}.} The Curi{\'o} family \citep{almeida2025buildinghighqualitydatasetsportuguese} represents a continued pretraining approach that adapts English LLaMA models (TinyLLaMA-1T and LLaMA-2-7B) to Portuguese. Curi{\'o}-1B and Curi{\'o}-7B are trained on approximately 150 billion tokens drawn primarily from the ClassiCC-PT corpus, leveraging cross-lingual transfer to inherit structural knowledge from their English base models. Meanwhile, Curi{\'o}-Edu \citep{almeida2025curioedu7bexaminingdata} focuses training on educational and STEM-filtered subsets of the corpus (roughly 20 billion tokens). Despite using only 10\% of the data and 20\% of the compute, Curi{\'o}-Edu-7B reportedly outperforms the full-corpus Curi{\'o}-7B on Portuguese benchmarks, highlighting the value of domain-focused curation. Both model families are openly released on Hugging Face under the ClassiCC-PT project.

\paragraph{GAIA.} GAIA \citep{gaia-gemma-3-4b-2025} is a 4-billion-parameter Brazilian Portuguese model derived from Google's Gemma-3-4b-pt through continual pretraining on approximately 13 billion tokens of high-quality Portuguese data (scientific articles, Wikipedia, and curated web sources). The model card mentions the use of weight-merging techniques across training stages, though detailed dataset composition and training logs are not publicly disclosed. GAIA targets Brazilian Portuguese applications and has been evaluated on language-specific benchmarks such as the ENEM \citep{silveira2017university} and OAB exam \citep{d2017passing}.

\paragraph{Jurema-7B.} Targeting the Brazilian legal domain, Jurema-7B \citep{jurema_7b_2025} fine-tunes Qwen2.5-7B-Instruct on a synthetic question-answer dataset generated from high-quality legal documents. The model's specialization yields substantial gains on domain-specific tasks. Jurema-7B is released under the Apache 2.0 license for research use, but no accompanying academic publication provides training details.

\paragraph{Amadeus-Verbo.} Cruz-Castañeda and Amadeus \citep{cruzcastaneda2025amadeusverbotechnicalreportpowerful} introduce Amadeus-Verbo, a family of Brazilian Portuguese models spanning 0.5B to 72B parameters. Built on the Qwen2.5 series, these models include base-tuned, merged, and instruction-tuned variants. Although the models are publicly released on Hugging Face, the composition of the Brazilian Portuguese training corpus is not disclosed.

\paragraph{Carvalho.} Carvalho \citep{gamallo2024galician} is a bilingual Galician–European Portuguese decoder model (1.3B parameters) developed via continual pretraining from Cerebras-GPT-1.3B \cite{dey2023cerebrasgptopencomputeoptimallanguage}. Trained on over 6 billion words of mixed Galician and Portuguese text with a jointly learned tokenizer, Carvalho demonstrates strong performance on Galician generation tasks and provides a valuable resource for closely related Ibero-Romance languages.

\subsection{Multilingual Models with Portuguese Coverage}

The multilingual frontier in 2025 is characterized by significant scale increases and architectural innovations that bring open-weight models (supporting Portuguese) closer to state-of-the-art performance across languages.

\paragraph{Qwen3.} Alibaba's Qwen3 family \citep{yang2025qwen3} represents a landmark release, with models spanning from 0.6B to 235B parameters (including mixture-of-experts variants). Pretrained on approximately 36 trillion tokens covering 119 languages, Qwen3 achieves competitive results with leading proprietary models on coding, mathematics, and general reasoning benchmarks. The dense models support up to 128K tokens and are released under the Apache 2.0 license. Notably, Qwen3-4B rivals the much larger Qwen2.5-72B on several tasks, illustrating substantial efficiency gains. The series also introduces dual-mode inference---a \textit{think mode} for step-by-step reasoning and \textit{non-think} for rapid responses---enabling flexible deployment under latency and quality constraints.

\paragraph{Gemma 3.} Google DeepMind's Gemma 3 series \citep{gemmateam2025gemma3technicalreport} comprises five core variants (270M, 1B, 4B, 12B, and 27B parameters), with multimodal capabilities (text and image) in the larger models and support for over 140 languages. Gemma3-4B matches Gemma 2's 27B model on key benchmarks, and Gemma3-27B rivals Gemini 1.5 Pro. All models support 128K token contexts and are released under the ``Gemma Terms of Use''.

\paragraph{Falcon 3.} The Technology Innovation Institute's Falcon 3 family \citep{Falcon3} includes models at 1B, 3B, 7B, and 10B scales, trained on approximately 14 trillion tokens of English, Spanish, Portuguese, and French data. Falcon3-10B achieves state-of-the-art results among sub-13B models, with particularly strong performance on mathematical reasoning. The models support up to 32K token contexts and are released under TII's permissive Falcon License.

\paragraph{Other Notable Releases.} Beyond domain or language-specific models, two fully open releases merit attention for their transparency and multilingual scope. SmolLM3 \citep{bakouch2025smollm3}, a 3-billion-parameter model from Hugging Face, supports six languages (including Portuguese) and offers dual-mode instruct/reasoning control with explicit \texttt{think} and \texttt{no\_think} flags. It supports 128K token contexts and is released with complete training blueprints. OLMo 3 \citep{olmo2025olmo3}, developed by the Allen Institute for AI, provides 7B- and 32B-parameter models with full transparency---data, checkpoints, and training recipes are all openly released. OLMo 3-Think variants emphasize intermediate reasoning and are designed to facilitate custom-domain adaptation and post-training experimentation.

\newpage
\section{Pretraining Data}
\label{sec:pretraining_data}

\begin{tcolorbox}[colback=gray!5!white, colframe=gray!75!black, title=Summary / TL;DR]
\small

We introduce \gigaverbo, a substantially improved Portuguese corpus comprising $\sim$ 320 billion tokens across 372 million documents, built using a language-specific filtering pipeline inspired by FineWeb 2. We employ Qwen2.5-32B as an LLM judge to annotate 700K documents for educational quality and toxicity, then distill these annotations into lightweight classifiers for corpus-scale filtering. The resulting corpus provides rich metadata (educational and toxicity scores) and is complemented by \gigaverbosynth, a 9.3-billion-token synthetic dataset generated using the Qwen2.5 series. Ablation studies demonstrate that models trained on a mixture of educational and synthetic data substantially outperform those trained on non-educational web data alone, achieving competitive performance with other models trained on significantly larger compute budgets.
\end{tcolorbox}

A high-quality pretraining corpus is foundational to effective language model development. In our prior work with Tucano, we introduced \textit{GigaVerbo}, a large-scale Portuguese text corpus comprising over 200 billion tokens across 145 million documents \citep{correa2025tucanoadvancingneuraltext}. GigaVerbo was created by aggregating a wide range of openly available Portuguese datasets hosted on Hugging Face, including web crawls, encyclopedic articles, blogs, translated conversational datasets, legal documents, and instruction-following corpora. To ensure consistency and quality, the corpus underwent exact hash deduplication and was filtered using a fine-tuned BERTimbau classifier trained on 110K GPT-4o-annotated samples, thereby removing low-quality text. At the time of its release, GigaVerbo was among the largest openly available Portuguese text corpora, alongside other notable efforts such as ClassiCC-PT \citep{almeida2025buildinghighqualitydatasetsportuguese}, Jabuticaba \citep{Cruz-Castaneda_Amadeus_2025}, and CrawlPT \citep{garcia2024crawlptlargescalecleanedportugueseweb}.

In this work, we introduce \gigaverbo, an updated and substantially improved version of the original corpus. The development of \gigaverbo was informed by recent advances in corpus construction methodology, particularly the emphasis on language-specific filtering, learned quality classifiers, and synthetic data augmentation. We first survey related efforts in Portuguese corpus construction to contextualize our approach, then detail the technical pipeline used to build \gigaverbo.

\subsection{Related Data Efforts}

\paragraph{CrawlPT.} Developed by \cite{garcia2024crawlptlargescalecleanedportugueseweb}, CrawlPT is a large-scale Portuguese corpus constructed by aggregating three major open-source datasets: brWaC, the Portuguese subset of CC100, and the Portuguese portion of OSCAR-2301. The resulting collection comprises over 52 million deduplicated documents, providing a broad, general-purpose snapshot of Portuguese web text, primarily derived from the Common Crawl. The preprocessing pipeline includes tokenization, language filtering, and extensive document-level deduplication. CrawlPT was used to train RoBERTaLexPT-base and RoBERTaCrawlPT-base. Additionally, the authors released LegalPT, a legal-domain dataset comprising more than 24 million documents aggregated from publicly available Portuguese-language legal data. All resources are openly available on Hugging Face.

\paragraph{ClassiCC-PT.} Developed by \cite{almeida2025buildinghighqualitydatasetsportuguese}, ClassiCC-PT is built directly from raw Common Crawl snapshots using a principled and fully auditable pipeline inspired by the FineWeb series \citep{penedo2024finewebdatasetsdecantingweb, penedo2025fineweb2pipelinescale}. The resulting corpus contains approximately 120 billion tokens and achieves performance comparable to industrial-grade corpora such as ClueWeb22, despite being constructed entirely from open infrastructure using publicly reproducible methods. ClassiCC-PT and its accompanying resources, including learned filters for Portuguese and the Curi{\'o} model family, are openly released on Hugging Face.

\paragraph{Jabuticaba.} Developed by \cite{Cruz-Castaneda_Amadeus_2025}, Jabuticaba is the largest commercial corpus of Portuguese text, comprising 669 GB (approximately 139 billion tokens) of curated data primarily sourced from Brazilian web content. The corpus employs a comprehensive methodological pipeline that includes language detection, content filtering, quality assessment, toxicity removal, normalization, and deduplication. Unlike GigaVerbo and ClassiCC-PT, the implementation details of Jabuticaba's pipeline are not fully open-sourced, though the authors provide a thorough description of their approach. The dataset itself is not publicly available due to its commercial nature, and to our knowledge, no published ablation studies or trained models using Jabuticaba as the primary pretraining corpus have been released.

\subsection{GigaVerbo-v2: Construction}

The recent landscape of corpus construction reveals several converging methodological insights that informed our development of \gigaverbo:

\begin{itemize}
    \item \textbf{Language-specific filtering:} Works such as ClassiCC-PT and FineWeb 2 demonstrate that language-specific filtering pipelines outperform multilingual, English-centric approaches \citep{almeida2025buildinghighqualitydatasetsportuguese, penedo2025fineweb2pipelinescale}.
    \item \textbf{Quality over quantity:} Both ClassiCC-PT and FineWeb-Edu show that data selection and quality filtering are more important than raw scale for effective model pretraining, as measured by downstream performance on ablation studies \citep{penedo2024finewebdatasetsdecantingweb}.
    \item \textbf{Learned filters:} The use of learned filters, particularly LLM-as-a-Judge approaches \citep{gunasekar2023textbooksneed, penedo2024finewebdatasetsdecantingweb}, has become increasingly popular for refining dataset quality beyond heuristic methods.
    \item \textbf{Synthetic data augmentation:} There is a growing trend toward augmenting datasets with synthesized data. Works such as BeyondWeb \citep{maini2025beyondweb} and Cosmopedia \citep{allal2024cosmopedia} show that incorporating model-generated content can significantly enhance dataset quality.
\end{itemize}

With these insights in mind, our goals for \gigaverbo were to (1) increase the overall size of the corpus, (2) improve data quality through both heuristic and learned filtering, (3) augment the corpus with high-quality synthesized data, and (4) maintain full openness and reproducibility of the entire pipeline.
    
\subsubsection{Data Extraction and Language Identification}

To ensure the creation of a high-quality Portuguese corpus, we implemented a data processing pipeline inspired by the FineWeb 2 methodology \citep{penedo2025fineweb2pipelinescale} and used the Datatrove library \citep{penedo2024datatrove}. This pipeline combines text extraction, language identification, quality filtering, and deduplication to produce a clean text corpus.

For web-crawled data sourced from Common Crawl WARC files, we begin by extracting text content using the Trafilatura library \citep{barbaresi-2021-trafilatura}. This extraction step is bypassed for datasets sourced from Hugging Face, which often include pre-cleaned text, allowing direct input into subsequent filtering stages. As in the first version of GigaVerbo, we sought to consolidate a wide range of openly available Portuguese datasets hosted on Hugging Face, since these datasets have often already undergone some degree of cleaning and curation.\footnote{Since most datasets obtained from the Hub have already been processed from earlier Common Crawl dumps, reprocessing the same data would result in substantial redundant and wasteful resource allocation. Therefore, we limited our processing to Common Crawl dumps released after the publication dates of the external datasets we used, with a primary focus on the 2025 crawl releases. In total, after running all our quality filters, we retained $\sim$ 6 billion tokens of timely web content.}

To ensure relevance and baseline quality, we apply an initial URL filter to remove documents from undesirable sources, using blocklists\footnote{\href{https://github.com/maravento/blackweb}{github.com/maravento/blackweb}.} to exclude low-quality or inappropriate websites. This is followed by a language identification step using FastText (\texttt{FT176}) \citep{joulin2016fasttextzipcompressingtextclassification}. Documents are retained only if they are confidently classified as Portuguese, with a default confidence threshold of 0.65. To improve language identification accuracy, we perform a second round using GlotLID \citep{Kargaran_2023} as the backend, setting the language score threshold to 0.87.\footnote{These configurations were imported from the settings used by \cite{penedo2025fineweb2pipelinescale}.}

\subsubsection{Data Filtering and Formatting}

While the definition of document ``quality'' is inherently subjective and context-dependent, we operationalize quality based on several heuristic criteria that reflect linguistic coherence and structural integrity. Specifically, we import the filters developed by \citet{penedo2025fineweb2pipelinescale} (FineWeb 2) and \citet{rae2022scalinglanguagemodelsmethods} (MassiveText), tuning them to be more sensitive to Portuguese's linguistic characteristics, including script, custom stop words, standard word lengths, and punctuation norms.

\begin{itemize}
    \item \textbf{Repetition filtering} (\texttt{GopherRepetitionFilter}): This filter removes documents with excessive repetition, which is particularly relevant for web data prone to boilerplate or auto-generated content.
    \item \textbf{Quality filtering} (\texttt{FineWebQualityFilter} and \texttt{GopherQualityFilter}): These filters enforce constraints on document structure and content quality to ensure documents have adequate punctuation and natural text flow. They also evaluate lexical properties, such as average word length and the presence of Portuguese stop words (e.g., \textit{como}, \textit{que}, \textit{para}, \textit{por}, \textit{com}).
\end{itemize}

Post-filtering, we apply formatting steps to correct encoding issues, remove personally identifiable information, and eliminate or replace undesirable patterns (e.g., excessive symbols).

\subsubsection{Data Deduplication}

According to \citet{lee2022deduplicatingtrainingdatamakes}, ``deduplication allows us to train models that emit memorized text ten times less frequently and require fewer train steps to achieve the same or better accuracy.'' We therefore implement a deduplication pipeline using the MinHash algorithm to address redundancy. MinHash scales efficiently across multiple CPU nodes and enables tuning of similarity thresholds.\footnote{Following \citet{penedo2025fineweb2pipelinescale}, we use 14 buckets, 8 hashes per bucket, and 5-grams, employing \texttt{xxHash} for hashing.}

For Hugging Face data sources, we expected substantial overlap, given that most available datasets originate from similar sources (e.g., Common Crawl). Therefore, for these datasets, we performed deduplication globally. For Common Crawl data, we performed deduplication on a per-snapshot basis, following \cite{penedo2024finewebdatasetsdecantingweb}, as most of the snapshots we processed were relatively recent and distinct.

\subsubsection{Quality Annotation via LLM Judges}

Recent work has demonstrated that learned quality filters, particularly those based on LLM judgments, can provide more nuanced quality control than heuristic methods alone \citep{gunasekar2023textbooksneed, penedo2024finewebdatasetsdecantingweb}. Following this approach, we augment our heuristic filtering pipeline with two complementary dimensions of learned filtering: educational quality and toxicity.

We employ Qwen/Qwen2.5-32B-Instruct \citep{qwen2025qwen25technicalreport} as an LLM judge to annotate document quality. This model was selected for its strong multilingual performance, support for the Portuguese language, and a permissive Apache-2.0 license, which enables open redistribution of the resulting annotations. We developed two evaluation protocols: an educational quality assessment inspired by FineWeb-Edu \citep{penedo2024finewebdatasetsdecantingweb}, which ranks documents on a 5-point Likert scale according to their suitability for educational use (primary through secondary level), and a toxicity assessment that evaluates the presence of harmful, offensive, or inappropriate content on a similar 5-point scale. Both prompts elicit structured JSON responses to facilitate automated processing. The complete prompts are provided in Appendix~\ref{appendix:llm-judge-prompts}.

We applied these evaluation protocols to a stratified random sample of 700,000 documents drawn from the deduplicated corpus, sampling proportionally across all data sources to ensure representativeness.

To enable efficient corpus-scale filtering, we distilled these LLM annotations into lightweight classification models. We fine-tuned several Portuguese encoder-based models---including BERTimbau \citep{souza2020bertimbau}, BERTabaporu \citep{costa-etal-2023-bertabaporu}, RoBERTaLexPT \citep{garcia2024crawlptlargescalecleanedportugueseweb}, RoBERTaCrawlPT \citep{garcia2024crawlptlargescalecleanedportugueseweb}, and DeBERTinha \citep{campiotti2023debertinhamultistepapproachadapt}---on the annotated datasets following the training procedure of \citet{penedo2024finewebdatasetsdecantingweb}. Each candidate model was evaluated on a held-out test set of 20,000 examples, with model selection based on macro F1 score. Detailed training configurations and full results are provided in Appendix~\ref{appendix:classifier-training}.

The best-performing models achieved competitive results: BERTimbau-large-cased obtained a macro F1 score of 0.57 on the 5-class educational quality task (0.79 F1 when binarized), while BERTabaporu-large-uncased achieved a macro F1 score of 0.60 on the 5-class toxicity task (0.70 F1 when binarized). These results are comparable to those reported by \citet{penedo2024finewebdatasetsdecantingweb} and \citet{almeida2025buildinghighqualitydatasetsportuguese} for similar filtering tasks in English and Portuguese, respectively. We release the top two performing models for each task.

\subsubsection{Final Data Composition}

After applying the complete filtering pipeline---including text extraction, language identification, heuristic quality filtering, deduplication, and learned filtering using our trained classifiers---we performed two final post-processing steps: (1) removal of documents shorter than 50 tokens, and (2) separation of documents with a toxicity score greater than 3. Given that toxicity is a challenging issue in Portuguese web data, we retained the filtered documents as a separate subset (\texttt{excluded}) within our corpus, which may be valuable for future research on toxicity detection and mitigation in Portuguese NLP. The main subset (\texttt{default}) contains the final curated corpus used for model pretraining.

The resulting corpus, \gigaverbo, comprises 372 million documents in the \texttt{default} subset, totaling approximately 834 GB and 318 billion tokens. The \texttt{excluded} subset contains an additional 2.9 million documents (7.8 GB, 3 billion tokens). Combined, the corpus provides 841.8 GB of Portuguese text spanning 320 billion tokens. Of the tokens in the \texttt{default} subset, approximately 119 billion (37\%) are classified as educational content (\texttt{edu\_int\_score} $\geq$ 3), while the remaining 199 billion tokens (63\%) comprise the remaining portion.

The corpus aggregates data from three primary source categories: Common Crawl snapshots (primarily from 2020--2025), curated datasets from Hugging Face (including web crawls, encyclopedic content, conversational data, and domain-specific corpora), and other openly available Portuguese resources (academic theses and public domain literature). Each document is annotated with educational quality and toxicity scores, enabling flexible filtering for downstream applications. Detailed information about data sources, licenses, token distributions per subset, and quality score distributions is provided in Appendix~\ref{appendix:base-data-sources-statistics}.

\subsection{GigaVerbo-v2-Synth: Synthetic Augmentation}

To complement our web-sourced corpus, we generated a synthetic dataset named \gigaverbosynth, inspired by approaches such as Cosmopedia \citep{allal2024cosmopedia}. The goal was to generate high-quality, diverse text data to fill gaps in domains where web data are scarce or of low quality.

\paragraph{Generation Pipeline.} Our pipeline consisted of four stages: (1) carefully crafted prompts to steer generation, (2) a diverse set of seed datasets to provide context and variety, (3) state-of-the-art LLMs to perform generation, and (4) filtering to ensure quality. We employed three models from the Qwen2.5 family \citep{qwen2025qwen25technicalreport}: Qwen2.5-32B-Instruct for complex tasks, such as generating detailed reasoning traces for mathematical problems; Qwen2.5-14B-Instruct for intermediate tasks; and Qwen2.5-7B-Instruct for simpler generation tasks, such as producing structured summaries of low-quality web samples.

For seed data, we selected 15 diverse datasets spanning education, science, technology, law, literature, and general knowledge (including FineWeb-Edu, Cosmopedia-v2, FineMath, Wikipedia, LegalPT, CodeParrot, and others---see Appendix~\ref{appendix:synth-data-sources-statistics} for the complete list). Domain coverage in the final dataset includes educational texts, short academic articles, tutorials, WikiHow-style guides, blog posts, legal text summaries, mathematical walk-throughs with step-by-step reasoning, programming tutorials, philosophy articles, short stories, children's stories, synthetic email exchanges, and extensive sequences of multiple-choice QA pairs. Because many evaluation benchmarks rely heavily on multiple-choice formats, we intentionally included a substantial volume of such data, as prior work has shown that exposure to this format can improve LLM performance on similarly framed tasks \citep{kang2025demystifyingsyntheticdatallm, yang2024syntheticcontinuedpretraining}.

\paragraph{Quality Filtering and Decontamination.} All samples were post-processed to ensure quality and avoid benchmark contamination. We implemented a decontamination method based on contiguous token-level $k$-gram matching, rather than word-level n-grams, to detect overlaps between synthetic data and reference benchmark datasets. For each synthetic sample, we extracted all contiguous sequences of tokenizer output tokens ($k$-grams) of length $k \in [8, 32]$ and checked their membership in a benchmark index constructed in the same token space; any samples flagged as non-members were removed. This approach is similar to that of \citet{muennighoff2025s1simpletesttimescaling}. We also applied approximate matching in token space to account for minor variations (e.g., punctuation differences or single-token substitutions). Finally, we removed samples containing characters outside the Portuguese (Latin script) Unicode range, thereby ensuring high confidence that models did not drift into other languages.\footnote{It was common to find samples where models trailed off to write passages in Chinese.}

\paragraph{Dataset Statistics.} The generation process spanned 4 months (January--June 2025) using 16 NVIDIA A40 GPUs distributed across two nodes, leveraging the vLLM library \citep{kang2025demystifyingsyntheticdatallm} as our inference engine. We estimate approximately 48,000 GPU hours were consumed, corresponding to approximately 14,400 kWh of energy, yielding a carbon footprint of 5.3 metric tons of CO$_2$e (see Section~\ref{sec:resource_consumption}). The final \gigaverbosynth dataset comprises 11,237,546 samples totaling approximately 9.3 billion tokens. Detailed statistics by generator and seed dataset are provided in Appendix~\ref{appendix:synth-data-sources-statistics}.

\subsection{Data Ablation Study}

To evaluate the quality of the different portions of our dataset, we conducted an ablation study to assess the downstream impact of these subsets on benchmark performance. More specifically, we prepared four different data mixtures:

\begin{enumerate}
    \item \textbf{Edu:} Consisting solely of the educational portion of our web-sourced corpus (samples with educational quality score $\geq$ 3).
    \item \textbf{Synth:} Consisting solely of our synthetic dataset.
    \item \textbf{Edu+Synth:} A combination of both the educational portion of our web-sourced corpus and our synthetic dataset.
    \item \textbf{Non-Edu:} Consisting solely of the non-educational portion of our web-sourced corpus (samples with educational quality score $<$ 3).
\end{enumerate}

To measure the impact of each data mixture on model performance, we trained separate 1.5-billion-parameter language models on each dataset, using the same architecture and training hyperparameters across all runs. All models were trained on a total of 46 billion tokens, corresponding to a compute budget of approximately 4.14$\times$10$^{20}$ FLOPs. For the Edu+Synth mixture, we sampled from both datasets equally during training, ensuring the model saw 23 billion tokens from each subset. For the pure Synth mix, we had to repeat the entire dataset $\sim$5x. The model architecture, training hyperparameters, and detailed experimental configuration are provided in Appendix~\ref{appendix:ablations}.

For evaluation, we employed the easy set of our evaluation suite, which includes benchmarks such as ARC Challenge \citep{clark2018thinksolvedquestionanswering}, Calame \citep{lopes2024gloriagenerativeopen}, Global PIQA \citep{chang2025globalpiqaevaluatingphysical}, HellaSwag \citep{zellers2019hellaswagmachinereallyfinish}, and Lambada \citep{lambada2016}. These benchmarks were selected because they provide a good signal of model performance even at early training stages, allowing us to observe the impact of different data mixtures without requiring extensive training (see Section~\ref{sec:building-evaluation-harness}). We also include results from the Tucano models \citep{correa2025tucanoadvancingneuraltext} for comparison, as they were trained on Portuguese data of a similar nature and provide multiple checkpoints at different training stages. To keep our comparison fair, we selected Tucano checkpoints with the closest training-token counts (52 billion) to our ablation models (46 billion).

\begin{figure}[h]
\centering
\includegraphics[width=0.95\linewidth]{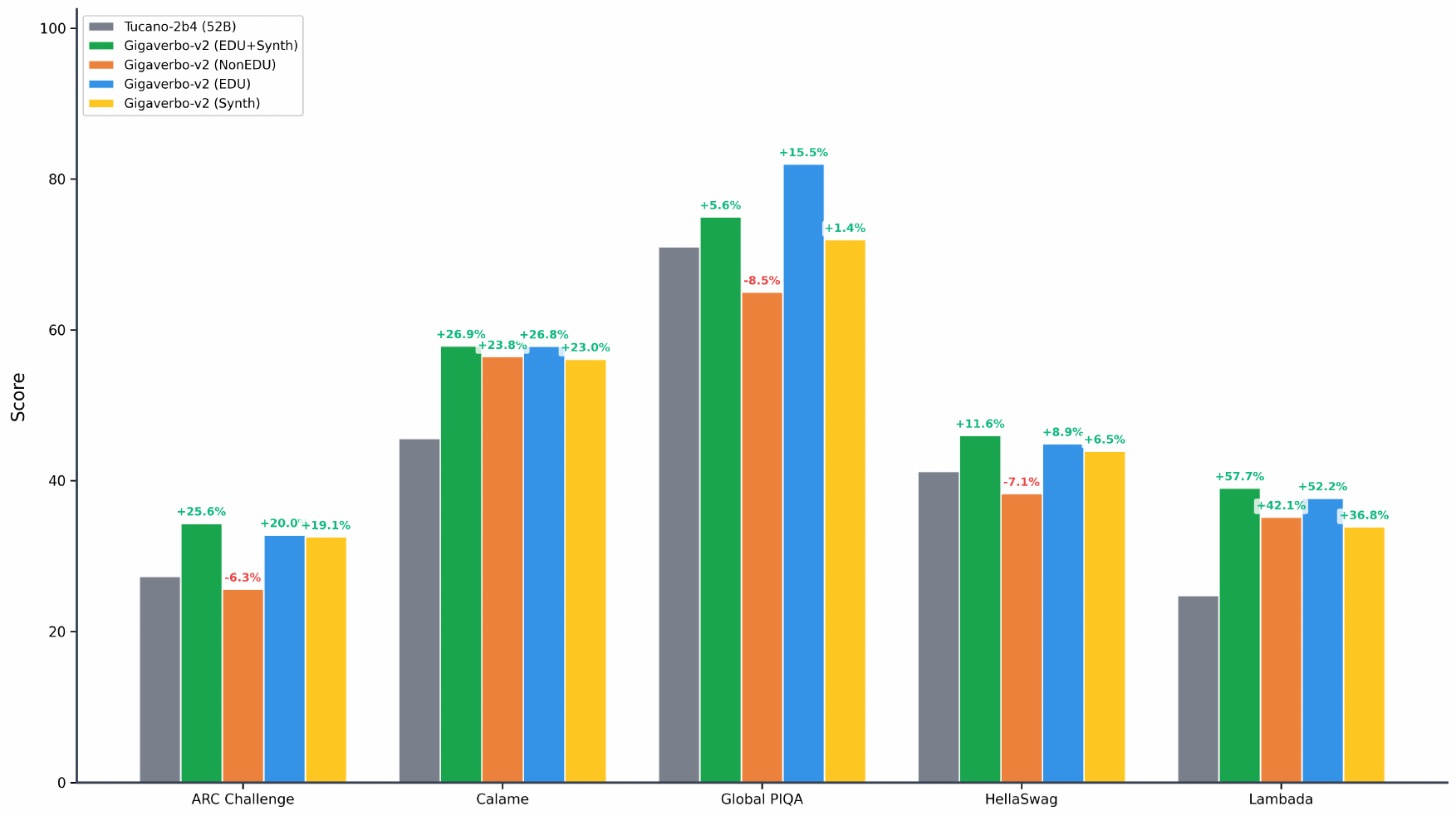}
\caption{Impact of Educational \& Synthetic Data (46B tokens). The Edu+Synth and Edu mixtures achieve the best performance across benchmarks, substantially outperforming the Non-Edu mixture and the Tucano-2b4 baseline. The percentage values represent the relative increase/decrease in performance with regard to the Tucano-2b4 baseline.}
\label{fig:gigaverbo-v2-ablation}
\end{figure}

Results show a significant improvement in downstream performance across benchmarks when compared to the first version of GigaVerbo. As illustrated in Figure~\ref{fig:gigaverbo-v2-ablation}, the Edu+Synth model achieves the best performance on ARC Challenge (34.4\%), substantially outperforming the Non-Edu mixture (25.6\%), and also surpassing both the Edu-only (32.8\%) and Synth-only (32.6\%) mixtures. Similar patterns emerge on HellaSwag, where the Edu+Synth mixture (46.0\%) substantially outperforms the Non-Edu model (38.3\%). On Global PIQA, the Edu-only mixture achieves the best performance (82.0\%). On Lambada, the Edu+Synth mixture achieves the highest score (39.0\%), followed closely by the Edu-only mixture (37.7\%). Detailed per-benchmark plots and a comprehensive comparison table including other relevant models are provided in Appendix~\ref{appendix:ablations}.

Even with a much smaller compute budget ($\sim$46B tokens) compared to other models (e.g., Curi{\'o} seen 1 trillion tokens of pretraining and 150 billion tokens of continual pretraining, Llama-3.2-1B seen 9 trillion tokens of pretraining), the models trained on \gigaverbo and \gigaverbosynth achieve competitive performance across all evaluated benchmarks (see Appendix~\ref{appendix:ablations}). In total, this ablation study involved training 4 separate 1.5 billion-parameter models for a total of 184 billion tokens, which translates to approximately 1.65$\times$10$^{21}$ FLOPs of compute. We estimate that the total energy consumed during these experiments was approximately 1,600 kWh, resulting in a carbon footprint of approximately 607 kg CO$_2$e (see Section~\ref{sec:resource_consumption}).

\newpage
\section{Tokenization}
\label{sec:tokenization}

\begin{tcolorbox}[colback=gray!5!white, colframe=gray!75!black, title=Summary / TL;DR]
\small

We train a custom SentencePiece tokenizer on a 40-40-20 mixture of Portuguese, English, and code data (49,152 vocabulary size). An evaluation on 600,000 Portuguese words demonstrates that our tokenizer achieves the lowest subword fertility (1.51) and the highest compression efficiency (2.88 characters per token) among all models compared, including Qwen3, Llama-3.2, and SmolLM3. Our tokenizer yields approximately 30\% compute savings relative to Qwen3's tokenizer in our pretraining setting, with efficiency gains compounding across pretraining, fine-tuning, and inference. Detailed evaluations of mixed-language data and analyses of computational costs are provided in Appendix~\ref{appendix:tokenization}.
\end{tcolorbox}

As previously noted in prior work \citep{finardi2021berta, larcher2023cabrita, correa2024teenytinyllama}, the efficiency of a tokenization scheme in compressing a given language directly affects the training cost of the corresponding language model. A tokenizer specifically tailored to a target domain can significantly reduce the number of tokens required to encode text \citep{larcher2023cabrita, correa2024teenytinyllama}, thereby reducing training time, inference latency, and computational costs.

Building on our previous experience with Portuguese-only tokenizers, we extended our focus to include a multilingual mixture of Portuguese, English, and code. This decision was motivated by our training mixture, which incorporates high-quality English and code data alongside Portuguese content (see Section~\ref{sec:pretraining}). An efficient tokenizer that handles all three domains enables us to leverage these resources without introducing data outside our tokenizer's domain.

\subsection{Training Procedure}

We trained our tokenizer using the SentencePiece library \citep{kudo2018sentencepiece}, an unsupervised text tokenizer that enables purely end-to-end systems without language-specific pre- or post-processing. SentencePiece is implemented in C++ with multi-threaded training support, allowing efficient processing of large-scale corpora. We constructed a training corpus with a 40-40-20 mixture of Portuguese, English, and code:

\begin{itemize}
    \item \textbf{Portuguese:} 2,000,000 text samples (sourced from \gigaverbo).
    \item \textbf{English:} 2,000,000 text samples (sourced from FineWeb Edu \citep{penedo2024finewebdatasetsdecantingweb}).
    \item \textbf{Code:} 975,000 text samples, covering 36 programming languages (sourced from the Starcoder dataset \citep{li2023starcodersourceyou}).
\end{itemize}

The resulting tokenizer has a vocabulary size of 49,152 tokens and employs the Byte-Pair Encoding (BPE) algorithm, with SentencePiece normalization and boundary handling.

\subsection{Evaluation Results}

To assess tokenizer quality, we employ two standard metrics \citep{rust2021goodtokenizermonolingualperformance}:

\begin{itemize}
    \item \textbf{Subword Fertility (SF):} The average number of tokens per word. Lower values indicate more efficient encoding, with a theoretical minimum of 1.0 (each word is a single token).
    \item \textbf{Proportion of Continued Words (PCW):} The ratio of words encoded with two or more tokens. Lower values indicate less aggressive word splitting.
\end{itemize}

We evaluated our tokenizer alongside several recent multilingual and Portuguese-adapted models on a corpus of 600,000 Portuguese words. Table~\ref{tab:tokenizer-eval-pt} presents a summary of key results. Our \tucano tokenizer achieves the lowest fertility (1.51) and highest compression efficiency (2.88 characters per token), outperforming all compared tokenizers. Detailed evaluation results on mixed (Portuguese, English, and code) data, as well as compute cost estimates based on fertility values, are provided in Appendix~\ref{appendix:tokenization}.

\begin{table}[h]
\centering
\small
\caption{Tokenizer evaluation on Portuguese text (600,000 words). Our tokenizer achieves the best compression efficiency across all metrics.}
\label{tab:tokenizer-eval-pt}
\begin{tabular}{lrrrrr}
\toprule
\textbf{Model} & \textbf{Tokens} & \textbf{Vocab Size} & \textbf{SF} & \textbf{PCW} & \textbf{Chars/Token} \\
\midrule
Tucano2-0.6B-Base   & 906,049   & 49,152  & 1.51 & 0.56 & 2.88 \\
GlorIA-1.3B         & 950,349   & 50,258  & 1.58 & 0.47 & 3.04 \\
Llama-3.2-1B        & 1,147,831 & 128,256 & 1.91 & 0.61 & 2.72 \\
SmolLM3-3B-Base     & 1,147,831 & 128,256 & 1.91 & 0.61 & 2.72 \\
OLMo-2-0425-1B      & 1,150,948 & 100,278 & 1.92 & 0.61 & 2.71 \\
Qwen3-0.6B          & 1,155,951 & 151,669 & 1.93 & 0.61 & 2.68 \\
Curi{\'o}-1.1b          & 1,304,472 & 32,000  & 2.17 & 0.55 & 3.09 \\
Granite-3.3-2b-base & 1,371,057 & 49,152  & 2.29 & 0.63 & 2.52 \\
\bottomrule
\end{tabular}
\end{table}

Reduced fertility directly translates into computational savings. Using the compute estimation heuristics proposed by \citet{ali2024tokenizer}, we estimate that training a $\sim$-billion-parameter model  (28 layers, hidden size 1,536, maximum sequence length 4,096, batch size 512) on 500 billion tokens with our tokenizer would require approximately $7.26 \times 10^{21}$ FLOPs, compared to $1.02 \times 10^{22}$ FLOPs for Qwen3's tokenizer---a reduction of approximately 30\%. These efficiency gains compound across pretraining, fine-tuning, and inference, making our tokenizer a cost-effective option for Portuguese language modeling.

\newpage
\section{Evaluation Framework}
\label{sec:building-evaluation-harness}

\begin{tcolorbox}[colback=gray!5!white, colframe=gray!75!black, title=Summary / TL;DR]
\small

Evaluations are essential for tracking model improvement during training, yet many existing Portuguese benchmarks exhibit high noise and volatility. We systematically analyze existing Portuguese evaluation suites using signal-quality metrics (mean absolute change, signal-to-noise ratio, and Spearman correlation) across checkpoints from SmolLM3-3b. Our analysis reveals that benchmarks implemented as generative tasks often require 1T+ tokens before showing meaningful improvement, making them impractical for short ablation studies or small pretraining runs. We construct a new Portuguese evaluation suite combining log-likelihood evaluations (Multiple-Choice format and Cloze formulation) with select generative tasks, yielding a two-tier suite: an \textit{Easy Set} providing reliable signals at $<$200B tokens and a \textit{Hard Set} for advanced evaluation at $>$1T tokens. We also adapt IFEval, GSM8K, RULER, and HumanEval for post-training assessment of instruction-following, mathematical reasoning, long-context, and coding capabilities. All benchmarks, implementations, and evaluation logs are released openly.
\end{tcolorbox}

\subsection{Motivation}

Evaluations guide LLM development by revealing model capabilities and learning progress~\citep{fourrier2025_the_llm_evaluation_guidebook}. However, not all benchmarks provide equally reliable signals, particularly during pretraining. In our previous work with Tucano~\citep{correa2025tucanoadvancingneuraltext}, we observed that certain Portuguese benchmarks exhibited minimal improvement as models were trained on more data, raising questions about their effectiveness for tracking pretraining progress.

To investigate this systematically, we analyze the widely used Portuguese evaluation suite developed by \cite{open-pt-llm-leaderboard}, which includes nine benchmarks spanning exam-based question answering (ENEM, BLUEX, OAB), natural language inference (ASSIN2 RTE, FAQUAD NLI), semantic similarity (ASSIN2 STS), and classification (HateBR, PT Hate Speech, TweetSentBR). All benchmarks are implemented as generative tasks that require the model to produce free-form text responses. While this setting is appropriate for evaluating instruction-tuned models, free-form generation requires substantial latent knowledge and is often too challenging for models during short pretraining runs~\citep{fourrier2025_the_llm_evaluation_guidebook}, especially in the early stages.

\subsection{Signal Analysis}
\label{subsec:signal-analysis}

To quantify benchmark reliability, we evaluate SmolLM3-3b~\citep{bakouch2025smollm3}, a 3-billion-parameter multilingual model trained on 11.5 trillion tokens---including Portuguese, albeit in a smaller and more diluted proportion---across multiple checkpoints from its training run. Following \citet{penedo2025fineweb2pipelinescale}, we define three signal-quality metrics:

\begin{itemize}
    \item \textbf{Mean Absolute Change:} Average absolute difference between consecutive evaluation points. \textit{Measures step-to-step instability.}
    \item \textbf{Signal-to-Noise Ratio (SNR):} Ratio of mean score to standard deviation. \textit{Higher values indicate that the signal dominates the noise.}
    \item \textbf{Spearman Correlation:} Correlation between benchmark scores and training steps. \textit{Higher positive values indicate the benchmark reliably tracks improvement.}
\end{itemize}

We also adopt the Normalized Preferred Metric (NPM) from \citet{Pires_2023} to aggregate performance across benchmarks with different random baselines:

$$
\text{NPM} = \frac{1}{N} \sum_{i=1}^{N} 100 \times \frac{\text{Preferred Metric}_i - \text{Random Score}_i}{\text{Max Score}_i - \text{Random Score}_i}
$$

Our analysis reveals substantial challenges with the existing suite. For SmolLM3-3b, certain benchmarks require 755B--1T+ tokens before surpassing their baselines by 5\% (Table~\ref{tab:baseline-surpass-old}). Meanwhile, benchmarks like ASSIN2 RTE, FAQUAD NLI, HateBR, and TweetSentBR exhibit extremely high volatility. Only ENEM, BLUEX, and OAB exhibit reasonable signal-to-noise ratios ($>$ 4) and strong Spearman correlations ($>$ 0.8), but these still require substantial training budgets before yielding meaningful signals. Detailed per-benchmark plots and signal statistics are provided in Appendix~\ref{appendix:garcia-harness-analysis}.

\subsection{Portuguese Suite Construction}

Motivated by these findings, we construct an improved Portuguese evaluation suite by integrating tasks from the LM-Evaluation-Harness~\citep{eval-harness} that have demonstrated reliable pretraining signals in prior work~\citep{Corr_a_2025}. Unlike the generative-task-only design of \cite{open-pt-llm-leaderboard} suite, we employ \textit{log-likelihood evaluations} in two formats:

\begin{itemize}
    \item \textbf{Multiple-Choice Format (MCF):} Choices are explicitly presented in the prompt and prefixed with labels (e.g., A/B/C/D). Each candidate's answer is scored using its conditional log-likelihood given the prompt, and the option with the highest log-likelihood is selected.

    \item \textbf{Cloze Formulation (CF):} The model completes a passage by predicting the most likely continuation, without explicit choice labels. As in MCF, candidate completions are evaluated via their conditional log-likelihood under the model, and the continuation with the highest log-likelihood is chosen. CF tasks typically provide earlier learning signals than MCF tasks~\citep{fourrier2025_the_llm_evaluation_guidebook}.
\end{itemize}

Our new suite includes: ARC Challenge~\citep{clark2018thinksolvedquestionanswering}, ASSIN2 Entailment and Paraphrase (log-likelihood variants)~\citep{fonseca2016assin}, Belebele~\citep{bandarkar-etal-2024-belebele}, Calame~\citep{lopes2024gloriagenerativeopen}, Global PIQA~\citep{chang2025globalpiqaevaluatingphysical}, HellaSwag~\citep{zellers2019hellaswagmachinereallyfinish}, LAMBADA~\citep{lambada2016}, and MMLU~\citep{hendryckstest2021}. All tasks are evaluated at a 5-shot regime.

Evaluating SmolLM3-3b on this suite yields substantial improvements in signal quality. Most benchmarks surpass their baselines within 94B--660B tokens (Table~\ref{tab:baseline-surpass-new}), and the aggregated NPM score exhibits a clearer monotonic upward trend across checkpoints (Appendix~\ref{appendix:new-harness-analysis}). Crucially, benchmarks such as ARC Challenge, Global PIQA, HellaSwag, LAMBADA, and Calame provide meaningful signals at $<$200B tokens, enabling cost-effective ablation studies.

The signal-quality analysis confirms these improvements: most benchmarks exhibit Spearman correlations $>$0.5, and CF tasks (Calame, HellaSwag, Global PIQA) achieve signal-to-noise ratios $>$20, substantially higher than MCF tasks such as Belebele and MMLU (SNR $<$7). Detailed statistics are provided in Appendix~\ref{appendix:new-harness-analysis}.

\subsection{Unified Suite Design}

We combine the most reliable benchmarks from both groups into a unified two-tier evaluation suite:

\begin{itemize}
    \item \textbf{Easy Set:} ARC Challenge, Calame, Global PIQA, HellaSwag, LAMBADA. These benchmarks provide reliable signals at early training stages ($<$200B tokens) with high signal-to-noise ratios ($>$10) and strong Spearman correlations ($>$0.57).
    \item \textbf{Hard Set:} ENEM, BLUEX, OAB, Belebele, MMLU. These benchmarks require more training data ($>$ 660B tokens) to demonstrate improvement, but they also provide complementary evaluation of advanced capabilities. All exhibit strong Spearman correlations ($>$0.8) despite low signal-to-noise ratios ($<$10).
\end{itemize}

This two-tier design enables researchers to track model progress efficiently during early pretraining (Easy Set) while also assessing advanced capabilities as training progresses (Hard Set). Comprehensive evaluation results and comparisons are provided in Appendix~\ref{appendix:unified-harness}.

\subsection{Post-Training Evaluation}

For assessing instruction-tuned models, we adapt four additional benchmarks:

\paragraph{IFEval-PT.} We manually translate and adapt 300 samples from the IFEval dataset~\citep{zhou2023instructionfollowingevaluationlargelanguage} into Portuguese, preserving multi-instruction complexity. The benchmark assesses whether models adhere to explicit formatting and content instructions (e.g., ``write at least 6 paragraphs with a title in all caps''). We report both strict and loose accuracy at the prompt and instruction levels.

\paragraph{GSM8K-PT.} We translate the GSM8K mathematical reasoning benchmark~\citep{cobbe2021trainingverifierssolvemath} using Qwen2.5-32B-Instruct, followed by manual review and correction, yielding 1,295 Portuguese math problems. Evaluation uses exact-match scoring with both strict and flexible numeric extraction.

\paragraph{RULER-PT.} We adapt the RULER long-context benchmark~\citep{hsieh2024rulerwhatsrealcontext, kim2025rulermeasureallbenchmarking} to Portuguese, including needle-in-a-haystack (NIAH) retrieval tasks, variable tracking, and common word extraction across multiple context lengths (1K--128K tokens). We leverage Portuguese texts from the OneRuler repository~\citep{kim2025rulermeasureallbenchmarking} for NIAH tasks.

\paragraph{HumanEval.} We evaluate code generation using the HumanEval benchmark~\citep{chen2021codex}, following the original implementation provided in the LM-Evaluation-Harness. We simply adapt the instruction prompts to Portuguese while keeping the original problems, test cases, and evaluation protocol unchanged. All models are evaluated in a zero-shot setting, and performance is reported using pass@1.

All four benchmarks are implemented in the LM-Evaluation-Harness and are released openly. Implementation details and full task descriptions are provided in Appendix~\ref{appendix:post-training-eval}.

\newpage
\section{Training Infrastructure}
\label{sec:infrastructure-software-stack}

\begin{tcolorbox}[colback=gray!5!white, colframe=gray!75!black, title=Summary / TL;DR]
\small

All training was conducted on Marvin, a state-of-the-art Tier-3 HPC cluster at the University of Bonn, featuring NVIDIA A100 and A40 GPUs and an InfiniBand NDR at 200 Gb/s. We leverage a custom open-source stack built on PyTorch (for distributed training) and Hugging Face Transformers (for architecture specification), employing DDP and FSDP for efficient large-scale training. Key optimizations include BF16 mixed precision, FlashAttention2, activation recomputation, and fused Triton kernels (Liger). Scaling experiments on a 7B-parameter reference model demonstrate near-linear speedup from 4 to 64 nodes (16--256 A100 GPUs), achieving 774K tokens/second and 144 TFLOPS/GPU at 64 nodes. Full infrastructure details, scaling analyses, and training configurations are provided in Appendix~\ref{appendix:infrastructure-software-stack}.
\end{tcolorbox}

\subsection{Hardware Infrastructure}

Our training infrastructure is Marvin, a Tier 3 HPC cluster at the University of Bonn. The system comprises 32 A100 GPU nodes (4$\times$80GB A100 per node, 128 GPUs total) optimized for highly scalable distributed training, and 24 A40 GPU nodes (8$\times$48GB A40 per node, 192 GPUs total) for machine learning workloads. All nodes are interconnected via Mellanox InfiniBand NDR at 200Gb/s, enabling efficient collective communication for multi-node training. Data processing leveraged CPU nodes (Intel Xeon Sapphire Rapids, 48--96 cores per node) with up to 4TB RAM per node, while the 5.6PB Lustre file system provided high-throughput storage for datasets and checkpoints. Detailed hardware specifications, node configurations, and task-level resource allocations are provided in Appendix~\ref{appendix:infrastructure-marvin-cluster}.

\subsection{Software Framework}

Our software stack is built on PyTorch~\citep{paszke2019pytorch} and Hugging Face Transformers~\citep{wolf-etal-2020-transformers}, ensuring compatibility with the broader open-source ecosystem. We employ two parallelism strategies depending on model size, Distributed Data Parallelism (DDP) \cite{li2020pytorchdistributedexperiencesaccelerating} and Fully Sharded Data Parallelism (FSDP)~\citep{zhao2023pytorchfsdpexperiencesscaling}. For all models below the 3-billion-parameter threshold, we use DDP; for our 3.7-billion-parameter model, we employ FSDP.

We integrate several performance optimizations into our stack:
\begin{itemize}
    \item \textbf{Mixed-precision training:} BF16 reduces memory footprint and accelerates training without compromising stability. TF32 tensor cores further enhance GEMM throughput on NVIDIA Ampere GPUs~\citep{valero2023mixed}.
    \item \textbf{FlashAttention2:} This memory-efficient attention implementation~\citep{dao2023flashattention2fasterattentionbetter} reduces memory complexity from $O(N^2)$ to $O(N)$ for sequence length $N$, achieving up to 73\% of theoretical peak FLOPS on A100 GPUs.
    \item \textbf{Fused Triton kernels:} We use the Liger Kernel library~\citep{hsu2025ligerkernel} for fused operations (RMSNorm, RoPE, SwiGLU, cross-entropy), yielding 20\% higher throughput and 60\% memory savings compared to standard PyTorch implementations.
    \item \textbf{Activation recomputation:} Re-materialization~\citep{chen2016trainingdeepnetssublinear} trades computation for memory, enabling training with longer sequences, larger batch sizes, and bigger models.
\end{itemize}

Comprehensive details on our codebase, library integrations, distributed training strategies (DDP and FSDP), and performance-optimization implementations are provided in Appendix~\ref{appendix:infrastructure-software-stack-foundry}.

\subsection{Scaling Analysis}

Scaling experiments on a 7B-parameter reference model demonstrate efficient multi-node scaling with our stack. Fixing the micro-batch size to 2 samples per GPU and using gradient accumulation to maintain a global batch size of 1024 samples ($\sim$4M tokens), we observe a near-linear speedup from 4 nodes (16 GPUs) to 64 nodes (256 GPUs). At 64 nodes, the system processes 774K tokens/second, achieving 144 TFLOPS/GPU---approximately 46\% of A100's theoretical peak for mixed-precision training. Detailed experimental setup, scaling curves, throughput analysis, scaling efficiency metrics, and per-node configurations are provided in Appendix~\ref{appendix:infrastructure-scaling-analysis}. Training system configuration details, including software versions and FSDP/NCCL parameters, are also documented in Appendix~\ref{appendix:infrastructure-scaling-analysis}.

\newpage
\section{Pretraining}
\label{sec:pretraining}

\begin{tcolorbox}[colback=gray!5!white, colframe=gray!75!black, title=Summary / TL;DR]
\small

We train \tucanoptsmall, a 670M-parameter model using a Llama-based architecture optimized for efficient training. Employing a hybrid AdamW+Muon optimizer and a three-stage warmup-stable-decay schedule, we train on approximately 408 billion tokens combining educational Portuguese, English, and synthetic data. The multi-stage curriculum progressively refines data mixtures, up-sampling higher-quality educational and reasoning-focused subsets in later stages while maintaining Portuguese as the majority language. Despite a substantially smaller compute budget than comparable models, \tucanoptsmall achieves strong performance on early-signal benchmarks (Easy Set NPM: 40.28) while consuming 92\% less energy than the original Tucano-2b4. Complete training configurations, stage-by-stage data mixtures, and ablation studies are provided in Appendix~\ref{appendix:pretraining}.
\end{tcolorbox}

\subsection{Model Architecture}

For the pretrained model in the \tucano series, we used the 0.6B-parameter architecture (\tucanoptsmall) as our primary experimental platform. This scale enables rapid experimentation while remaining comparable to baseline models such as Qwen2.5-0.5B and Qwen3-0.6B-Base. Following our prior work with the original Tucano series~\citep{correa2025tucanoadvancingneuraltext}, we maintain a Llama-based architecture~\citep{touvron2023llama} incorporating standard components: RMSnorm for normalization~\citep{zhang2019rootmeansquarelayer}, RoPE positional embeddings~\citep{su2023roformerenhancedtransformerrotary}, and SwiGLU activations~\citep{shazeer2020gluvariantsimprovetransformer}.

Three principles guided our design choices: \textit{GPU efficiency}, \textit{model expressiveness}, and \textit{saturation resistance}. All key dimensions (hidden size, intermediate size, vocabulary size) are aligned with GPU-friendly dimensions, thereby optimizing tensor operations and minimizing GPU under-utilization. We adopt a ``deep and slim'' configuration, consistent with findings that increasing depth (while keeping width moderate) enhances generalization in small-to-mid-sized transformers~\citep{liu2024mobilellmoptimizingsubbillionparameter, allal2025smollm2smolgoesbig, bakouch2025smollm3, warner2024smarterbetterfasterlonger}. Finally, evidence from Pythia~\citep{biderman2023pythia} and Llama~\citep{touvron2023llama} suggests that models with more than 400M parameters continue to benefit from prolonged training without early saturation, thereby justifying our moderate parameter count and extended training budget. Complete architectural specifications are provided in Appendix~\ref{appendix:pretraining-architecture}.

\subsection{Optimization Strategy}

Recent work on the Muon optimizer (\textit{MomentUm Orthogonalized by Newton-Schulz})~\citep{jordan2024muon} has shown promising results for LLM training, with anecdotal evidence from large-scale deployments~\citep{nanochat, kimiteam2026kimik2openagentic} and emerging empirical validation~\citep{liu2025muonscalablellmtraining, chang2025convergencemuon}. Muon is designed as a drop-in replacement for Adam that can be applied selectively to different parameter groups, making it attractive for experimentation without full commitment to a new optimization strategy.

To assess Muon's suitability for our setup, we conducted a small-scale pilot study comparing AdamW and AdamW+Muon on approximately 25 billion tokens. Both runs used a warmup-stable-decay (WSD) learning rate schedule~\citep{hagele2024scalinglawscomputeoptimaltraining} with 2,000 warmup steps, 9,000 stable steps, and approximately 1,000 cooldown steps (1-sqrt decay), matching the learning schedule of our planned pretraining recipe. We measured two gradient stability metrics: \textit{variability} (coefficient of variation, $\sigma/\mu$) and \textit{volatility} (standard deviation of consecutive changes in the gradient norm). The AdamW+Muon configuration exhibited substantially lower volatility (0.077 vs. 0.191) and reduced variability (1.61 vs. 1.79), suggesting smoother optimization trajectories. Based on these preliminary results, we adopted AdamW+Muon for our full training recipe, applying Muon to attention and feed-forward weights while using AdamW for embeddings and normalization layers. Detailed ablation results and gradient norm plots are provided in Appendix~\ref{appendix:pretraining-optimizer-ablation}.

\subsection{Training Configuration}

We derive optimal hyperparameters using the empirical scaling laws from DeepSeek LLM~\citep{deepseekai2024deepseekllmscalingopensource}, which relate compute budget ($C$) to batch size and learning rate via predictable power-law relationships. Using the adjusted DeepSeek formulation, we estimate our compute budget and apply the scaling heuristics to obtain a batch size of 2,097,152 tokens ($2^{21}$) and a maximum learning rate of $7 \times 10^{-4}$ for AdamW (and $7 \times 10^{-3}$ for Muon, following the 10$\times$ scaling recommended for this optimizer). Following recent best practices~\citep{olmo20252olmo2furious, bakouch2025smollm3}, we exclude embeddings from weight decay to enhance training stability. We perform checkpointing every 5,000 optimization steps, enabling regular performance evaluation and flexible resumption. Complete hyperparameter settings and scaling law derivations are provided in Appendix~\ref{appendix:pretraining-hyperparameters}.

\subsection{Training Recipe}

Unlike the single-stage warmup-decay recipe used in our original Tucano experiments, we designed a three-stage training curriculum inspired by OLMo2~\citep{olmo20252olmo2furious} and SmolLM~\citep{allal2025smollm2smolgoesbig, bakouch2025smollm3}. Each stage employs a specific phase of the WSD learning rate schedule~\citep{hagele2024scalinglawscomputeoptimaltraining}, progressively refining the data mixture to upsample higher-quality subsets. Throughout all stages, we maintain a Portuguese-dominant mixture (63\% overall) and supplement with high-quality English educational content (FineWeb-Edu~\citep{penedo2024finewebdatasetsdecantingweb}, FineMath), synthetic data (Cosmopedia-v2~\citep{allal2024cosmopedia}, \gigaverbosynth), and reasoning-focused datasets (OpenScience, Big Reasoning Traces, Math Meta Reasoning Filtered). The total training volume is approximately 408 billion tokens, distributed across 195,000 optimization steps.

\paragraph{Stage 1 (Warmup+Stable):} 100,000 steps, $\sim$292B tokens. The learning rate increases linearly over the first 2,000 steps to a peak, then remains constant for the remaining 98,000 steps. The data mixture comprises 61\% Portuguese educational content (\gigaverbo, education score $\geq$ 3, repeated 2$\times$) and 39\% English content (30\% FineWeb-Edu, 9\% FineMath).

\paragraph{Stage 2 (Stable):} 60,000 steps, $\sim$140B tokens. The learning rate remains stable at the peak value. The mixture becomes more diverse: 54\% Portuguese web text (\gigaverbo, education score $\geq$ 4, repeated 2$\times$), 14\% Portuguese synthetic data (\gigaverbosynth, repeated 2$\times$), 21\% English synthetic data (Cosmopedia-v2), 10\% English educational content (FineWeb-Edu), 8\% reasoning traces (OpenScience, Big Reasoning Traces, Math Meta Reasoning), and 5\% mathematical content (FineMath).

\paragraph{Stage 3 (1-sqrt Decay):} 35,000 steps, $\sim$74B tokens. The learning rate is annealed from the peak value using a 1-sqrt decay schedule. We further increase Portuguese content to 75\%: 36\% web text (\gigaverbo, education score $\geq$ 4) and 39\% synthetic data (\gigaverbosynth, repeated 3$\times$), complemented by 15\% English reasoning traces and 10\% English mathematical content. This final stage emphasizes the use of high-quality Portuguese data and reasoning capabilities.

The multi-stage design enables flexible checkpoint resumption and supports our ablation studies. Data mixture proportions and repetition factors were iteratively refined during early training by monitoring the aggregate NPM score on the Easy Set evaluations (Section~\ref{sec:building-evaluation-harness}). Complete per-stage data tables, token distributions, and mixing strategies are provided in Appendix~\ref{appendix:pretraining-data-mixtures}.

\subsection{Results}

\begin{figure}[h]
\centering
\includegraphics[width=0.95\linewidth]{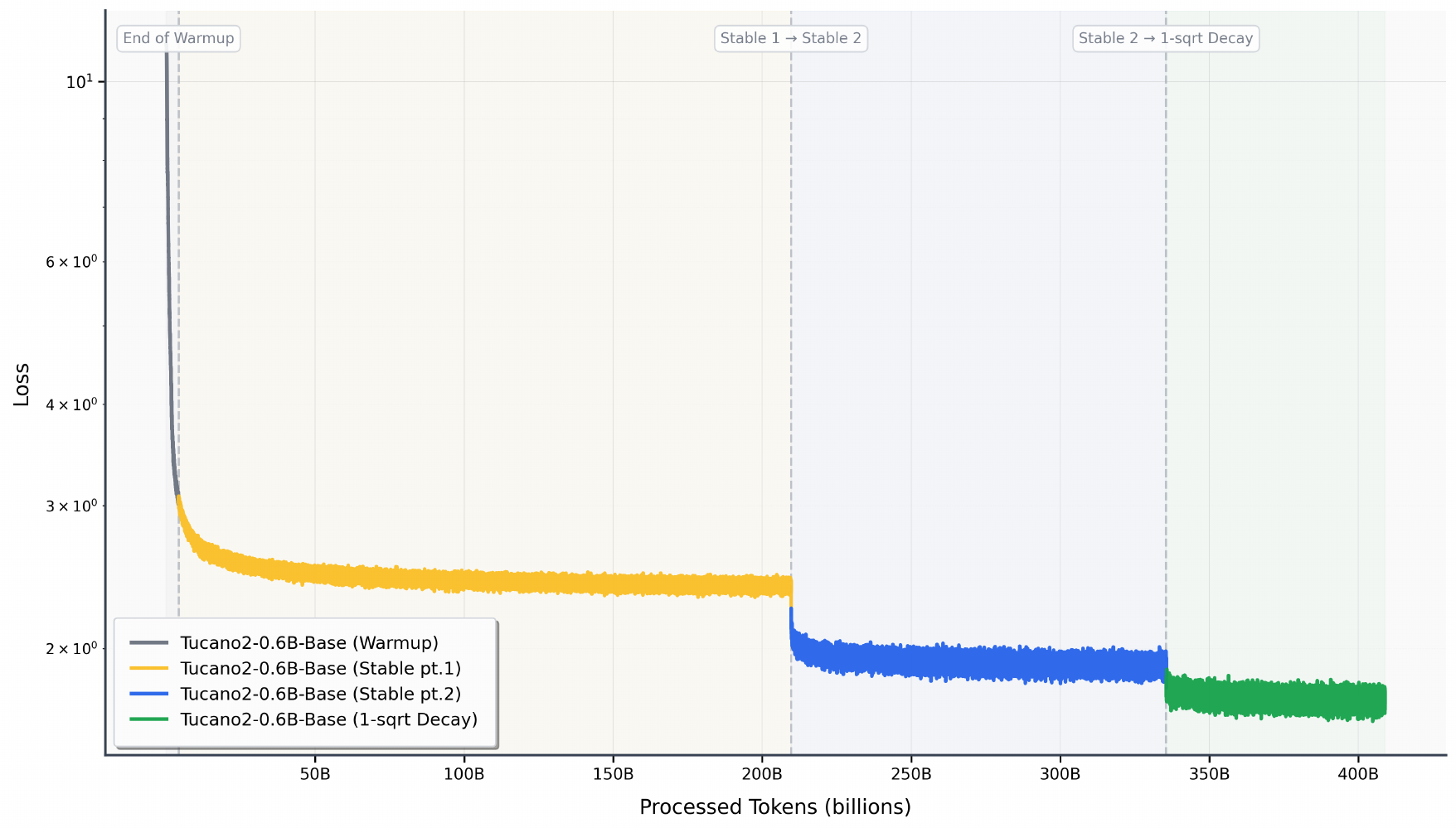}
\caption{Pretraining loss curve across 195,000 steps ($\sim$408B tokens).}
\label{fig:pretraining-loss-curve}
\end{figure}

\paragraph{Learning Curves.} The training run proceeded smoothly across all three stages, with no signs of gradient explosion, vanishing gradients, or overfitting. Figure~\ref{fig:pretraining-loss-curve} shows the pretraining loss curve, exhibiting the expected discontinuous jumps at stage transitions due to changes in data mixtures and associated differences in average cross-entropy. Similar behaviors have been observed in other multi-stage training setups~\citep{zhang2024tinyllamaopensourcesmalllanguage, apertus2025apertusdemocratizingopencompliant}. Gradient norm analysis (Appendix~\ref{appendix:pretraining-gradient-norms}) confirms stable optimization dynamics throughout the entire training run.

\begin{figure}[h]
\centering
\includegraphics[width=0.95\linewidth]{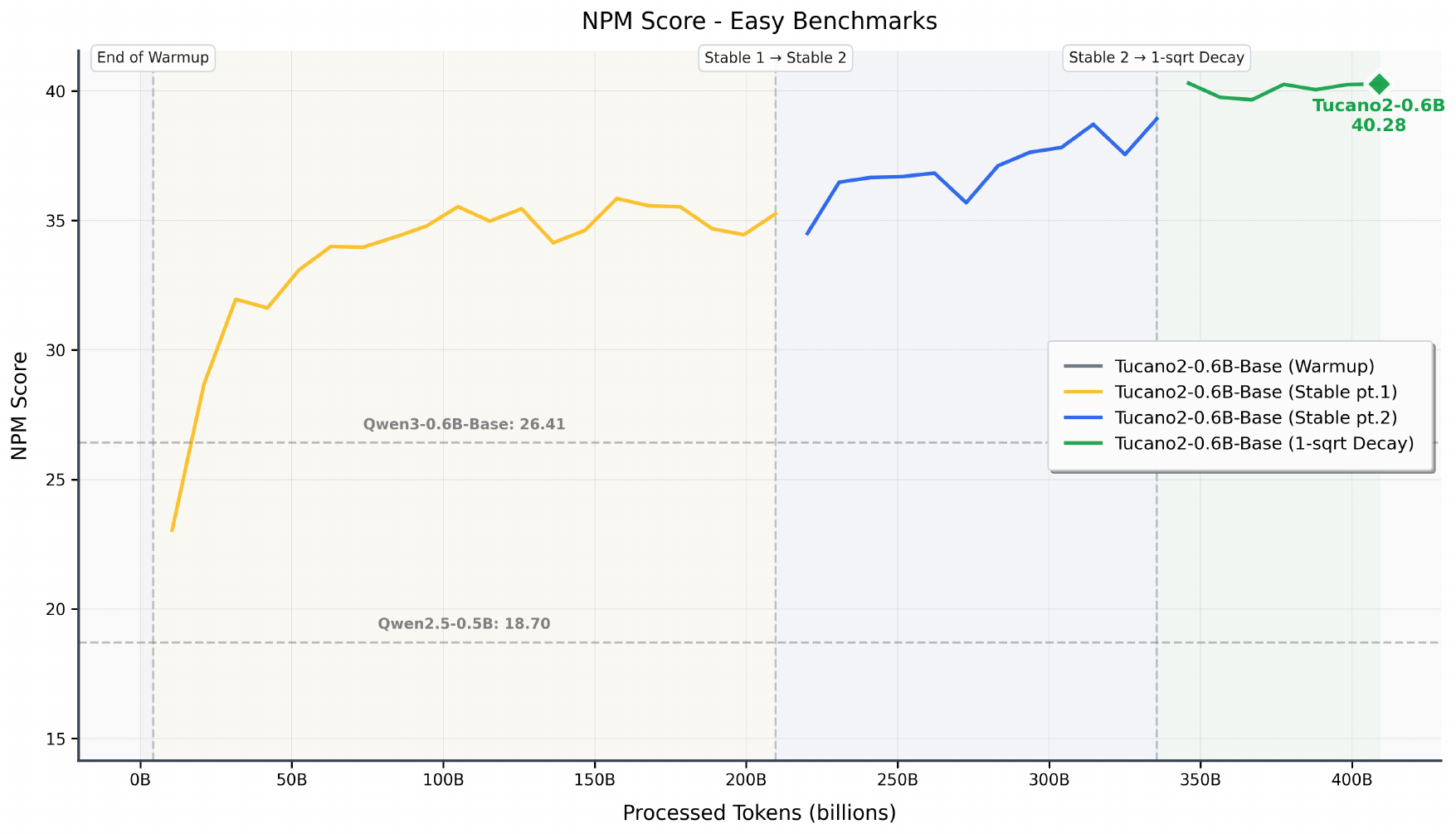}
\caption{Normalized Preferred Metric (NPM) scores on the Easy Set evaluations.}
\label{fig:pretraining-easy-npm}
\end{figure}

\paragraph{Benchmark Performance.} We evaluated \tucanoptsmall at regular intervals ($\sim$10B token checkpoints) using our two-tier evaluation suite (Section~\ref{sec:building-evaluation-harness}). On the Easy Set---which provides reliable signals at early training stages---the model exhibits steady improvement throughout training (Figure~\ref{fig:pretraining-easy-npm}). The final checkpoint achieves an aggregate NPM score of 40.28, substantially outperforming Qwen3-0.6B-Base (26.41) and Qwen2.5-0.5B (18.70) despite using a much smaller Portuguese-specific training corpus. Notably, \tucanoptsmall surpasses other Portuguese models of comparable or larger size, including Curi{\'o}-1.1B (39.16), Tucano-2b4 (33.55), and Curi{\'o}-Edu-1B (34.77).

\begin{figure}[h]
\centering
\includegraphics[width=0.95\linewidth]{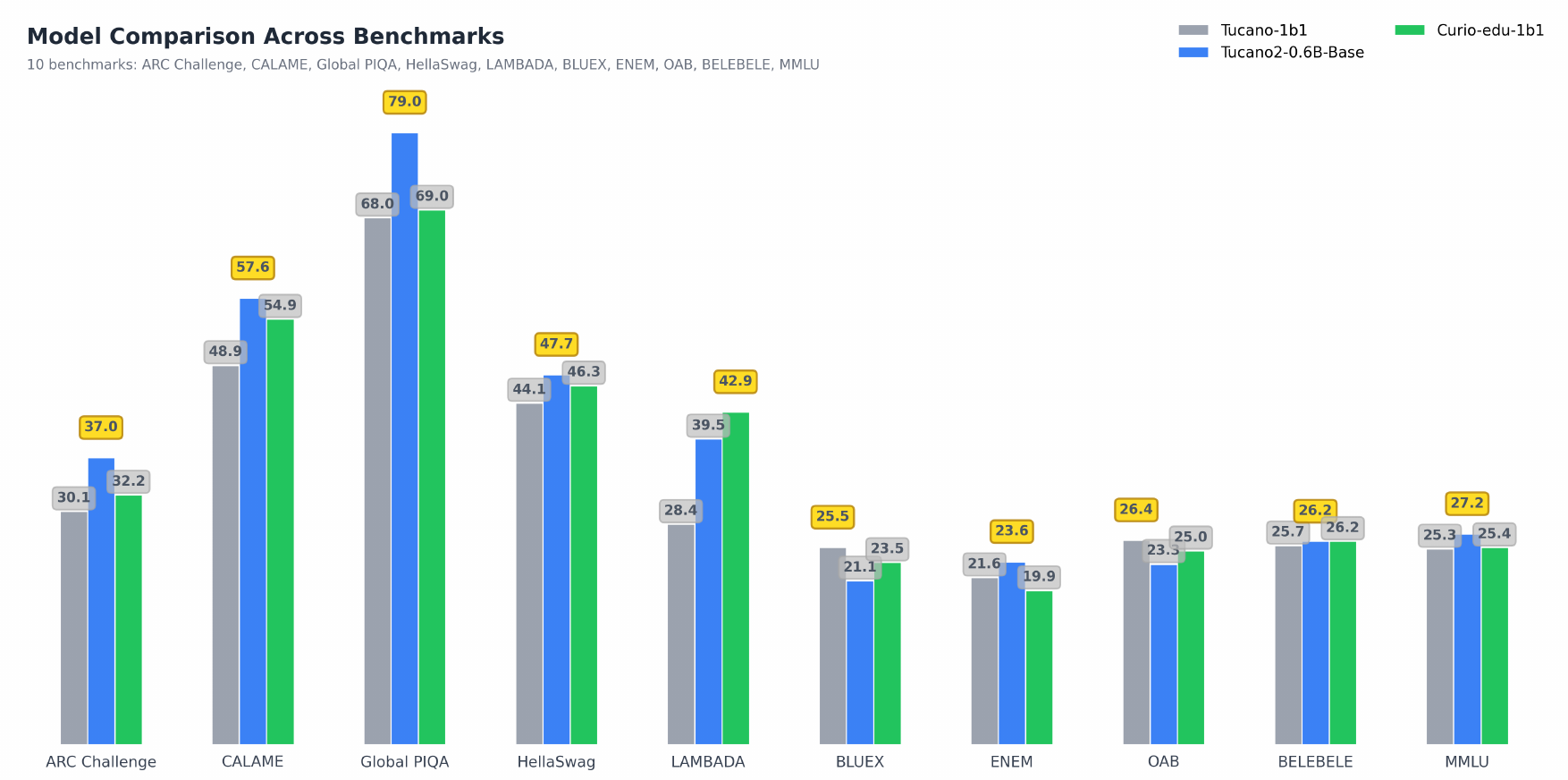}
\caption{Comparison of our 0.6-Base model against Tucano-1b1 and Curi{\'o}-edu-1b1.}
\label{fig:pretraining-model-comparison}
\end{figure}

As expected, performance on the Hard Set---which requires $>$660B tokens to show meaningful improvement (Section~\ref{sec:building-evaluation-harness})---remains close to random baselines. Nevertheless, when combining both Easy and Hard Set scores, \tucanoptsmall achieves a total average NPM of 20.64, outperforming Qwen2.5-0.5B (19.89), Curi{\'o}-1.1B (19.23), Tucano-2b4 (17.88), and other Portuguese baselines (see Table~\ref{tab:pretraining-results-comparison} and Figure~\ref{fig:pretraining-model-comparison}).

\begin{table}[h]
\centering
\small
\caption{Aggregate benchmark (NPM) results.}
\label{tab:pretraining-results-comparison}
\begin{tabular}{lccc}
\toprule
\textbf{Model} & \textbf{Total Avg.} & \textbf{Easy Set} & \textbf{Hard Set} \\
\midrule
Qwen3-0.6B-Base       & \textbf{29.40} & 26.41            & \textbf{32.38}           \\
\textbf{\tucanoptsmall} & 20.64          & \textbf{40.28}   & 0.99                   \\
Qwen2.5-0.5B          & 19.89          & 18.70            & 21.09                    \\
Curi{\'o}-1.1B            & 19.23          & 39.16            & $-0.69$              \\
Tucano-2b4            & 17.88          & 33.55            & 2.20                     \\
Curi{\'o}-Edu-1B          & 17.72          & 34.77            & 0.67                 \\
Carvalho-pt-gl-1.3B   & 12.54          & 26.75            & $-1.66$                  \\
GlorIA-1.3B           & 5.93           & 27.27            & $-15.42$                 \\
\bottomrule
\end{tabular}
\end{table}

\paragraph{Efficiency and Environmental Impact.} Training \tucanoptsmall required approximately 1.47$\times$10$^{22}$ FLOPs across 195,000 steps, consuming 872.61 kWh of energy and resulting in an estimated carbon footprint of 332.42 kg CO$_2$e (see Section~\ref{sec:resource_consumption}). Compared to the original Tucano-2b4 model, which required 11,749 kWh and produced 4,475 kg CO$_2$e, this represents a 92\% reduction in both energy consumption and carbon emissions while achieving superior aggregate benchmark performance. Detailed per-benchmark learning curves are provided in Appendix~\ref{appendix:pretraining-detailed-results}.

\subsection{Toward Continual Pretraining}

While our pretraining recipe has yielded substantial improvements---particularly on the Easy Set evaluations---it also highlights a clear limitation: certain capabilities of interest for Portuguese LLMs (e.g., advanced reasoning, exam-based question answering) do not emerge unless the model is trained on substantially larger volumes of high-quality data. As demonstrated in Section~\ref{sec:building-evaluation-harness}, benchmarks in the Hard Set require 660B--1T+ tokens before showing meaningful signals (for a 3B parameter model). This volume is difficult to reach for low-resource languages when computational resources are constrained.

To bridge this capability gap, continual pretraining---adapting larger multilingual base models to Portuguese using our curated datasets---represents a promising direction for future work. This approach has been extensively explored in the Portuguese NLP literature~\citep{Pires_2023, almeida2024sabia2newgenerationportuguese, almeida2025buildinghighqualitydatasetsportuguese}, though with varying degrees of success. While models like Sabiá and Curi{\'o} have reported significant performance gains, other efforts have shown only modest improvements or even performance degradation on certain tasks~\citep{cruzcastaneda2025amadeusverbotechnicalreportpowerful, gaia-gemma-3-4b-2025}. These mixed results suggest that continual pretraining is not guaranteed to succeed when applied naively, and that careful design of data mixtures, learning rate schedules, and training duration is critical for effective adaptation.

\newpage
\section{Continual Pretraining}
\label{sec:continual_pretraining}

\begin{tcolorbox}[colback=gray!5!white, colframe=gray!75!black, title=Summary / TL;DR]
\small

While \tucanoptsmall demonstrates strong Easy Set performance, Hard Set capabilities require substantially larger data volumes than feasible for from-scratch pretraining under constrained budgets. We adopt a continual pretraining strategy that adapts the Qwen3 base models to the Portuguese language. Using training-free OMP-based tokenizer transplantation, we replace the 151K Qwen3 vocabulary with our own 49K-token Portuguese-optimized tokenizer, reducing the embedding parameter count by $\sim$68\% while preserving pretrained capabilities. We then continually pretrain three model variants (0.5B, 1.5B, 3.7B) on purely Portuguese data, with token counts ranging from 50 to 100B, using AdamW with cosine decay schedules. The resulting models achieve state-of-the-art Portuguese performance for their size range: \tucanocptlarge attains an aggregate NPM of 59.21, surpassing Qwen3-4B-Base (57.86) and Qwen2.5-7B (57.97), while using negligible additional compute ($<$0.13\% of total pretraining FLOPs). Complete hyperparameter configurations, data mixtures, and per-benchmark results are provided in Appendix~\ref{appendix:continual_pretraining}.
\end{tcolorbox}

\subsection{Motivation}

While our multi-stage training recipe for \tucanoptsmall yields strong Easy Set performance (Section~\ref{sec:pretraining}), Hard Set capabilities---such as advanced reasoning and exam-based question answering---require training volumes exceeding 660B--1T tokens (Section~\ref{sec:building-evaluation-harness}), a scale difficult to achieve from scratch for low-resource languages under constrained compute budgets. Continual pretraining offers a pragmatic alternative: by adapting strong multilingual base models to the target domain, one can leverage the rich representations already learned during large-scale pretraining while focusing adaptation effort on Portuguese~\citep{ke2023continualpretraininglanguagemodels}.

We select the Qwen3 series (Base versions) as our starting point for three reasons. First, Qwen3 models consistently outperform other similarly sized baselines on our evaluation suite, particularly on Hard Set benchmarks (Section~\ref{sec:pretraining}). Second, all Qwen3 models within our target size range are released under the Apache-2.0 license, enabling unrestricted modification and redistribution. Third, the Qwen3 architecture covers a broad parameter range, enabling a family of Portuguese-adapted models at multiple scales.

However, the Qwen3 series employs a vocabulary of 151K tokens---approximately three times that of our Portuguese-optimized tokenizer (Section~\ref{sec:tokenization}). This large, multilingual vocabulary introduces significant inefficiencies for Portuguese-focused tasks: many tokens are irrelevant to Portuguese, inflating memory consumption during training and reducing inference throughput. We address this challenge by transplanting the tokenizer, replacing the Qwen3 vocabulary with our more efficient tokenizer while preserving the pretrained weights.

\subsection{Tokenizer Transplantation (OMP)}

We adopt the Orthogonal Matching Pursuit (OMP) approach to training-free tokenizer transplantation~\citep{goddard2025trainingfreetokenizertransplantationorthogonal}. OMP is a greedy sparse approximation algorithm that reconstructs each donor-model embedding as a $k$-sparse linear combination of shared-token embeddings, then transfers the same coefficients to the target model's embedding space. Because pretrained LLM embedding spaces are approximately orthogonally aligned on shared tokens, these coefficients transfer meaningfully, preserving geometric and semantic structure without requiring gradient updates.

Using the \texttt{mergekit-tokensurgeon} utility~\citep{goddard2025arceesmergekittoolkitmerging}, we transplant our 49K-token Portuguese-optimized tokenizer into three Qwen3 base models (0.6B, 1.7B, 4B). The vocabulary reduction from 151K to 49K removes approximately 68\% of embedding parameters, yielding the Tucano2-qwen models: \tucanocptsmall (from Qwen3-0.6B-Base), \tucanocptmedium (from Qwen3-1.7B-Base), and \tucanocptlarge (from Qwen3-4B-Base). Post-transplantation evaluation reveals that the adapted models retain substantial performance on Hard Set benchmarks while exhibiting expected degradation on language-modeling-sensitive Easy Set tasks. Detailed transplantation results are provided in Appendix~\ref{appendix:cpt-omp-results}.

\subsection{Training Configuration}

Since continual pretraining adapts existing pretrained weights, hyperparameter selection requires careful balancing between adaptation and knowledge preservation. Following heuristics from the literature~\citep{parmar2024reusedontretrainrecipe, wang2024learningratepathswitching}, we set peak learning rates at a fraction of the inferred original pretraining rates, with values determined through systematic sweeps of 10,000 steps ($\sim$10.4B tokens each). We adopt AdamW with a simple warmup-cosine-decay schedule, as small-scale experiments with the AdamW+Muon variant used for \tucanoptsmall yielded inferior performance---likely due to an optimizer-state mismatch with the Qwen3 pretrained weights.

All models use a total batch size of 1,048,576 tokens (reduced from the 2M used during from-scratch pretraining to accommodate budget constraints) and a maximum context length of 4,096 tokens. Peak learning rates decrease with model size: $2.5\times10^{-4}$ (0.5B), $1\times10^{-4}$ (1.5B), and $7.5\times10^{-5}$ (3.7B). Training budgets range from 50B tokens (0.5B and 3.7B) to 100B tokens (1.5B), representing $<$0.3\% of the original Qwen3 pretraining volume. Complete hyperparameter tables are provided in Appendix~\ref{appendix:cpt-hyperparameters}.

\subsection{Data Mixtures}

Unlike the multilingual mixtures used during from-scratch pretraining, continual pretraining employs purely Portuguese data. This deliberate shift maximizes Portuguese-specific adaptation within the constrained compute budget. Each mixture combines curated web text from \gigaverbo (education score $\geq$ 4) with synthetic data from \gigaverbosynth, maintaining a web-to-synthetic ratio of approximately 60:40 for the 0.5B and 3.7B models and 70:30 for the 1.5B model. Full data mixture compositions are provided in Appendix~\ref{appendix:cpt-data-mixtures}.

\subsection{Results}

Table~\ref{tab:cpt-results-comparison} presents aggregate results across our evaluation suite. The continual pretraining strategy produces consistent, substantial gains across all model scales.

\begin{table}[h]
\centering
\small
\caption{Aggregate benchmark results (NPM) comparisons against other baselines.}
\label{tab:cpt-results-comparison}
\begin{tabular}{lccc}
\toprule
\textbf{Model} & \textbf{Total Avg.} & \textbf{Easy Set} & \textbf{Hard Set} \\
\midrule
\textbf{\tucanocptlarge} & \textbf{59.21} & \textbf{57.41} & 61.00 \\
Qwen2.5-7B                       & 57.97           & 54.12 & 61.83 \\
Qwen3-4B-Base                    & 57.86           & 52.52 & \textbf{63.20} \\
SmolLM3-3B-Base                  & 50.25           & 54.06 & 46.44 \\
Qwen2.5-3B                      & 50.16           & 47.69 & 52.62 \\
\textbf{\tucanocptmedium}       & 47.90  & 47.97 & 47.82 \\
Curi\'{o}-Edu-7B                 & 45.66           & 57.46 & 33.87 \\
Qwen3-1.7B-Base                 & 44.48           & 40.94 & 48.03 \\
Curi\'{o}-7B                    & 42.79           & 58.97 & 26.60 \\
Llama-3.2-3B                    & 40.50           & 43.79 & 37.21 \\
Granite-3.3-2B                  & 39.97           & 45.31 & 34.63 \\
\textbf{\tucanocptsmall}        & 35.36  & 39.93 & 30.79 \\
Qwen3-0.6B-Base                 & 29.40           & 26.41 & 32.38 \\
\textbf{\tucanoptsmall}               & 20.64           & 40.28 & 0.99 \\
Qwen2.5-0.5B                    & 19.89           & 18.70 & 21.09 \\
\bottomrule
\end{tabular}
\end{table}

At the largest scale, \tucanocptlarge achieves the highest total average NPM (59.21), surpassing both Qwen3-4B-Base (57.86) and even Qwen2.5-7B (57.97), a model with nearly twice the parameters. At the 1.5B scale, \tucanocptmedium (NPM: 47.90) improves by +3.42 points over Qwen3-1.7B-Base (44.48) and substantially outperforms domain-adapted models of comparable or larger size, such as Curi{\'o}-7B (42.79) and Curi{\'o}-Edu-7B (45.66). At the smallest scale, \tucanocptsmall (NPM: 35.36) achieves a +5.96-point improvement over Qwen3-0.6B-Base (29.40) while dramatically outperforming \tucanoptsmall (20.64).

\begin{figure}[h]
\centering
\includegraphics[width=0.95\linewidth]{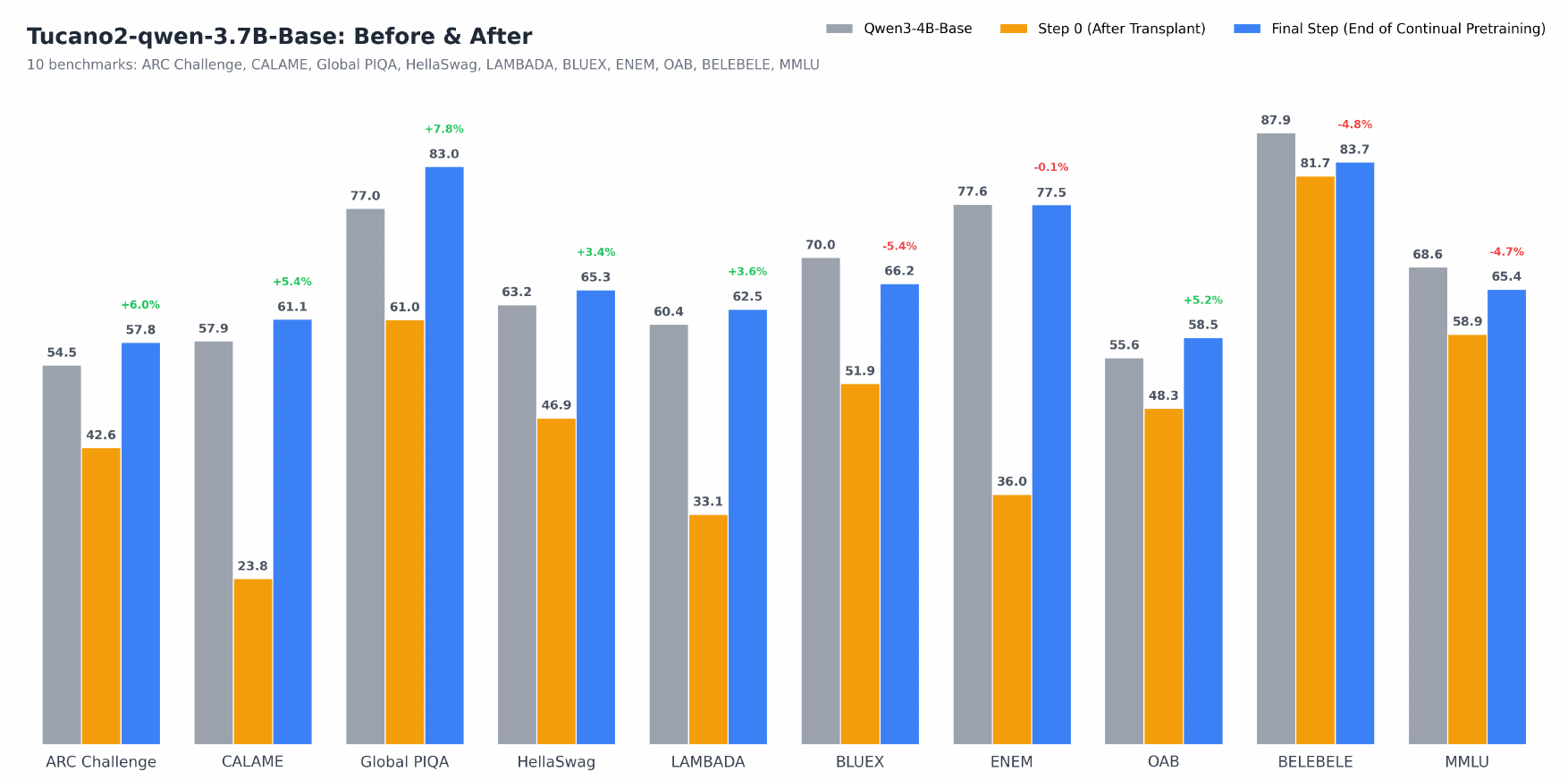}
\caption{Per-benchmark performance comparison. Bars indicate the absolute difference in evaluation scores between the continually pretrained model and its base.}
\label{fig:cpt-before-after-3.7B}
\end{figure}

Figure~\ref{fig:cpt-before-after-3.7B} illustrates the per-benchmark performance shift for the 3.7B model. Per-benchmark comparisons for all model sizes are provided in Appendix~\ref{appendix:cpt-detailed-results}.

\newpage
\subsection{Compute-Performance Tradeoff}

\begin{figure}[h]
\centering
\includegraphics[width=0.95\linewidth]{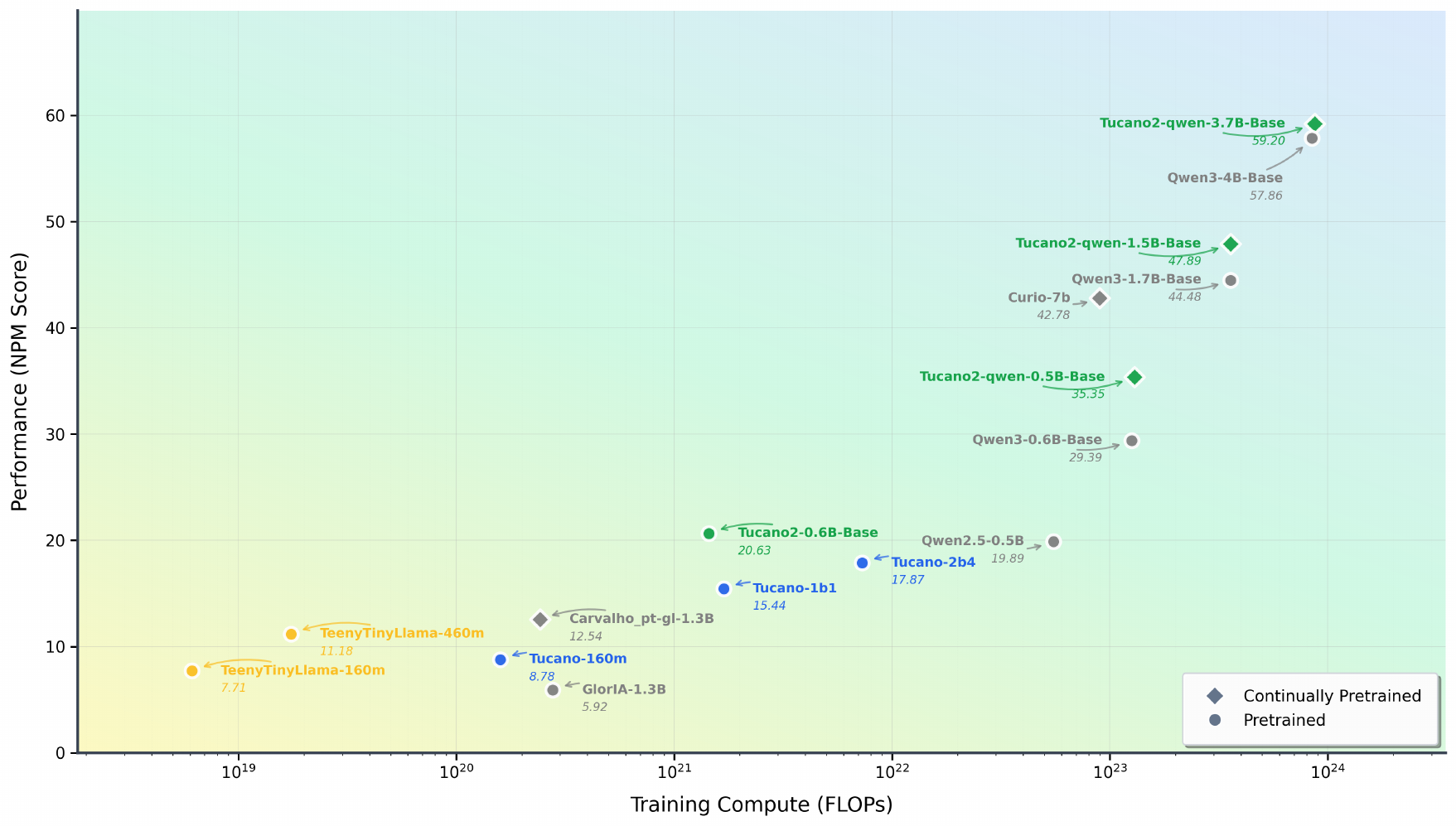}
\caption{Aggregate NPM score vs.\ total compute (FLOPs).}
\label{fig:cpt-performance-vs-compute}
\end{figure}

Figure~\ref{fig:cpt-performance-vs-compute} plots aggregate NPM against total compute ($C = 6ND$, where $N$ is the parameter count and $D$ is the total number of tokens processed, including both pretraining and continual pretraining). These results underscore the cost-effectiveness of continual pretraining as a strategy for adapting strong multilingual foundations to low-resource languages: rather than training from scratch, a modest additional investment on top of a well-pretrained base yields substantial gains. A detailed breakdown of compute for all models is provided in Appendix~\ref{appendix:cpt-performance-vs-compute}.

\newpage
\section{Post-Training}
\label{sec:post_training}

\begin{tcolorbox}[colback=gray!5!white, colframe=gray!75!black, title=Summary / TL;DR]
\small
We construct two Portuguese alignment datasets---\gigaverbosft (${\sim}$4M examples, ${\sim}$2.1B tokens spanning 12 task types) and \gigaverbopreferences (${\sim}$28K contrastive pairs)---using the Qwen2.5 model family for synthetic generation, with a dedicated quality-assessment classifier fine-tuned from Qwen3-4B (F1-macro 0.80) for filtering. Due to the high cost and quality challenges of generating Portuguese reasoning traces, we train separate \textit{Instruct} and \textit{Think} model families rather than a single hybrid model. Each base model undergoes Supervised Fine-Tuning (SFT) followed by Anchored Preference Optimization (APO). \tucanocptlargeinstruct achieves the highest Knowledge \& Reasoning score (NPM 56.22) among all models in the 3--4B range, surpassing Qwen3-4B, SmolLM3-3B, and Gemma-3-Gaia-PT-BR-4b-it. \tucanocptlargethink attains the top Knowledge \& Reasoning NPM (54.07) among reasoning models, while generating chain-of-thought traces entirely in Portuguese. All datasets, quality-assessment models, training configurations, and evaluation logs are released openly. Complete training hyperparameters, data mixture details, and per-benchmark breakdowns are provided in Appendix~\ref{appendix:post_training}.
\end{tcolorbox}

After developing our base models, we proceed to the post-training phase, which comprises two stages: \textit{Supervised Fine-Tuning} (SFT) and \textit{Preference Optimization} (PO). SFT enhances instruction-following capabilities by fine-tuning on a large, diverse set of instruction-response pairs in Portuguese. PO further refines alignment by explicitly capturing both quality-focused (helpfulness and reasoning) and safety-focused (refusal and risk mitigation) signals.

\subsection{GigaVerbo-v2 SFT}

\paragraph{Task taxonomy.}
We define a comprehensive taxonomy of 12 task types: code generation, function calling, general instruction following, mathematical problem solving, mathematical problem solving with chain-of-thought (CoT), reasoning (with explicit \texttt{<think>}$\ldots$\texttt{</think>} traces), retrieval-augmented generation (RAG), rewriting, structured output generation (JSON), summarization, system prompts, and translation. This taxonomy is designed to prepare models for a broad range of downstream applications.

\paragraph{Dataset construction.}
Prompts were sourced from multiple public instruction-tuning datasets in English and Portuguese, with all English content translated to Portuguese using Qwen2.5-32B-Instruct. Responses were generated synthetically using the Qwen2.5 model family, with model capacity matched to task complexity: Qwen2.5-7B-Instruct and Qwen2.5-14B-Instruct for simpler tasks (summarization, rewriting, translation) and Qwen2.5-32B-Instruct for complex tasks (reasoning, Math CoT). For reasoning-intensive tasks, we employed a two-stage pipeline: (1)~Qwen2.5-7B-Instruct generated initial direct responses, then (2)~Qwen2.5-32B-Instruct produced detailed reasoning traces appended in \texttt{<think>}$\ldots$\texttt{</think>} format. A key finding during this process is that generating high-quality reasoning traces in Portuguese remains challenging. Experiments with Qwen3 and SmolLM3 revealed that these models frequently default to English during reasoning, produce inconsistent reasoning that trails off without conclusions, or switch languages mid-generation. This motivated our use of Qwen2.5-32B-Instruct and highlights an important ecosystem gap that our Think models aim to address.

\paragraph{Quality filtering.}
Since \gigaverbosft is entirely LLM-generated, we implemented a dedicated quality filtering pipeline. We first annotated 500K randomly sampled interactions using an LLM-as-a-judge prompt (Qwen2.5-32B-Instruct scoring instruction adherence on a 1--5 scale) (see Section~\ref{appendix:llm-judge-prompts}), then fine-tuned Qwen3-4B-Base into a regression-based quality classifier, achieving F1-macro of 0.80 (F1 of 0.98 at the $\geq$3 acceptability threshold) (see Section~\ref{appendix:instruct-quality-annotators} for details on the training o these annotators). Samples scoring below 3.5 were removed, followed by benchmark decontamination and language filtering. The resulting dataset comprises ${\sim}$4.1M examples (${\sim}$2.15B tokens); full statistics are in Table~\ref{tab:gigaverbo-sft-summary}.

\begin{table}[h]
\centering
\small
\caption{Dataset composition after filtering.}
\label{tab:gigaverbo-sft-summary}
\begin{tabular}{lrr}
\toprule
\textbf{Subset} & \textbf{Samples} & \textbf{Tokens} \\
\midrule
Retrieval       & 1,977,667 & 1,013,172,488 \\
General         & 1,235,976 &   700,483,545 \\
Math            &   220,042 &    83,234,218 \\
Structured      &   163,542 &    70,632,221 \\
Summarization   &   128,669 &    90,310,108 \\
Code            &    80,774 &    84,389,567 \\
Reasoning       &    78,249 &    34,786,173 \\
Math CoT        &    63,413 &    27,107,305 \\
Function Call   &    45,891 &    28,712,972 \\
Translation     &    45,204 &     7,877,426 \\
Rewriting       &    29,150 &     3,674,384 \\
System Prompts  &    20,512 &     7,261,615 \\
\midrule
\textbf{Total}  & \textbf{4,089,089} & \textbf{2,151,642,022} \\
\bottomrule
\end{tabular}
\end{table}

Notably, reasoning samples constitute $<$2\% of the total, a direct consequence of the high cost and quality challenges described above. This imbalance motivated our decision to train separate Instruct and Think models rather than a single hybrid model (Section~\ref{sec:sft_po_training}).

\subsection{GigaVerbo-v2 Preferences}

To further align our models, we construct a preference dataset that spans both quality- and safety-focused alignment objectives. Prompts were sourced from UltraFeedback (diverse instruction-following) and HarmfulQA (adversarial/NSFW prompts), covering two core alignment challenges: (1)~benign requests requiring helpful, well-reasoned responses, and (2)~harmful requests requiring consistent refusal behavior.

Response generation followed a Constitutional AI approach with distinct constitutions for each subset. For the \textit{harmless} subset, chosen responses were generated by Qwen2.5-32B-Instruct with chain-of-thought reasoning, while rejected responses came from Qwen2.5-7B-Instruct. For the \textit{harmful} subset, chosen (refusal) responses used safety-oriented constitutions applied to Qwen2.5-32B-Instruct, while rejected (compliant) responses were generated by an abliterated variant of Qwen2.5-32B-Instruct~\citep{arditi2024refusallanguagemodelsmediated}. After decontamination and quality filtering, the final dataset contains ${\sim}$28K preference pairs (${\sim}$28M tokens), with a balanced split between reasoning and non-reasoning formats. URLs to the used constitutions are provided in Appendix~\ref{appendix:preferences_details} and~\ref{appendix:constitutional_ai}.

\subsection{Training Configuration}
\label{sec:sft_po_training}

Each base model (0.5B, 1.5B, 3.7B) was used to produce two chat variants:

\begin{itemize}
    \item \textbf{Instruct}: trained on the full multi-task \gigaverbosft mixture + the non-reasoning part of \gigaverbopreferences.
    \item \textbf{Think}: trained exclusively on reasoning-intensive tasks from both our post-training datasets.
\end{itemize}

Both variants follow a two-stage recipe: SFT with assistant-only loss masking (5 epochs, context length 4,096, AdamW with cosine schedule), followed by Anchored Preference Optimization~\citep{doosterlinck2024anchoredpreferenceoptimizationcontrastive} using the \texttt{apo\_zero} loss with $\beta = 0.5$ (5 epochs). Full hyperparameter tables are provided in Appendix~\ref{appendix:post_training_configs}.

\subsection{Results}

\subsubsection{Instruct Variants}

Table~\ref{tab:instruct-results} compares the Instruct variants against chat models of similar size across three evaluation dimensions: Knowledge \& Reasoning (ARC-Challenge, ENEM, BLUEX, OAB Exams, BELEBELE, MMLU, GSM8K-PT), Instruction Following (IFEval-PT), and Coding (HumanEval).

\begin{table}[h]
\centering
\small
\caption{Post-training results for Instruct variants.}
\label{tab:instruct-results}
\begin{tabular}{lcccc}
\toprule
 & \textbf{Total Avg.} & \textbf{K\&R (NPM)} & \textbf{Instruct} & \textbf{Coding} \\
\midrule
\textbf{\tucanocptlargeinstruct} & 53.64 & 56.22 & 41.67 & 47.56 \\
Jurema-7B                            & 53.03          & 50.66          & 47.00 & 75.61 \\
Qwen2.5-3B-Instruct                  & 51.71          & 47.34          & 63.33 & 70.73 \\
Qwen3-4B                             & 51.36          & 42.33          & 79.33 & 86.59 \\
Gemma-3-Gaia-PT-BR-4b-it             & 49.93          & 45.00          & 70.33 & 64.02 \\
SmolLM3-3B                           & 49.54          & 43.99          & 69.67 & 68.29 \\
Llama-3.2-3B-Instruct                & 45.82          & 43.08          & 62.67 & 48.17 \\
Qwen2.5-1.5B-Instruct                & 41.39          & 40.25          & 42.00 & 48.78 \\
\textbf{\tucanocptmediuminstruct} & 37.54 & 39.61          & 34.33 & 26.22 \\
Qwen3-1.7B                           & 36.30          & 28.24          & 65.00 & 64.02 \\
\textbf{\tucanocptsmallinstruct} & 26.08 & 27.77          & 30.00 & 10.37 \\
Qwen3-0.6B                           & 22.21          & 15.13          & 55.00 & 39.02 \\
Llama-3.2-1B-Instruct                & 20.14         & 15.37          & 44.33 & 29.27 \\
Qwen2.5-0.5B-Instruct                & 17.8          & 14.98          & 31  & 24.39 \\
\bottomrule
\end{tabular}
\end{table}

\tucanocptlargeinstruct achieves the highest Total Average (53.64) and the highest Knowledge \& Reasoning score (56.22) among all models in the 3--4B range, surpassing Qwen3-4B (42.33), SmolLM3-3B (43.99), and the Portuguese-specialized Gemma-3-Gaia-PT-BR-4b-it (45.00). These gains span multiple domain-specific benchmarks, like BLUEX, ENEM, OAB, ARC-Challenge, BELEBELE, and MMLU. Mathematical reasoning is a particular strength: on GSM8K-PT, \tucanocptlargeinstruct scores 53.81, outperforming both Qwen3-4B (39.88) and Gemma-3-Gaia-PT-BR-4b-it (51.29).

\begin{figure}[h]
\centering
\includegraphics[width=0.95\linewidth]{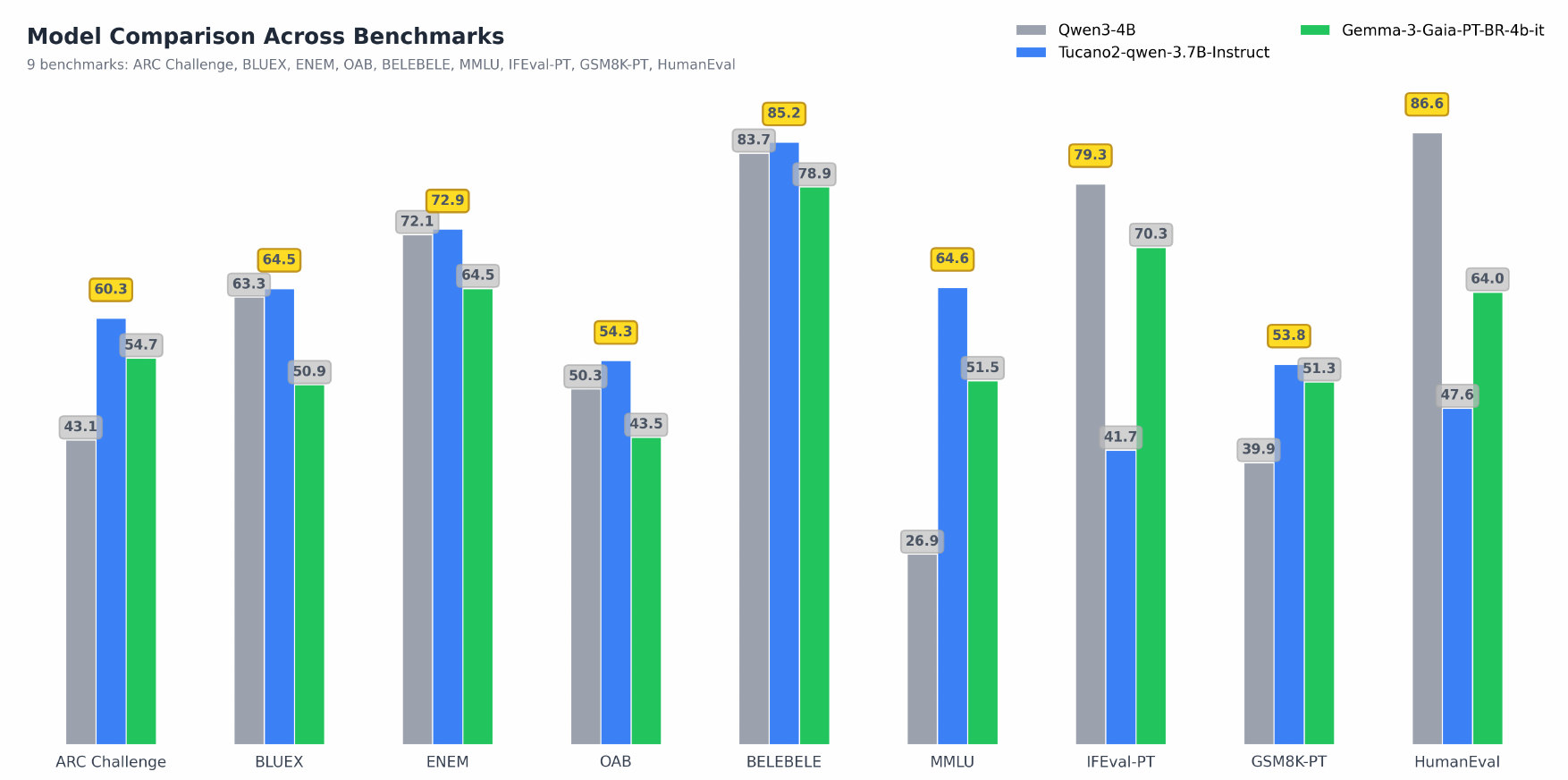}
\caption{Per-benchmark comparison of \tucanocptlargeinstruct against Qwen3-4B and Gemma-3-Gaia-PT-BR-4b-it.}
\label{fig:instruct-3.7B-comparison}
\end{figure}

However, Tucano2 models lag behind on Instruction Following (IFEval-PT) and Coding (HumanEval), where models such as Qwen3-4B (79.33 IF, 86.59 Code) maintain clear advantages. This gap reflects limitations in our data mixture. For instance, \gigaverbosft does not contain samples designed to "benchmax" IFEval-style prompts, and has a limited volume of coding-specific training samples ($\sim$4\%). Figure~\ref{fig:instruct-3.7B-comparison} presents per-benchmark comparisons for the 3.7B Instruct model; comparisons at other scales are provided in Appendix~\ref{appendix:nstruct-think-results}.

\subsubsection{Think Variants}

Table~\ref{tab:think-results} compares the Think variants against reasoning models of similar size. Coding benchmarks are excluded because our Think models were not trained on coding data post-training.

\begin{table}[h]
\centering
\small
\caption{Post-training results for Think variants.}
\label{tab:think-results}
\begin{tabular}{lccc}
\toprule
 & \textbf{Total Avg.} & \textbf{K\&R (NPM)} & \textbf{Instruct} \\
\midrule
\textbf{\tucanocptlargethink} & 51.27 & 54.07 & 31.67 \\
SmolLM3-3B                        & 48.58          & 46.28          & 64.67 \\
Qwen3-4B                          & 46.35          & 40.97          & 84.00 \\
Qwen3-1.7B                        & 36.54          & 32.00          & 68.33 \\
\textbf{\tucanocptmediumthink}  & 27.54          & 26.67          & 33.67 \\
Qwen3-0.6B                        & 24.11          & 19.22          & 58.33 \\
\textbf{\tucanocptsmallthink}  & 14.41          & 12.52          & 27.67 \\
\bottomrule
\end{tabular}
\end{table}

Despite training on substantially less reasoning data and operating under a constrained 4,096-token context window (roughly half the inference budget used to evaluate the competing models), \tucanocptlargethink achieves the highest Knowledge \& Reasoning NPM (54.07), surpassing both Qwen3-4B (40.97) and SmolLM3-3B (46.28) on benchmarks including OAB, ARC-Challenge, BELEBELE, and MMLU---while reasoning entirely in Portuguese.

\begin{figure}[h]
\centering
\includegraphics[width=0.95\linewidth]{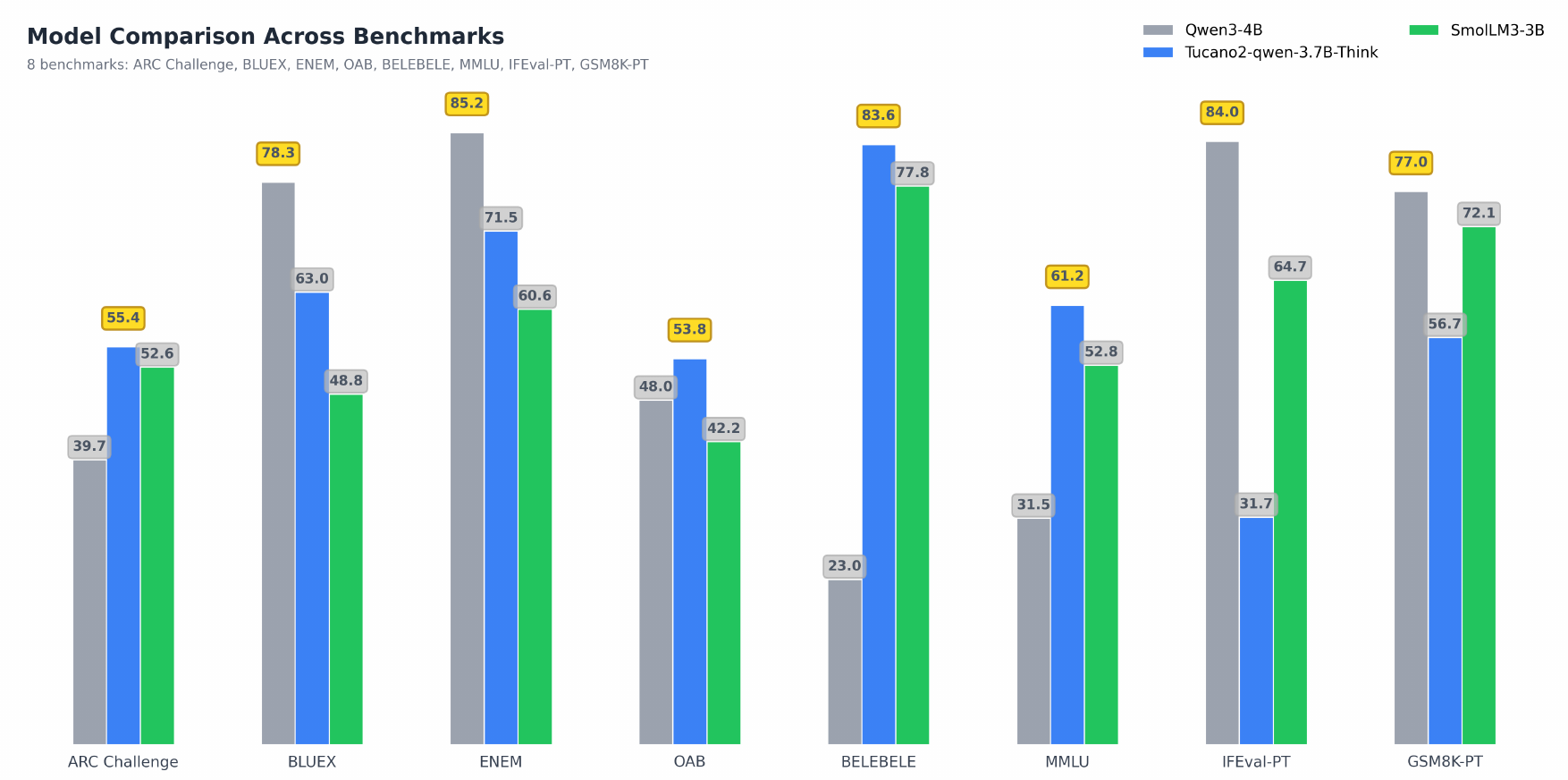}
\caption{Per-benchmark comparison for the Think variant.}
\label{fig:think-3.7B-comparison}
\end{figure}

We noticed that the evaluated reasoning models underperform on certain evaluation metrics, such as IFEval-PT and HumanEval (see Appendix~\ref{appendix:instruct-think-highlights}). This is consistent with the specialization trade-off: reasoning models optimize for extended chain-of-thought problem-solving rather than format-adherent instruction-following. Figure~\ref{fig:think-3.7B-comparison} presents per-benchmark comparisons for the 3.7B Think model.

\subsubsection{Long-Context Evaluation}

We evaluated all post-trained models on RULER-PT at context lengths of 1,024, 2,048, and 4,096 tokens. Tucano2 models fall substantially behind Qwen3 counterparts across all lengths, with the gap widening at longer contexts (Table~\ref{tab:ruler-all-lengths}). We attribute this primarily to the absence of long-context reasoning and retrieval samples in our SFT and preference data---tasks such as multi-key KV retrieval and common word extraction require robust long-range attention that must be explicitly cultivated during training. Addressing this limitation through long-context data augmentation and potential extensions of context length is a priority for future work. Detailed per-task RULER breakdowns are provided in Appendix~\ref{appendix:ruler-details}.

\begin{table}[h]
\centering
\small
\caption{RULER-PT aggregate scores across context lengths.}
\label{tab:ruler-all-lengths}
\begin{tabular}{lccc}
\toprule
\textbf{Model} & \textbf{1024} & \textbf{2048} & \textbf{4096} \\
\midrule
Qwen3-4B                          & 0.966 & 0.984 & 0.979 \\
Qwen3-1.7B                        & 0.977 & 0.930 & 0.961 \\
Qwen3-0.6B                        & 0.924 & 0.861 & 0.885 \\
Tucano2-qwen-3.7B-Think           & 0.817 & 0.765 & 0.707 \\
Tucano2-qwen-3.7B-Instruct        & 0.795 & 0.710 & 0.686 \\
Tucano2-qwen-1.5B-Think           & 0.603 & 0.501 & 0.440 \\
Tucano2-qwen-1.5B-Instruct        & 0.573 & 0.494 & 0.421 \\
Tucano2-qwen-0.5B-Instruct        & 0.607 & 0.481 & 0.382 \\
Tucano2-qwen-0.5B-Think           & 0.471 & 0.364 & 0.293 \\
\bottomrule
\end{tabular}
\end{table}

\newpage
\section{Resource Consumption}
\label{sec:resource_consumption}

\begin{tcolorbox}[colback=gray!5!white, colframe=gray!75!black, title=Summary / TL;DR]
\small

All energy and carbon estimates were tracked using the CodeCarbon library, using the average carbon intensity of NRW, Germany (0.38 kg CO$_2$e/kWh) as a baseline. The dominant cost is synthetic data generation (\gigaverbosynth: 14,400 kWh, 5,472 kg CO$_2$e), which alone accounts for roughly 73\% of the total tracked energy. Pretraining \tucanoptsmall consumed 873 kWh (332 kg CO$_2$e), while continual pretraining of the full Tucano-qwen family required 2,326 kWh (884 kg CO$_2$e). Across all phases, total tracked consumption amounts to approximately 20,856 kWh and 7,929 kg CO$_2$e. These figures represent a lower bound: energy consumed by CPU-based data preprocessing, cluster cooling, and other peripheral systems is not captured by CodeCarbon, so the true footprint is higher. Additionally, the GPU-based aggregated material footprint of all training runs in the \tucano series was estimated using a FLOP-based resource allocation methodology. The total material footprint across 10 training runs (covering the Base, Instruct, and Think models) amounts to 0.307 kg of copper, followed by 0.010 kg of iron, $4.53 \times 10^{-3}$ kg of tin, $2.99 \times 10^{-3}$ kg of silicon, and $2.47 \times 10^{-3}$ kg of nickel.

\end{tcolorbox}

\subsection{Energy and Carbon Analysis}

In Table~\ref{tab:resource-consumption}, we summarize the estimated energy consumption and carbon footprint for each phase of the project, as reported by CodeCarbon~\citep{codecarbon}. All projected carbon emissions estimates used the average energy grid carbon intensity for North Rhine-Westphalia (0.38 kg CO$_2$e/kWh), Germany.

\begin{table}[h]
\centering
\small
\caption{Estimated energy consumption and carbon footprint per project phase.}
\label{tab:resource-consumption}
\begin{tabular}{lrr}
\toprule
\textbf{Phase} & \textbf{Energy (kWh)} & \textbf{Carbon (kg CO$_2$e)} \\
\midrule
\gigaverbosynth (data generation) & 14,400 &  5,472 \\
Continual Pretraining             &  2,326 &    884 \\
\gigaverbo Ablations              &  1,600 &    608 \\
Evaluations (all models)          &  1,000 &    380 \\
Post-Training (SFT + PO)          &    530 &    201 \\
Pretraining                       &    873 &    332 \\
\midrule
\textbf{Total (tracked)}               & \textbf{20,729} & \textbf{7,877} \\
\bottomrule
\end{tabular}
\end{table}

\subsection{Resource Consumption Beyond Carbon}

Training AI models not only consumes energy during computation but also requires substantial material resources for manufacturing the training hardware, in this case, NVIDIA A100 SXM GPUs. Based on the material composition of the GPUs, their assumed operational lifespan, and the model FLOP utilization achieved during training runs, the share of hardware-related resource consumption attributable to each training run can be estimated. Using the mineral composition analysis reported in \cite{falk2025morethancarbon} and the FLOP-based resource allocation methodology introduced in \cite{falk2025flops}, we estimate the aggregated material footprint of all training runs in the \tucano series. This includes all pre-training and post-training runs across all model sizes (see Table \ref{tab:modelOverview}).

\begin{table}[ht]
\centering
\caption{Model configurations and training budgets (tokens) used for resource footprint estimation.}
\label{tab:modelOverview}
\begin{tabular}{llccc}
\toprule
\textbf{Family} & \textbf{Params} & \textbf{Base} & \textbf{Instruct} & \textbf{Think} \\
\midrule
Tucano2-0.6      & 670M & 408B & --   & --   \\
Tucano2-qwen-0.5 & 490M & 50B  & 4.5B & 250M \\
Tucano2-qwen-1.5 & 1.5B & 100B & 4.5B & 250M \\
Tucano2-qwen-3.7 & 3.7B & 50B  & 4.5B & 250M \\
\bottomrule
\end{tabular}
\end{table}

Using an average MFU of $\sim$60\%, assuming a GPU lifespan of three years, and training performed on NVIDIA A100 SXM GPUs, the total material footprint across all 10 training runs (covering all Base, Instruct, and Think models) amounts to 0.307 kg of copper, followed by 0.010 kg of iron, $4.53 \times 10^{-3}$ kg of tin, $2.99 \times 10^{-3}$ kg of silicon, and $2.47 \times 10^{-3}$ kg of nickel. Smaller amounts include aluminum, calcium, chromium, and barium, while all remaining elements are below the mg scale. Copper clearly dominates the overall elemental footprint, primarily due to its large share in the heatsink.

For three of our base models, we use Qwen3 as a foundation (Qwen3-0.6B-Base, Qwen3-1.7B-Base, Qwen3-4B-Base). Taking the aggregated mineral resource consumption of these three base models into account as well would add an additional 105.63 kg of copper, 3.49 kg of iron, 1.56 kg of tin, 1.03 kg of silicon, and 0.85 kg of nickel --- increasing the total footprint by roughly two orders of magnitude relative to the development of our models.

These estimates account only for the material content of the manufactured accelerators and do not include additional resource losses occurring during raw material extraction, processing, manufacturing, or production waste, and should therefore be interpreted as lower-bound estimates.

\begin{figure} [H]
    \centering
    \includegraphics[width=0.75\linewidth]{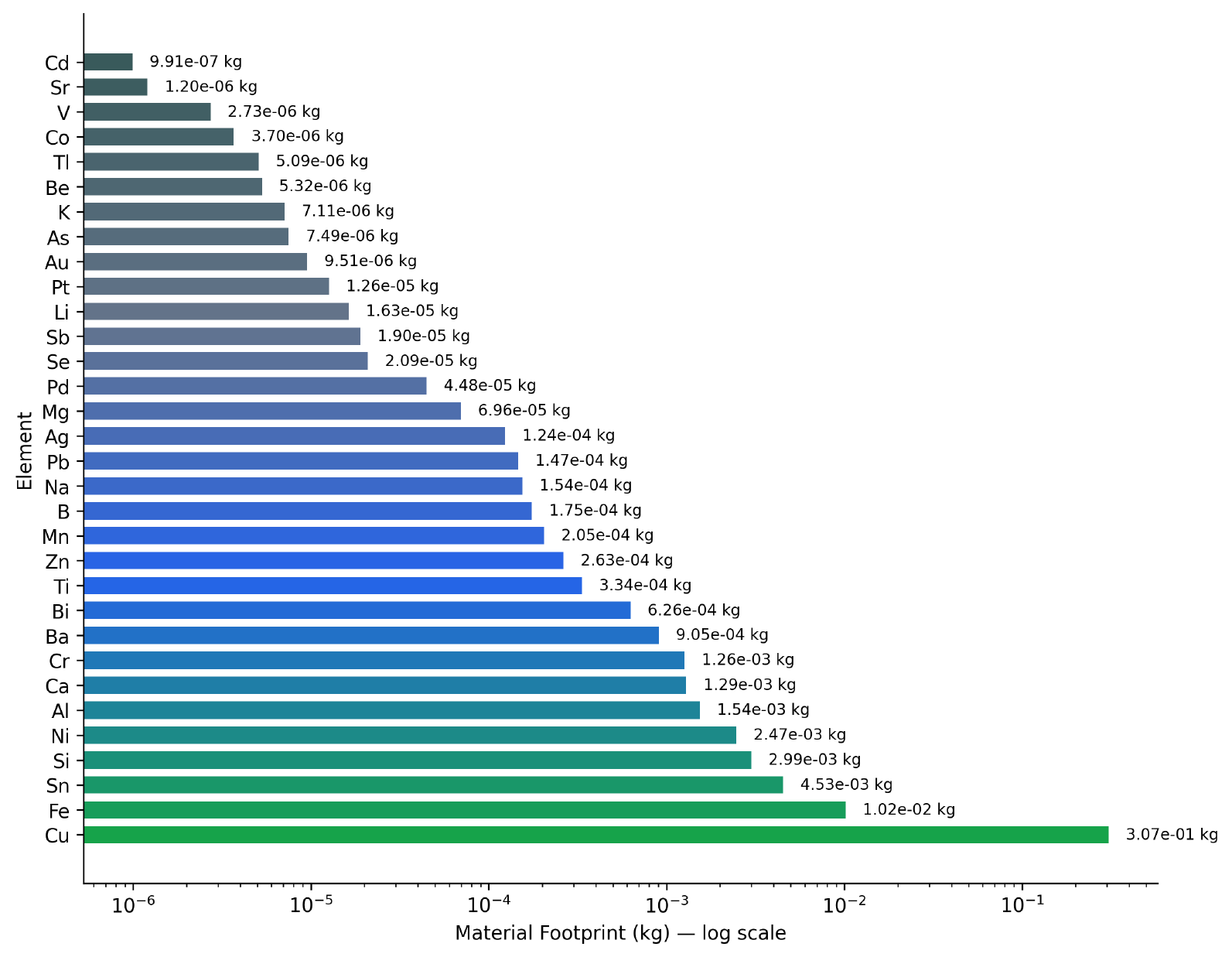}
    \caption{Aggregated material footprint tied to the training of our models. Estimations were made using the parameter count of each model, the amount of tokens each run took, an average MFU of 60\%, and an assumed GPU lifespan of 3 years.}
    \label{fig:placeholder}
\end{figure}

\subsection{Sustainability Discussion}

The resource accounting above surfaces several points relevant to sustainability in LLM research.

\paragraph{Synthetic data generation dominates total cost.}
Over 69\% of all tracked energy was consumed during the construction of \gigaverbosynth, i.e., generating responses at scale with Qwen2.5-32B-Instruct. This highlights a structural tension in the current paradigm: while synthetic data enables high-quality supervision without expensive human annotation, inference at a billion-parameter scale is itself a significant compute sink. Future work should prioritize low-cost generation strategies---such as those proposed by \cite{maini2025beyondweb}---to reduce this overhead.

\paragraph{Continual pretraining is a compute-efficient path to strong performance.}
The entire Tucano-qwen family (three model scales, 50--100B tokens each) was trained for 2,326 kWh, which is only $\sim$2.7$\times$ the cost of training the 670M \tucanoptsmall from scratch. Given that the 3.7B Tucano-qwen models surpass substantially larger baselines on Portuguese benchmarks (Section~\ref{sec:continual_pretraining}), continual pretraining with tokenizer transplantation is a promising, energy-efficient strategy for language adaptation.

\paragraph{These estimates are a lower bound.}
CodeCarbon tracks only GPU device power draw during instrumented runs. Energy consumed by CPU-only preprocessing pipelines (tokenization, deduplication, quality filtering), cluster interconnects, storage I/O, cooling infrastructure, and idle reservation periods is not reflected in Table~\ref{tab:resource-consumption}. The actual carbon footprint of this work is therefore higher than reported, and we encourage future work to adopt whole-system energy auditing where possible. In addition, these estimates exclude embodied carbon emissions outside the use phase, i.e., emissions that arise from raw material extraction, component manufacturing, transportation, and end-of-life disposal.

\paragraph{Contextualising the carbon footprint.}
The total tracked footprint of $\sim$7,900 kg CO$_2$e is comparable to roughly 2--3 transatlantic round-trip flights per person and is several orders of magnitude lower than the reported costs of training frontier models. For instance, the energy consumption for Llama 4 during training is estimated at 7.38 million GPU hours of computation, which translates to approximately 1,999 tons of CO$_2$e emissions~\citep{arxiv2026llama4herdarchitecture}. While this does not diminish the responsibility to minimize unnecessary computation, it illustrates that targeted, linguistically-motivated model development can be conducted at a comparatively modest environmental cost, particularly when training is co-located with low-carbon energy grids.

\newpage
\section{Conclusion}
\label{sec:conclusion}

In this work, we presented \tucano, a fully open suite of Portuguese large language models spanning 0.5B to 3.5B parameters, accompanied by the complete stack of artifacts required for reproducible development: large-scale pretraining corpora (\gigaverbo, \gigaverbosynth), annotation and filtering models, supervised fine-tuning and preference datasets (\gigaverbosft, \gigaverbopreferences), a Portuguese evaluation suite, and all training recipes and code---released under permissive licenses. Our pretraining and continual pretraining strategies demonstrate that carefully curated, language-specific data pipelines and efficient tokenization can yield models that outperform or match multilingual baselines. In particular, \tucanocptlargeinstruct achieves the highest Knowledge \& Reasoning score among all models in the 3--4B parameter range, while \tucanocptlargethink produces chain-of-thought reasoning entirely in Portuguese, a capability largely absent from prior open Portuguese models. Together, these contributions narrow persistent gaps in Portuguese NLP and provide a concrete, reproducible blueprint for community-driven LLM development in low-resource languages.

\paragraph{Limitations and Future Work.}
Despite these advances, several limitations point toward important directions for future research. First, our synthetic data generation pipeline, while effective, accounted for the largest share of total energy consumption (${\sim}$73\%), underscoring the need for more efficient, scalable generation methods that reduce costs without sacrificing diversity. Second, our post-training pipeline relies on SFT followed by Anchored Preference Optimization, but the preference dataset (\gigaverbopreferences) remains relatively small (${\sim}$28K pairs). Scaling preference data by an order of magnitude and investigating reinforcement learning--based post-training methods---such as GRPO and other reward-driven optimization frameworks---could substantially improve alignment quality, particularly for reasoning and agentic capabilities. Moreover, introducing a ``mid-training'' stage that targets areas where we currently lack (e.g., long-context reasoning, code generation) could be an effective step before returning to post-training, as demonstrated by previous studies~\citep{howard2018universallanguagemodelfinetuning, faircodegenteam2025cwmopenweightsllmresearch, xu2025phi4minireasoningexploringlimitssmall, bakouch2025smollm3}.

Looking ahead, we identify several high-impact opportunities. On the data side, developing more diverse and challenging SFT samples---especially for multi-step reasoning, structured output generation, and tool use---would further close the gap to frontier models. Expanding the volume and diversity of reasoning-focused training data, including both native Portuguese and carefully translated mathematical and scientific corpora, is critical for advancing reasoning capabilities. Context extension techniques and the construction of high-quality long-context samples for post-training represent another promising axis, given the increasing importance of long-horizon tasks in practical applications. Finally, designing datasets and training protocols tailored for agentic settings---including multi-turn user--agent interactions, long-horizon planning, and dynamic tool orchestration---would position Portuguese LLMs to participate in the next wave of autonomous AI systems. We hope that the open release of all artifacts produced in this work will lower barriers to entry and catalyze sustained, community-driven progress for Portuguese and other low-resource languages.

\newpage
\section*{Acknowledgements}

Polyglot is a project funded by the Federal Ministry of Education and Research (BMBF) and the Ministry of Culture and Science of the State of North Rhine-Westphalia (MWK) as part of TRA Sustainable Futures (University of Bonn) and the Excellence Strategy of the federal and state governments.

A.S. acknowledges funding by the Deutsche Forschungsgemeinschaft (DFG, German Research Foundation) as part of the CRC 1639 NuMeriQS – project No. 511713970.

We also gratefully acknowledge the granted access to the \href{https://www.hpc.uni-bonn.de/en/systems/marvin}{Marvin cluster} hosted by \href{https://www.uni-bonn.de/en}{University of Bonn} along with the support provided by its High Performance Computing \& Analytics Lab.

\newpage
\section*{Authors Contribution}

The corresponding author is \textbf{Nicholas Kluge Corrêa}. He is a postdoc researcher at the Bonn-Aachen International Center for Information Technology (b-it) / CAISA Lab, as part of the Lamarr Institute for Machine Learning and Artificial Intelligence, University of Bonn (Bonn, NRW, Germany). His contact email is \href{mailto:kluge@uni-bonn.de}{kluge@uni-bonn.de}. \textbf{N.K.C.} contributed to the project's idealization, development of the software stack, dataset curation, training, and evaluation of the models, as well as writing the article and documenting the repositories. \textbf{N.K.C.} is one of the Principal Investigators of this project.

\textbf{Aniket Sen} is a postdoc researcher at the Helmholtz-Institut für Strahlen und Kernphysik, University of Bonn, and the Bethe Center for Theoretical Physics, University of Bonn (Bonn, NRW, Germany). His contact email is \href{mailto:sen@hiskp.uni-bonn.de}{sen@hiskp.uni-bonn.de}. \textbf{A.S.} contributed to the optimization of the software stack, as well as the article's writing. \textbf{A.S.} is one of the Principal Investigators of this project.

\textbf{Shiza Fatimah} is a master's student working at the Bonn-Aachen International Center for Information Technology (b-it) / CAISA Lab, as part of the Lamarr Institute for Machine Learning and Artificial Intelligence, University of Bonn (Bonn, NRW, Germany). Her contact email is \href{mailto:s39sfati@uni-bonn.de}{s39sfati@uni-bonn.de}. \textbf{Sh.F.} contributed to the project's idealization, development of the software stack, dataset curation, training, and evaluation of the models, as well as writing the article and documenting the repositories.

\textbf{Sophia Falk} is a PhD researcher at the Bonn Sustainable AI Lab, Institute for Science and Ethics, University of Bonn (Bonn, NRW, Germany). Her contact email is \href{mailto:falk@iwe.uni-bonn.de}{falk@iwe.uni-bonn.de}. \textbf{So.F.} contributed to implementing the carbon tracking methodology, monitoring training runs, and writing the article.

\textbf{Lucie Flek} is a full professor at the University of Bonn, leading the Data Science and Language Technologies group. Her contact email is \href{mailto:flek@bit.uni-bonn.de}{flek@bit.uni-bonn.de}. \textbf{L.F.} contributed to the project's idealization, as well as writing and reviewing the article.

\textbf{Lennard Landgraf} is a research assistant at the Center for Science and Thought (CST),  University of Bonn (Bonn, NRW, Germany). His contact email is \href{mailto:lanlen@uni-bonn.de}{lanlen@uni-bonn.de}. \textbf{L.L.} contributed to the project's idealization, documentation, organization, as well as writing and reviewing the article.

\textbf{Julia Kastner} is a research assistant at the Center for Science and Thought (CST), University of Bonn (Bonn, NRW, Germany). Her contact email is \href{mailto:jkastner@uni-bonn.de}{jkastner@uni-bonn.de}. \textbf{J.K.} contributed to the project's idealization, documentation, organization, as well as writing and reviewing the article.

\newpage
\begin{small}
\bibliography{biblio}

@misc{cohere2024gap,
    author={{Cohere For AI team}},
    title={Policy Primer - The AI Language Gap},
    howpublished={\url{https://cohere.com/research/papers/policy-primer-the-ai-language-gap-2024-06-27}},
    year={2024},
    publisher={Cohere For AI}
}

@inproceedings{hedderich-etal-2021-survey,
    title = "A Survey on Recent Approaches for Natural Language Processing in Low-Resource Scenarios",
    author = {Hedderich, Michael A.  and
      Lange, Lukas  and
      Adel, Heike  and
      Str{\"o}tgen, Jannik  and
      Klakow, Dietrich},
    editor = "Toutanova, Kristina  and
      Rumshisky, Anna  and
      Zettlemoyer, Luke  and
      Hakkani-Tur, Dilek  and
      Beltagy, Iz  and
      Bethard, Steven  and
      Cotterell, Ryan  and
      Chakraborty, Tanmoy  and
      Zhou, Yichao",
    booktitle = "Proceedings of the 2021 Conference of the North American Chapter of the Association for Computational Linguistics: Human Language Technologies",
    month = jun,
    year = "2021",
    address = "Online",
    publisher = "Association for Computational Linguistics",
    url = "https://aclanthology.org/2021.naacl-main.201/",
    doi = "10.18653/v1/2021.naacl-main.201",
    pages = "2545--2568"
}

@misc{ranathunga2021neuralmachinetranslationlowresource,
      title={Neural Machine Translation for Low-Resource Languages: A Survey}, 
      author={Surangika Ranathunga and En-Shiun Annie Lee and Marjana Prifti Skenduli and Ravi Shekhar and Mehreen Alam and Rishemjit Kaur},
      year={2021},
      eprint={2106.15115},
      archivePrefix={arXiv},
      primaryClass={cs.CL},
      url={https://arxiv.org/abs/2106.15115}, 
}

@inproceedings{joshi-etal-2020-state,
    title = "The State and Fate of Linguistic Diversity and Inclusion in the {NLP} World",
    author = "Joshi, Pratik  and
      Santy, Sebastin  and
      Budhiraja, Amar  and
      Bali, Kalika  and
      Choudhury, Monojit",
    editor = "Jurafsky, Dan  and
      Chai, Joyce  and
      Schluter, Natalie  and
      Tetreault, Joel",
    booktitle = "Proceedings of the 58th Annual Meeting of the Association for Computational Linguistics",
    month = jul,
    year = "2020",
    address = "Online",
    publisher = "Association for Computational Linguistics",
    url = "https://aclanthology.org/2020.acl-main.560/",
    doi = "10.18653/v1/2020.acl-main.560",
    pages = "6282--6293"
}

@misc{virtanen2019multilingualenoughbertfinnish,
      title={Multilingual is not enough: BERT for Finnish}, 
      author={Antti Virtanen and Jenna Kanerva and Rami Ilo and Jouni Luoma and Juhani Luotolahti and Tapio Salakoski and Filip Ginter and Sampo Pyysalo},
      year={2019},
      eprint={1912.07076},
      archivePrefix={arXiv},
      primaryClass={cs.CL},
      url={https://arxiv.org/abs/1912.07076}, 
}

@inproceedings{Martin_2020,
   title={CamemBERT: a Tasty French Language Model},
   url={http://dx.doi.org/10.18653/v1/2020.acl-main.645},
   DOI={10.18653/v1/2020.acl-main.645},
   booktitle={Proceedings of the 58th Annual Meeting of the Association for Computational Linguistics},
   publisher={Association for Computational Linguistics},
   author={Martin, Louis and Muller, Benjamin and Ortiz Suárez, Pedro Javier and Dupont, Yoann and Romary, Laurent and de la Clergerie, Éric and Seddah, Djamé and Sagot, Benoît},
   year={2020},
   pages={7203–7219} 
}

@misc{armengolestape2021multilingualmodelsbestchoice,
      title={Are Multilingual Models the Best Choice for Moderately Under-resourced Languages? A Comprehensive Assessment for Catalan}, 
      author={Jordi Armengol-Estapé and Casimiro Pio Carrino and Carlos Rodriguez-Penagos and Ona de Gibert Bonet and Carme Armentano-Oller and Aitor Gonzalez-Agirre and Maite Melero and Marta Villegas},
      year={2021},
      eprint={2107.07903},
      archivePrefix={arXiv},
      primaryClass={cs.CL},
      url={https://arxiv.org/abs/2107.07903}, 
}

@article{Corr_a_2025,
   title={Tucano: Advancing neural text generation for Portuguese},
   volume={6},
   ISSN={2666-3899},
   url={http://dx.doi.org/10.1016/j.patter.2025.101325},
   DOI={10.1016/j.patter.2025.101325},
   number={11},
   journal={Patterns},
   publisher={Elsevier BV},
   author={Corr{\^e}a, Nicholas Kluge and Sen, Aniket and Falk, Sophia and Fatimah, Shiza},
   year={2025},
   month=nov, pages={101325} 
}

@misc{silva2025evaluationaiethicstools,
      title={Evaluation of AI Ethics Tools in Language Models: A Developers' Perspective Case Study}, 
      author={Jhessica Silva and Diego A. B. Moreira and Gabriel O. dos Santos and Alef Ferreira and Helena Maia and Sandra Avila and Helio Pedrini},
      year={2025},
      eprint={2512.15791},
      archivePrefix={arXiv},
      primaryClass={cs.CY},
      url={https://arxiv.org/abs/2512.15791}, 
}

@misc{gaia-gemma-3-4b-2025,
    title={GAIA: An Open Language Model for Brazilian Portuguese},
    author={Celso Gon\c{c}alves Camilo{-}Junior and S\'avio Salvarino Teles de Oliveira and Lucas Araujo Pereira and Marcellus Amadeus and Rodrigo Scotti and Daniel Fazzioni and Artur Matos Andrade Novais and Salatiel Abra\~{a}o Avelar Jord\~{a}o},
    year={2025},
    publisher={Hugging Face},
    journal={Hugging Face repository},
    howpublished={\url{https://huggingface.co/CEIA-UFG/Gemma-3-Gaia-PT-BR-4b-it}}
}

@misc{jurema_7b_2025,
    title        = {Jurema-7B: Modelo de Linguagem Especializado no Domínio Jurídico Brasileiro},
    author       = {NeuralMind and Escavador},
    year         = {2025},
    publisher    = {Hugging Face},
    journal      = {Hugging Face repository},
    howpublished = {\url{https://huggingface.co/Jurema-br/Jurema-7B}},
    note         = {Desenvolvido com apoio e financiamento da FINEP}
}

@article{Cruz-Castaneda_Amadeus_2025,
  title        = {Large Languages Models in Brazilian Portuguese: A Chronological Survey},
  author       = {Cruz-Castañeda, William Alberto and Amadeus, Marcellus},
  journal      = {Journal of the Brazilian Computer Society},
  volume       = {31},
  number       = {1},
  pages        = {1167--1186},
  year         = {2025},
  month        = oct,
  doi          = {10.5753/jbcs.2025.5789},
  url          = {https://journals-sol.sbc.org.br/index.php/jbcs/article/view/5789}
}

@misc{cruzcastaneda2025amadeusverbotechnicalreportpowerful,
      title={Amadeus-Verbo Technical Report: The powerful Qwen2.5 family models trained in Portuguese}, 
      author={William Alberto Cruz-Castañeda and Marcellus Amadeus},
      year={2025},
      eprint={2506.00019},
      archivePrefix={arXiv},
      primaryClass={cs.CL},
      url={https://arxiv.org/abs/2506.00019}, 
}

@misc{abonizio2025sabia3technicalreport,
      title={Sabi\'a-3 Technical Report}, 
      author={Hugo Abonizio and Thales Sales Almeida and Thiago Laitz and Roseval Malaquias Junior and Giovana Kerche Bonás and Rodrigo Nogueira and Ramon Pires},
      year={2025},
      eprint={2410.12049},
      archivePrefix={arXiv},
      primaryClass={cs.CL},
      url={https://arxiv.org/abs/2410.12049}, 
}

@misc{deepseekai2025deepseekv3technicalreport,
      title={DeepSeek-V3 Technical Report}, 
      author={DeepSeek-AI and Aixin Liu and Bei Feng and Bing Xue and Bingxuan Wang and Bochao Wu and Chengda Lu and Chenggang Zhao and Chengqi Deng and Chenyu Zhang and Chong Ruan and Damai Dai and Daya Guo and Dejian Yang and Deli Chen and Dongjie Ji and Erhang Li and Fangyun Lin and Fucong Dai and Fuli Luo and Guangbo Hao and Guanting Chen and Guowei Li and H. Zhang and Han Bao and Hanwei Xu and Haocheng Wang and Haowei Zhang and Honghui Ding and Huajian Xin and Huazuo Gao and Hui Li and Hui Qu and J. L. Cai and Jian Liang and Jianzhong Guo and Jiaqi Ni and Jiashi Li and Jiawei Wang and Jin Chen and Jingchang Chen and Jingyang Yuan and Junjie Qiu and Junlong Li and Junxiao Song and Kai Dong and Kai Hu and Kaige Gao and Kang Guan and Kexin Huang and Kuai Yu and Lean Wang and Lecong Zhang and Lei Xu and Leyi Xia and Liang Zhao and Litong Wang and Liyue Zhang and Meng Li and Miaojun Wang and Mingchuan Zhang and Minghua Zhang and Minghui Tang and Mingming Li and Ning Tian and Panpan Huang and Peiyi Wang and Peng Zhang and Qiancheng Wang and Qihao Zhu and Qinyu Chen and Qiushi Du and R. J. Chen and R. L. Jin and Ruiqi Ge and Ruisong Zhang and Ruizhe Pan and Runji Wang and Runxin Xu and Ruoyu Zhang and Ruyi Chen and S. S. Li and Shanghao Lu and Shangyan Zhou and Shanhuang Chen and Shaoqing Wu and Shengfeng Ye and Shengfeng Ye and Shirong Ma and Shiyu Wang and Shuang Zhou and Shuiping Yu and Shunfeng Zhou and Shuting Pan and T. Wang and Tao Yun and Tian Pei and Tianyu Sun and W. L. Xiao and Wangding Zeng and Wanjia Zhao and Wei An and Wen Liu and Wenfeng Liang and Wenjun Gao and Wenqin Yu and Wentao Zhang and X. Q. Li and Xiangyue Jin and Xianzu Wang and Xiao Bi and Xiaodong Liu and Xiaohan Wang and Xiaojin Shen and Xiaokang Chen and Xiaokang Zhang and Xiaosha Chen and Xiaotao Nie and Xiaowen Sun and Xiaoxiang Wang and Xin Cheng and Xin Liu and Xin Xie and Xingchao Liu and Xingkai Yu and Xinnan Song and Xinxia Shan and Xinyi Zhou and Xinyu Yang and Xinyuan Li and Xuecheng Su and Xuheng Lin and Y. K. Li and Y. Q. Wang and Y. X. Wei and Y. X. Zhu and Yang Zhang and Yanhong Xu and Yanhong Xu and Yanping Huang and Yao Li and Yao Zhao and Yaofeng Sun and Yaohui Li and Yaohui Wang and Yi Yu and Yi Zheng and Yichao Zhang and Yifan Shi and Yiliang Xiong and Ying He and Ying Tang and Yishi Piao and Yisong Wang and Yixuan Tan and Yiyang Ma and Yiyuan Liu and Yongqiang Guo and Yu Wu and Yuan Ou and Yuchen Zhu and Yuduan Wang and Yue Gong and Yuheng Zou and Yujia He and Yukun Zha and Yunfan Xiong and Yunxian Ma and Yuting Yan and Yuxiang Luo and Yuxiang You and Yuxuan Liu and Yuyang Zhou and Z. F. Wu and Z. Z. Ren and Zehui Ren and Zhangli Sha and Zhe Fu and Zhean Xu and Zhen Huang and Zhen Zhang and Zhenda Xie and Zhengyan Zhang and Zhewen Hao and Zhibin Gou and Zhicheng Ma and Zhigang Yan and Zhihong Shao and Zhipeng Xu and Zhiyu Wu and Zhongyu Zhang and Zhuoshu Li and Zihui Gu and Zijia Zhu and Zijun Liu and Zilin Li and Ziwei Xie and Ziyang Song and Ziyi Gao and Zizheng Pan},
      year={2025},
      eprint={2412.19437},
      archivePrefix={arXiv},
      primaryClass={cs.CL},
      url={https://arxiv.org/abs/2412.19437}, 
}

@misc{kimiteam2026kimik2openagentic,
      title={Kimi K2: Open Agentic Intelligence}, 
      author={Yifan Bai and Yiping Bao and Y. Charles and Cheng Chen and Guanduo Chen and Haiting Chen and Huarong Chen and Jiahao Chen and Ningxin Chen and Ruijue Chen and Yanru Chen and Yuankun Chen and Yutian Chen and Zhuofu Chen and Jialei Cui and Hao Ding and Mengnan Dong and Angang Du and Chenzhuang Du and Dikang Du and Yulun Du and Yu Fan and Yichen Feng and Kelin Fu and Bofei Gao and Chenxiao Gao and Hongcheng Gao and Peizhong Gao and Tong Gao and Yuyao Ge and Shangyi Geng and Qizheng Gu and Xinran Gu and Longyu Guan and Haiqing Guo and Jianhang Guo and Xiaoru Hao and Tianhong He and Weiran He and Wenyang He and Yunjia He and Chao Hong and Hao Hu and Yangyang Hu and Zhenxing Hu and Weixiao Huang and Zhiqi Huang and Zihao Huang and Tao Jiang and Zhejun Jiang and Xinyi Jin and Yongsheng Kang and Guokun Lai and Cheng Li and Fang Li and Haoyang Li and Ming Li and Wentao Li and Yang Li and Yanhao Li and Yiwei Li and Zhaowei Li and Zheming Li and Hongzhan Lin and Xiaohan Lin and Zongyu Lin and Chengyin Liu and Chenyu Liu and Hongzhang Liu and Jingyuan Liu and Junqi Liu and Liang Liu and Shaowei Liu and T. Y. Liu and Tianwei Liu and Weizhou Liu and Yangyang Liu and Yibo Liu and Yiping Liu and Yue Liu and Zhengying Liu and Enzhe Lu and Haoyu Lu and Lijun Lu and Yashuo Luo and Shengling Ma and Xinyu Ma and Yingwei Ma and Shaoguang Mao and Jie Mei and Xin Men and Yibo Miao and Siyuan Pan and Yebo Peng and Ruoyu Qin and Zeyu Qin and Bowen Qu and Zeyu Shang and Lidong Shi and Shengyuan Shi and Feifan Song and Jianlin Su and Zhengyuan Su and Lin Sui and Xinjie Sun and Flood Sung and Yunpeng Tai and Heyi Tang and Jiawen Tao and Qifeng Teng and Chaoran Tian and Chensi Wang and Dinglu Wang and Feng Wang and Hailong Wang and Haiming Wang and Jianzhou Wang and Jiaxing Wang and Jinhong Wang and Shengjie Wang and Shuyi Wang and Si Wang and Xinyuan Wang and Yao Wang and Yejie Wang and Yiqin Wang and Yuxin Wang and Yuzhi Wang and Zhaoji Wang and Zhengtao Wang and Zhengtao Wang and Zhexu Wang and Chu Wei and Qianqian Wei and Haoning Wu and Wenhao Wu and Xingzhe Wu and Yuxin Wu and Chenjun Xiao and Jin Xie and Xiaotong Xie and Weimin Xiong and Boyu Xu and Jinjing Xu and L. H. Xu and Lin Xu and Suting Xu and Weixin Xu and Xinran Xu and Yangchuan Xu and Ziyao Xu and Jing Xu and Jing Xu and Junjie Yan and Yuzi Yan and Hao Yang and Xiaofei Yang and Yi Yang and Ying Yang and Zhen Yang and Zhilin Yang and Zonghan Yang and Haotian Yao and Xingcheng Yao and Wenjie Ye and Zhuorui Ye and Bohong Yin and Longhui Yu and Enming Yuan and Hongbang Yuan and Mengjie Yuan and Siyu Yuan and Haobing Zhan and Dehao Zhang and Hao Zhang and Wanlu Zhang and Xiaobin Zhang and Yadong Zhang and Yangkun Zhang and Yichi Zhang and Yizhi Zhang and Yongting Zhang and Yu Zhang and Yutao Zhang and Yutong Zhang and Zheng Zhang and Haotian Zhao and Yikai Zhao and Zijia Zhao and Huabin Zheng and Shaojie Zheng and Longguang Zhong and Jianren Zhou and Xinyu Zhou and Zaida Zhou and Jinguo Zhu and Zhen Zhu and Weiyu Zhuang and Xinxing Zu},
      year={2026},
      eprint={2507.20534},
      archivePrefix={arXiv},
      primaryClass={cs.LG},
      url={https://arxiv.org/abs/2507.20534}, 
}

@misc{5team2025glm45agenticreasoningcoding,
      title={GLM-4.5: Agentic, Reasoning, and Coding (ARC) Foundation Models}, 
      author={Aohan Zeng and Xin Lv and Qinkai Zheng and Zhenyu Hou and Bin Chen and Chengxing Xie and Cunxiang Wang and Da Yin and Hao Zeng and Jiajie Zhang and Kedong Wang and Lucen Zhong and Mingdao Liu and Rui Lu and Shulin Cao and Xiaohan Zhang and Xuancheng Huang and Yao Wei and Yean Cheng and Yifan An and Yilin Niu and Yuanhao Wen and Yushi Bai and Zhengxiao Du and Zihan Wang and Zilin Zhu and Bohan Zhang and Bosi Wen and Bowen Wu and Bowen Xu and Can Huang and Casey Zhao and Changpeng Cai and Chao Yu and Chen Li and Chendi Ge and Chenghua Huang and Chenhui Zhang and Chenxi Xu and Chenzheng Zhu and Chuang Li and Congfeng Yin and Daoyan Lin and Dayong Yang and Dazhi Jiang and Ding Ai and Erle Zhu and Fei Wang and Gengzheng Pan and Guo Wang and Hailong Sun and Haitao Li and Haiyang Li and Haiyi Hu and Hanyu Zhang and Hao Peng and Hao Tai and Haoke Zhang and Haoran Wang and Haoyu Yang and He Liu and He Zhao and Hongwei Liu and Hongxi Yan and Huan Liu and Huilong Chen and Ji Li and Jiajing Zhao and Jiamin Ren and Jian Jiao and Jiani Zhao and Jianyang Yan and Jiaqi Wang and Jiayi Gui and Jiayue Zhao and Jie Liu and Jijie Li and Jing Li and Jing Lu and Jingsen Wang and Jingwei Yuan and Jingxuan Li and Jingzhao Du and Jinhua Du and Jinxin Liu and Junkai Zhi and Junli Gao and Ke Wang and Lekang Yang and Liang Xu and Lin Fan and Lindong Wu and Lintao Ding and Lu Wang and Man Zhang and Minghao Li and Minghuan Xu and Mingming Zhao and Mingshu Zhai and Pengfan Du and Qian Dong and Shangde Lei and Shangqing Tu and Shangtong Yang and Shaoyou Lu and Shijie Li and Shuang Li and Shuang-Li and Shuxun Yang and Sibo Yi and Tianshu Yu and Wei Tian and Weihan Wang and Wenbo Yu and Weng Lam Tam and Wenjie Liang and Wentao Liu and Xiao Wang and Xiaohan Jia and Xiaotao Gu and Xiaoying Ling and Xin Wang and Xing Fan and Xingru Pan and Xinyuan Zhang and Xinze Zhang and Xiuqing Fu and Xunkai Zhang and Yabo Xu and Yandong Wu and Yida Lu and Yidong Wang and Yilin Zhou and Yiming Pan and Ying Zhang and Yingli Wang and Yingru Li and Yinpei Su and Yipeng Geng and Yitong Zhu and Yongkun Yang and Yuhang Li and Yuhao Wu and Yujiang Li and Yunan Liu and Yunqing Wang and Yuntao Li and Yuxuan Zhang and Zezhen Liu and Zhen Yang and Zhengda Zhou and Zhongpei Qiao and Zhuoer Feng and Zhuorui Liu and Zichen Zhang and Zihan Wang and Zijun Yao and Zikang Wang and Ziqiang Liu and Ziwei Chai and Zixuan Li and Zuodong Zhao and Wenguang Chen and Jidong Zhai and Bin Xu and Minlie Huang and Hongning Wang and Juanzi Li and Yuxiao Dong and Jie Tang},
      year={2025},
      eprint={2508.06471},
      archivePrefix={arXiv},
      primaryClass={cs.CL},
      url={https://arxiv.org/abs/2508.06471}, 
}

@misc{allal2025smollm2smolgoesbig,
      title={SmolLM2: When Smol Goes Big -- Data-Centric Training of a Small Language Model}, 
      author={Loubna Ben Allal and Anton Lozhkov and Elie Bakouch and Gabriel Martín Blázquez and Guilherme Penedo and Lewis Tunstall and Andrés Marafioti and Hynek Kydlíček and Agustín Piqueres Lajarín and Vaibhav Srivastav and Joshua Lochner and Caleb Fahlgren and Xuan-Son Nguyen and Clémentine Fourrier and Ben Burtenshaw and Hugo Larcher and Haojun Zhao and Cyril Zakka and Mathieu Morlon and Colin Raffel and Leandro von Werra and Thomas Wolf},
      year={2025},
      eprint={2502.02737},
      archivePrefix={arXiv},
      primaryClass={cs.CL},
      url={https://arxiv.org/abs/2502.02737}, 
}

@article{bakouch2025smollm3,
  title={SmolLM3: smol, multilingual, long-context reasoner},
  author={Bakouch, Elie and Allal, Loubna Ben and Lozhkov, Anton and Tazi, Nouamane and Tunstall, Lewis and Patino, Carlos Miguel and Beeching, Edward and Roucher, Aymeric and Reedi, Aksel Joonas and Gallou{\'e}dec, Quentin and others},
  journal={Hugging Face Blog},
  year={2025}
}

@misc{groeneveld2024olmoacceleratingsciencelanguage,
      title={OLMo: Accelerating the Science of Language Models}, 
      author={Dirk Groeneveld and Iz Beltagy and Pete Walsh and Akshita Bhagia and Rodney Kinney and Oyvind Tafjord and Ananya Harsh Jha and Hamish Ivison and Ian Magnusson and Yizhong Wang and Shane Arora and David Atkinson and Russell Authur and Khyathi Raghavi Chandu and Arman Cohan and Jennifer Dumas and Yanai Elazar and Yuling Gu and Jack Hessel and Tushar Khot and William Merrill and Jacob Morrison and Niklas Muennighoff and Aakanksha Naik and Crystal Nam and Matthew E. Peters and Valentina Pyatkin and Abhilasha Ravichander and Dustin Schwenk and Saurabh Shah and Will Smith and Emma Strubell and Nishant Subramani and Mitchell Wortsman and Pradeep Dasigi and Nathan Lambert and Kyle Richardson and Luke Zettlemoyer and Jesse Dodge and Kyle Lo and Luca Soldaini and Noah A. Smith and Hannaneh Hajishirzi},
      year={2024},
      eprint={2402.00838},
      archivePrefix={arXiv},
      primaryClass={cs.CL},
      url={https://arxiv.org/abs/2402.00838}, 
}

@misc{olmo20252olmo2furious,
      title={2 OLMo 2 Furious}, 
      author={Pete Walsh and Luca Soldaini and Dirk Groeneveld and Kyle Lo and Shane Arora and Akshita Bhagia and Yuling Gu and Shengyi Huang and Matt Jordan and Nathan Lambert and Dustin Schwenk and Oyvind Tafjord and Taira Anderson and David Atkinson and Faeze Brahman and Christopher Clark and Pradeep Dasigi and Nouha Dziri and Allyson Ettinger and Michal Guerquin and David Heineman and Hamish Ivison and Pang Wei Koh and Jiacheng Liu and Saumya Malik and William Merrill and Lester James V. Miranda and Jacob Morrison and Tyler Murray and Crystal Nam and Jake Poznanski and Valentina Pyatkin and Aman Rangapur and Michael Schmitz and Sam Skjonsberg and David Wadden and Christopher Wilhelm and Michael Wilson and Luke Zettlemoyer and Ali Farhadi and Noah A. Smith and Hannaneh Hajishirzi},
      year={2025},
      eprint={2501.00656},
      archivePrefix={arXiv},
      primaryClass={cs.CL},
      url={https://arxiv.org/abs/2501.00656}, 
}

@misc{olmo2025olmo3,
      title={Olmo 3}, 
      author={Allyson Ettinger and Amanda Bertsch and Bailey Kuehl and David Graham and David Heineman and Dirk Groeneveld and Faeze Brahman and Finbarr Timbers and Hamish Ivison and Jacob Morrison and Jake Poznanski and Kyle Lo and Luca Soldaini and Matt Jordan and Mayee Chen and Michael Noukhovitch and Nathan Lambert and Pete Walsh and Pradeep Dasigi and Robert Berry and Saumya Malik and Saurabh Shah and Scott Geng and Shane Arora and Shashank Gupta and Taira Anderson and Teng Xiao and Tyler Murray and Tyler Romero and Victoria Graf and Akari Asai and Akshita Bhagia and Alexander Wettig and Alisa Liu and Aman Rangapur and Chloe Anastasiades and Costa Huang and Dustin Schwenk and Harsh Trivedi and Ian Magnusson and Jaron Lochner and Jiacheng Liu and Lester James V. Miranda and Maarten Sap and Malia Morgan and Michael Schmitz and Michal Guerquin and Michael Wilson and Regan Huff and Ronan Le Bras and Rui Xin and Rulin Shao and Sam Skjonsberg and Shannon Zejiang Shen and Shuyue Stella Li and Tucker Wilde and Valentina Pyatkin and Will Merrill and Yapei Chang and Yuling Gu and Zhiyuan Zeng and Ashish Sabharwal and Luke Zettlemoyer and Pang Wei Koh and Ali Farhadi and Noah A. Smith and Hannaneh Hajishirzi},
      year={2025},
      eprint={2512.13961},
      archivePrefix={arXiv},
      primaryClass={cs.CL},
      url={https://arxiv.org/abs/2512.13961}, 
}

@misc{k2team2025k2v2360openreasoningenhancedllm,
      title={K2-V2: A 360-Open, Reasoning-Enhanced LLM}, 
      author={Zhengzhong Liu and Liping Tang and Linghao Jin and Haonan Li and Nikhil Ranjan and Desai Fan and Shaurya Rohatgi and Richard Fan and Omkar Pangarkar and Huijuan Wang and Zhoujun Cheng and Suqi Sun and Seungwook Han and Bowen Tan and Gurpreet Gosal and Xudong Han and Varad Pimpalkhute and Shibo Hao and Ming Shan Hee and Joel Hestness and Haolong Jia and Liqun Ma and Aaryamonvikram Singh and Daria Soboleva and Natalia Vassilieva and Renxi Wang and Yingquan Wu and Yuekai Sun and Taylor Killian and Alexander Moreno and John Maggs and Hector Ren and Guowei He and Hongyi Wang and Xuezhe Ma and Yuqi Wang and Mikhail Yurochkin and Eric P. Xing},
      year={2025},
      eprint={2512.06201},
      archivePrefix={arXiv},
      primaryClass={cs.LG},
      url={https://arxiv.org/abs/2512.06201}, 
}

@article{zhou2025megamath,
  title={Megamath: Pushing the limits of open math corpora},
  author={Zhou, Fan and Wang, Zengzhi and Ranjan, Nikhil and Cheng, Zhoujun and Tang, Liping and He, Guowei and Liu, Zhengzhong and Xing, Eric P},
  journal={arXiv preprint arXiv:2504.02807},
  year={2025}
}

@misc{liu2023llm360,
      title={LLM360: Towards Fully Transparent Open-Source LLMs}, 
      author={Zhengzhong Liu and Aurick Qiao and Willie Neiswanger and Hongyi Wang and Bowen Tan and Tianhua Tao and Junbo Li and Yuqi Wang and Suqi Sun and Omkar Pangarkar and Richard Fan and Yi Gu and Victor Miller and Yonghao Zhuang and Guowei He and Haonan Li and Fajri Koto and Liping Tang and Nikhil Ranjan and Zhiqiang Shen and Xuguang Ren and Roberto Iriondo and Cun Mu and Zhiting Hu and Mark Schulze and Preslav Nakov and Tim Baldwin and Eric P. Xing},
      year={2023},
      eprint={2312.06550},
      archivePrefix={arXiv},
      primaryClass={cs.CL}
}

@misc{apertus2025apertusdemocratizingopencompliant,
      title={Apertus: Democratizing Open and Compliant LLMs for Global Language Environments}, 
      author={Alejandro Hernández-Cano and Alexander Hägele and Allen Hao Huang and Angelika Romanou and Antoni-Joan Solergibert and Barna Pasztor and Bettina Messmer and Dhia Garbaya and Eduard Frank Ďurech and Ido Hakimi and Juan García Giraldo and Mete Ismayilzada and Negar Foroutan and Skander Moalla and Tiancheng Chen and Vinko Sabolčec and Yixuan Xu and Michael Aerni and Badr AlKhamissi and Inés Altemir Mariñas and Mohammad Hossein Amani and Matin Ansaripour and Ilia Badanin and Harold Benoit and Emanuela Boros and Nicholas Browning and Fabian Bösch and Maximilian Böther and Niklas Canova and Camille Challier and Clement Charmillot and Jonathan Coles and Jan Deriu and Arnout Devos and Lukas Drescher and Daniil Dzenhaliou and Maud Ehrmann and Dongyang Fan and Simin Fan and Silin Gao and Miguel Gila and María Grandury and Diba Hashemi and Alexander Hoyle and Jiaming Jiang and Mark Klein and Andrei Kucharavy and Anastasiia Kucherenko and Frederike Lübeck and Roman Machacek and Theofilos Manitaras and Andreas Marfurt and Kyle Matoba and Simon Matrenok and Henrique Mendonça and Fawzi Roberto Mohamed and Syrielle Montariol and Luca Mouchel and Sven Najem-Meyer and Jingwei Ni and Gennaro Oliva and Matteo Pagliardini and Elia Palme and Andrei Panferov and Léo Paoletti and Marco Passerini and Ivan Pavlov and Auguste Poiroux and Kaustubh Ponkshe and Nathan Ranchin and Javi Rando and Mathieu Sauser and Jakhongir Saydaliev and Muhammad Ali Sayfiddinov and Marian Schneider and Stefano Schuppli and Marco Scialanga and Andrei Semenov and Kumar Shridhar and Raghav Singhal and Anna Sotnikova and Alexander Sternfeld and Ayush Kumar Tarun and Paul Teiletche and Jannis Vamvas and Xiaozhe Yao and Hao Zhao and Alexander Ilic and Ana Klimovic and Andreas Krause and Caglar Gulcehre and David Rosenthal and Elliott Ash and Florian Tramèr and Joost VandeVondele and Livio Veraldi and Martin Rajman and Thomas Schulthess and Torsten Hoefler and Antoine Bosselut and Martin Jaggi and Imanol Schlag},
      year={2025},
      eprint={2509.14233},
      archivePrefix={arXiv},
      primaryClass={cs.CL},
      url={https://arxiv.org/abs/2509.14233}, 
}

@article{correa2024teenytinyllama,
  title={\textbf{TeenyTinyLlama: open-source tiny language models trained in Brazilian Portuguese}},
  author={Corr{\^e}a, Nicholas Kluge and Falk, Sophia and Fatimah, Shiza and Sen, Aniket and De Oliveira, Nythamar},
  journal={Machine Learning with Applications},
  volume={16},
  pages={100558},
  year={2024},
  doi={10.1016/j.mlwa.2024.100558},
  publisher={Elsevier}
}

@article{hoffmann2022training,
  title={Training compute-optimal large language models},
  author={Hoffmann, Jordan and Borgeaud, Sebastian and Mensch, Arthur and Buchatskaya, Elena and Cai, Trevor and Rutherford, Eliza and Casas, Diego de Las and Hendricks, Lisa Anne and Welbl, Johannes and Clark, Aidan and others},
  journal={arXiv preprint arXiv:2203.15556},
  year={2022}
}

@article{correa2025tucanoadvancingneuraltext,
    title={{Tucano: Advancing Neural Text Generation for Portuguese}},
    author={Corr{\^e}a, Nicholas Kluge and Sen, Aniket and Falk, Sophia and Fatimah, Shiza},
    journal={Patterns},
    publisher={Elsevier},
    year={2025},
    doi={10.1016/j.patter.2025.101325},
    url={https://doi.org/10.1016/j.patter.2025.101325},
    issn={2666-3899}
}

@article{workshop2022bloom,
  title={Bloom: A 176b-parameter open-access multilingual language model},
  author={Workshop, BigScience and Scao, Teven Le and Fan, Angela and Akiki, Christopher and Pavlick, Ellie and Ili{\'c}, Suzana and Hesslow, Daniel and Castagn{\'e}, Roman and Luccioni, Alexandra Sasha and Yvon, Fran{\c{c}}ois and others},
  journal={arXiv preprint arXiv:2211.05100},
  year={2022}
}

@article{bai2023qwen,
  title={Qwen technical report},
  author={Bai, Jinze and Bai, Shuai and Chu, Yunfei and Cui, Zeyu and Dang, Kai and Deng, Xiaodong and Fan, Yang and Ge, Wenbin and Han, Yu and Huang, Fei and others},
  journal={arXiv preprint arXiv:2309.16609},
  year={2023}
}

@inproceedings{biderman2023pythia,
  title={Pythia: A suite for analyzing large language models across training and scaling},
  author={Biderman, Stella and Schoelkopf, Hailey and Anthony, Quentin Gregory and Bradley, Herbie and O’Brien, Kyle and Hallahan, Eric and Khan, Mohammad Aflah and Purohit, Shivanshu and Prashanth, USVSN Sai and Raff, Edward and others},
  booktitle={International Conference on Machine Learning},
  pages={2397--2430},
  year={2023},
  organization={PMLR}
}

@article{touvron2023llama,
  title={Llama 2: Open foundation and fine-tuned chat models},
  author={Touvron, Hugo and Martin, Louis and Stone, Kevin and Albert, Peter and Almahairi, Amjad and Babaei, Yasmine and Bashlykov, Nikolay and Batra, Soumya and Bhargava, Prajjwal and Bhosale, Shruti and others},
  journal={arXiv preprint arXiv:2307.09288},
  year={2023}
}

@inproceedings{silveira2017university,
  title={University entrance exam as a guiding test for artificial intelligence},
  author={Silveira, Igor Cataneo and Mau{\'a}, Denis Deratani},
  booktitle={2017 Brazilian Conference on Intelligent Systems (BRACIS)},
  pages={426--431},
  year={2017},
  organization={IEEE}
}

@inproceedings{d2017passing,
  title={Passing the Brazilian OAB Exam: Data Preparation and Some Experiments1},
  author={d RADEMAKER, Alexandre},
  booktitle={Legal Knowledge and Information Systems: JURIX 2017: The Thirtieth Annual Conference},
  volume={302},
  pages={89},
  year={2017},
  organization={IOS Press}
}

@misc{almeida2025buildinghighqualitydatasetsportuguese,
      title={Building High-Quality Datasets for Portuguese LLMs: From Common Crawl Snapshots to Industrial-Grade Corpora}, 
      author={Thales Sales Almeida and Rodrigo Nogueira and Helio Pedrini},
      year={2025},
      eprint={2509.08824},
      archivePrefix={arXiv},
      primaryClass={cs.CL},
      url={https://arxiv.org/abs/2509.08824}, 
}

@misc{almeida2025curioedu7bexaminingdata,
      title={Curi\'o-Edu 7B: Examining Data Selection Impacts in LLM Continued Pretraining}, 
      author={Thales Sales Almeida and Rodrigo Nogueira and Hélio Pedrini},
      year={2025},
      eprint={2512.12770},
      archivePrefix={arXiv},
      primaryClass={cs.CL},
      url={https://arxiv.org/abs/2512.12770}, 
}

@inproceedings{gamallo2024galician,
  title={A galician-portuguese generative model},
  author={Gamallo, Pablo and Rodr{\'\i}guez, Pablo and Santos, Daniel and Sotelo, Susana and Miquelina, Nuno and Paniagua, Silvia and Schmidt, Daniela and de-Dios-Flores, Iria and Quaresma, Paulo and Bardanca, Daniel and others},
  booktitle={EPIA Conference on Artificial Intelligence},
  pages={292--304},
  year={2024},
  organization={Springer}
}

@misc{dey2023cerebrasgptopencomputeoptimallanguage,
      title={Cerebras-GPT: Open Compute-Optimal Language Models Trained on the Cerebras Wafer-Scale Cluster}, 
      author={Nolan Dey and Gurpreet Gosal and Zhiming and Chen and Hemant Khachane and William Marshall and Ribhu Pathria and Marvin Tom and Joel Hestness},
      year={2023},
      eprint={2304.03208},
      archivePrefix={arXiv},
      primaryClass={cs.LG},
      url={https://arxiv.org/abs/2304.03208}, 
}

@misc{qwen2025qwen25technicalreport,
      title={Qwen2.5 Technical Report}, 
      author={An Yang and Baosong Yang and Beichen Zhang and Binyuan Hui and Bo Zheng and Bowen Yu and Chengyuan Li and Dayiheng Liu and Fei Huang and Haoran Wei and Huan Lin and Jian Yang and Jianhong Tu and Jianwei Zhang and Jianxin Yang and Jiaxi Yang and Jingren Zhou and Junyang Lin and Kai Dang and Keming Lu and Keqin Bao and Kexin Yang and Le Yu and Mei Li and Mingfeng Xue and Pei Zhang and Qin Zhu and Rui Men and Runji Lin and Tianhao Li and Tianyi Tang and Tingyu Xia and Xingzhang Ren and Xuancheng Ren and Yang Fan and Yang Su and Yichang Zhang and Yu Wan and Yuqiong Liu and Zeyu Cui and Zhenru Zhang and Zihan Qiu},
      year={2025},
      eprint={2412.15115},
      archivePrefix={arXiv},
      primaryClass={cs.CL},
      url={https://arxiv.org/abs/2412.15115}, 
}

@article{yang2025qwen3,
  title={Qwen3 technical report},
  author={Yang, An and Li, Anfeng and Yang, Baosong and Zhang, Beichen and Hui, Binyuan and Zheng, Bo and Yu, Bowen and Gao, Chang and Huang, Chengen and Lv, Chenxu and others},
  journal={arXiv preprint arXiv:2505.09388},
  year={2025}
}

@misc{gemmateam2025gemma3technicalreport,
      title={Gemma 3 Technical Report}, 
      author={Aishwarya Kamath and Johan Ferret and Shreya Pathak and Nino Vieillard and Ramona Merhej and Sarah Perrin and Tatiana Matejovicova and Alexandre Ramé and Morgane Rivière and Louis Rouillard and Thomas Mesnard and Geoffrey Cideron and Jean-bastien Grill and Sabela Ramos and Edouard Yvinec and Michelle Casbon and Etienne Pot and Ivo Penchev and Gaël Liu and Francesco Visin and Kathleen Kenealy and Lucas Beyer and Xiaohai Zhai and Anton Tsitsulin and Robert Busa-Fekete and Alex Feng and Noveen Sachdeva and Benjamin Coleman and Yi Gao and Basil Mustafa and Iain Barr and Emilio Parisotto and David Tian and Matan Eyal and Colin Cherry and Jan-Thorsten Peter and Danila Sinopalnikov and Surya Bhupatiraju and Rishabh Agarwal and Mehran Kazemi and Dan Malkin and Ravin Kumar and David Vilar and Idan Brusilovsky and Jiaming Luo and Andreas Steiner and Abe Friesen and Abhanshu Sharma and Abheesht Sharma and Adi Mayrav Gilady and Adrian Goedeckemeyer and Alaa Saade and Alex Feng and Alexander Kolesnikov and Alexei Bendebury and Alvin Abdagic and Amit Vadi and András György and André Susano Pinto and Anil Das and Ankur Bapna and Antoine Miech and Antoine Yang and Antonia Paterson and Ashish Shenoy and Ayan Chakrabarti and Bilal Piot and Bo Wu and Bobak Shahriari and Bryce Petrini and Charlie Chen and Charline Le Lan and Christopher A. Choquette-Choo and CJ Carey and Cormac Brick and Daniel Deutsch and Danielle Eisenbud and Dee Cattle and Derek Cheng and Dimitris Paparas and Divyashree Shivakumar Sreepathihalli and Doug Reid and Dustin Tran and Dustin Zelle and Eric Noland and Erwin Huizenga and Eugene Kharitonov and Frederick Liu and Gagik Amirkhanyan and Glenn Cameron and Hadi Hashemi and Hanna Klimczak-Plucińska and Harman Singh and Harsh Mehta and Harshal Tushar Lehri and Hussein Hazimeh and Ian Ballantyne and Idan Szpektor and Ivan Nardini and Jean Pouget-Abadie and Jetha Chan and Joe Stanton and John Wieting and Jonathan Lai and Jordi Orbay and Joseph Fernandez and Josh Newlan and Ju-yeong Ji and Jyotinder Singh and Kat Black and Kathy Yu and Kevin Hui and Kiran Vodrahalli and Klaus Greff and Linhai Qiu and Marcella Valentine and Marina Coelho and Marvin Ritter and Matt Hoffman and Matthew Watson and Mayank Chaturvedi and Michael Moynihan and Min Ma and Nabila Babar and Natasha Noy and Nathan Byrd and Nick Roy and Nikola Momchev and Nilay Chauhan and Noveen Sachdeva and Oskar Bunyan and Pankil Botarda and Paul Caron and Paul Kishan Rubenstein and Phil Culliton and Philipp Schmid and Pier Giuseppe Sessa and Pingmei Xu and Piotr Stanczyk and Pouya Tafti and Rakesh Shivanna and Renjie Wu and Renke Pan and Reza Rokni and Rob Willoughby and Rohith Vallu and Ryan Mullins and Sammy Jerome and Sara Smoot and Sertan Girgin and Shariq Iqbal and Shashir Reddy and Shruti Sheth and Siim Põder and Sijal Bhatnagar and Sindhu Raghuram Panyam and Sivan Eiger and Susan Zhang and Tianqi Liu and Trevor Yacovone and Tyler Liechty and Uday Kalra and Utku Evci and Vedant Misra and Vincent Roseberry and Vlad Feinberg and Vlad Kolesnikov and Woohyun Han and Woosuk Kwon and Xi Chen and Yinlam Chow and Yuvein Zhu and Zichuan Wei and Zoltan Egyed and Victor Cotruta and Minh Giang and Phoebe Kirk and Anand Rao and Kat Black and Nabila Babar and Jessica Lo and Erica Moreira and Luiz Gustavo Martins and Omar Sanseviero and Lucas Gonzalez and Zach Gleicher and Tris Warkentin and Vahab Mirrokni and Evan Senter and Eli Collins and Joelle Barral and Zoubin Ghahramani and Raia Hadsell and Yossi Matias and D. Sculley and Slav Petrov and Noah Fiedel and Noam Shazeer and Oriol Vinyals and Jeff Dean and Demis Hassabis and Koray Kavukcuoglu and Clement Farabet and Elena Buchatskaya and Jean-Baptiste Alayrac and Rohan Anil and Dmitry and Lepikhin and Sebastian Borgeaud and Olivier Bachem and Armand Joulin and Alek Andreev and Cassidy Hardin and Robert Dadashi and Léonard Hussenot},
      year={2025},
      eprint={2503.19786},
      archivePrefix={arXiv},
      primaryClass={cs.CL},
      url={https://arxiv.org/abs/2503.19786}, 
}

@misc{Falcon3,
    title = {The Falcon 3 Family of Open Models},
    url = {https://huggingface.co/blog/falcon3},
    author = {Falcon-LLM Team},
    month = {December},
    year = {2024}
}

@misc{penedo2024datatrove,
  author = {Penedo, Guilherme and Kydlíček, Hynek and Cappelli, Alessandro and Sasko, Mario and Wolf, Thomas},
  title = {DataTrove: large scale data processing},
  year = {2024},
  publisher = {GitHub},
  journal = {GitHub repository},
  url = {https://github.com/huggingface/datatrove}
}

@inproceedings{barbaresi-2021-trafilatura,
    title = "Trafilatura: {A} Web Scraping Library and Command-Line Tool for Text Discovery and Extraction",
    author = "Barbaresi, Adrien",
    editor = "Ji, Heng  and
      Park, Jong C.  and
      Xia, Rui",
    booktitle = "Proceedings of the 59th Annual Meeting of the Association for Computational Linguistics and the 11th International Joint Conference on Natural Language Processing: System Demonstrations",
    month = aug,
    year = "2021",
    address = "Online",
    publisher = "Association for Computational Linguistics",
    url = "https://aclanthology.org/2021.acl-demo.15/",
    doi = "10.18653/v1/2021.acl-demo.15",
    pages = "122--131"
}

@misc{joulin2016fasttextzipcompressingtextclassification,
      title={FastText.zip: Compressing text classification models}, 
      author={Armand Joulin and Edouard Grave and Piotr Bojanowski and Matthijs Douze and Hérve Jégou and Tomas Mikolov},
      year={2016},
      eprint={1612.03651},
      archivePrefix={arXiv},
      primaryClass={cs.CL},
      url={https://arxiv.org/abs/1612.03651}, 
}

@inproceedings{Kargaran_2023,
   title={GlotLID: Language Identification for Low-Resource Languages},
   url={http://dx.doi.org/10.18653/v1/2023.findings-emnlp.410},
   DOI={10.18653/v1/2023.findings-emnlp.410},
   booktitle={Findings of the Association for Computational Linguistics: EMNLP 2023},
   publisher={Association for Computational Linguistics},
   author={Kargaran, Amir and Imani, Ayyoob and Yvon, François and Schuetze, Hinrich},
   year={2023},
   pages={6155–6218}
}

@misc{rae2022scalinglanguagemodelsmethods,
      title={Scaling Language Models: Methods, Analysis \& Insights from Training Gopher}, 
      author={Jack W. Rae and Sebastian Borgeaud and Trevor Cai and Katie Millican and Jordan Hoffmann and Francis Song and John Aslanides and Sarah Henderson and Roman Ring and Susannah Young and Eliza Rutherford and Tom Hennigan and Jacob Menick and Albin Cassirer and Richard Powell and George van den Driessche and Lisa Anne Hendricks and Maribeth Rauh and Po-Sen Huang and Amelia Glaese and Johannes Welbl and Sumanth Dathathri and Saffron Huang and Jonathan Uesato and John Mellor and Irina Higgins and Antonia Creswell and Nat McAleese and Amy Wu and Erich Elsen and Siddhant Jayakumar and Elena Buchatskaya and David Budden and Esme Sutherland and Karen Simonyan and Michela Paganini and Laurent Sifre and Lena Martens and Xiang Lorraine Li and Adhiguna Kuncoro and Aida Nematzadeh and Elena Gribovskaya and Domenic Donato and Angeliki Lazaridou and Arthur Mensch and Jean-Baptiste Lespiau and Maria Tsimpoukelli and Nikolai Grigorev and Doug Fritz and Thibault Sottiaux and Mantas Pajarskas and Toby Pohlen and Zhitao Gong and Daniel Toyama and Cyprien de Masson d'Autume and Yujia Li and Tayfun Terzi and Vladimir Mikulik and Igor Babuschkin and Aidan Clark and Diego de Las Casas and Aurelia Guy and Chris Jones and James Bradbury and Matthew Johnson and Blake Hechtman and Laura Weidinger and Iason Gabriel and William Isaac and Ed Lockhart and Simon Osindero and Laura Rimell and Chris Dyer and Oriol Vinyals and Kareem Ayoub and Jeff Stanway and Lorrayne Bennett and Demis Hassabis and Koray Kavukcuoglu and Geoffrey Irving},
      year={2022},
      eprint={2112.11446},
      archivePrefix={arXiv},
      primaryClass={cs.CL},
      url={https://arxiv.org/abs/2112.11446}, 
}

@misc{lee2022deduplicatingtrainingdatamakes,
      title={Deduplicating Training Data Makes Language Models Better}, 
      author={Katherine Lee and Daphne Ippolito and Andrew Nystrom and Chiyuan Zhang and Douglas Eck and Chris Callison-Burch and Nicholas Carlini},
      year={2022},
      eprint={2107.06499},
      archivePrefix={arXiv},
      primaryClass={cs.CL},
      url={https://arxiv.org/abs/2107.06499}, 
}

@misc{gunasekar2023textbooksneed,
      title={Textbooks Are All You Need}, 
      author={Suriya Gunasekar and Yi Zhang and Jyoti Aneja and Caio César Teodoro Mendes and Allie Del Giorno and Sivakanth Gopi and Mojan Javaheripi and Piero Kauffmann and Gustavo de Rosa and Olli Saarikivi and Adil Salim and Shital Shah and Harkirat Singh Behl and Xin Wang and Sébastien Bubeck and Ronen Eldan and Adam Tauman Kalai and Yin Tat Lee and Yuanzhi Li},
      year={2023},
      eprint={2306.11644},
      archivePrefix={arXiv},
      primaryClass={cs.CL},
      url={https://arxiv.org/abs/2306.11644}, 
}

@article{maini2025beyondweb,
  title={Beyondweb: Lessons from scaling synthetic data for trillion-scale pretraining},
  author={Maini, Pratyush and Dorna, Vineeth and Doshi, Parth and Carranza, Aldo and Pan, Fan and Urbanek, Jack and Burstein, Paul and Fang, Alex and Deng, Alvin and Abbas, Amro and others},
  journal={arXiv preprint arXiv:2508.10975},
  year={2025}
}

@article{allal2024cosmopedia,
  title={Cosmopedia: how to create large-scale synthetic data for pre-training},
  author={Allal, Loubna Ben and Lozhkov, Anton and van Strien, Daniel},
  journal={Hugging Face Blog},
  pages={56},
  year={2024}
}

@inproceedings{garcia2024crawlptlargescalecleanedportugueseweb,
    title = "{R}o{BERT}a{L}ex{PT}: A Legal {R}o{BERT}a Model pretrained with deduplication for {P}ortuguese",
    author = "Garcia, Eduardo A. S.  and
      Silva, Nadia F. F.  and
      Siqueira, Felipe  and
      Albuquerque, Hidelberg O.  and
      Gomes, Juliana R. S.  and
      Souza, Ellen  and
      Lima, Eliomar A.",
    editor = "Gamallo, Pablo  and
      Claro, Daniela  and
      Teixeira, Ant{\'o}nio  and
      Real, Livy  and
      Garcia, Marcos  and
      Oliveira, Hugo Gon{\c{c}}alo  and
      Amaro, Raquel",
    booktitle = "Proceedings of the 16th International Conference on Computational Processing of Portuguese",
    month = mar,
    year = "2024",
    address = "Santiago de Compostela, Galicia/Spain",
    publisher = "Association for Computational Lingustics",
    url = "https://aclanthology.org/2024.propor-1.38",
    pages = "374--383",
}

@misc{penedo2024finewebdatasetsdecantingweb,
      title={The FineWeb Datasets: Decanting the Web for the Finest Text Data at Scale}, 
      author={Guilherme Penedo and Hynek Kydlíček and Loubna Ben allal and Anton Lozhkov and Margaret Mitchell and Colin Raffel and Leandro Von Werra and Thomas Wolf},
      year={2024},
      eprint={2406.17557},
      archivePrefix={arXiv},
      primaryClass={cs.CL},
      url={https://arxiv.org/abs/2406.17557}, 
}

@misc{penedo2025fineweb2pipelinescale,
      title={FineWeb2: One Pipeline to Scale Them All -- Adapting Pre-Training Data Processing to Every Language}, 
      author={Guilherme Penedo and Hynek Kydlíček and Vinko Sabolčec and Bettina Messmer and Negar Foroutan and Amir Hossein Kargaran and Colin Raffel and Martin Jaggi and Leandro Von Werra and Thomas Wolf},
      year={2025},
      eprint={2506.20920},
      archivePrefix={arXiv},
      primaryClass={cs.CL},
      url={https://arxiv.org/abs/2506.20920}, 
}

@inproceedings{souza2020bertimbau,
  author    = {F{\'a}bio Souza and
               Rodrigo Nogueira and
               Roberto Lotufo},
  title     = {{BERT}imbau: pretrained {BERT} models for {B}razilian {P}ortuguese},
  booktitle = {9th Brazilian Conference on Intelligent Systems, {BRACIS}, Rio Grande do Sul, Brazil, October 20-23 (to appear)},
  year      = {2020}
}

@inproceedings{costa-etal-2023-bertabaporu,
    title = "{BERT}abaporu: Assessing a Genre-Specific Language Model for {P}ortuguese {NLP}",
    author = "Costa, Pablo Botton  and
      Pavan, Matheus Camasmie  and
      Santos, Wesley Ramos  and
      Silva, Samuel Caetano  and
      Paraboni, Ivandr{\'e}",
    editor = "Mitkov, Ruslan  and
      Angelova, Galia",
    booktitle = "Proceedings of the 14th International Conference on Recent Advances in Natural Language Processing",
    month = sep,
    year = "2023",
    address = "Varna, Bulgaria",
    publisher = "INCOMA Ltd., Shoumen, Bulgaria",
    url = "https://aclanthology.org/2023.ranlp-1.24/",
    pages = "217--223"
}

@misc{campiotti2023debertinhamultistepapproachadapt,
      title={DeBERTinha: A Multistep Approach to Adapt DebertaV3 XSmall for Brazilian Portuguese Natural Language Processing Task}, 
      author={Israel Campiotti and Matheus Rodrigues and Yuri Albuquerque and Rafael Azevedo and Alyson Andrade},
      year={2023},
      eprint={2309.16844},
      archivePrefix={arXiv},
      primaryClass={cs.CL},
      url={https://arxiv.org/abs/2309.16844}, 
}

@inproceedings{kwon2023efficient,
  title={Efficient Memory Management for Large Language Model Serving with PagedAttention},
  author={Woosuk Kwon and Zhuohan Li and Siyuan Zhuang and Ying Sheng and Lianmin Zheng and Cody Hao Yu and Joseph E. Gonzalez and Hao Zhang and Ion Stoica},
  booktitle={Proceedings of the ACM SIGOPS 29th Symposium on Operating Systems Principles},
  year={2023}
}

@inproceedings{wolf-etal-2020-transformers,
    title = "Transformers: State-of-the-Art Natural Language Processing",
    author = "Thomas Wolf and Lysandre Debut and Victor Sanh and Julien Chaumond and Clement Delangue and Anthony Moi and Pierric Cistac and Tim Rault and Rémi Louf and Morgan Funtowicz and Joe Davison and Sam Shleifer and Patrick von Platen and Clara Ma and Yacine Jernite and Julien Plu and Canwen Xu and Teven Le Scao and Sylvain Gugger and Mariama Drame and Quentin Lhoest and Alexander M. Rush",
    booktitle = "Proceedings of the 2020 Conference on Empirical Methods in Natural Language Processing: System Demonstrations",
    month = oct,
    year = "2020",
    address = "Online",
    publisher = "Association for Computational Linguistics",
    url = "https://www.aclweb.org/anthology/2020.emnlp-demos.6",
    pages = "38--45"
}

@misc{loshchilov2019decoupledweightdecayregularization,
      title={Decoupled Weight Decay Regularization}, 
      author={Ilya Loshchilov and Frank Hutter},
      year={2019},
      eprint={1711.05101},
      archivePrefix={arXiv},
      primaryClass={cs.LG},
      url={https://arxiv.org/abs/1711.05101}, 
}

@misc{kang2025demystifyingsyntheticdatallm,
      title={Demystifying Synthetic Data in LLM Pre-training: A Systematic Study of Scaling Laws, Benefits, and Pitfalls}, 
      author={Feiyang Kang and Newsha Ardalani and Michael Kuchnik and Youssef Emad and Mostafa Elhoushi and Shubhabrata Sengupta and Shang-Wen Li and Ramya Raghavendra and Ruoxi Jia and Carole-Jean Wu},
      year={2025},
      eprint={2510.01631},
      archivePrefix={arXiv},
      primaryClass={cs.LG},
      url={https://arxiv.org/abs/2510.01631}, 
}

@misc{yang2024syntheticcontinuedpretraining,
      title={Synthetic continued pretraining}, 
      author={Zitong Yang and Neil Band and Shuangping Li and Emmanuel Candès and Tatsunori Hashimoto},
      year={2024},
      eprint={2409.07431},
      archivePrefix={arXiv},
      primaryClass={cs.LG},
      url={https://arxiv.org/abs/2409.07431}, 
}

@misc{muennighoff2025s1simpletesttimescaling,
      title={s1: Simple test-time scaling}, 
      author={Niklas Muennighoff and Zitong Yang and Weijia Shi and Xiang Lisa Li and Li Fei-Fei and Hannaneh Hajishirzi and Luke Zettlemoyer and Percy Liang and Emmanuel Candès and Tatsunori Hashimoto},
      year={2025},
      eprint={2501.19393},
      archivePrefix={arXiv},
      primaryClass={cs.CL},
      url={https://arxiv.org/abs/2501.19393}, 
}

@misc{clark2018thinksolvedquestionanswering,
      title={Think you have Solved Question Answering? Try ARC, the AI2 Reasoning Challenge}, 
      author={Peter Clark and Isaac Cowhey and Oren Etzioni and Tushar Khot and Ashish Sabharwal and Carissa Schoenick and Oyvind Tafjord},
      year={2018},
      eprint={1803.05457},
      archivePrefix={arXiv},
      primaryClass={cs.AI},
      url={https://arxiv.org/abs/1803.05457}, 
}

@misc{lopes2024gloriagenerativeopen,
      title={Gl\'orIA -- A Generative and Open Large Language Model for Portuguese}, 
      author={Ricardo Lopes and João Magalhães and David Semedo},
      year={2024},
      eprint={2402.12969},
      archivePrefix={arXiv},
      primaryClass={cs.CL},
      url={https://arxiv.org/abs/2402.12969}, 
}

@misc{chang2025globalpiqaevaluatingphysical,
      title={Global PIQA: Evaluating Physical Commonsense Reasoning Across 100+ Languages and Cultures}, 
      author={Tyler A. Chang and Catherine Arnett and Abdelrahman Eldesokey and Abdelrahman Sadallah and Abeer Kashar and Abolade Daud and Abosede Grace Olanihun and Adamu Labaran Mohammed and Adeyemi Praise and Adhikarinayum Meerajita Sharma and collaborators},
      year={2025},
      eprint={2510.24081},
      archivePrefix={arXiv},
      primaryClass={cs.CL},
      url={https://arxiv.org/abs/2510.24081}, 
}

@misc{zellers2019hellaswagmachinereallyfinish,
      title={HellaSwag: Can a Machine Really Finish Your Sentence?}, 
      author={Rowan Zellers and Ari Holtzman and Yonatan Bisk and Ali Farhadi and Yejin Choi},
      year={2019},
      eprint={1905.07830},
      archivePrefix={arXiv},
      primaryClass={cs.CL},
      url={https://arxiv.org/abs/1905.07830}, 
}

@misc{lambada2016,
    author={Paperno, Denis and Kruszewski, Germán and Lazaridou, Angeliki and Pham, Quan Ngoc and Bernardi, Raffaella and Pezzelle, Sandro and Baroni, Marco and Boleda, Gemma and Fernández, Raquel},
    title={The LAMBADA dataset},
    DOI={10.5281/zenodo.2630551},
    publisher={Zenodo},
    year={2016},
    month={Aug}
}

@inbook{Pires_2023,
   title={Sabiá: Portuguese Large Language Models},
   ISBN={9783031453922},
   ISSN={1611-3349},
   url={http://dx.doi.org/10.1007/978-3-031-45392-2_15},
   DOI={10.1007/978-3-031-45392-2_15},
   booktitle={Intelligent Systems},
   publisher={Springer Nature Switzerland},
   author={Pires, Ramon and Abonizio, Hugo and Almeida, Thales Sales and Nogueira, Rodrigo},
   year={2023},
   pages={226–240} }

@inproceedings{fonseca2016assin,
  title={ASSIN: Avaliacao de similaridade semantica e inferencia textual},
  author={Fonseca, E and Santos, L and Criscuolo, Marcelo and Aluisio, S},
  booktitle={Computational Processing of the Portuguese Language-12th International Conference, Tomar, Portugal},
  pages={13--15},
  year={2016}
}

@article{hendryckstest2021,
  title={Measuring Massive Multitask Language Understanding},
  author={Dan Hendrycks and Collin Burns and Steven Basart and Andy Zou and Mantas Mazeika and Dawn Song and Jacob Steinhardt},
  journal={Proceedings of the International Conference on Learning Representations (ICLR)},
  year={2021}
}

@inproceedings{bandarkar-etal-2024-belebele,
    title = "The Belebele Benchmark: a Parallel Reading Comprehension Dataset in 122 Language Variants",
    author = "Bandarkar, Lucas  and
      Liang, Davis  and
      Muller, Benjamin  and
      Artetxe, Mikel  and
      Shukla, Satya Narayan  and
      Husa, Donald  and
      Goyal, Naman  and
      Krishnan, Abhinandan  and
      Zettlemoyer, Luke  and
      Khabsa, Madian",
    booktitle = "Proceedings of the 62nd Annual Meeting of the Association for Computational Linguistics (Volume 1: Long Papers)",
    month = aug,
    year = "2024",
    address = "Bangkok, Thailand and virtual meeting",
    publisher = "Association for Computational Linguistics",
    url = "https://aclanthology.org/2024.acl-long.44",
    pages = "749--775",
}

@inproceedings{kudo2018sentencepiece,
  title={SentencePiece: A simple and language independent subword tokenizer and detokenizer for neural text processing},
  author={Kudo, Taku and Richardson, John},
  booktitle={Proceedings of the 2018 conference on empirical methods in natural language processing: System demonstrations},
  pages={66--71},
  year={2018}
}

@misc{rust2021goodtokenizermonolingualperformance,
      title={How Good is Your Tokenizer? On the Monolingual Performance of Multilingual Language Models}, 
      author={Phillip Rust and Jonas Pfeiffer and Ivan Vulić and Sebastian Ruder and Iryna Gurevych},
      year={2021},
      eprint={2012.15613},
      archivePrefix={arXiv},
      primaryClass={cs.CL},
      url={https://arxiv.org/abs/2012.15613}, 
}

@inproceedings{ali2024tokenizer,
  title={Tokenizer choice for llm training: Negligible or crucial?},
  author={Ali, Mehdi and Fromm, Michael and Thellmann, Klaudia and Rutmann, Richard and L{\"u}bbering, Max and Leveling, Johannes and Klug, Katrin and Ebert, Jan and Doll, Niclas and Buschhoff, Jasper and others},
  booktitle={Findings of the Association for Computational Linguistics: NAACL 2024},
  pages={3907--3924},
  year={2024}
}

@article{larcher2023cabrita,
  title={Cabrita: closing the gap for foreign languages},
  author={Larcher, Celio and Piau, Marcos and Finardi, Paulo and Gengo, Pedro and Esposito, Piero and Carid{\'a}, Vinicius},
  journal={Preprint at arXiv. \url{https://doi.org/10.48550/arXiv.2308.11878}},
  year={2023}
}

@article{finardi2021berta,
  title={BERTa$\backslash$'u: Ita$\backslash$'u BERT for digital customer service},
  author={Finardi, Paulo and Viegas, Jos{\'e} Di{\'e} and Ferreira, Gustavo T and Mansano, Alex F and Carid{\'a}, Vinicius F},
  journal={Preprint at arXiv. \url{https://doi.org/10.48550/arXiv.2101.12015}},
  year={2021}
}

@misc{li2023starcodersourceyou,
      title={StarCoder: may the source be with you!}, 
      author={Raymond Li and Loubna Ben Allal and Yangtian Zi and Niklas Muennighoff and Denis Kocetkov and Chenghao Mou and Marc Marone and Christopher Akiki and Jia Li and Jenny Chim and Qian Liu and Evgenii Zheltonozhskii and Terry Yue Zhuo and Thomas Wang and Olivier Dehaene and Mishig Davaadorj and Joel Lamy-Poirier and João Monteiro and Oleh Shliazhko and Nicolas Gontier and Nicholas Meade and Armel Zebaze and Ming-Ho Yee and Logesh Kumar Umapathi and Jian Zhu and Benjamin Lipkin and Muhtasham Oblokulov and Zhiruo Wang and Rudra Murthy and Jason Stillerman and Siva Sankalp Patel and Dmitry Abulkhanov and Marco Zocca and Manan Dey and Zhihan Zhang and Nour Fahmy and Urvashi Bhattacharyya and Wenhao Yu and Swayam Singh and Sasha Luccioni and Paulo Villegas and Maxim Kunakov and Fedor Zhdanov and Manuel Romero and Tony Lee and Nadav Timor and Jennifer Ding and Claire Schlesinger and Hailey Schoelkopf and Jan Ebert and Tri Dao and Mayank Mishra and Alex Gu and Jennifer Robinson and Carolyn Jane Anderson and Brendan Dolan-Gavitt and Danish Contractor and Siva Reddy and Daniel Fried and Dzmitry Bahdanau and Yacine Jernite and Carlos Muñoz Ferrandis and Sean Hughes and Thomas Wolf and Arjun Guha and Leandro von Werra and Harm de Vries},
      year={2023},
      eprint={2305.06161},
      archivePrefix={arXiv},
      primaryClass={cs.CL},
      url={https://arxiv.org/abs/2305.06161}, 
}

@misc{open-pt-llm-leaderboard,
  author = {Garcia, Eduardo A. S.},
  title = {Open Portuguese LLM Leaderboard},
  year = {2024},
  publisher = {Hugging Face},
  howpublished = "\url{https://huggingface.co/spaces/eduagarcia/open_pt_llm_leaderboard}"
}

@misc{fourrier2025_the_llm_evaluation_guidebook,
  title={The LLM Evaluation Guidebook},
  author={Clémentine Fourrier and Thibaud Frere and Guilherme Penedo and Thomas Wolf},
  year={2025},
  
}

@misc{eval-harness,
  author       = {Gao, Leo and Tow, Jonathan and Abbasi, Baber and Biderman, Stella and Black, Sid and DiPofi, Anthony and Foster, Charles and Golding, Laurence and Hsu, Jeffrey and Le Noac'h, Alain and Li, Haonan and McDonell, Kyle and Muennighoff, Niklas and Ociepa, Chris and Phang, Jason and Reynolds, Laria and Schoelkopf, Hailey and Skowron, Aviya and Sutawika, Lintang and Tang, Eric and Thite, Anish and Wang, Ben and Wang, Kevin and Zou, Andy},
  title        = {The Language Model Evaluation Harness},
  month        = 07,
  year         = 2024,
  publisher    = {Zenodo},
  version      = {v0.4.3},
  doi          = {10.5281/zenodo.12608602},
  url          = {https://zenodo.org/records/12608602}
}

@misc{zhou2023instructionfollowingevaluationlargelanguage,
      title={Instruction-Following Evaluation for Large Language Models}, 
      author={Jeffrey Zhou and Tianjian Lu and Swaroop Mishra and Siddhartha Brahma and Sujoy Basu and Yi Luan and Denny Zhou and Le Hou},
      year={2023},
      eprint={2311.07911},
      archivePrefix={arXiv},
      primaryClass={cs.CL},
      url={https://arxiv.org/abs/2311.07911}, 
}

@misc{cobbe2021trainingverifierssolvemath,
      title={Training Verifiers to Solve Math Word Problems}, 
      author={Karl Cobbe and Vineet Kosaraju and Mohammad Bavarian and Mark Chen and Heewoo Jun and Lukasz Kaiser and Matthias Plappert and Jerry Tworek and Jacob Hilton and Reiichiro Nakano and Christopher Hesse and John Schulman},
      year={2021},
      eprint={2110.14168},
      archivePrefix={arXiv},
      primaryClass={cs.LG},
      url={https://arxiv.org/abs/2110.14168}, 
}

@misc{hsieh2024rulerwhatsrealcontext,
      title={RULER: What's the Real Context Size of Your Long-Context Language Models?}, 
      author={Cheng-Ping Hsieh and Simeng Sun and Samuel Kriman and Shantanu Acharya and Dima Rekesh and Fei Jia and Yang Zhang and Boris Ginsburg},
      year={2024},
      eprint={2404.06654},
      archivePrefix={arXiv},
      primaryClass={cs.CL},
      url={https://arxiv.org/abs/2404.06654}, 
}

@misc{kim2025rulermeasureallbenchmarking,
      title={One ruler to measure them all: Benchmarking multilingual long-context language models}, 
      author={Yekyung Kim and Jenna Russell and Marzena Karpinska and Mohit Iyyer},
      year={2025},
      eprint={2503.01996},
      archivePrefix={arXiv},
      primaryClass={cs.CL},
      url={https://arxiv.org/abs/2503.01996}, 
}

@article{paszke2019pytorch,
  title={Pytorch: An imperative style, high-performance deep learning library},
  author={Paszke, Adam and Gross, Sam and Massa, Francisco and Lerer, Adam and Bradbury, James and Chanan, Gregory and Killeen, Trevor and Lin, Zeming and Gimelshein, Natalia and Antiga, Luca and others},
  journal={Advances in neural information processing systems},
  volume={32},
  year={2019}
}

@inproceedings{valero2023mixed,
  title={Mixed-precision s/dgemm using the tf32 and tf64 frameworks on low-precision ai tensor cores},
  author={Valero-Lara, Pedro and Jorquera, Ian and Lui, Frank and Vetter, Jeffrey},
  booktitle={Proceedings of the SC'23 Workshops of the International Conference on High Performance Computing, Network, Storage, and Analysis},
  pages={179--186},
  year={2023}
}

@misc{dao2023flashattention2fasterattentionbetter,
      title={FlashAttention-2: Faster Attention with Better Parallelism and Work Partitioning}, 
      author={Tri Dao},
      year={2023},
      eprint={2307.08691},
      archivePrefix={arXiv},
      primaryClass={cs.LG},
      url={https://arxiv.org/abs/2307.08691}, 
}

@inproceedings{hsu2025ligerkernel,
    title={Liger-Kernel: Efficient Triton Kernels for {LLM} Training},
    author={Pin-Lun Hsu and Yun Dai and Vignesh Kothapalli and Qingquan Song and Shao Tang and Siyu Zhu and Steven Shimizu and Shivam Sahni and Haowen Ning and Yanning Chen and Zhipeng Wang},
    booktitle={Championing Open-source DEvelopment in ML Workshop @ ICML25},
    year={2025},
    url={https://openreview.net/forum?id=36SjAIT42G}
}

@misc{chen2016trainingdeepnetssublinear,
      title={Training Deep Nets with Sublinear Memory Cost}, 
      author={Tianqi Chen and Bing Xu and Chiyuan Zhang and Carlos Guestrin},
      year={2016},
      eprint={1604.06174},
      archivePrefix={arXiv},
      primaryClass={cs.LG},
      url={https://arxiv.org/abs/1604.06174}, 
}

@misc{zhao2023pytorchfsdpexperiencesscaling,
      title={PyTorch FSDP: Experiences on Scaling Fully Sharded Data Parallel}, 
      author={Yanli Zhao and Andrew Gu and Rohan Varma and Liang Luo and Chien-Chin Huang and Min Xu and Less Wright and Hamid Shojanazeri and Myle Ott and Sam Shleifer and Alban Desmaison and Can Balioglu and Pritam Damania and Bernard Nguyen and Geeta Chauhan and Yuchen Hao and Ajit Mathews and Shen Li},
      year={2023},
      eprint={2304.11277},
      archivePrefix={arXiv},
      primaryClass={cs.DC},
      url={https://arxiv.org/abs/2304.11277}, 
}

@misc{rajbhandari2020zeromemoryoptimizationstraining,
      title={ZeRO: Memory Optimizations Toward Training Trillion Parameter Models}, 
      author={Samyam Rajbhandari and Jeff Rasley and Olatunji Ruwase and Yuxiong He},
      year={2020},
      eprint={1910.02054},
      archivePrefix={arXiv},
      primaryClass={cs.LG},
      url={https://arxiv.org/abs/1910.02054}, 
}

@misc{ainslie2023gqatraininggeneralizedmultiquery,
      title={GQA: Training Generalized Multi-Query Transformer Models from Multi-Head Checkpoints}, 
      author={Joshua Ainslie and James Lee-Thorp and Michiel de Jong and Yury Zemlyanskiy and Federico Lebrón and Sumit Sanghai},
      year={2023},
      eprint={2305.13245},
      archivePrefix={arXiv},
      primaryClass={cs.CL},
      url={https://arxiv.org/abs/2305.13245}, 
}

@misc{li2020pytorchdistributedexperiencesaccelerating,
      title={PyTorch Distributed: Experiences on Accelerating Data Parallel Training}, 
      author={Shen Li and Yanli Zhao and Rohan Varma and Omkar Salpekar and Pieter Noordhuis and Teng Li and Adam Paszke and Jeff Smith and Brian Vaughan and Pritam Damania and Soumith Chintala},
      year={2020},
      eprint={2006.15704},
      archivePrefix={arXiv},
      primaryClass={cs.DC},
      url={https://arxiv.org/abs/2006.15704}, 
}

@misc{vonwerra2020trl,
  title   = {{TRL: Transformers Reinforcement Learning}},
  author  = {von Werra, Leandro and Belkada, Younes and Tunstall, Lewis and Beeching, Edward and Thrush, Tristan and Lambert, Nathan and Huang, Shengyi and Rasul, Kashif and Gallouédec, Quentin},
  license = {Apache-2.0},
  url     = {https://github.com/huggingface/trl},
  year    = {2020}
}

@inproceedings{lhoest-etal-2021-datasets,
    title = "Datasets: A Community Library for Natural Language Processing",
    author = "Lhoest, Quentin  and
      Villanova del Moral, Albert  and
      Jernite, Yacine  and
      Thakur, Abhishek  and
      von Platen, Patrick  and
      Patil, Suraj  and
      Chaumond, Julien  and
      Drame, Mariama  and
      Plu, Julien  and
      Tunstall, Lewis  and
      Davison, Joe  and
      {\v{S}}a{\v{s}}ko, Mario  and
      Chhablani, Gunjan  and
      Malik, Bhavitvya  and
      Brandeis, Simon  and
      Le Scao, Teven  and
      Sanh, Victor  and
      Xu, Canwen  and
      Patry, Nicolas  and
      McMillan-Major, Angelina  and
      Schmid, Philipp  and
      Gugger, Sylvain  and
      Delangue, Cl{\'e}ment  and
      Matussi{\`e}re, Th{\'e}o  and
      Debut, Lysandre  and
      Bekman, Stas  and
      Cistac, Pierric  and
      Goehringer, Thibault  and
      Mustar, Victor  and
      Lagunas, Fran{\c{c}}ois  and
      Rush, Alexander  and
      Wolf, Thomas",
    booktitle = "Proceedings of the 2021 Conference on Empirical Methods in Natural Language Processing: System Demonstrations",
    month = nov,
    year = "2021",
    address = "Online and Punta Cana, Dominican Republic",
    publisher = "Association for Computational Linguistics",
    url = "https://aclanthology.org/2021.emnlp-demo.21",
    pages = "175--184",
    eprint={2109.02846},
    archivePrefix={arXiv},
    primaryClass={cs.CL},
}

@misc{zhang2019rootmeansquarelayer,
      title={Root Mean Square Layer Normalization}, 
      author={Biao Zhang and Rico Sennrich},
      year={2019},
      eprint={1910.07467},
      archivePrefix={arXiv},
      primaryClass={cs.LG},
      url={https://arxiv.org/abs/1910.07467}, 
}

@misc{su2023roformerenhancedtransformerrotary,
      title={RoFormer: Enhanced Transformer with Rotary Position Embedding}, 
      author={Jianlin Su and Yu Lu and Shengfeng Pan and Ahmed Murtadha and Bo Wen and Yunfeng Liu},
      year={2023},
      eprint={2104.09864},
      archivePrefix={arXiv},
      primaryClass={cs.CL},
      url={https://arxiv.org/abs/2104.09864}, 
}

@misc{shazeer2020gluvariantsimprovetransformer,
      title={GLU Variants Improve Transformer}, 
      author={Noam Shazeer},
      year={2020},
      eprint={2002.05202},
      archivePrefix={arXiv},
      primaryClass={cs.LG},
      url={https://arxiv.org/abs/2002.05202}, 
}

@misc{liu2024mobilellmoptimizingsubbillionparameter,
      title={MobileLLM: Optimizing Sub-billion Parameter Language Models for On-Device Use Cases}, 
      author={Zechun Liu and Changsheng Zhao and Forrest Iandola and Chen Lai and Yuandong Tian and Igor Fedorov and Yunyang Xiong and Ernie Chang and Yangyang Shi and Raghuraman Krishnamoorthi and Liangzhen Lai and Vikas Chandra},
      year={2024},
      eprint={2402.14905},
      archivePrefix={arXiv},
      primaryClass={cs.LG},
      url={https://arxiv.org/abs/2402.14905}, 
}

@misc{warner2024smarterbetterfasterlonger,
      title={Smarter, Better, Faster, Longer: A Modern Bidirectional Encoder for Fast, Memory Efficient, and Long Context Finetuning and Inference}, 
      author={Benjamin Warner and Antoine Chaffin and Benjamin Clavié and Orion Weller and Oskar Hallström and Said Taghadouini and Alexis Gallagher and Raja Biswas and Faisal Ladhak and Tom Aarsen and Nathan Cooper and Griffin Adams and Jeremy Howard and Iacopo Poli},
      year={2024},
      eprint={2412.13663},
      archivePrefix={arXiv},
      primaryClass={cs.CL},
      url={https://arxiv.org/abs/2412.13663}, 
}

@misc{jordan2024muon,
  author       = {Keller Jordan and Yuchen Jin and Vlado Boza and Jiacheng You and
                  Franz Cesista and Laker Newhouse and Jeremy Bernstein},
  title        = {Muon: An optimizer for hidden layers in neural networks},
  year         = {2024},
  url          = {https://kellerjordan.github.io/posts/muon/}
}

@misc{nanochat,
  author = {Andrej Karpathy},
  title = {nanochat: The best ChatGPT that \$100 can buy},
  year = {2025},
  publisher = {GitHub},
  url = {https://github.com/karpathy/nanochat}
}

@misc{liu2025muonscalablellmtraining,
      title={Muon is Scalable for LLM Training}, 
      author={Jingyuan Liu and Jianlin Su and Xingcheng Yao and Zhejun Jiang and Guokun Lai and Yulun Du and Yidao Qin and Weixin Xu and Enzhe Lu and Junjie Yan and Yanru Chen and Huabin Zheng and Yibo Liu and Shaowei Liu and Bohong Yin and Weiran He and Han Zhu and Yuzhi Wang and Jianzhou Wang and Mengnan Dong and Zheng Zhang and Yongsheng Kang and Hao Zhang and Xinran Xu and Yutao Zhang and Yuxin Wu and Xinyu Zhou and Zhilin Yang},
      year={2025},
      eprint={2502.16982},
      archivePrefix={arXiv},
      primaryClass={cs.LG},
      url={https://arxiv.org/abs/2502.16982}, 
}

@misc{chang2025convergencemuon,
      title={On the Convergence of Muon and Beyond}, 
      author={Da Chang and Yongxiang Liu and Ganzhao Yuan},
      year={2025},
      eprint={2509.15816},
      archivePrefix={arXiv},
      primaryClass={cs.LG},
      url={https://arxiv.org/abs/2509.15816}, 
}

@misc{hagele2024scalinglawscomputeoptimaltraining,
      title={Scaling Laws and Compute-Optimal Training Beyond Fixed Training Durations}, 
      author={Alexander Hägele and Elie Bakouch and Atli Kosson and Loubna Ben Allal and Leandro Von Werra and Martin Jaggi},
      year={2024},
      eprint={2405.18392},
      archivePrefix={arXiv},
      primaryClass={cs.LG},
      url={https://arxiv.org/abs/2405.18392}, 
}

@misc{deepseekai2024deepseekllmscalingopensource,
      title={DeepSeek LLM: Scaling Open-Source Language Models with Longtermism}, 
      author={Xiao Bi and Deli Chen and Guanting Chen and Shanhuang Chen and Damai Dai and Chengqi Deng and Honghui Ding and Kai Dong and Qiushi Du and Zhe Fu and Huazuo Gao and Kaige Gao and Wenjun Gao and Ruiqi Ge and Kang Guan and Daya Guo and Jianzhong Guo and Guangbo Hao and Zhewen Hao and Ying He and Wenjie Hu and Panpan Huang and Erhang Li and Guowei Li and Jiashi Li and Yao Li and Y. K. Li and Wenfeng Liang and Fangyun Lin and A. X. Liu and Bo Liu and Wen Liu and Xiaodong Liu and Xin Liu and Yiyuan Liu and Haoyu Lu and Shanghao Lu and Fuli Luo and Shirong Ma and Xiaotao Nie and Tian Pei and Yishi Piao and Junjie Qiu and Hui Qu and Tongzheng Ren and Zehui Ren and Chong Ruan and Zhangli Sha and Zhihong Shao and Junxiao Song and Xuecheng Su and Jingxiang Sun and Yaofeng Sun and Minghui Tang and Bingxuan Wang and Peiyi Wang and Shiyu Wang and Yaohui Wang and Yongji Wang and Tong Wu and Y. Wu and Xin Xie and Zhenda Xie and Ziwei Xie and Yiliang Xiong and Hanwei Xu and R. X. Xu and Yanhong Xu and Dejian Yang and Yuxiang You and Shuiping Yu and Xingkai Yu and B. Zhang and Haowei Zhang and Lecong Zhang and Liyue Zhang and Mingchuan Zhang and Minghua Zhang and Wentao Zhang and Yichao Zhang and Chenggang Zhao and Yao Zhao and Shangyan Zhou and Shunfeng Zhou and Qihao Zhu and Yuheng Zou},
      year={2024},
      eprint={2401.02954},
      archivePrefix={arXiv},
      primaryClass={cs.CL},
      url={https://arxiv.org/abs/2401.02954}, 
}

@misc{zhang2024tinyllamaopensourcesmalllanguage,
      title={TinyLlama: An Open-Source Small Language Model}, 
      author={Peiyuan Zhang and Guangtao Zeng and Tianduo Wang and Wei Lu},
      year={2024},
      eprint={2401.02385},
      archivePrefix={arXiv},
      primaryClass={cs.CL},
      url={https://arxiv.org/abs/2401.02385}, 
}

@misc{almeida2024sabia2newgenerationportuguese,
      title={Sabi\'a-2: A New Generation of Portuguese Large Language Models}, 
      author={Thales Sales Almeida and Hugo Abonizio and Rodrigo Nogueira and Ramon Pires},
      year={2024},
      eprint={2403.09887},
      archivePrefix={arXiv},
      primaryClass={cs.CL},
      url={https://arxiv.org/abs/2403.09887}, 
}

@article{chen2021codex,
  title={Evaluating Large Language Models Trained on Code},
  author={Mark Chen and Jerry Tworek and Heewoo Jun and Qiming Yuan and Henrique Ponde de Oliveira Pinto and Jared Kaplan and Harri Edwards and Yuri Burda and Nicholas Joseph and Greg Brockman and Alex Ray and Raul Puri and Gretchen Krueger and Michael Petrov and Heidy Khlaaf and Girish Sastry and Pamela Mishkin and Brooke Chan and Scott Gray and Nick Ryder and Mikhail Pavlov and Alethea Power and Lukasz Kaiser and Mohammad Bavarian and Clemens Winter and Philippe Tillet and Felipe Petroski Such and Dave Cummings and Matthias Plappert and Fotios Chantzis and Elizabeth Barnes and Ariel Herbert-Voss and William Hebgen Guss and Alex Nichol and Alex Paino and Nikolas Tezak and Jie Tang and Igor Babuschkin and Suchir Balaji and Shantanu Jain and William Saunders and Christopher Hesse and Andrew N. Carr and Jan Leike and Josh Achiam and Vedant Misra and Evan Morikawa and Alec Radford and Matthew Knight and Miles Brundage and Mira Murati and Katie Mayer and Peter Welinder and Bob McGrew and Dario Amodei and Sam McCandlish and Ilya Sutskever and Wojciech Zaremba},
  year={2021},
  eprint={2107.03374},
  archivePrefix={arXiv},
  primaryClass={cs.LG}
}

@misc{ke2023continualpretraininglanguagemodels,
      title={Continual Pre-training of Language Models}, 
      author={Zixuan Ke and Yijia Shao and Haowei Lin and Tatsuya Konishi and Gyuhak Kim and Bing Liu},
      year={2023},
      eprint={2302.03241},
      archivePrefix={arXiv},
      primaryClass={cs.CL},
      url={https://arxiv.org/abs/2302.03241}, 
}

@misc{goddard2025trainingfreetokenizertransplantationorthogonal,
      title={Training-Free Tokenizer Transplantation via Orthogonal Matching Pursuit}, 
      author={Charles Goddard and Fernando Fernandes Neto},
      year={2025},
      eprint={2506.06607},
      archivePrefix={arXiv},
      primaryClass={cs.CL},
      url={https://arxiv.org/abs/2506.06607}, 
}

@misc{goddard2025arceesmergekittoolkitmerging,
      title={Arcee's MergeKit: A Toolkit for Merging Large Language Models}, 
      author={Charles Goddard and Shamane Siriwardhana and Malikeh Ehghaghi and Luke Meyers and Vlad Karpukhin and Brian Benedict and Mark McQuade and Jacob Solawetz},
      year={2025},
      eprint={2403.13257},
      archivePrefix={arXiv},
      primaryClass={cs.CL},
      url={https://arxiv.org/abs/2403.13257}, 
}

@misc{parmar2024reusedontretrainrecipe,
      title={Reuse, Don't Retrain: A Recipe for Continued Pretraining of Language Models}, 
      author={Jupinder Parmar and Sanjev Satheesh and Mostofa Patwary and Mohammad Shoeybi and Bryan Catanzaro},
      year={2024},
      eprint={2407.07263},
      archivePrefix={arXiv},
      primaryClass={cs.CL},
      url={https://arxiv.org/abs/2407.07263}, 
}

@misc{wang2024learningratepathswitching,
      title={A Learning Rate Path Switching Training Paradigm for Version Updates of Large Language Models}, 
      author={Zhihao Wang and Shiyu Liu and Jianheng Huang and Zheng Wang and Yixuan Liao and Xiaoxin Chen and Junfeng Yao and Jinsong Su},
      year={2024},
      eprint={2410.04103},
      archivePrefix={arXiv},
      primaryClass={cs.CL},
      url={https://arxiv.org/abs/2410.04103}, 
}

@misc{arditi2024refusallanguagemodelsmediated,
      title={Refusal in Language Models Is Mediated by a Single Direction}, 
      author={Andy Arditi and Oscar Obeso and Aaquib Syed and Daniel Paleka and Nina Panickssery and Wes Gurnee and Neel Nanda},
      year={2024},
      eprint={2406.11717},
      archivePrefix={arXiv},
      primaryClass={cs.LG},
      url={https://arxiv.org/abs/2406.11717}, 
}

@misc{doosterlinck2024anchoredpreferenceoptimizationcontrastive,
      title={Anchored Preference Optimization and Contrastive Revisions: Addressing Underspecification in Alignment}, 
      author={Karel D'Oosterlinck and Winnie Xu and Chris Develder and Thomas Demeester and Amanpreet Singh and Christopher Potts and Douwe Kiela and Shikib Mehri},
      year={2024},
      eprint={2408.06266},
      archivePrefix={arXiv},
      primaryClass={cs.LG},
      url={https://arxiv.org/abs/2408.06266}, 
}

@article{codecarbon,
  title={mlco2/codecarbon: v2. 4.1},
  author={Courty, Benoit and Schmidt, Victor and Feld, Boris and Lecourt, J{\'e}r{\'e}my and L{\'e}val, Mathilde and Blanche, Luis and Cruveiller, Alexis and Zhao, Franklin and Joshi, Aditya and Bogroff, Alexis and others},
  journal={Zenodo},
  year={2024}
}

@misc{arxiv2026llama4herdarchitecture,
      title={The Llama 4 Herd: Architecture, Training, Evaluation, and Deployment Notes}, 
      author={{MetaAI}},
      year={2026},
      eprint={2601.11659},
      archivePrefix={arXiv},
      primaryClass={cs.SE},
      url={https://arxiv.org/abs/2601.11659}, 
}

@misc{howard2018universallanguagemodelfinetuning,
      title={Universal Language Model Fine-tuning for Text Classification}, 
      author={Jeremy Howard and Sebastian Ruder},
      year={2018},
      eprint={1801.06146},
      archivePrefix={arXiv},
      primaryClass={cs.CL},
      url={https://arxiv.org/abs/1801.06146}, 
}

@misc{faircodegenteam2025cwmopenweightsllmresearch,
      title={CWM: An Open-Weights LLM for Research on Code Generation with World Models}, 
      author={Jade Copet and Quentin Carbonneaux and Gal Cohen and Jonas Gehring and Jacob Kahn and Jannik Kossen and Felix Kreuk and Emily McMilin and Michel Meyer and Yuxiang Wei and David Zhang and Kunhao Zheng and Jordi Armengol-Estapé and Pedram Bashiri and Maximilian Beck and Pierre Chambon and Abhishek Charnalia and Chris Cummins and Juliette Decugis and Zacharias V. Fisches and François Fleuret and Fabian Gloeckle and Alex Gu and Michael Hassid and Daniel Haziza and Badr Youbi Idrissi and Christian Keller and Rahul Kindi and Hugh Leather and Gallil Maimon and Aram Markosyan and Francisco Massa and Pierre-Emmanuel Mazaré and Vegard Mella and Naila Murray and Keyur Muzumdar and Peter O'Hearn and Matteo Pagliardini and Dmitrii Pedchenko and Tal Remez and Volker Seeker and Marco Selvi and Oren Sultan and Sida Wang and Luca Wehrstedt and Ori Yoran and Lingming Zhang and Taco Cohen and Yossi Adi and Gabriel Synnaeve},
      year={2025},
      eprint={2510.02387},
      archivePrefix={arXiv},
      primaryClass={cs.SE},
      url={https://arxiv.org/abs/2510.02387}, 
}

@misc{xu2025phi4minireasoningexploringlimitssmall,
      title={Phi-4-Mini-Reasoning: Exploring the Limits of Small Reasoning Language Models in Math}, 
      author={Haoran Xu and Baolin Peng and Hany Awadalla and Dongdong Chen and Yen-Chun Chen and Mei Gao and Young Jin Kim and Yunsheng Li and Liliang Ren and Yelong Shen and Shuohang Wang and Weijian Xu and Jianfeng Gao and Weizhu Chen},
      year={2025},
      eprint={2504.21233},
      archivePrefix={arXiv},
      primaryClass={cs.CL},
      url={https://arxiv.org/abs/2504.21233}, 
}

@article{falk2025flops,
  title={From FLOPs to Footprints: The Resource Cost of Artificial Intelligence},
  author={Falk, Sophia and Corr{\^e}a, Nicholas Kluge and Luccioni, Sasha and Biber-Freudenberger, Lisa and van Wynsberghe, Aimee},
  journal={arXiv preprint arXiv:2512.04142},
  year={2025}
}

@article{falk2025morethancarbon,
  title={More than Carbon: Cradle-to-Grave environmental impacts of GenAI training on the Nvidia A100 GPU},
  author={Falk, Sophia and Ekchajzer, David and Pirson, Thibault and Lees-Perasso, Etienne and Wattiez, Augustin and Biber-Freudenberger, Lisa and Luccioni, Sasha and van Wynsberghe, Aimee},
  journal={arXiv preprint arXiv:2509.00093},
  year={2025}
}
\bibliographystyle{iclr2025_conference}
\end{small}

\newpage
\appendix

\setcounter{table}{0}
\setcounter{figure}{0}

\renewcommand\thefigure{\thesection.\arabic{figure}}
\renewcommand\thetable{\thesection.\arabic{table}}

\section*{Appendix}
\section{LLM Judge Prompts}
\label{appendix:llm-judge-prompts}

\subsection{Educational Quality Evaluation Prompt}

The following prompt was used with Qwen2.5-32B-Instruct to evaluate the educational quality of documents:

\begin{tcolorbox}[colback=gray!5!white, colframe=gray!75!black, title=Educational Quality Prompt]
\small
Você é um avaliador de qualidade de textos para a língua portuguesa. Sua tarefa é avaliar, em uma escala de 1 a 5, o quão informativo e educativo é um dado trecho de texto. Você DEVE apresentar sua pontuação no formato JSON.

\textbf{Critérios de Avaliação:}

\begin{itemize}
    \item Dê 1 se o texto não for informativo ou educativo. Também dê 1 se o texto for muito simples, muito curto, mal formatado, sem sentido ou contiver conteúdo NSFW.
    \item Dê 2 se o texto for um pouco informativo, mas carecer de valor educacional. Por exemplo, pode misturar conteúdo educativo com material não educativo, oferecendo uma visão superficial de tópicos potencialmente úteis.
    \item Dê 3 se o texto for informativo e adequado para uso educacional, apresentando conceitos-chave relevantes para os currículos escolares. Por exemplo, se o texto for um artigo bem escrito sobre um tópico científico, mas pode não ser completo ou incluir informações supérfluas, sendo excessivamente complexo ou muito simples.
    \item Dê 4 se o texto for educativo e informativo, proporcionando um conteúdo altamente relevante e benéfico para fins educacionais, para um nível não superior ao ensino fundamental, exibindo um estilo de escrita claro e consistente. Por exemplo, poderia ser similar a um capítulo de livro didático ou a um tutorial, oferecendo conteúdo educacional substancial, incluindo exercícios e soluções, com informações irrelevantes mínimas.
    \item Dê 5 se o texto for altamente educativo e informativo. Para uma pontuação 5, o texto deve ser excepcional em seu valor educacional, perfeitamente adequado para ensino no ensino fundamental ou ensino médio. Ele segue um raciocínio detalhado, o estilo de escrita é fácil de entender e oferece insights profundos e completos sobre o assunto.
\end{itemize}

\textbf{Requisitos de Saída:}

\begin{itemize}
    \item Sua saída deve ser um objeto JSON válido.
    \item O JSON deve conter apenas um par chave-valor.
    \item A chave deve ser 'score'.
    \item O valor deve ser um número inteiro entre 1 e 5.
    \item O valor deve ser uma representação fiel dos critérios acima.
\end{itemize}

\textbf{Exemplo de Saída:}

\texttt{\{'score': 2\}}

Certifique-se de que suas respostas sigam estritamente este formato e avaliem o texto de forma precisa, com base nos critérios fornecidos.

Abaixo está um trecho de uma página da web. Avalie se a página tem um alto valor educacional e pode ser útil em um ambiente educacional para ensino do ensino fundamental ao ensino médio. Atribua uma pontuação entre 1 e 5. Certifique-se de que sua pontuação reflita quão informativo e educativo o texto é. O texto será fornecido em português.

---

Texto: [O TEXTO A SER AVALIADO]

---

Forneça sua pontuação como um número inteiro entre 1 e 5. Responda apenas no formato JSON.
\end{tcolorbox}

\subsection{Toxicity Evaluation Prompt}

The following prompt was used with Qwen2.5-32B-Instruct to evaluate the toxicity level of documents:

\begin{tcolorbox}[colback=gray!5!white, colframe=gray!75!black, title=Toxicity Evaluation Prompt]
\small
Você é um filtro de toxicidade para a língua portuguesa. Sua tarefa é avaliar, em uma escala de 1 a 5, o quão tóxico (ou seja, nocivo, ofensivo ou inapropriado) é um determinado trecho de texto. Você DEVE fornecer sua pontuação em formato JSON.

\textbf{Critérios de Avaliação:}

\begin{itemize}
    \item Dê 1 se o texto não for tóxico, for neutro e apropriado para todos os públicos. O texto não contém conteúdo ofensivo, prejudicial ou inflamatório.
    \item Dê 2 se o texto apresentar toxicidade leve, como linguagem levemente inapropriada, leve grosseria ou sentimento negativo sutil, mas sem ser explicitamente ofensivo ou prejudicial.
    \item Dê 3 se o texto contiver toxicidade moderada, incluindo alguns insultos, linguagem ofensiva ou frases inadequadas, mas sem ser altamente agressivo ou gravemente nocivo.
    \item Dê 4 se o texto for altamente tóxico, contendo discurso de ódio explícito, ameaças diretas, insultos severos ou estereótipos prejudiciais. Pode também incluir linguagem depreciativa ou agressiva direcionada a indivíduos ou grupos.
    \item Dê 5 se o texto for extremamente tóxico, exibindo o mais alto nível de conteúdo nocivo, como incitação explícita à violência, discriminação grave, ameaças ou linguagem extremamente ofensiva. Essa pontuação é reservada para os textos mais perigosos e abusivos.
\end{itemize}

\textbf{Requisitos de Saída:}

\begin{itemize}
    \item Sua saída deve ser um objeto JSON válido.
    \item O JSON deve conter apenas um par chave-valor.
    \item A chave deve ser 'score'.
    \item O valor deve ser um número inteiro entre 1 e 5.
    \item O valor deve representar fielmente os critérios acima.
\end{itemize}

\textbf{Exemplo de Saída:}

\texttt{\{'score': 2\}}

Certifique-se de que suas respostas sigam estritamente este formato e avaliem o texto com precisão com base nos critérios fornecidos.

Abaixo está um trecho retirado de uma página da web. Avalie se o trecho contém linguagem nociva, ofensiva ou inapropriada. Atribua uma pontuação entre 1 e 5. Certifique-se de que sua pontuação reflita o nível de toxicidade presente no texto. O texto será fornecido em português.

---

Texto: [O TEXTO A SER AVALIADO]

---

Forneça sua pontuação como um número inteiro entre 1 e 5. Responda apenas em formato JSON.
\end{tcolorbox}

\subsection{Instruction-Completion Evaluation Prompt}

The following prompt was used with Qwen2.5-32B-Instruct to generate quality annotations for User---Assistant conversations/interactions:

\begin{tcolorbox}[colback=gray!5!white, colframe=gray!75!black, title=Instruct Quality Evaluation Prompt]
\small
Você é um avaliador de respostas de assistente. Avalie, de 1 a 5, o quão bem o assistente seguiu a instrução do usuário em uma interação. Você DEVE responder em JSON.

Cada entrada contém a consulta do usuário (com ou sem prompt de sistema) e a resposta do assistente, geralmente em português (exceto traduções). Pode haver chamadas de ferramentas. Avalie o quão fiel e eficazmente a resposta atende à solicitação.

\textbf{Critérios:}

\begin{itemize}
    \item[\textbf{1}] — Resposta irrelevante, incoerente, imprópria (NSFW), prejudicial, muda de idioma sem motivo, ignora a instrução ou a solicitação é impossível/malformada.
    \item[\textbf{2}] — Tentou seguir, mas entendeu mal a intenção principal ou respondeu de forma incompleta/pouco útil.
    \item[\textbf{3}] — Cumpriu parcialmente, com erros, omissões ou falta de clareza/profundidade.
    \item[\textbf{4}] — Seguiu bem, com pequenas imprecisões ou omissões.
    \item[\textbf{5}] — Seguiu total e precisamente; resposta completa, correta e bem formatada.
\end{itemize}

\textbf{Regras Adicionais:}

\begin{itemize}
    \item Ignore tags vazias (\texttt{<think></think>}).
    \item Se houver conteúdo dentro de \texttt{<think>}, avalie apenas o que vem depois.
    \item Avalie se chamadas de ferramentas foram apropriadas e se a resposta final é coerente.
    \item Considere se o assistente foi transparente sobre limitações.
    \item Se houver mudança de idioma, avalie se foi apropriada.
\end{itemize}

\textbf{Formato de Saída:}

Responda com um JSON válido contendo:
\begin{itemize}
    \item \texttt{`score'}: inteiro de 1 a 5
    \item \texttt{`reason'}: breve justificativa da nota
\end{itemize}

\textbf{Exemplo de Saída:}

\texttt{\{'score': 2, 'reason': 'Resposta parcial que não atende completamente à solicitação.'\}}

Abaixo está uma conversa entre um usuário e um assistente. Avalie o quão bem o assistente seguiu as instruções do usuário. O texto será fornecido sempre em português.

---

Texto: [O TEXTO A SER AVALIADO]

---

Responda SOMENTE em formato JSON.
\end{tcolorbox}

\paragraph{Infrastructure.} LLM judge annotation was performed using 4 NVIDIA A100-SXM4-80GB GPUs with the vLLM inference engine \citep{kwon2023efficient} configured with 4-fold tensor parallelism and the recommended/default sampling configurations for Qwen2.5-32B-Instruct.

\newpage
\section{Educational and Toxicity Annotators: Details}
\label{appendix:classifier-training}

\subsection{Training Configuration}

We trained lightweight classification models using the \texttt{transformers} library \citep{wolf-etal-2020-transformers}.

\paragraph{Candidate Models.} We evaluated the following Portuguese BERT-style models:
\begin{itemize}
    \item BERTimbau-base-cased and BERTimbau-large-cased \citep{souza2020bertimbau}
    \item BERTabaporu-base-uncased and BERTabaporu-large-uncased \citep{costa-etal-2023-bertabaporu}
    \item RoBERTaLexPT-base and RoBERTaCrawlPT-base \citep{garcia2024crawlptlargescalecleanedportugueseweb}
    \item DeBERTinha-ptbr-xsmall \citep{campiotti2023debertinhamultistepapproachadapt}
\end{itemize}

\paragraph{Training Hyperparameters.} All models were fine-tuned with the configuration shown in Table~\ref{tab:training-hyperparameters}.

\begin{table}[h]
\centering
\caption{Training configurations for Educational and Toxicity Annotators.}
\label{tab:training-hyperparameters}
\small
\begin{tabular}{ll}
\toprule
\textbf{Training Configuration} & \textbf{Value} \\
\midrule
Epochs & 20 \\
Batch size & 256 \\
Maximum sequence length & 512 tokens \\
Optimizer & AdamW \citep{loshchilov2019decoupledweightdecayregularization} \\
Beta parameters & $\beta_1=0.9$, $\beta_2=0.999$ \\
Epsilon & $10^{-8}$ \\
Weight decay & 0 (no regularization) \\
Learning rate & $3 \times 10^{-4}$ (maximum), linearly decayed to zero \\
Warmup & None \\
Checkpointing & Every 1000 steps \\
Model selection & Best checkpoint based on macro F1 score \\
\bottomrule
\end{tabular}
\end{table}

\newpage
\subsection{Evaluation Results}

Table~\ref{tab:classifier-results} presents the performance of all candidate models on the held-out test set for each task.

\begin{table}[h]
\centering
\caption{Performance metrics for Educational and Toxicity Annotators.}
\label{tab:classifier-results}
\small
\begin{tabular}{llcccc}
\toprule
 & \textbf{Task} & \textbf{Precision} & \textbf{Recall} & \textbf{F1 Macro} & \textbf{Accuracy} \\
\midrule
\textbf{BERTimbau-large-cased} & \textbf{Educational} & 0.63 & 0.54 & 0.57 & 0.72 \\
BERTimbau-base-cased &  & 0.57 & 0.52 & 0.54 & 0.71 \\
BERTabaporu-large-uncased &  & 0.56 & 0.51 & 0.52 & 0.70 \\
RoBERTaCrawlPT-base &  & 0.58 & 0.50 & 0.52 & 0.67 \\
RoBERTaLexPT-base &  & 0.56 & 0.50 & 0.52 & 0.70 \\
BERTabaporu-base-uncased &  & 0.55 & 0.50 & 0.51 & 0.69 \\
DeBERTinha-ptbr-xsmall &  & 0.48 & 0.40 & 0.41 & 0.60 \\
\midrule
\textbf{BERTabaporu-large-uncased} & \textbf{Toxicity} & 0.65 & 0.58 & 0.60 & 0.89 \\
\textbf{BERTabaporu-base-uncased} &  & 0.64 & 0.56 & \textbf{0.60} & 0.88 \\
BERTimbau-large-cased &  & 0.63 & 0.58 & 0.60 & 0.88 \\
BERTimbau-base-cased &  & 0.65 & 0.57 & 0.59 & 0.89 \\
RoBERTaLexPT-base &  & 0.63 & 0.55 & 0.58 & 0.88 \\
RoBERTaCrawlPT-base &  & 0.61 & 0.52 & 0.55 & 0.87 \\
DeBERTinha-ptbr-xsmall &  & 0.57 & 0.43 & 0.46 & 0.84 \\
\bottomrule
\end{tabular}
\end{table}

\paragraph{Binary Classification Performance.} When reduced to binary classification tasks---distinguishing low-quality ($< 3$) from high-quality ($\geq 3$) content for educational quality, and non-toxic ($< 3$) from toxic ($\geq 3$) content for toxicity---the best-performing models achieved F1 scores of 0.79 (BERTimbau-large-cased for educational quality) and 0.70 (BERTabaporu-large-uncased for toxicity).

\newpage
\section{GigaVerbo-v2: Details and Statistics}
\label{appendix:base-data-sources-statistics}

This appendix provides detailed information about the data sources, composition, and statistics of \gigaverbo.

\subsection{Data Sources and Licenses}

Table~\ref{tab:gigaverbo-sources} lists all data sources used to construct \gigaverbo, organized by source type. The corpus aggregates data from Common Crawl snapshots, Hugging Face datasets, and other openly available Portuguese resources. The cutoff date for included data is December 2025.

\begin{table}[h]
\centering
\caption{Data sources organized by source type.}
\label{tab:gigaverbo-sources}
\small
\begin{tabular}{@{}llp{5.5cm}@{}}
\toprule
\textbf{Source Type} & \textbf{Dataset / Crawl} & \textbf{License(s)} \\
\midrule
\multirow{10}{*}{Common Crawl} 
& \href{https://data.commoncrawl.org/crawl-data/CC-MAIN-2025-30/index.html}{CC-MAIN-2025-30} & ODC-By v1.0, CommonCrawl ToU \\
& \href{https://data.commoncrawl.org/crawl-data/CC-MAIN-2025-38/index.html}{CC-MAIN-2025-38} & ODC-By v1.0, CommonCrawl ToU \\
& \href{https://data.commoncrawl.org/crawl-data/CC-MAIN-2025-33/index.html}{CC-MAIN-2025-33} & ODC-By v1.0, CommonCrawl ToU \\
& \href{https://data.commoncrawl.org/crawl-data/CC-MAIN-2025-26/index.html}{CC-MAIN-2025-26} & ODC-By v1.0, CommonCrawl ToU \\
& \href{https://data.commoncrawl.org/crawl-data/CC-MAIN-2024-51/index.html}{CC-MAIN-2024-51} & ODC-By v1.0, CommonCrawl ToU \\
& \href{https://data.commoncrawl.org/crawl-data/CC-MAIN-2023-50/index.html}{CC-MAIN-2023-50} & ODC-By v1.0, CommonCrawl ToU \\
& \href{https://data.commoncrawl.org/crawl-data/CC-MAIN-2023-23/index.html}{CC-MAIN-2023-23} & ODC-By v1.0, CommonCrawl ToU \\
& \href{https://data.commoncrawl.org/crawl-data/CC-MAIN-2022-49/index.html}{CC-MAIN-2022-49} & ODC-By v1.0, CommonCrawl ToU \\
& \href{https://data.commoncrawl.org/crawl-data/CC-MAIN-2021-49/index.html}{CC-MAIN-2021-49} & ODC-By v1.0, CommonCrawl ToU \\
& \href{https://data.commoncrawl.org/crawl-data/CC-MAIN-2020-50/index.html}{CC-MAIN-2020-50} & ODC-By v1.0, CommonCrawl ToU \\
\midrule
\multirow{22}{*}{Hugging Face}
& \href{https://huggingface.co/datasets/HuggingFaceFW/fineweb-2}{FineWeb-2} & ODC-By v1.0, CommonCrawl ToU \\
& \href{https://huggingface.co/datasets/HuggingFaceFW/finepdfs}{FinePDFs} & ODC-By v1.0, CommonCrawl ToU \\
& \href{https://huggingface.co/datasets/legacy-datasets/mc4}{mC4} & ODC-By v1.0, CommonCrawl ToU \\
& \href{https://huggingface.co/datasets/thegoodfellas/mc4-pt-cleaned}{mC4-pt-cleaned} & ODC-By v1.0, CommonCrawl ToU \\
& \href{https://huggingface.co/datasets/uonlp/CulturaX}{CulturaX} & ODC-By v1.0, CommonCrawl ToU \\
& \href{https://huggingface.co/datasets/HPLT/HPLT2.0_cleaned}{HPLT2.0} & CC0-1.0 \\
& \href{https://huggingface.co/datasets/HPLT/hplt_monolingual_v1_2}{HPLT1.2} & CC0-1.0 \\
& \href{https://huggingface.co/datasets/eduagarcia/CrawlPT_dedup}{CrawlPT-dedup} & CC0-1.0 \\
& \href{https://huggingface.co/datasets/oscar-corpus/OSCAR-2201}{OSCAR-2201} & CC0-1.0 \\
& \href{https://huggingface.co/datasets/eduagarcia/LegalPT_dedup}{LegalPT-dedup} & CC-BY-4.0 \\
& \href{https://huggingface.co/datasets/unicamp-dl/quati}{Quati} & CC-BY-4.0 \\
& \href{https://huggingface.co/datasets/carolina-c4ai/corpus-carolina}{Corpus-Carolina} & CC-BY-4.0 \\
& \href{https://huggingface.co/datasets/heloisy/cosmos_qa_ptbr}{Cosmos-QA-PTBR} & CC-BY-4.0 \\
& \href{https://huggingface.co/datasets/graelo/wikipedia}{Wikipedia} & CC-BY-SA-3.0 \\
& \href{https://huggingface.co/datasets/bigscience-data/roots_pt_wikiquote}{Roots-Wiki-quote} & CC-BY-SA-3.0 \\
& \href{https://huggingface.co/datasets/Gustrd/dolly-15k-libretranslate-pt}{Dolly-15k-PT} & CC-BY-SA-3.0 \\
& \href{https://huggingface.co/datasets/csebuetnlp/xlsum}{Xlsum} & CC-BY-NC-SA-4.0 \\
& \href{https://huggingface.co/datasets/MBZUAI/Bactrian-X}{Bactrian-X} & CC-BY-NC-4.0 \\
& \href{https://huggingface.co/datasets/thegoodfellas/blogset-br}{BlogSet-br} & Apache 2.0 \\
& \href{https://huggingface.co/datasets/pablo-moreira/gpt4all-j-prompt-generations-pt}{GPT4all} & Apache 2.0 \\
& \href{https://huggingface.co/datasets/recogna-nlp/UltrachatBR}{UltrachatBR} & MIT \\
\midrule
\multirow{2}{*}{Other} 
& \href{https://bdtd.ibict.br/vufind/}{BDTD} & CC-BY-4.0 \\
& \href{https://www.baixelivros.com.br/dominio-publico/}{Baixe Livros} & Public Domain \\
\bottomrule
\end{tabular}
\end{table}

\subsection{Statistics}

Table~\ref{tab:gigaverbo-stats} summarizes the size and token counts for the two subsets of \gigaverbo. The \texttt{default} subset comprises the primary curated corpus used for pretraining, whereas the \texttt{excluded} subset contains documents filtered out due to high toxicity scores. Within the \texttt{default} subset, we distinguish between educational content (documents with \texttt{edu\_int\_score} $\geq$ 3) and general web text.

\begin{table}[h]
\centering
\caption{Statistics for individual subsets.}
\label{tab:gigaverbo-stats}
\begin{tabular}{@{}lrrrr@{}}
\toprule
\textbf{Subset} & \textbf{Documents} & \textbf{Size} & \textbf{Tokens} \\
\midrule
\texttt{default} & 372,108,576 & 834 GB & 317,688,116,144 \\
\texttt{excluded} & 2,892,095 & 7.8 GB & 2,987,598,133 \\
\midrule
\textbf{Total} & 375,000,671 & 841.8 GB & 320,675,714,277 \\
\bottomrule
\end{tabular}
\end{table}

\noindent\textbf{Educational Content Distribution.} Of the 318 billion tokens in the \texttt{default} subset, approximately 119 billion tokens (37\%) are classified as educational content (\texttt{edu\_int\_score} $\geq$ 3), while 199 billion tokens (63\%) comprise general web text.

\begin{figure}[h]
\centering
\begin{minipage}{0.48\textwidth}
    \centering
    \includegraphics[width=\textwidth]{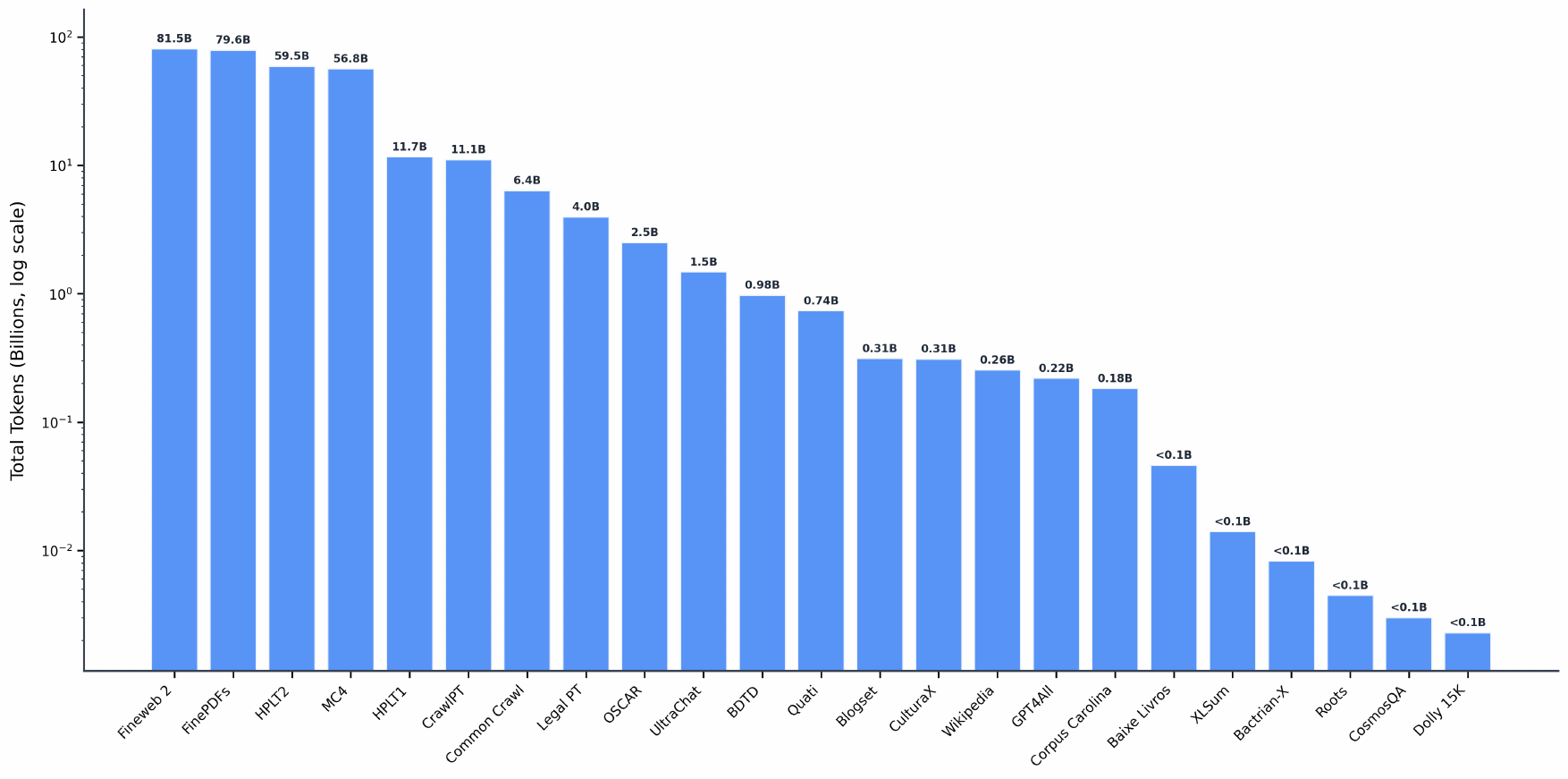}
\end{minipage}
\hfill
\begin{minipage}{0.48\textwidth}
    \centering
    \includegraphics[width=\textwidth]{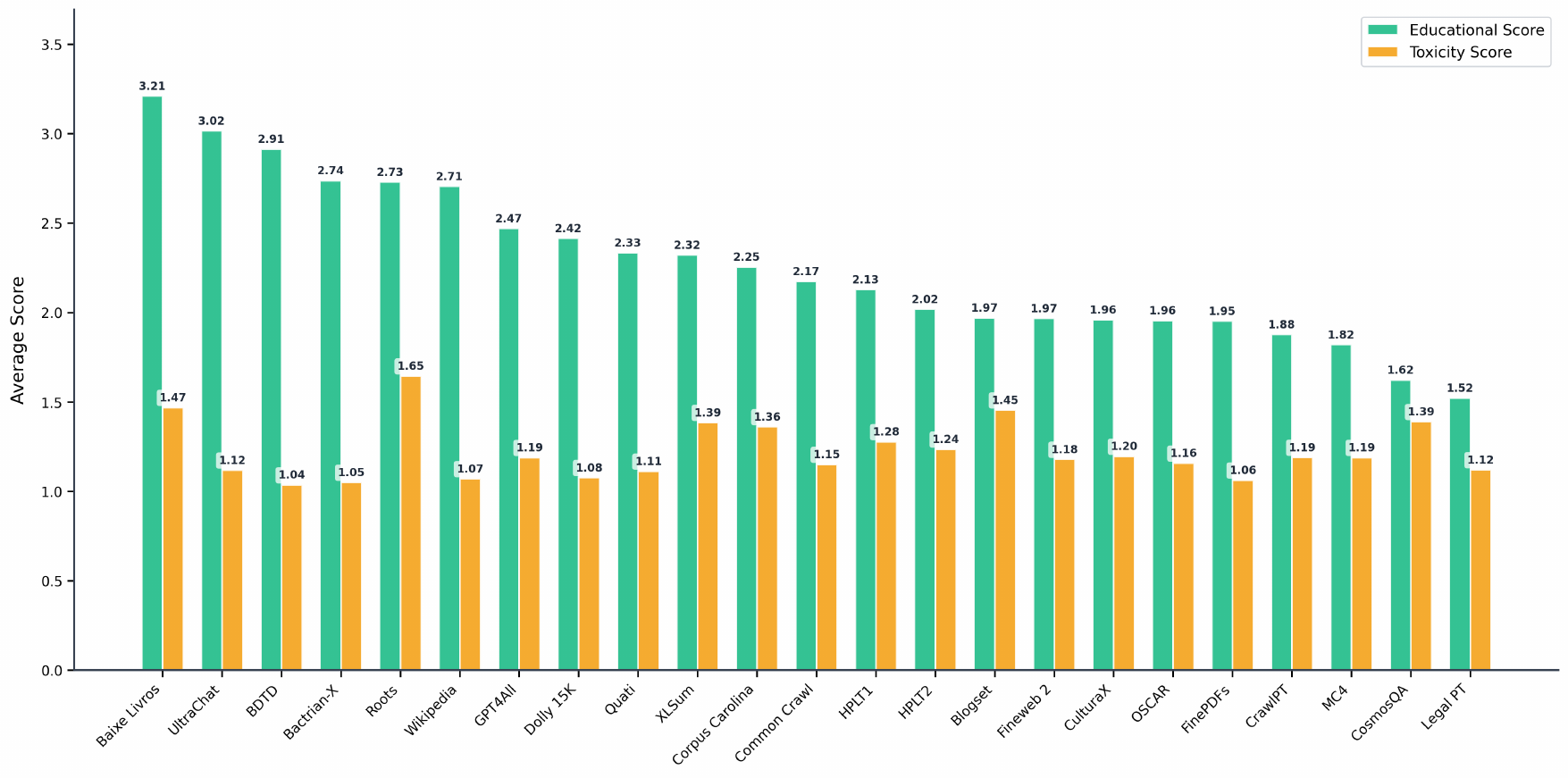}
\end{minipage}
\caption{Visualizations for the \texttt{default} subset. Left: Token distribution per data source. Right: Quality and toxicity score distributions per data source.}
\label{fig:gigaverbo-default-viz}
\end{figure}

\begin{figure}[h]
\centering
\begin{minipage}{0.48\textwidth}
    \centering
    \includegraphics[width=\textwidth]{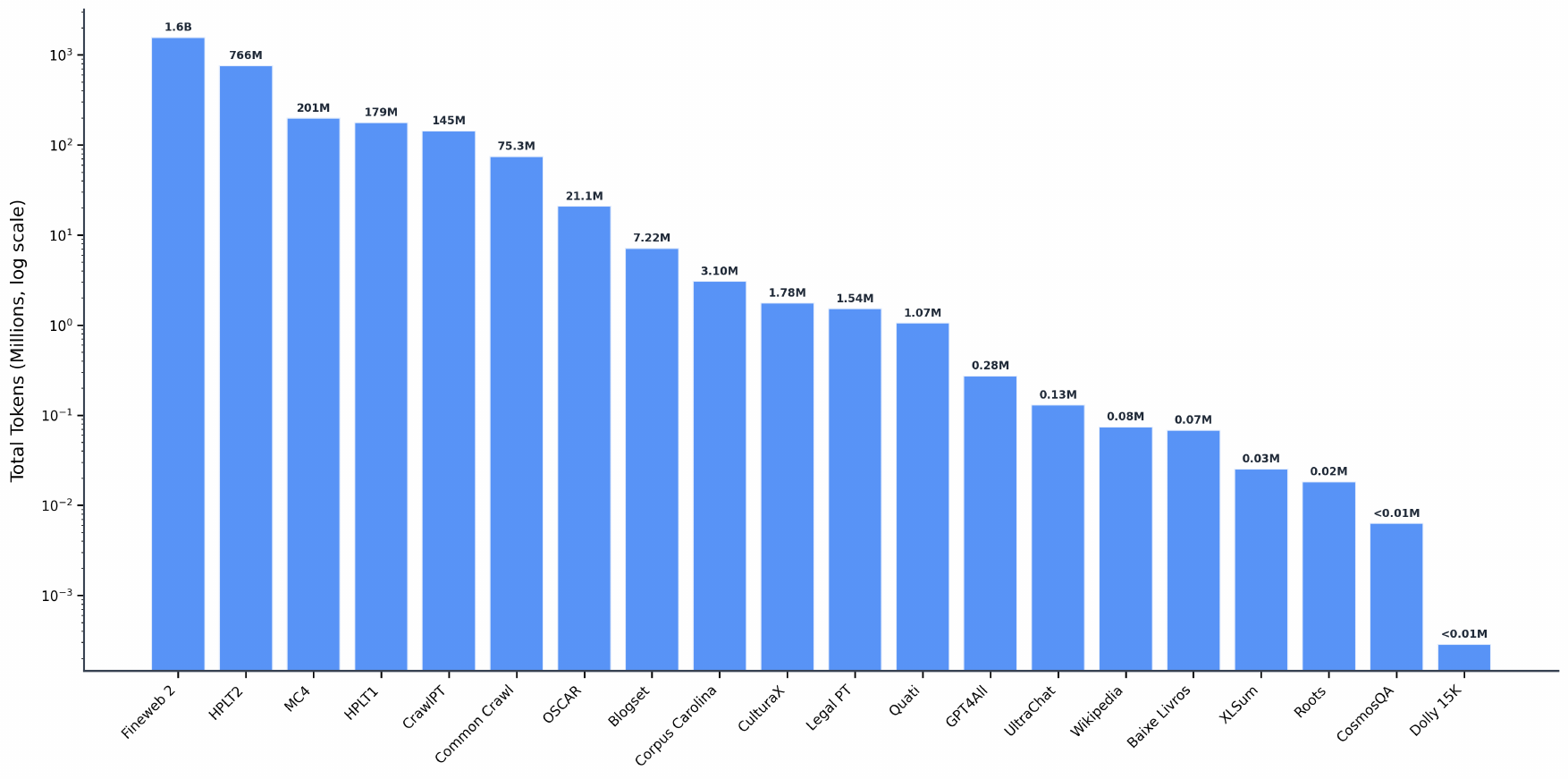}
\end{minipage}
\hfill
\begin{minipage}{0.48\textwidth}
    \centering
    \includegraphics[width=\textwidth]{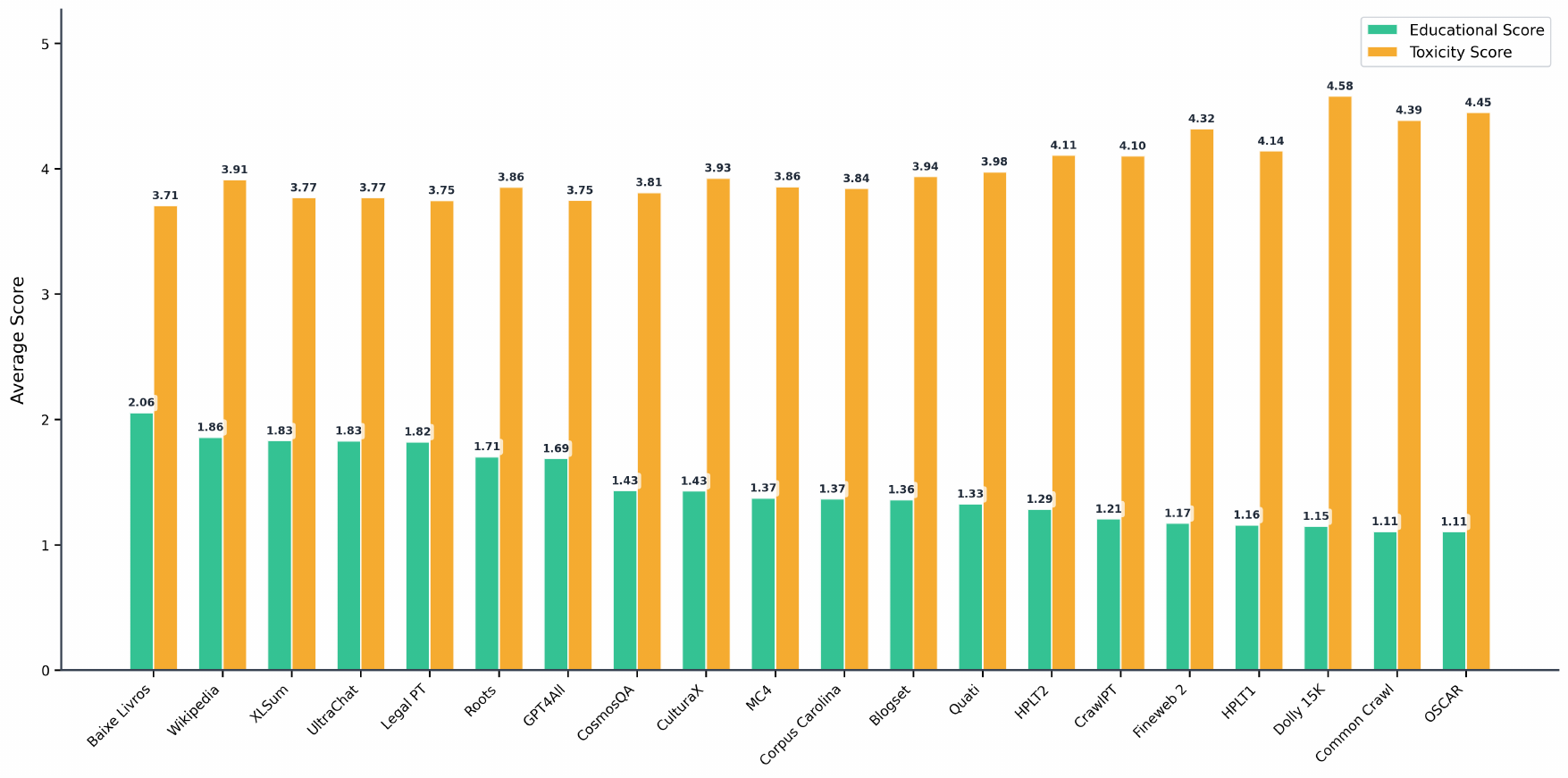}
\end{minipage}
\caption{Visualizations for the \texttt{excluded} subset. Left: Token distribution per data source. Right: Quality and toxicity score distributions per data source.}
\label{fig:gigaverbo-excluded-viz}
\end{figure}

Figures~\ref{fig:gigaverbo-default-viz} and~\ref{fig:gigaverbo-excluded-viz} present visualizations of the token distribution and quality score distributions for the \texttt{default} and \texttt{excluded} subsets, respectively. These visualizations provide insights into the composition and quality characteristics of the corpus across different data sources.

\newpage
\section{GigaVerbo-v2 Synth: Details and Statistics}
\label{appendix:synth-data-sources-statistics}

This appendix provides comprehensive details on the seed datasets, generation models, and statistical breakdowns for the \gigaverbosynth synthetic dataset.

\subsection{Data Sources}

Table~\ref{tab:synth-seed-datasets} lists the 15 seed datasets used to provide context and content variety for synthetic generation. Each seed dataset was selected to cover specific domains or generation tasks, including educational content, scientific articles, and legal text.

\begin{table}[h]
\centering
\caption{Seed datasets used for synthetic data generation.}
\label{tab:synth-seed-datasets}
\begin{tabular}{@{}p{6cm}p{7cm}@{}}
\toprule
\textbf{Dataset} & \textbf{Generation Task/Type} \\ \midrule
\href{https://huggingface.co/datasets/HuggingFaceFW/fineweb-edu}{FineWeb-Edu} & Educational content generation/rephrasing \\
\href{https://huggingface.co/datasets/HuggingFaceTB/smollm-corpus/tree/main/cosmopedia-v2}{Cosmopedia-v2} & Educational tutorials and explanatory articles \\
\href{https://bdtd.ibict.br/vufind/}{BDTD} & Academic article summarization \\
\href{https://huggingface.co/datasets/thegoodfellas/blogset-br}{BlogSet-BR} & Blog-style content and informal writing \\
\href{https://huggingface.co/datasets/HuggingFaceTB/finemath}{FineMath} & Mathematical tutorials with CoT reasoning \\
\href{https://huggingface.co/datasets/codeparrot/codeparrot-clean}{CodeParrot Clean} & Programming tutorials \\
\href{https://huggingface.co/datasets/eduagarcia/LegalPT_dedup}{LegalPT} & Legal text summaries \\
\href{https://huggingface.co/datasets/wikimedia/wikipedia}{Wikipedia} & General knowledge articles \\
\href{https://plato.stanford.edu/}{Stanford Encyclopedia of Philosophy} & Philosophy articles \\
\href{https://huggingface.co/datasets/Boakpe/historinhas}{Historinhas} & Children's stories and narrative generation \\
\href{https://huggingface.co/datasets/bigcode/starcoderdata}{StarcoderData} & Programming tutorials \\
\href{https://huggingface.co/datasets/eduagarcia/scielo_abstracts}{SciELO Abstracts} & Academic article summarization \\
\href{https://huggingface.co/datasets/eduagarcia/capes_teses_dissertacoes}{CAPES Theses and Dissertations} &  Academic article summarization \\
\href{https://www.baixelivros.com.br/dominio-publico/}{BaixeLivros Domínio Público} & Literary content and short stories \\
\href{https://huggingface.co/datasets/argilla/FinePersonas-Synthetic-Email-Conversations}{FinePersonas} & Conversational exchanges and dialogue \\
\bottomrule
\end{tabular}%
\end{table}

\newpage
\subsection{Statistics}

We employed three models from the Qwen2.5 family to balance generation quality and computational efficiency. Table~\ref{tab:synth-generator-stats} summarizes the number of examples and total tokens produced by each generator.

\begin{table}[h]
\centering
\caption{Statistics by generator.}
\label{tab:synth-generator-stats}
\begin{tabular}{@{}lrr@{}}
\toprule
\textbf{Generator} & \textbf{Examples} & \textbf{Total Tokens} \\ \midrule
Qwen2.5-32B-Instruct & 7,719,118 & 6,484,418,687 \\
Qwen2.5-14B-Instruct & 1,806,691 & 1,625,788,204 \\
Qwen2.5-7B-Instruct & 1,711,737 & 1,211,544,076 \\ \midrule
\textbf{Total} & \textbf{11,237,546} & \textbf{9,321,750,967} \\
\bottomrule
\end{tabular}
\end{table}

Table~\ref{tab:synth-seed-stats} provides a breakdown of the number of examples and total tokens generated from each seed dataset.

\begin{table}[h]
\centering
\caption{Statistics by seed dataset.}
\label{tab:synth-seed-stats}
\small
\begin{tabular}{@{}lrr@{}}
\toprule
\textbf{Seed Dataset} & \textbf{Examples} & \textbf{Total Tokens} \\ \midrule
Cosmopedia-v2 & 1,896,692 & 1,669,702,784 \\
FineMath & 1,308,713 & 1,361,573,271 \\
Wikipedia & 1,205,667 & 996,465,458 \\
LegalPT & 2,012,941 & 990,403,657 \\
FineWeb-Edu & 1,074,616 & 978,048,012 \\
CodeParrot & 512,331 & 785,518,858 \\
Blogset BR & 820,862 & 667,773,253 \\
StarCoder & 268,968 & 560,829,372 \\
CAPES Theses & 340,792 & 371,964,656 \\
BDTD & 342,428 & 369,321,918 \\
SciELO & 276,109 & 294,413,728 \\
Historinhas & 715,578 & 178,535,564 \\
FinePersonas & 443,729 & 85,060,998 \\
Baixe Livros & 16,370 & 9,712,263 \\
Stanford Encyclopedia of Philosophy & 1,750 & 2,427,175 \\ \midrule
\textbf{Total} & \textbf{11,237,546} & \textbf{9,321,750,967} \\
\bottomrule
\end{tabular}%
\end{table}

\begin{figure}[h]
\centering
\includegraphics[width=0.48\textwidth]{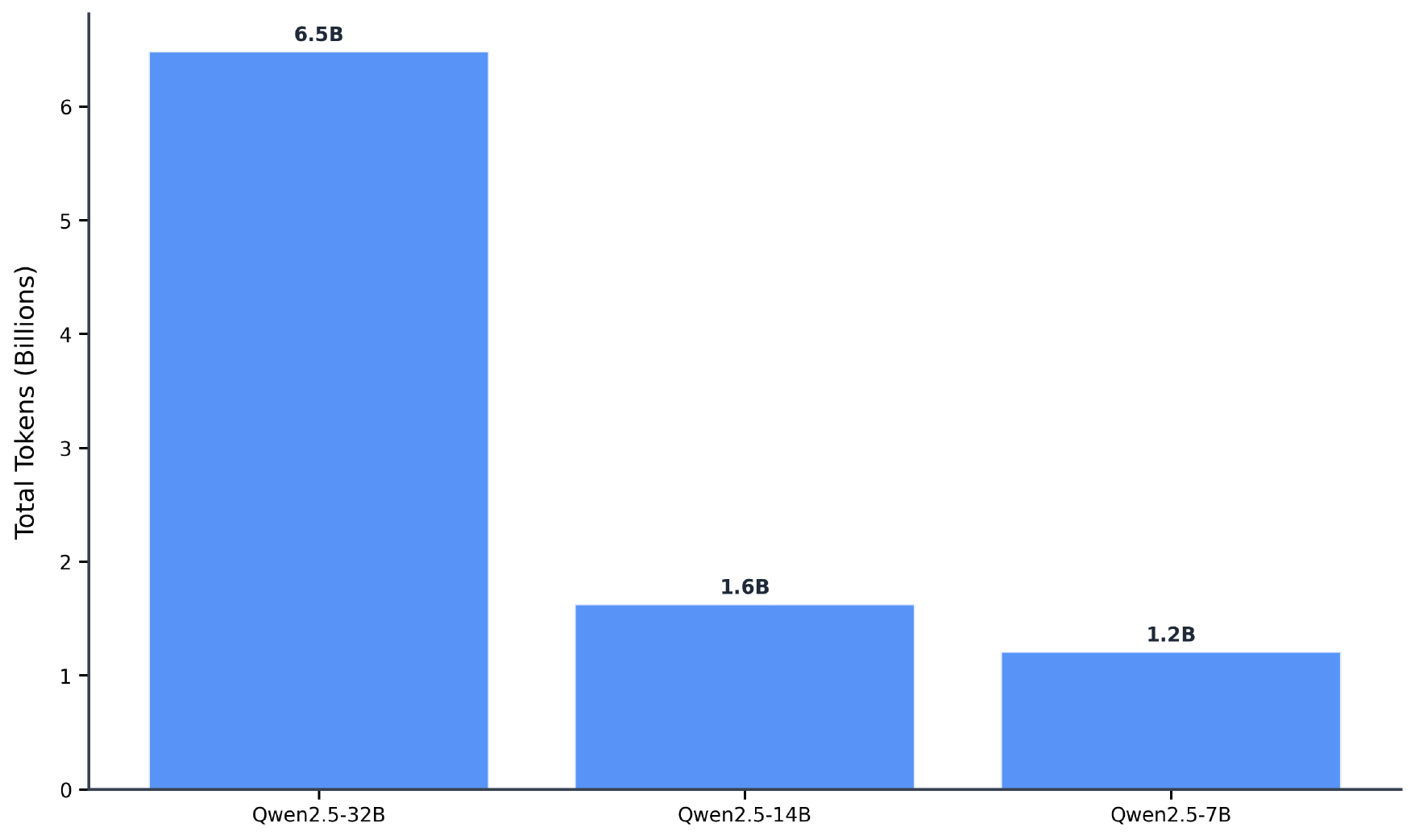}
\hfill
\includegraphics[width=0.48\textwidth]{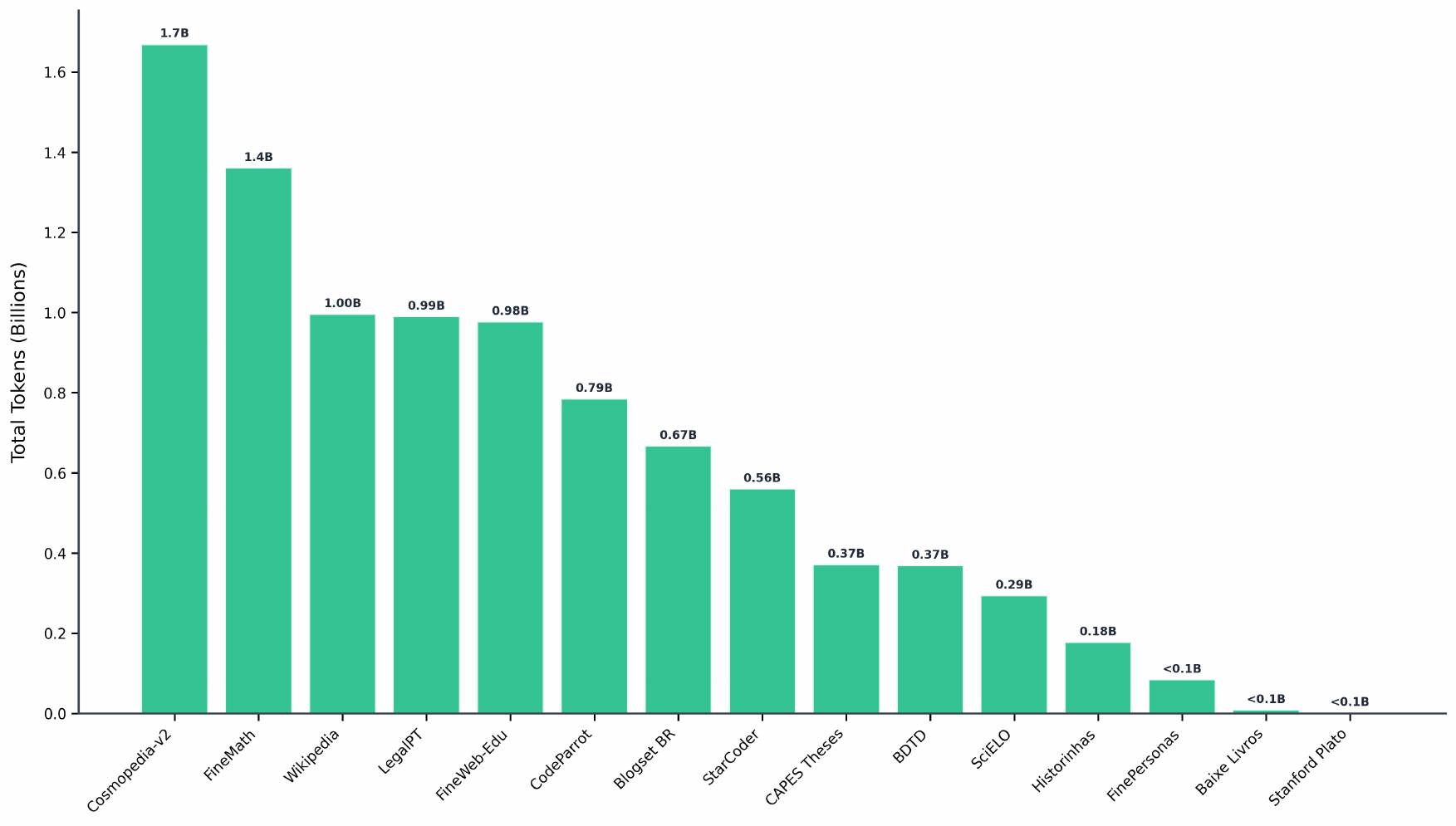}
\caption{Token distribution. (a) Distribution across the three Qwen2.5 generator models. (b) Distribution across the 15 seed datasets.}
\label{fig:synth-token-distributions}
\end{figure}

Figure~\ref{fig:synth-token-distributions} provides visual representations of the token distributions across generators and seed datasets.

\newpage
\section{Data Ablation Study: Details}
\label{appendix:ablations}

This appendix provides detailed experimental configurations, per-benchmark results, and comparisons for the \gigaverbo ablation study described in Section~\ref{sec:pretraining_data}.

\subsection{Model Architecture and Training Configuration}

Table~\ref{tab:ablation-hyperparams} presents the complete model architecture and training configuration used for all four ablation experiments. These values correspond to model configuration settings for a Llama model/architecture.\footnote{\href{https://huggingface.co/docs/transformers/main/en/model_doc/llama}{huggingface.co/docs/transformers/main/en/model\_doc/llama}.} All models share the same configuration, differing only in the training data mixture.

\begin{table}[h]
\centering
\caption{Model Architecture and Training Configuration.}
\label{tab:ablation-hyperparams}
\begin{tabular}{lll}
\toprule
\textbf{Category} & \textbf{Parameter} & \textbf{Value} \\
\midrule
\multirow{14}{*}{\textbf{Model Architecture}} & Vocabulary Size & 49,152 \\
 & Hidden Layers & 28 \\
 & Attention Heads & 16 \\
 & Key-Value Heads & 8 \\
 & Head Dimension & 128 \\
 & Hidden Size & 2,048 \\
 & Intermediate Size & 6,144 \\
 & Max Position Embeddings & 4,096 \\
 & Tie Word Embeddings & True \\
 & Hidden Activation & SwiGLU \\
 & RoPE Theta & 50,000.0 \\
 & RMS Norm Epsilon & 1e-6 \\
\midrule
\multirow{11}{*}{\textbf{Training Configuration}} & Parallelism Strategy & DDP \\
 & Activation Checkpointing & False \\
 & Total Batch Size & 2,097,152 \\
 & Warmup Steps & 2,000 \\
 & Optimizer(s) & AdamW+Muon \\
 & AdamW Max Learning Rate & 0.0008 \\
 & Muon Max Learning Rate & 0.008 \\
 & Min Learning Rate & 0.0 \\
 & Weight Decay & 0.1 \\
 & Beta1 & 0.9 \\
 & Beta2 & 0.95 \\
 & Epsilon & 1e-8 \\
 & LR Decay Type & Cosine \\
 & Max Steps & 22,000 \\
 & Max Gradient Norm & 1.0 \\
\bottomrule
\end{tabular}
\end{table}

\subsection{Detailed Results}

\begin{figure}[h]
\centering
\begin{subfigure}[b]{0.48\textwidth}
    \centering
    \includegraphics[width=\textwidth]{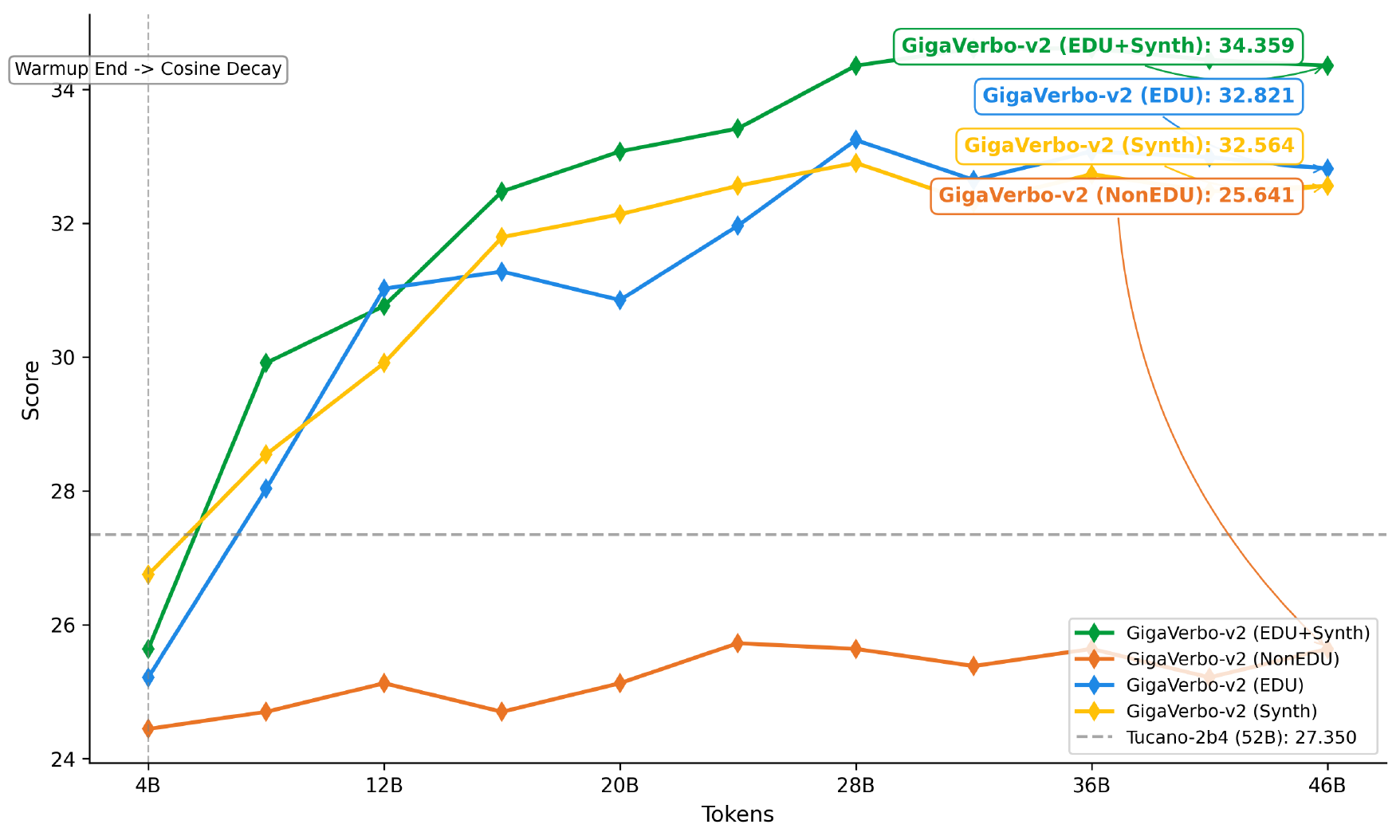}
    \caption{ARC Challenge}
    \label{fig:arc}
\end{subfigure}
\hfill
\begin{subfigure}[b]{0.48\textwidth}
    \centering
    \includegraphics[width=\textwidth]{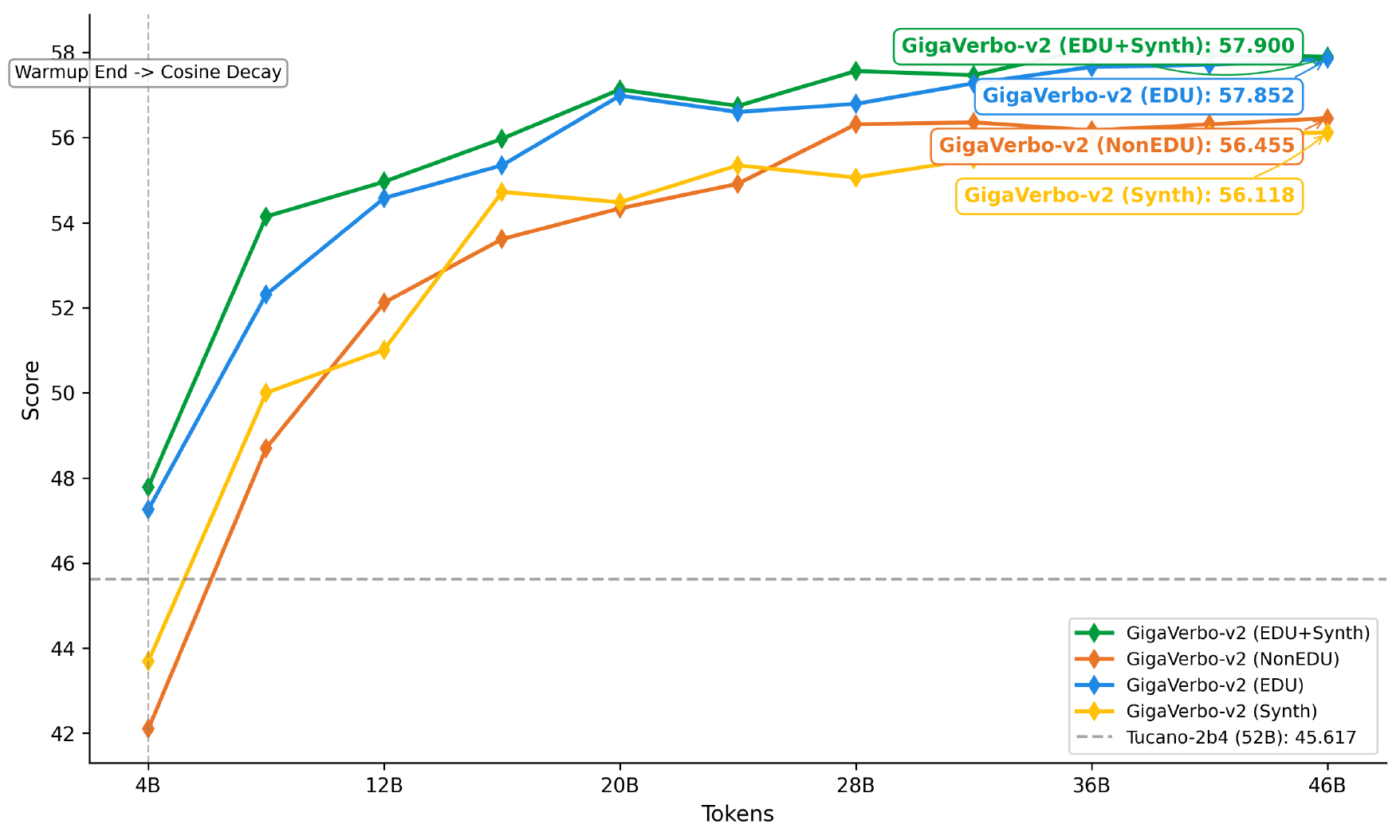}
    \caption{Calame}
    \label{fig:calame}
\end{subfigure}

\vspace{0.5cm}

\begin{subfigure}[b]{0.48\textwidth}
    \centering
    \includegraphics[width=\textwidth]{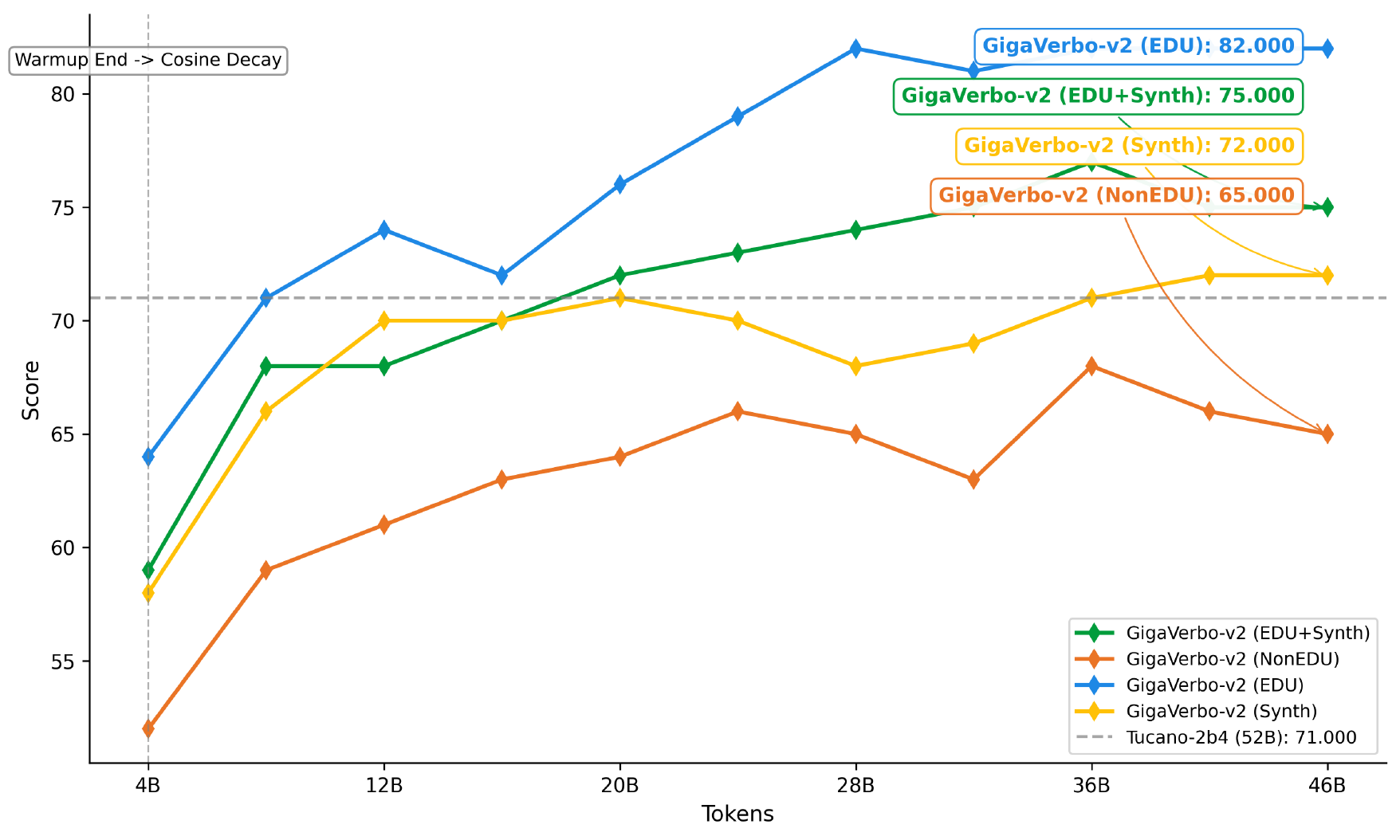}
    \caption{Global PIQA}
    \label{fig:global-piqa}
\end{subfigure}
\hfill
\begin{subfigure}[b]{0.48\textwidth}
    \centering
    \includegraphics[width=\textwidth]{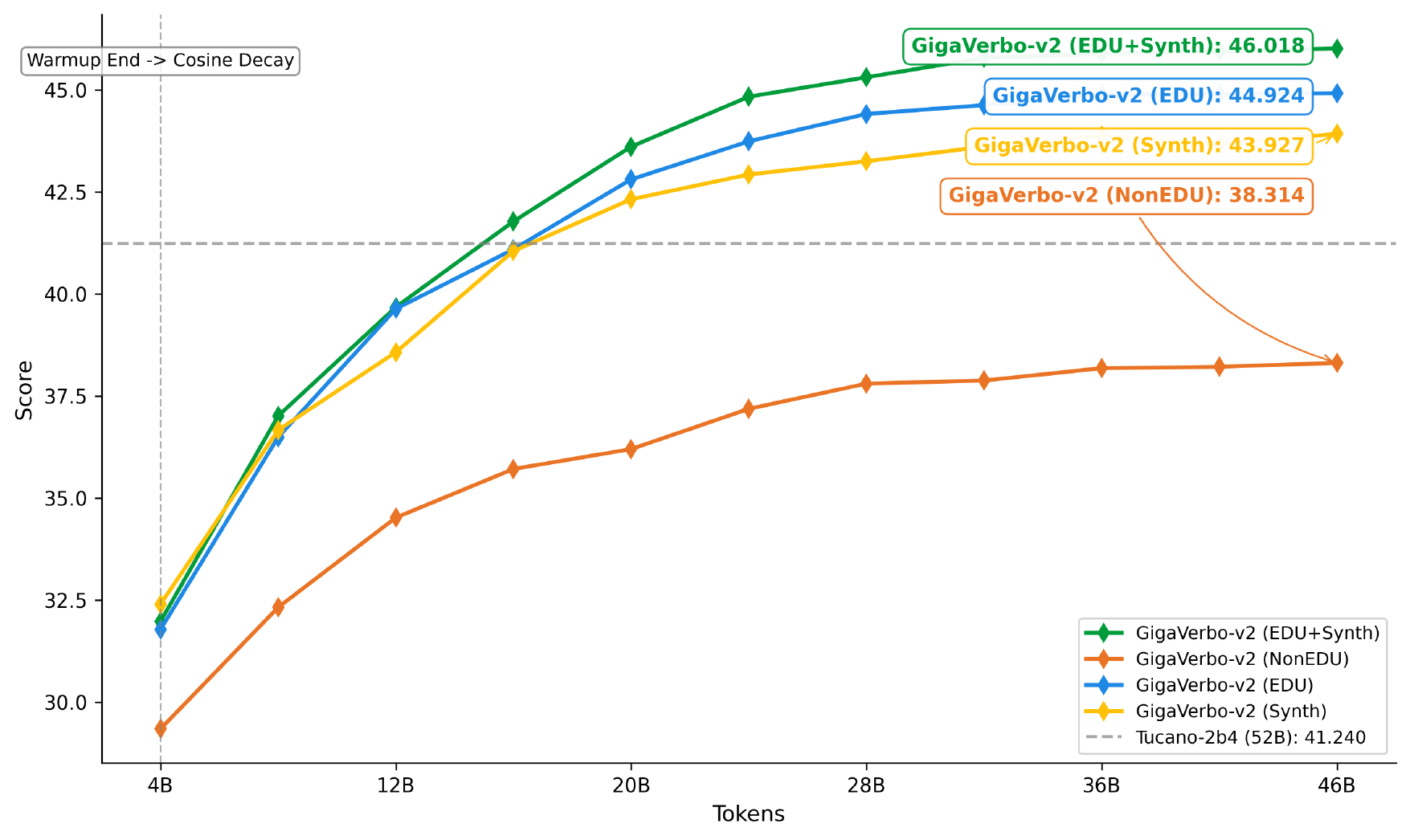}
    \caption{HellaSwag}
    \label{fig:hellaswag}
\end{subfigure}

\vspace{0.5cm}

\begin{subfigure}[b]{0.48\textwidth}
    \centering
    \includegraphics[width=\textwidth]{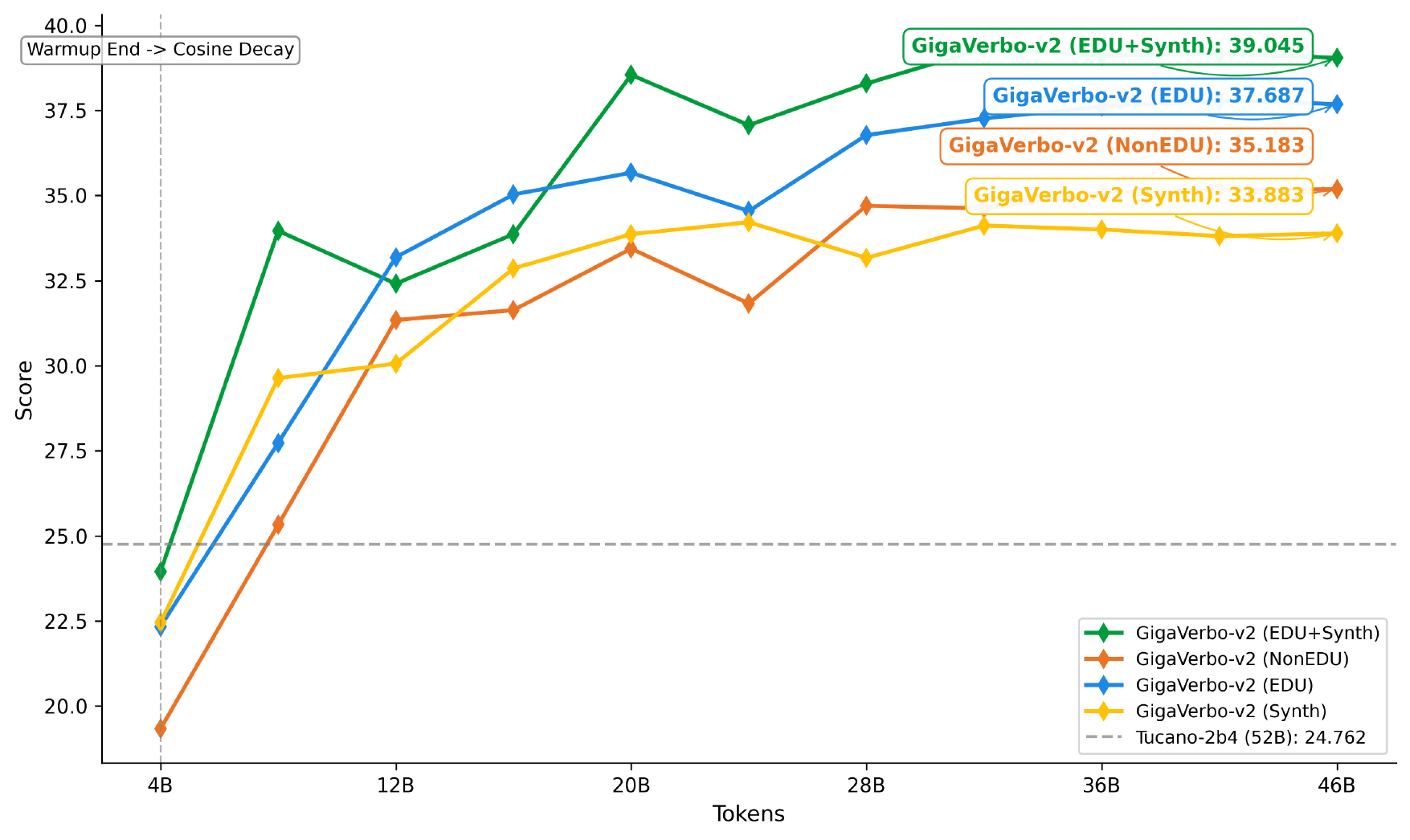}
    \caption{Lambada}
    \label{fig:lambada}
\end{subfigure}

\caption{Per-benchmark performance comparisons across the four data mixtures and the Tucano-2b4 baseline. The Edu and Edu+Synth mixtures consistently outperform the Non-Edu mixture across most tasks, with particularly strong gains on ARC Challenge and HellaSwag.}
\label{fig:ablation-detailed-benchmarks}
\end{figure}

Figure~\ref{fig:ablation-detailed-benchmarks} presents per-benchmark performance comparisons across the four data mixtures and the Tucano-2b4 baseline.

\paragraph{ARC Challenge.} ARC is perhaps where we observe the most pronounced differences among data mixtures. The Edu+Synth model achieves a score (34.4\%) that substantially outperforms the Non-Edu mixture (25.6\%), and also surpasses both the Edu-only (32.8\%) and Synth-only (32.6\%) mixtures.

\paragraph{Calame.} For Calame, which focuses on predicting the last word of a passage, we see a less pronounced advantage from the Edu portion of GigaVerbo-v2. We hypothesize that this is due to the nature of the benchmark, which may not benefit as much from educational content as other benchmarks focused on reasoning or commonsense understanding.

\paragraph{Global PIQA.} On Global PIQA, we observe that the addition of synthetic data to the educational portion of GigaVerbo-v2 yields a decrease in performance (75.0\% vs 82.0\%). However, both models still significantly outperform the Non-Edu model (65.0\%). Given that the synthetic data was not specifically tailored for physical commonsense reasoning tasks (or cultural knowledge tasks in general), this result is not entirely surprising. It does, however, highlight some shortcomings of our synthetic dataset that we aim to address in future iterations.

\paragraph{HellaSwag.} HellaSwag is yet another benchmark where the Edu+Synth mixture outperforms the Non-Edu model by a significant margin (46.0\% vs 38.3\%), and the Edu-only mixture by a smaller margin (46.0\% vs 44.9\%). This reinforces the finding that educational and synthetic data jointly improve performance on commonsense reasoning tasks.

\paragraph{Lambada.} Lambada is the only benchmark where the Non-Edu mixture outperforms the Synth-only mixture (35.2\% vs 33.9\%). However, the Edu+Synth mixture still achieves the best performance (39.0\%), followed closely by the Edu-only mixture (37.7\%). Again, we hypothesize that this is due to the nature of the benchmark, which focuses on predicting the last word of a passage, a task that may not benefit as much from the repetitive patterns found in our synthetic corpus.

\subsection{Baseline Comparison}

Table~\ref{tab:ablation-comprehensive-comparison} presents a comparison of the ablation models with other relevant Portuguese and multilingual baselines. Models are ranked by Normalized Performance Metric (NPM) \citep{Pires_2023} (see Section~\ref{sec:building-evaluation-harness}), computed as the average of normalized scores across all five benchmarks.

\begin{table}[h]
\centering
\caption{Comprehensive comparison of our ablation models with other Portuguese and multilingual language models.}
\label{tab:ablation-comprehensive-comparison}
\resizebox{\textwidth}{!}{%
\begin{tabular}{lrrrrrrrrr}
\toprule
\textbf{Model} & \textbf{Param.} & \textbf{Tokens} & \textbf{NPM} & \textbf{ARC} & \textbf{Calame} & \textbf{Global PIQA} & \textbf{HellaSwag} & \textbf{Lambada} \\
\midrule
GigaVerbo-v2 (Edu) & 1.5B & 46B & \textbf{39.31} & 32.80 & 57.90 & \textbf{82.00} & 44.90 & 37.70 \\
Curió-1.1b & 1.1B & 1.15T & 39.16 & 30.40 & \textbf{59.20} & 75.00 & \textbf{49.50} & \textbf{46.70} \\
GigaVerbo-v2 (Edu+Synth) & 1.5B & 46B & 37.49 & \textbf{34.40} & 57.90 & 75.00 & 46.00 & 39.00 \\
Curió-edu-1b1 & 1.1B & 1.02T & 34.77 & 32.20 & 54.90 & 69.00 & 46.30 & 42.90 \\
GigaVerbo-v2 (Synth) & 1.5B & 46B & 33.86 & 32.60 & 56.10 & 72.00 & 43.90 & 33.90 \\
Tucano-2b4 & 2.4B & 500B & 33.55 & 30.40 & 50.30 & 73.00 & 48.80 & 32.40 \\
Tucano-1b1 & 1.1B & 250B & 29.12 & 30.10 & 48.90 & 68.00 & 44.10 & 28.40 \\
Llama-3.2-1B (9T) & 1.0B & 9T & 28.32 & 31.70 & 50.00 & 55.00 & 45.30 & 45.60 \\
GigaVerbo-v2 (Non-Edu) & 1.5B & 46B & 28.05 & 25.60 & 56.50 & 65.00 & 38.30 & 35.20 \\
GlorIA-1.3B (35B) & 1.3B & 35B & 27.27 & 26.40 & 54.70 & 64.00 & 36.40 & 36.70 \\
Carvalho\_pt-gl-1.3B & 1.3B & 31B & 26.75 & 27.00 & 53.40 & 63.00 & 38.50 & 33.60 \\
\bottomrule
\end{tabular}
}
\end{table}

\newpage
\section{Tokenization: Details}
\label{appendix:tokenization}

This section provides detailed evaluation results for our tokenizer across multiple configurations, including comparisons with contemporary multilingual and Portuguese-adapted models.

\subsection{Evaluations on Portuguese-Only Text}

Table~\ref{tab:tokenizer-eval-pt-detailed} presents tokenization metrics on a corpus of 600,000 Portuguese words.

\begin{table}[h]
\centering
\small
\caption{Evaluation results on Portuguese text (600,000 words).}
\label{tab:tokenizer-eval-pt-detailed}
\begin{tabular}{lrrrr}
\toprule
\textbf{Model} & \textbf{Vocab Size} & \textbf{SF} & \textbf{PCW} & \textbf{Chars/Token} \\
\midrule
Tucano2-0.6B-Base   & 49,152  & 1.51 & 0.56 & 2.88 \\
GlorIA-1.3B         & 50,258  & 1.58 & 0.47 & 3.04 \\
Llama-3.2-1B        & 128,256 & 1.91 & 0.61 & 2.72 \\
SmolLM3-3B-Base     & 128,256 & 1.91 & 0.61 & 2.72 \\
OLMo-2-0425-1B      & 100,278 & 1.92 & 0.61 & 2.71 \\
Qwen3-0.6B          & 151,669 & 1.93 & 0.61 & 2.68 \\
Curi{\'o}-1.1b          & 32,000  & 2.17 & 0.55 & 3.09 \\
granite-3.3-2b-base & 49,152  & 2.29 & 0.63 & 2.52 \\
\bottomrule
\end{tabular}
\end{table}

\subsection{Evaluation on Mixed-Language Text}

To assess tokenizer performance on our intended pretraining mixture, we evaluated all tokenizers on a mixed corpus of 600,000 words comprising Portuguese, English, and code snippets in proportions representative of our training data. Results are presented in Table~\ref{tab:tokenizer-eval-mixed}.

\begin{table}[h]
\centering
\small
\caption{Evaluation results on mixed data (Portuguese, English, and code; 600,000 words).}
\label{tab:tokenizer-eval-mixed}
\begin{tabular}{lrrrr}
\toprule
\textbf{Model} & \textbf{Vocab Size} & \textbf{SF} & \textbf{PCW} & \textbf{Chars/Token} \\
\midrule
Tucano2-0.6B-Base   & 49,152  & 1.48 & 0.48 & 2.94 \\
Llama-3.2-1B        & 128,256 & 1.59 & 0.45 & 3.09 \\
SmolLM3-3B-Base     & 128,256 & 1.59 & 0.45 & 3.09 \\
OLMo-2-0425-1B      & 100,278 & 1.59 & 0.45 & 3.08 \\
Qwen3-0.6B          & 151,669 & 1.60 & 0.46 & 3.05 \\
Curi{\'o}-1.1b          & 32,000  & 1.84 & 0.40 & 3.57 \\
GlorIA-1.3B         & 50,258  & 1.87 & 0.53 & 2.73 \\
granite-3.3-2b-base & 49,152  & 1.91 & 0.48 & 2.86 \\
\bottomrule
\end{tabular}
\end{table}

\subsection{Computational Cost Analysis}

The reduced computational cost of our tokenizer yields substantial computational savings during training. Following the compute-estimation methodology of \citet{ali2024tokenizer}, we estimate the FLOPs required to train a model under a fixed token budget of 500 billion tokens. The compute cost $C_{\text{dataset}}$ is approximated as:

\begin{equation}
C_{\text{dataset}} = 96 \cdot l \cdot h^2 \cdot \left(1 + \frac{s}{6h} + \frac{V}{16lh}\right) \cdot D
\label{eq:training_cost}
\end{equation}

where:
\begin{itemize}
    \item $C_{\text{dataset}}$ is the total computational cost (in FLOPs) to train on a given token count,
    \item $l$ is the number of transformer layers,
    \item $h$ is the hidden (embedding) dimension of the model,
    \item $s$ is the input sequence length,
    \item $V$ is the vocabulary size, and
    \item $D$ is the total number of tokens in the training dataset.
\end{itemize}

The expression inside the parentheses in Equation~\eqref{eq:training_cost} decomposes the per-token cost into three components:
\begin{enumerate}
    \item The constant term $1$ captures the dominant cost of the feed-forward network and the attention projection matrices within each layer, which scales as $\mathcal{O}(h^2)$.
    \item The term $\dfrac{s}{6h}$ accounts for the self-attention mechanism (i.e., the $QK^\top$ and attention-value products), whose cost grows linearly with the sequence length $s$ relative to the hidden size $h$.
    \item The term $\dfrac{V}{16lh}$ represents the cost of the input embedding and output (unembedding) projection to the vocabulary, which is amortised across layers and scales with the vocabulary size $V$.
\end{enumerate}

The leading factor of $96$ arises from the combined cost of the forward and backward passes (approximately $3\times$ the forward pass cost) applied to the matrix multiplications within each layer~\cite{ali2024tokenizer}. Note that the batch size $B$ cancels when converting from the per-step cost $C$ to the per-token cost $C_{\text{token}} = C / (B \cdot s)$, making $C_{\text{dataset}}$ independent of the training batch configuration.

Table~\ref{tab:tokenizer-compute-cost} summarizes the estimated compute costs for training a model (28 layers, hidden size 1,536, maximum sequence length 4,096, batch size 512) on 500 billion tokens using different tokenizers. Our tokenizer requires approximately $7.26 \times 10^{21}$ FLOPs, compared to $1.02 \times 10^{22}$ FLOPs for Qwen3 and $1.10 \times 10^{22}$ FLOPs for Granite, corresponding to compute savings of roughly 30\% and 34\%, respectively. These reductions compound across multiple training runs, ablation studies, and continual pretraining phases, highlighting the importance of tokenizer efficiency in resource-constrained research scenarios.

\begin{table}[h]
\centering
\small
\caption{Estimated compute cost for training a $\sim$1B-parameter model on 500 billion tokens using different tokenizers.}
\label{tab:tokenizer-compute-cost}
\begin{tabular}{lrrr}
\toprule
\textbf{Tokenizer} & \textbf{SF} & \textbf{Vocab Size} & \textbf{Est. Compute (FLOPs)} \\
\midrule
Tucano2-0.6B-Base   & 1.51 & 49,152  & 7.26$\times$10$^{21}$ \\
GlorIA-1.3B         & 1.58 & 50,258  & 7.62$\times$10$^{21}$ \\
Llama-3.2-1B        & 1.91 & 128,256 & 9.89$\times$10$^{21}$ \\
SmolLM3-3B-Base     & 1.91 & 128,256 & 9.89$\times$10$^{21}$ \\
OLMo-2-0425-1B      & 1.92 & 100,278 & 9.67$\times$10$^{21}$ \\
Qwen3-0.6B          & 1.93 & 151,669 & 1.02$\times$10$^{22}$ \\
Curi{\'o}-1.1b          & 2.17 & 32,000  & 1.03$\times$10$^{22}$ \\
granite-3.3-2b-base & 2.29 & 49,152  & 1.10$\times$10$^{22}$ \\
\bottomrule
\end{tabular}
\end{table}

\newpage
\section{Evaluation Framework: Details}
\label{appendix:evaluations}

This appendix presents detailed results, visualizations, and the methodology for our analysis.

\subsection{Garcia et al. evaluation suite}
\label{appendix:garcia-harness-analysis}

\subsubsection{Evaluation Configuration}

Table~\ref{tab:garcia-config} summarizes the configuration of benchmarks in the Portuguese evaluation suite from \cite{open-pt-llm-leaderboard}.

\begin{table}[!htbp]
\centering
\small
\caption{Configuration of benchmarks in Garcia et al.'s evaluation suite. All benchmarks are implemented as generative (free-form generation) tasks.}
\label{tab:garcia-config}
\begin{tabular}{lllr}
\toprule
\textbf{Benchmark} & \textbf{n-shot} & \textbf{Type} & \textbf{Baseline (\%)} \\
\midrule
ENEM            & 3-shot  & MC-Q\&A         & 20.0  \\
BLUEX           & 3-shot  & MC-Q\&A         & 22.5  \\
OAB Exams       & 3-shot  & MC-Q\&A         & 25.0  \\
ASSIN2 RTE      & 15-shot & Entailment      & 50.0  \\
ASSIN2 STS      & 15-shot & Similarity      & 0.0   \\
FAQUAD NLI      & 15-shot & Entailment      & 45.6  \\
HateBR          & 25-shot & Classification  & 50.0  \\
PT Hate Speech  & 25-shot & Classification  & 47.9  \\
TweetSentBR     & 25-shot & Classification  & 32.8  \\
\bottomrule
\end{tabular}
\end{table}

\subsubsection{Baseline Comparison}

Table~\ref{tab:baseline-surpass-old} shows the number of training tokens required for SmolLM3-3B to surpass each benchmark's baseline by 5 percentage points.

\begin{table}[!htbp]
\centering
\small
\caption{Training tokens required for SmolLM3-3B to surpass baselines on Garcia et al.'s evaluation suite.}
\label{tab:baseline-surpass-old}
\begin{tabular}{lrrrrrr}
\toprule
\textbf{Benchmark} & \textbf{Baseline} & \textbf{Threshold} & \textbf{First Step} & \textbf{Tokens (B)} & \textbf{Score} & \textbf{Improvement} \\
\midrule
ASSIN2 RTE     & 50.00 & 55.00 & 80,000  & 188   & 71.08 & 21.10 \\
BLUEX          & 22.50 & 27.50 & 320,000 & 755   & 27.68 & 5.20  \\
ENEM           & 20.00 & 25.00 & 320,000 & 755   & 25.05 & 5.10  \\
FAQUAD NLI     & 45.60 & 50.60 & 40,000  & 94    & 53.74 & 8.10  \\
HateBR         & 50.00 & 55.00 & 40,000  & 94    & 58.75 & 8.80  \\
OAB            & 25.00 & 30.00 & 440,000 & 1,038 & 30.66 & 5.70  \\
PT Hate Speech & 47.90 & 52.90 & 40,000  & 94    & 55.55 & 7.70  \\
TweetSentBR    & 32.80 & 37.80 & 80,000  & 188   & 49.18 & 16.40 \\
\bottomrule
\end{tabular}
\end{table}

\newpage
\subsubsection{Signal Analysis}

Table~\ref{tab:garcia-signal-quality} presents the signal-quality metrics for each benchmark. Benchmarks such as ASSIN2 RTE, FAQUAD NLI, and HateBR exhibit high volatility and low signal-to-noise ratios.

\begin{table}[!htbp]
\centering
\small
\caption{Signal-quality metrics for Garcia et al.'s evaluation suite.}
\label{tab:garcia-signal-quality}
\begin{tabular}{lrrr}
\toprule
\textbf{Benchmark} & \textbf{Mean Abs Change} & \textbf{SNR} & \textbf{Spearman} \\
\midrule
ASSIN2 RTE     & 0.127 & 4.953  & 0.606  \\
ASSIN2 STS     & 0.085 & 2.492  & 0.792  \\
BLUEX          & 0.024 & 5.993  & 0.859  \\
ENEM           & 0.037 & 4.494  & 0.878  \\
FAQUAD NLI     & 0.025 & 12.155 & 0.032  \\
HateBR         & 0.092 & 6.266  & 0.290  \\
OAB            & 0.021 & 8.242  & 0.810  \\
PT Hate Speech & 0.109 & 4.710  & 0.206  \\
TweetSentBR    & 0.071 & 7.530  & -0.080 \\
\bottomrule
\end{tabular}
\end{table}

\subsubsection{Per-Benchmark Results}

\begin{figure}[!htbp]
\centering
\includegraphics[width=0.75\linewidth]{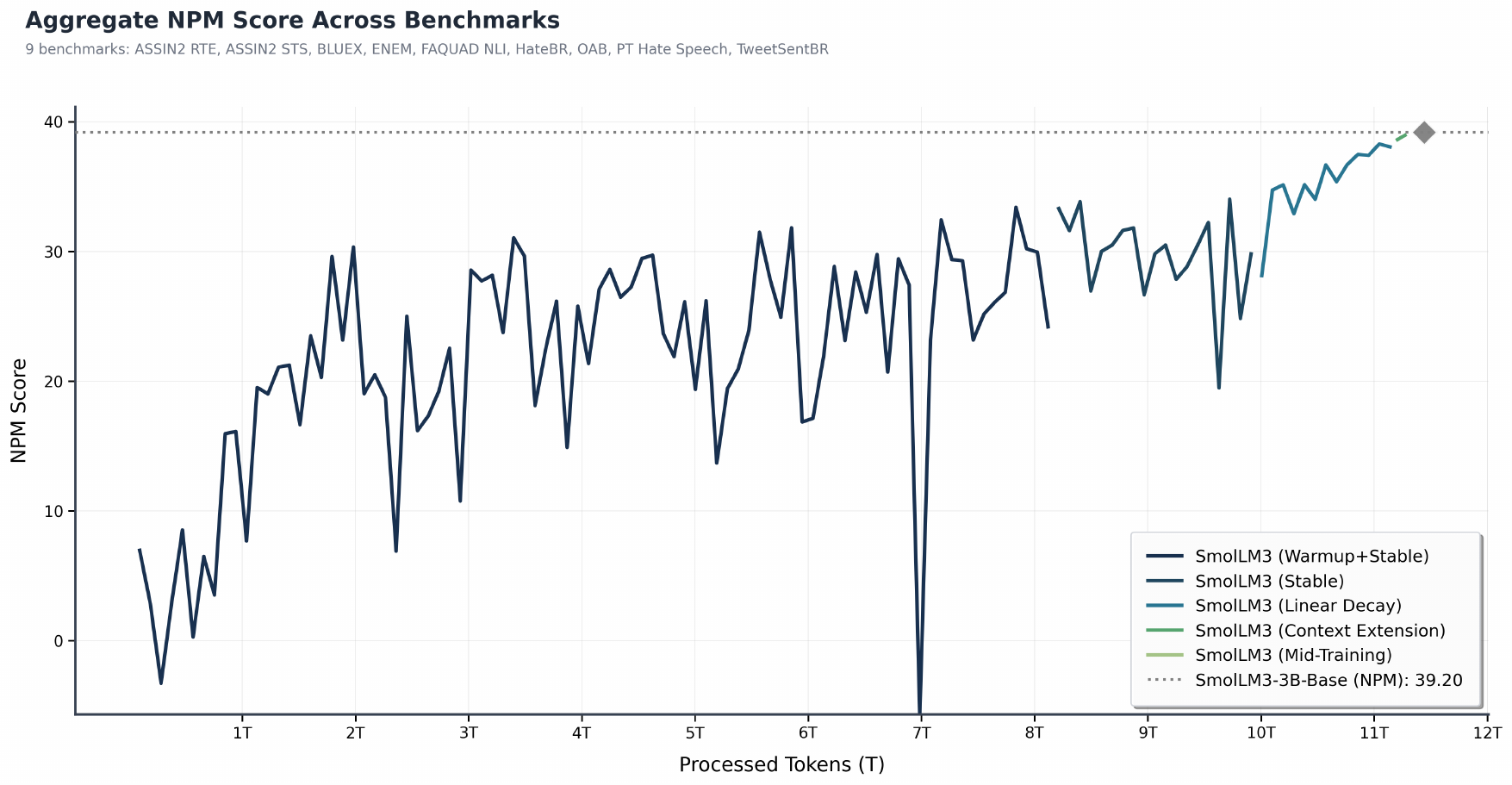}
\caption{NPM aggregated score for SmolLM3-3b on Garcia et al.'s evaluation suite. The score shows an upward trend but with substantial volatility across checkpoints.}
\label{fig:garcia-npm}
\end{figure}

\begin{figure}[!htbp]
\centering
\begin{tabular}{ccc}
\includegraphics[width=0.30\linewidth]{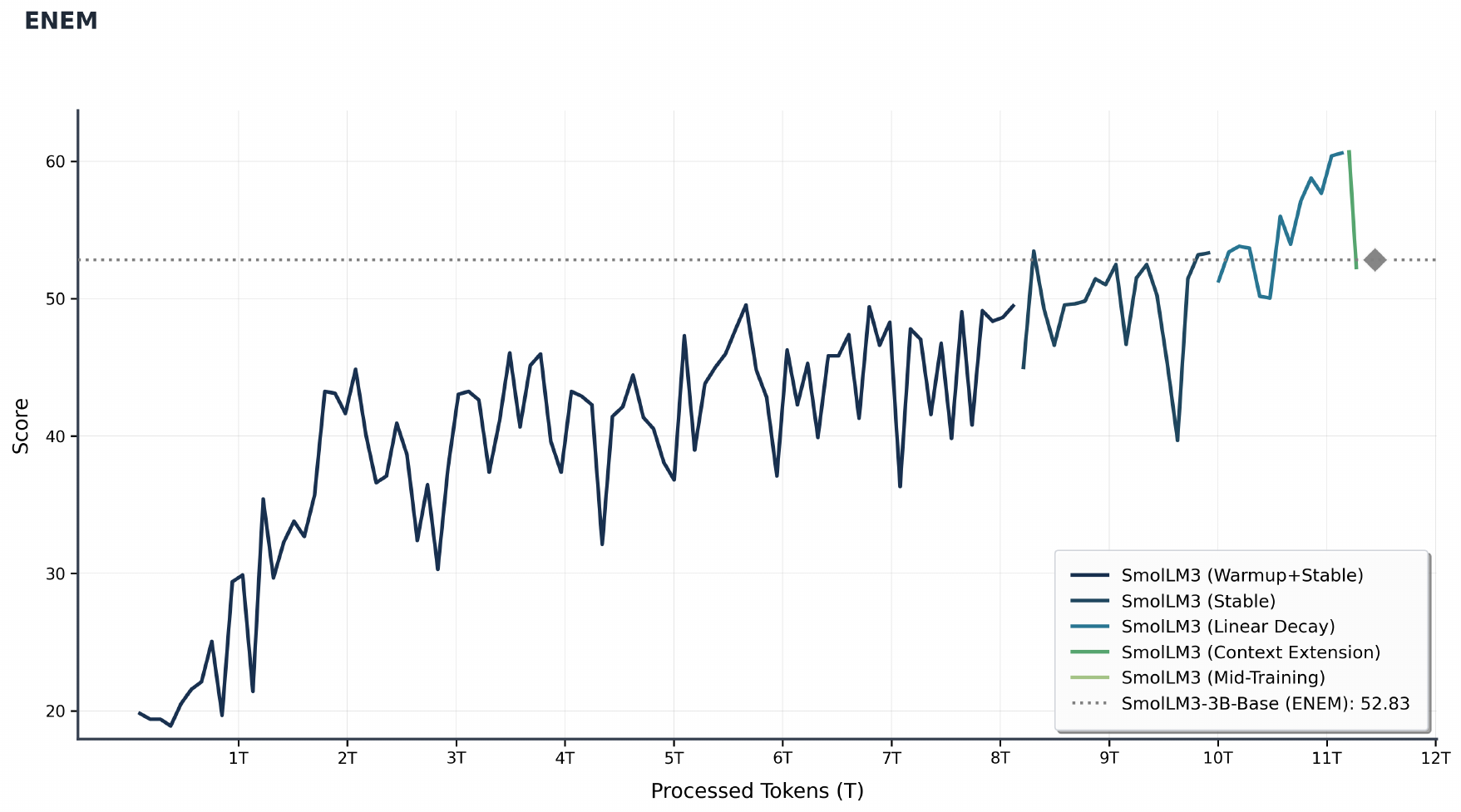} &
\includegraphics[width=0.30\linewidth]{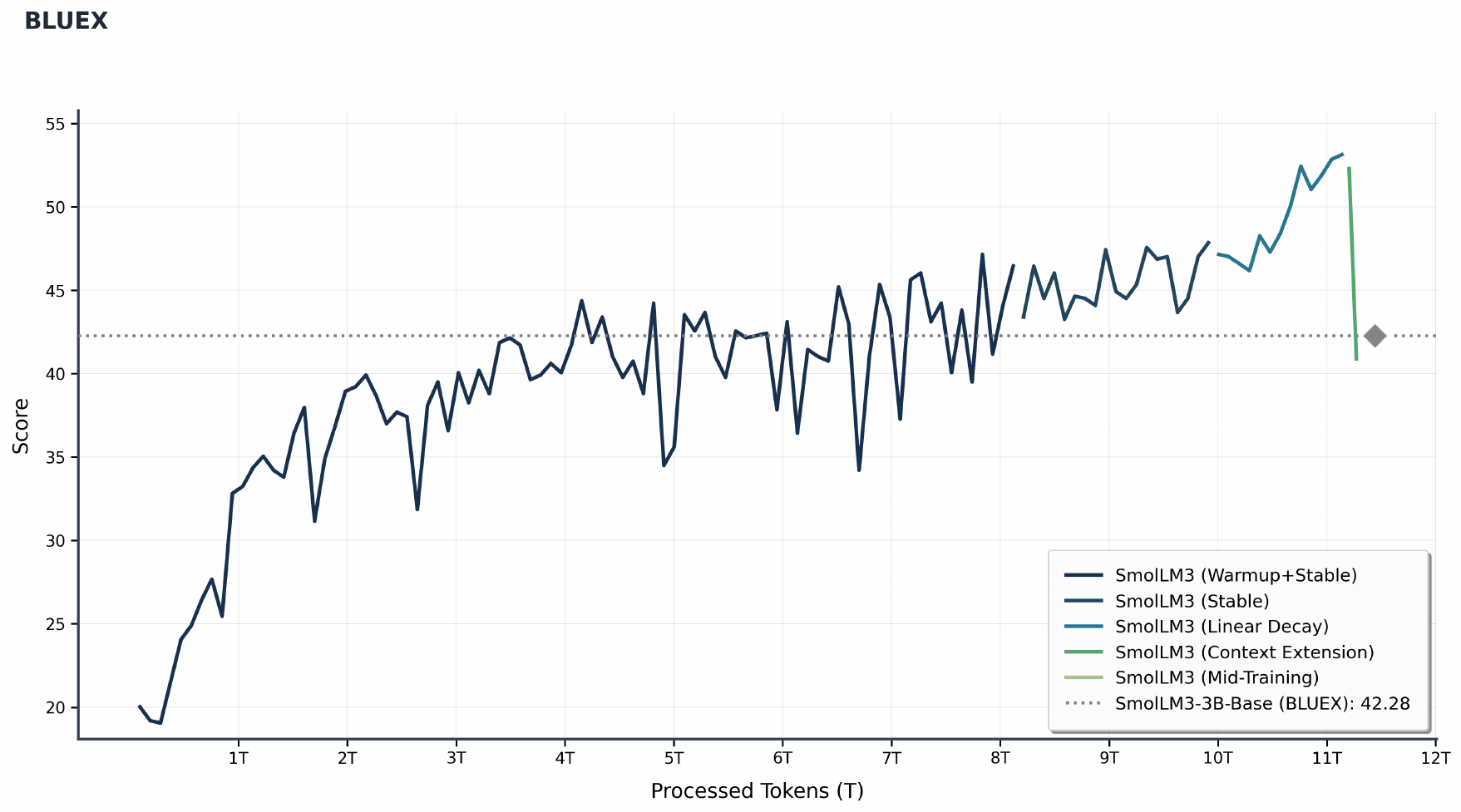} &
\includegraphics[width=0.30\linewidth]{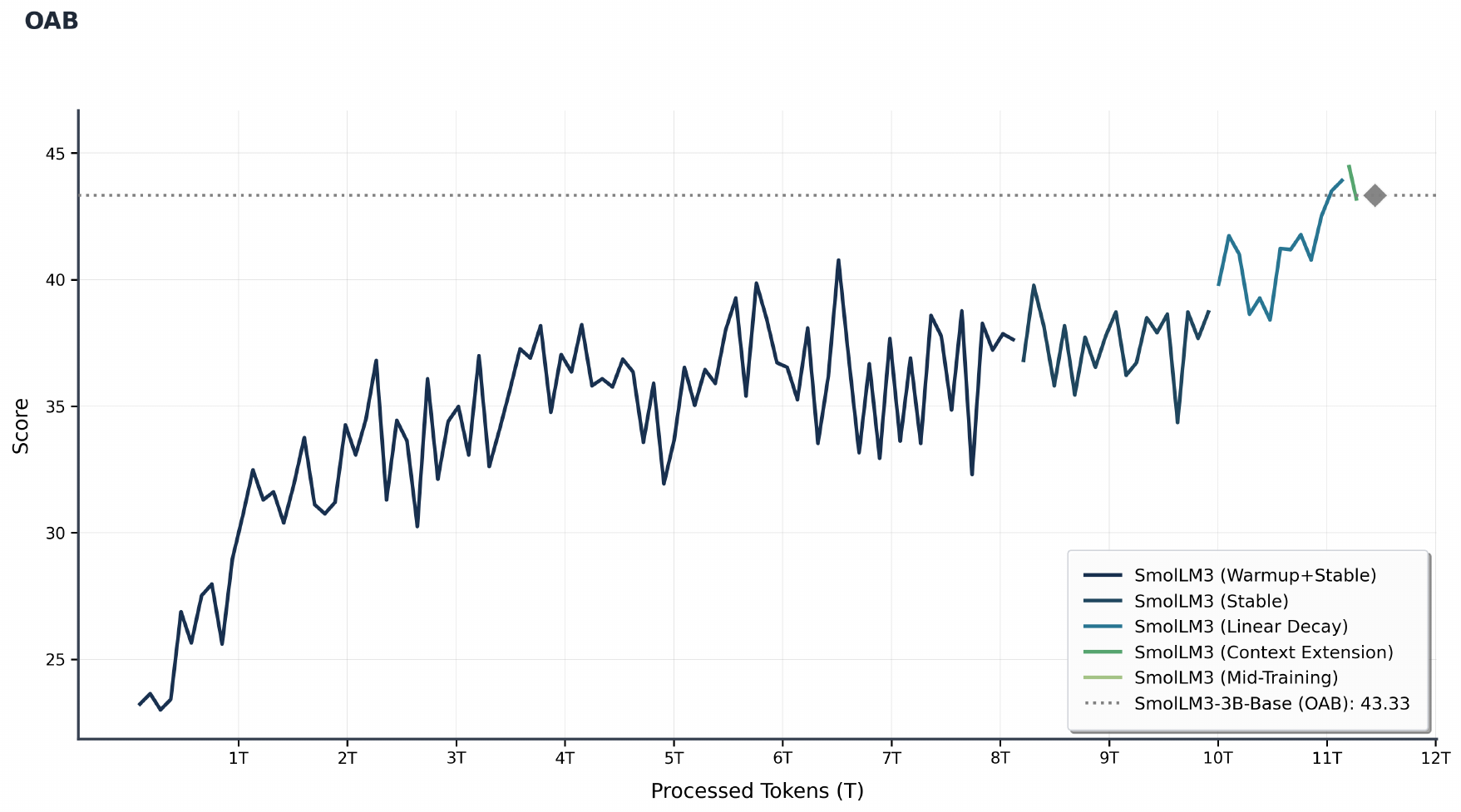} \\
\includegraphics[width=0.30\linewidth]{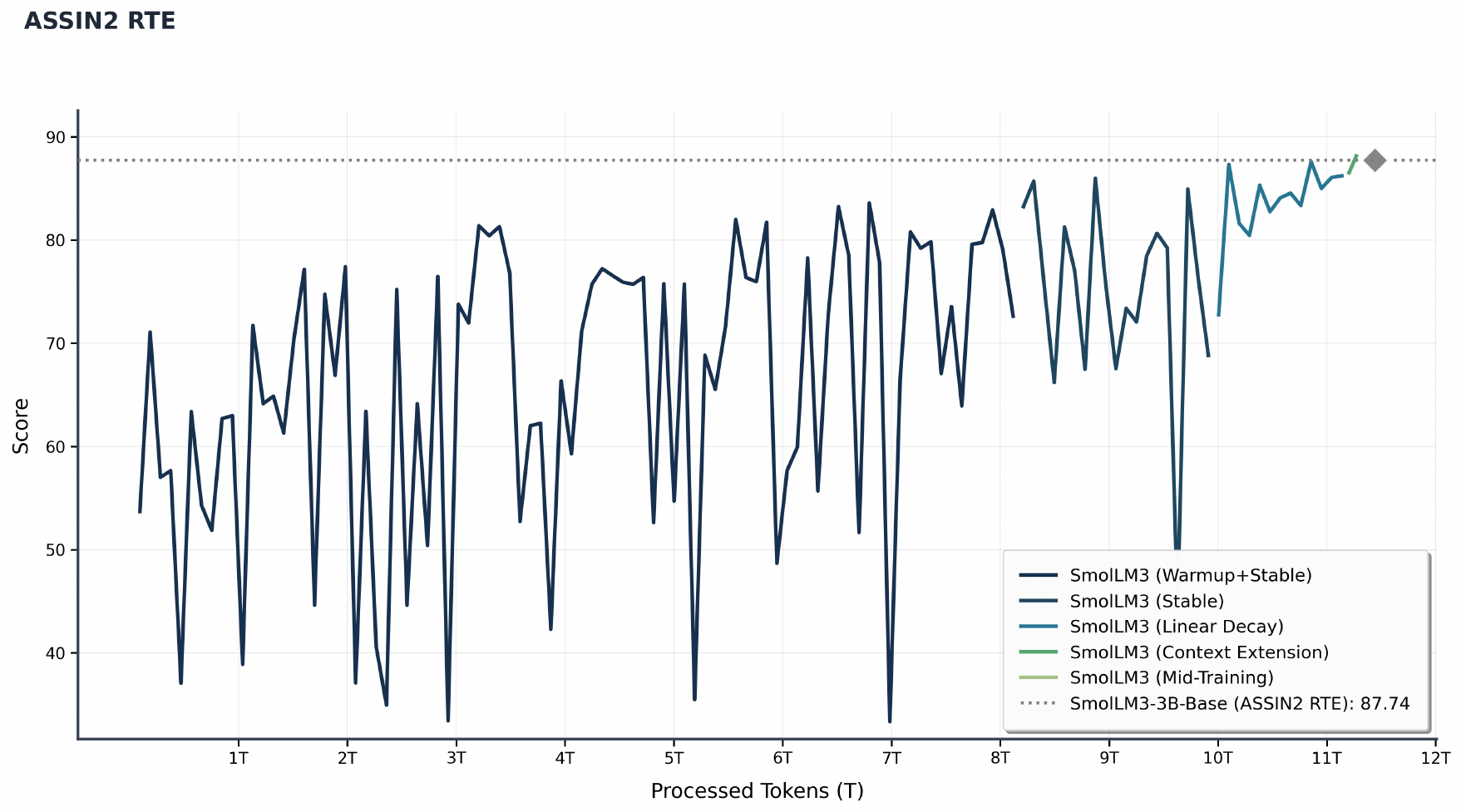} &
\includegraphics[width=0.30\linewidth]{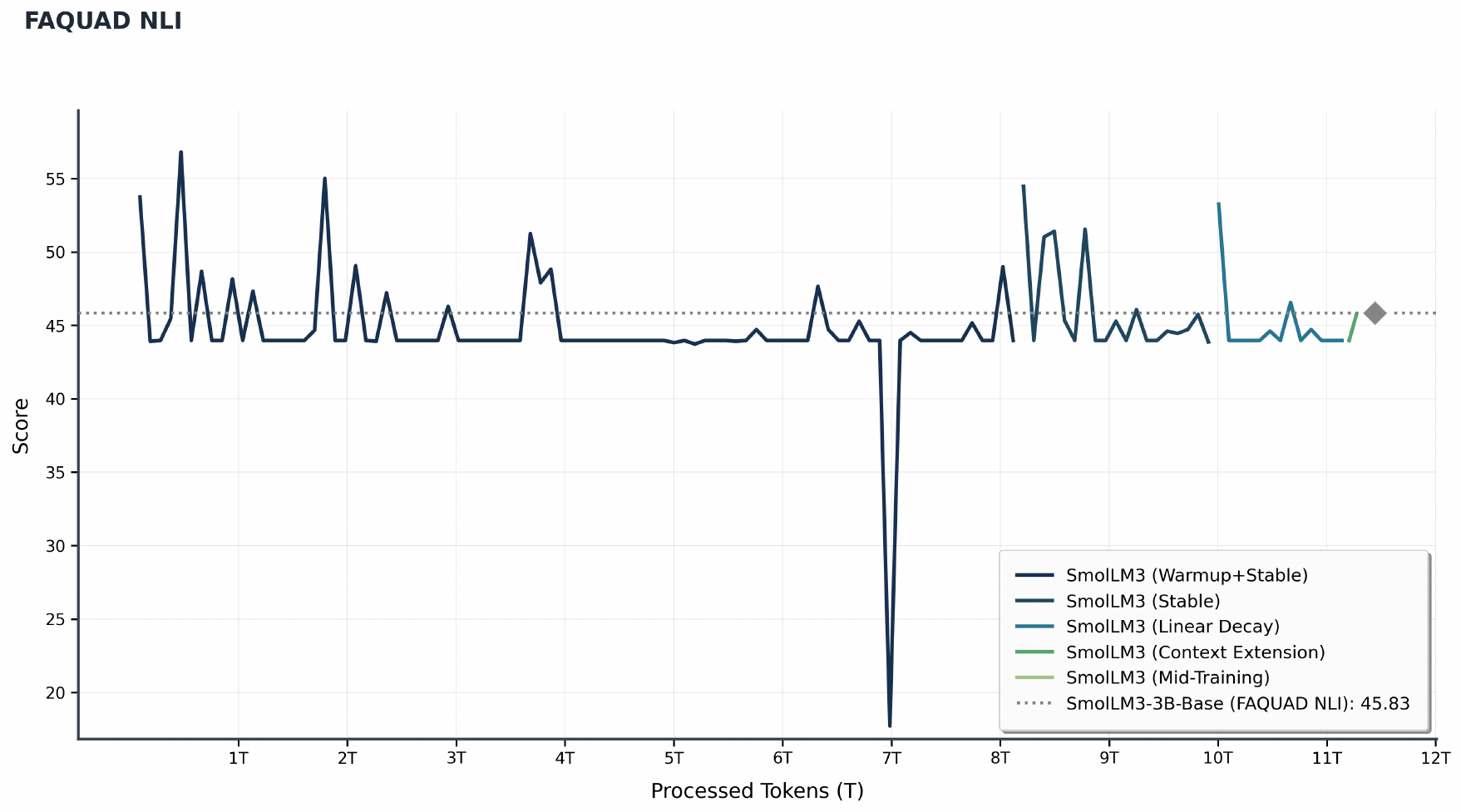} &
\includegraphics[width=0.30\linewidth]{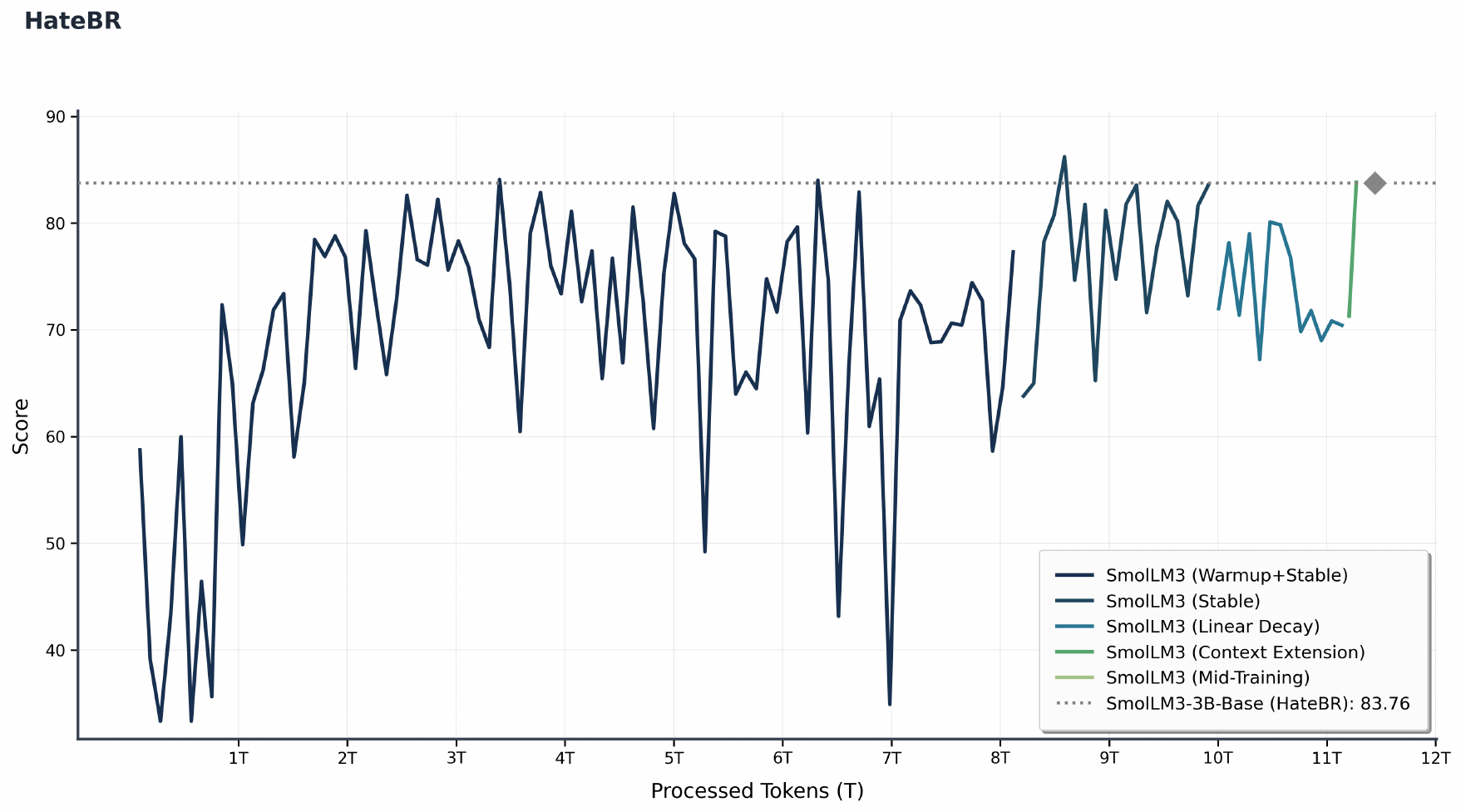} \\
\includegraphics[width=0.30\linewidth]{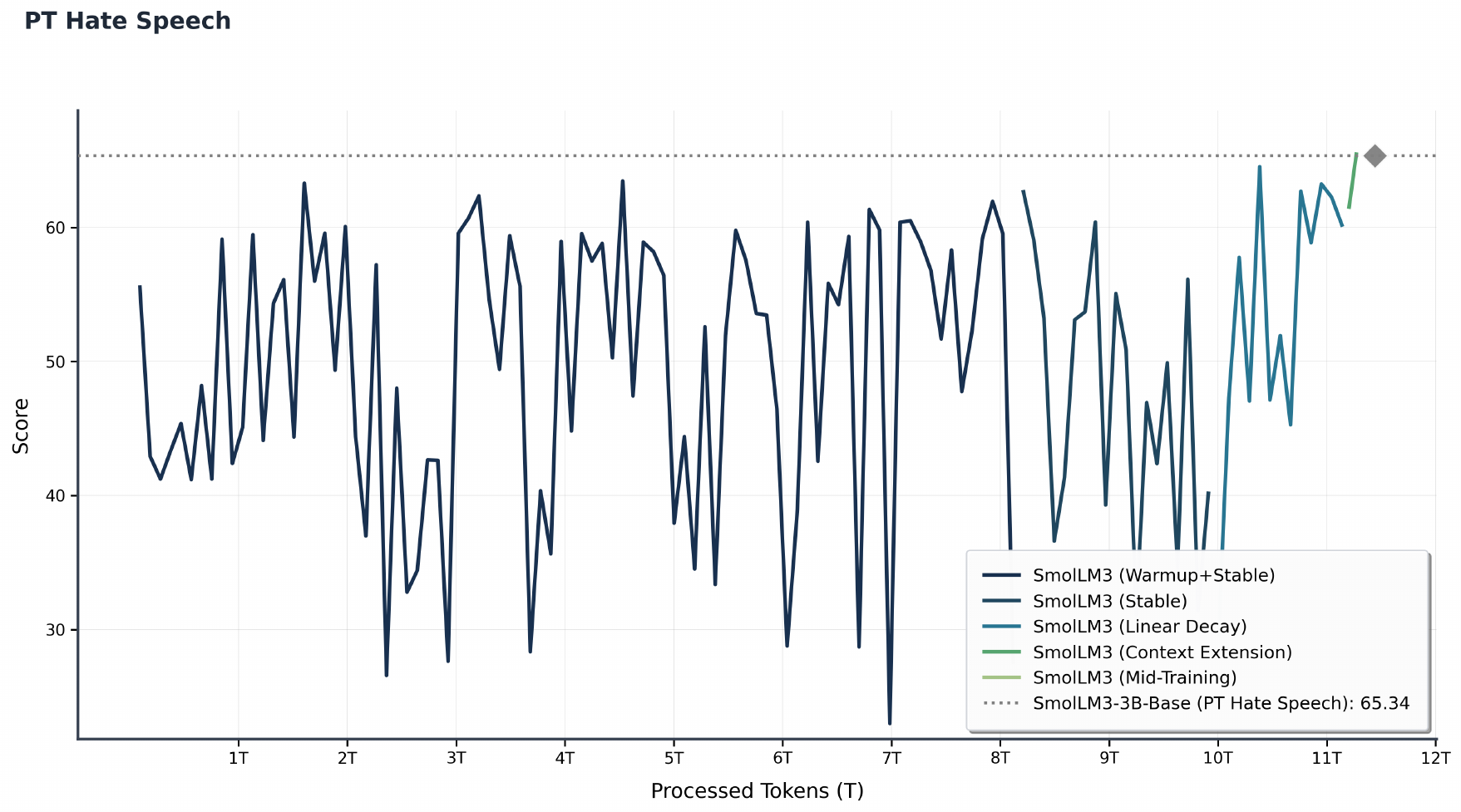} &
\includegraphics[width=0.30\linewidth]{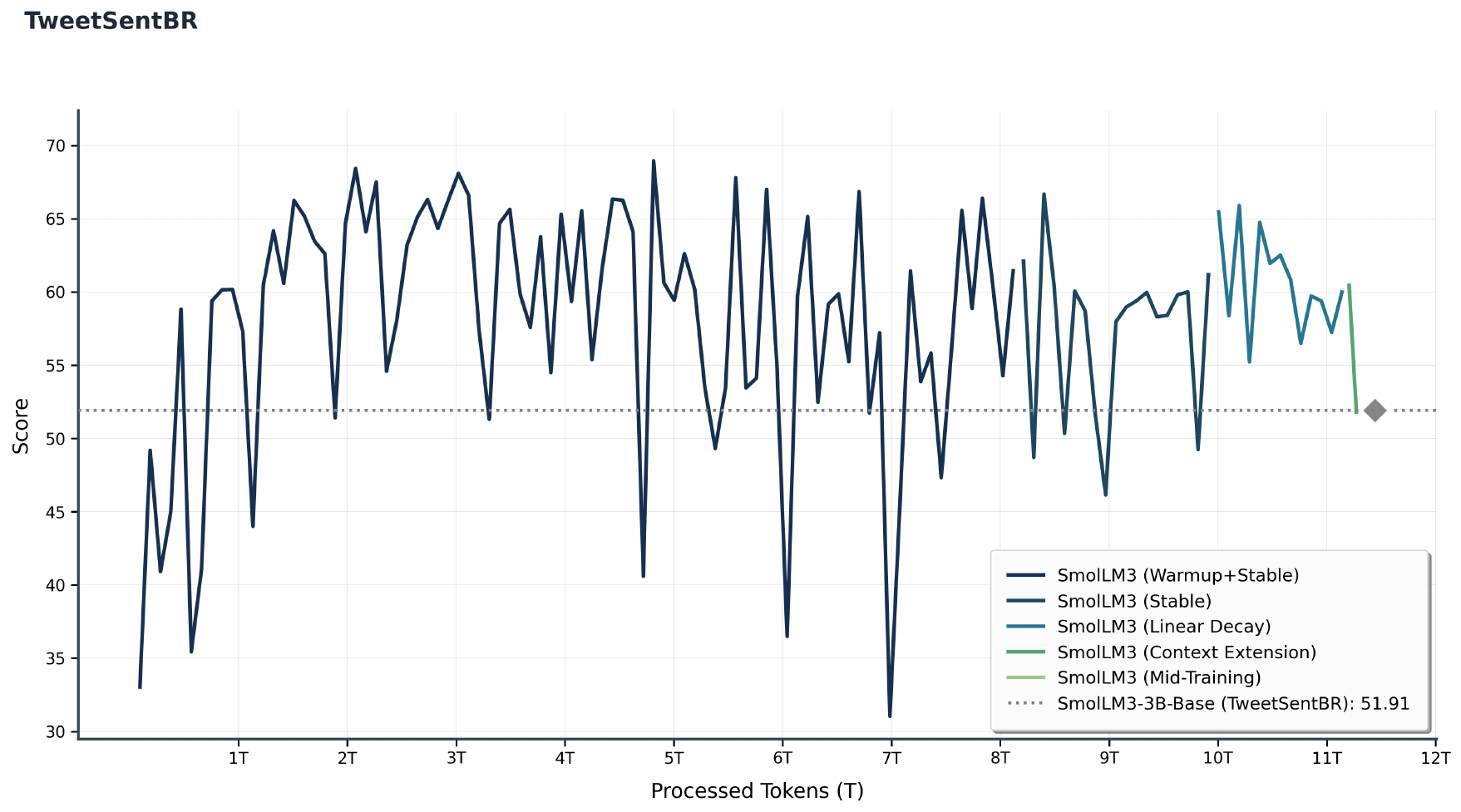} &
\includegraphics[width=0.30\linewidth]{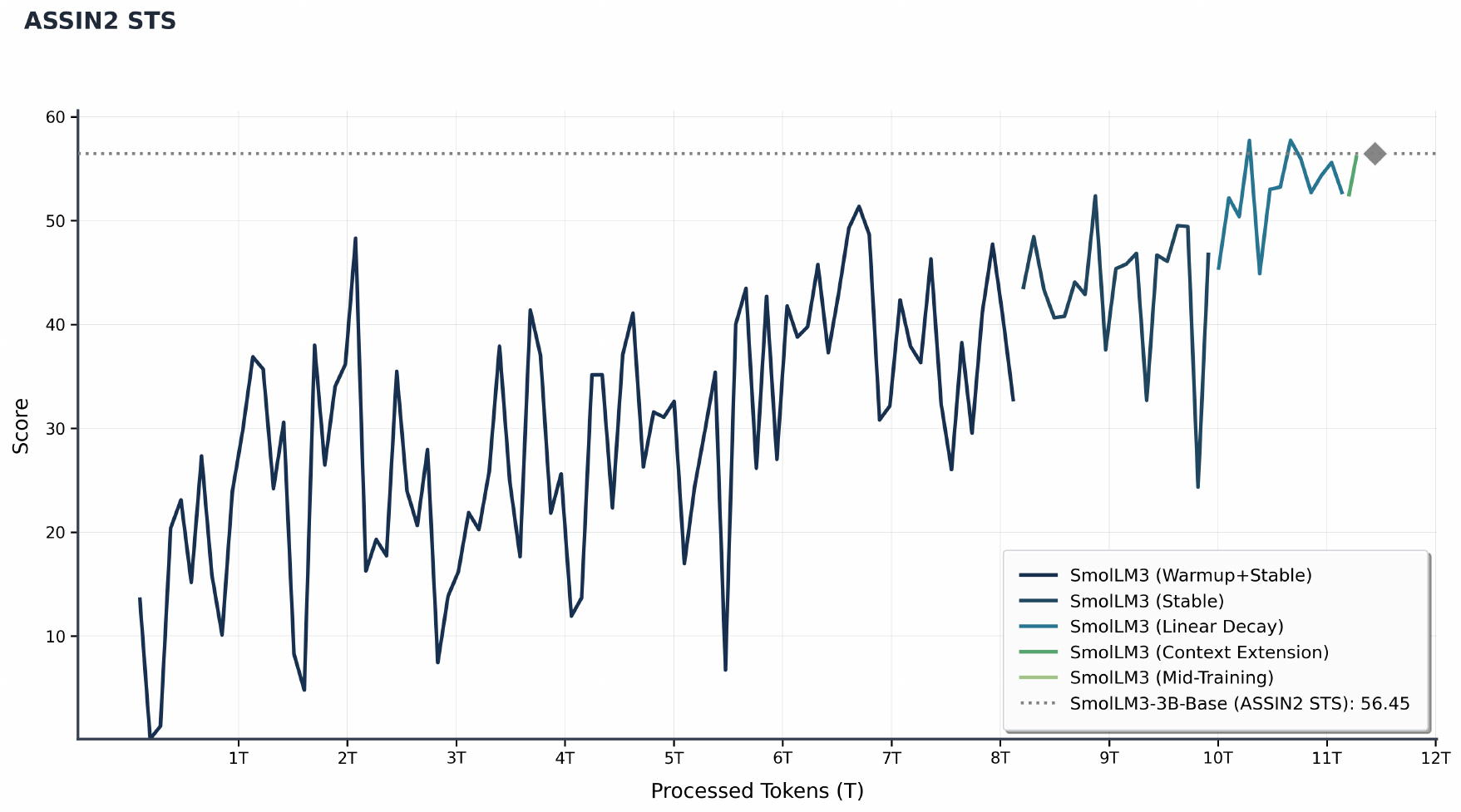} \\
\end{tabular}
\caption{Individual benchmark performance on Garcia et al.'s evaluation suite.}
\label{fig:garcia-benchmarks}
\end{figure}

Figures~\ref{fig:garcia-npm} and \ref{fig:garcia-benchmarks} show the NPM aggregated score and individual benchmark performance for SmolLM3-3b on Garcia et al.'s evaluation suite.

\newpage

\begin{figure}[!htbp]
\centering
\begin{tabular}{cc}
\includegraphics[width=0.45\linewidth]{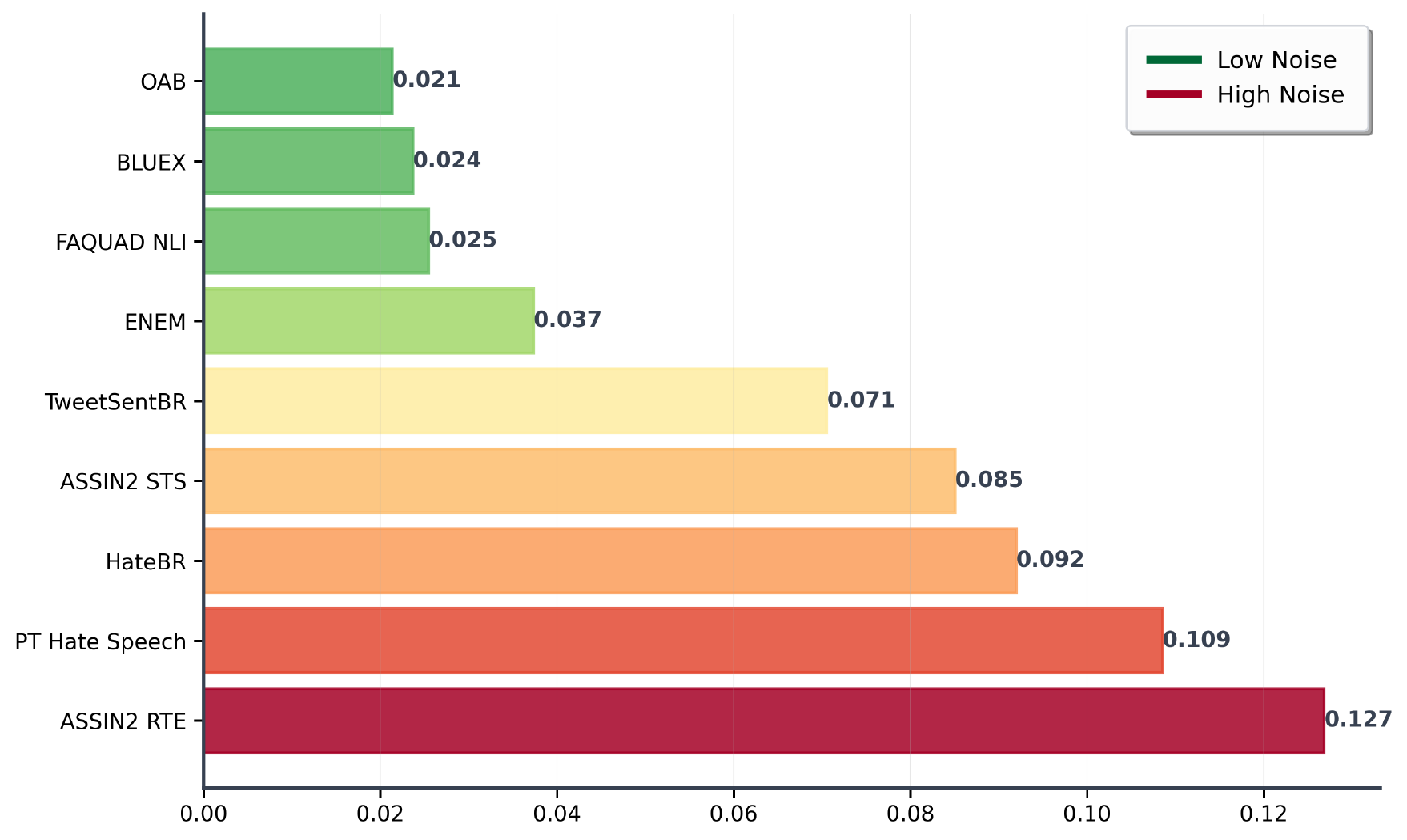} &
\includegraphics[width=0.45\linewidth]{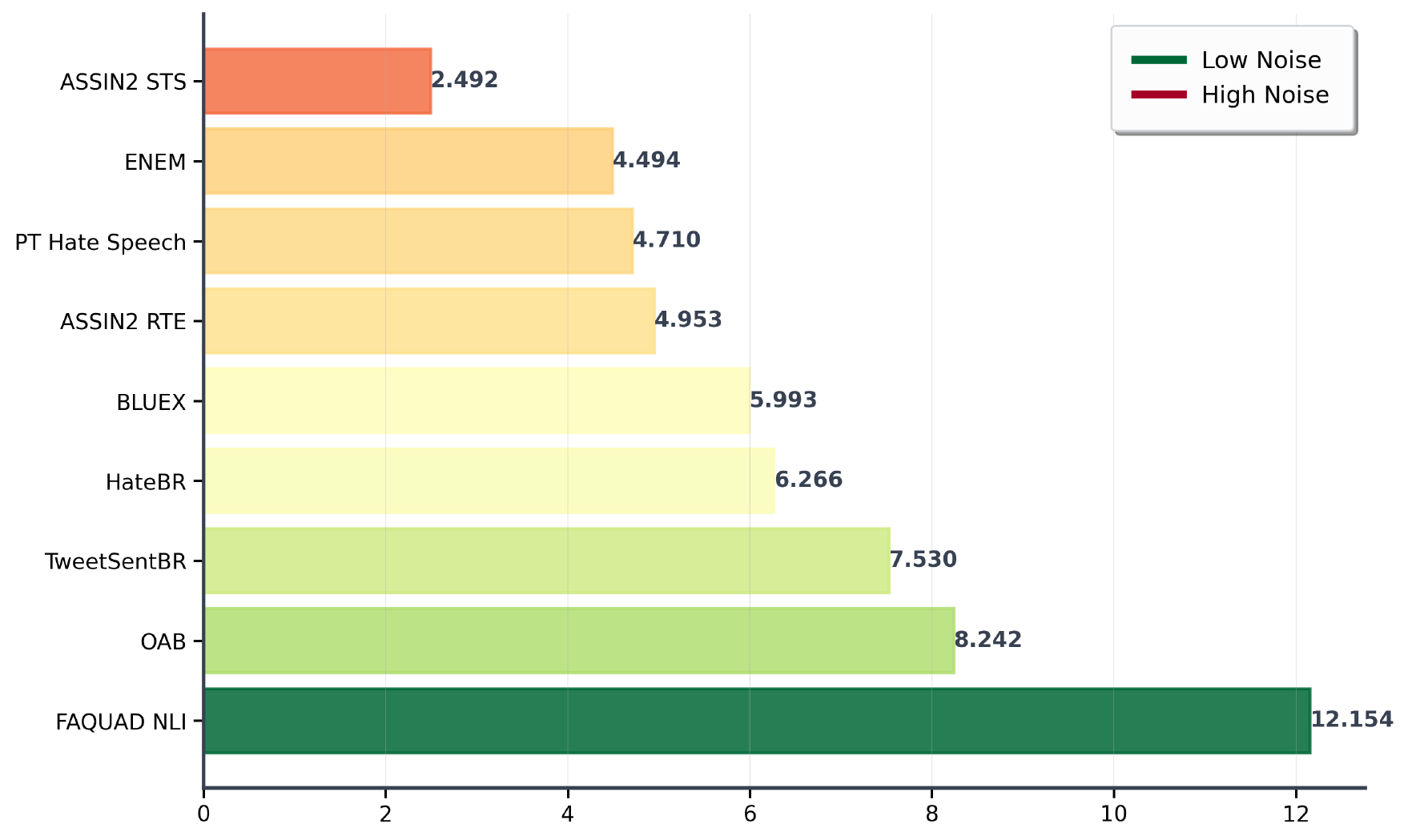} \\
(a) Mean Absolute Change & (b) Signal-to-Noise Ratio \\
\multicolumn{2}{c}{
\includegraphics[width=0.45\linewidth]{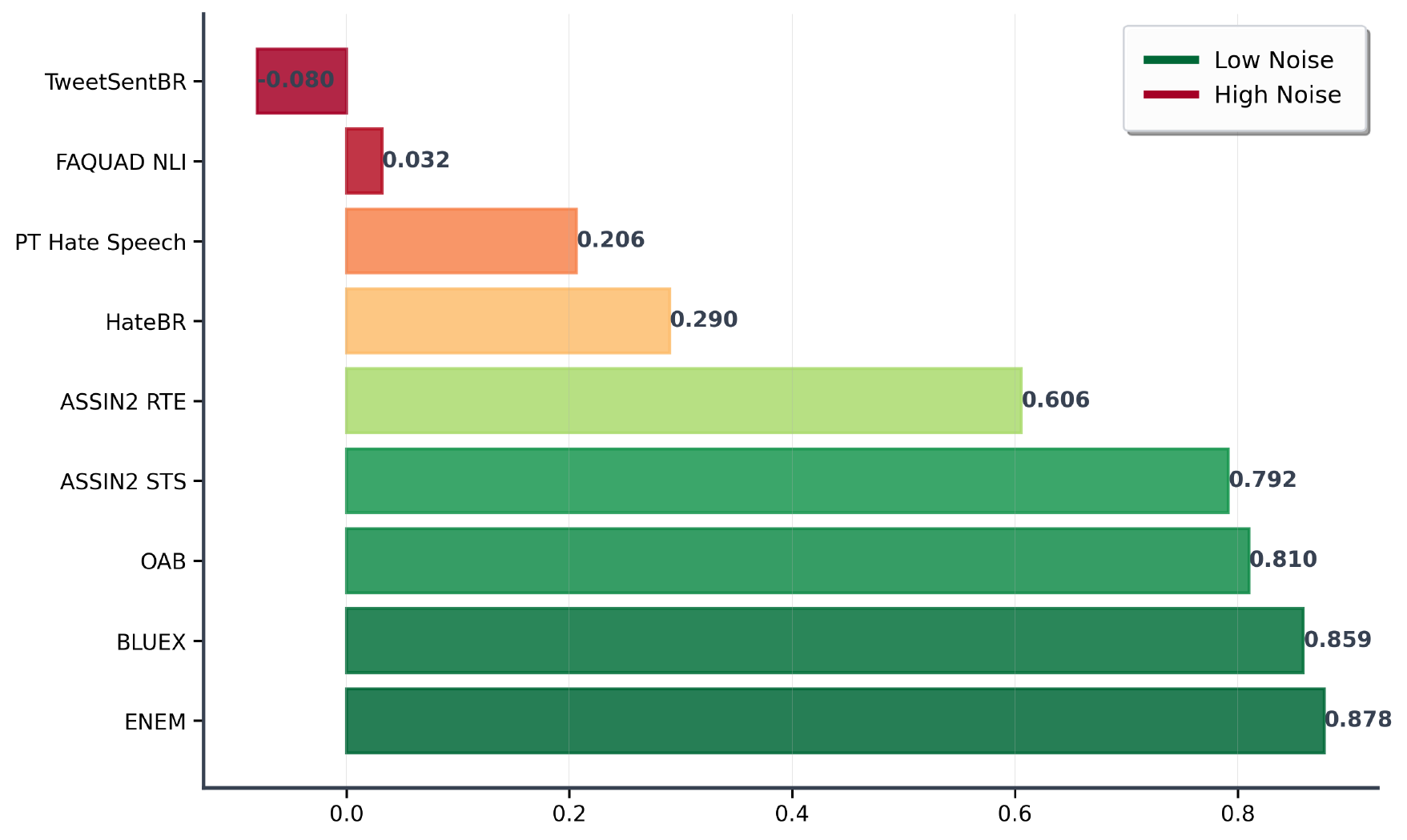}
} \\
\multicolumn{2}{c}{(c) Spearman Correlation}
\end{tabular}
\caption{Signal-quality metrics for Garcia et al.'s evaluation suite.}
\label{fig:garcia-noise}
\end{figure}

Figure~\ref{fig:garcia-noise} presents the three signal-quality metrics across all benchmarks in Garcia et al.'s harness.

\newpage
\subsection{New evaluation suite}
\label{appendix:new-harness-analysis}

\subsubsection{Evaluation Configuration}

Table~\ref{tab:new-config} summarizes the configuration of benchmarks in our new Portuguese evaluation suite.

\begin{table}[!htbp]
\centering
\small
\caption{Configuration of benchmarks in our new Portuguese evaluation suite. Benchmarks employ log-likelihood evaluation using either the Multiple-Choice Format (MCF) or the Cloze Formulation (CF).}
\label{tab:new-config}
\begin{tabular}{lllr}
\toprule
\textbf{Benchmark} & \textbf{n-shot} & \textbf{Type} & \textbf{Baseline (\%)} \\
\midrule
CALAME           & 5-shot & Completion (CF)           & 0   \\
Global PIQA      & 5-shot & Completion (CF)           & 50  \\
ASSIN2 Paraphrase & 5-shot & Paraphrase Detection (MCF) & 50  \\
ASSIN2 Entailment & 5-shot & Entailment (MCF)          & 50  \\
BELEBELE         & 5-shot & MC-Q\&A (MCF)             & 25  \\
LAMBADA          & 5-shot & Completion (CF)           & 0   \\
ARC-Challenge    & 5-shot & MC-Q\&A (MCF)             & 25  \\
MMLL             & 5-shot & MC-Q\&A (MCF)             & 25  \\
HellaSwag        & 5-shot & Completion (CF)           & 25  \\
\bottomrule
\end{tabular}
\end{table}

\subsubsection{Baseline Comparison}

Table~\ref{tab:baseline-surpass-new} shows the number of training tokens required for SmolLM3-3B to surpass each benchmark's baseline by 5 percentage points.

\begin{table}[!htbp]
\centering
\small
\caption{Training tokens required for SmolLM3-3B to surpass baselines on our new evaluation suite. Most benchmarks require 94B--660B tokens, substantially less than the 755B--1T+ required by Garcia et al.'s harness.}
\label{tab:baseline-surpass-new}
\begin{tabular}{lrrrrrr}
\toprule
\textbf{Benchmark} & \textbf{Baseline} & \textbf{Threshold} & \textbf{First Step} & \textbf{Tokens (B)} & \textbf{Score} & \textbf{Improvement} \\
\midrule
ARC Challenge & 25.00 & 30.00 & 80,000  & 188 & 30.77 & 5.80  \\
ASSIN2 ENT    & 50.00 & 55.00 & 40,000  & 94  & 58.08 & 8.10  \\
ASSIN2 PAR    & 50.00 & 55.00 & 40,000  & 94  & 63.65 & 13.60 \\
BELEBELE      & 25.00 & 30.00 & 280,000 & 660 & 30.00 & 5.00  \\
Global PIQA   & 50.00 & 55.00 & 40,000  & 94  & 65.00 & 15.00 \\
HellaSwag     & 25.00 & 30.00 & 40,000  & 94  & 36.50 & 11.50 \\
MMLU          & 25.00 & 30.00 & 320,000 & 755 & 32.59 & 7.60  \\
\bottomrule
\end{tabular}
\vspace{0.3cm}
\begin{flushleft}
\small
\textit{Note:} CALAME and LAMBADA are excluded because their baselines are 0\%, and SmolLM3-3b shows immediate improvement at the first evaluation checkpoint.
\end{flushleft}
\end{table}

\subsubsection{Signal Analysis}

Table~\ref{tab:new-signal-quality} presents the signal-quality metrics for each benchmark in our new evaluation suite. Most benchmarks show substantial improvements in SNR and Spearman's rank correlation compared with Garcia et al.'s evaluation suite.

\newpage
\begin{table}[!htbp]
\centering
\small
\caption{Signal-quality metrics for our new Portuguese evaluation suite. Most benchmarks exhibit strong Spearman correlations ($>$0.5) and acceptable SNR, with CF tasks showing particularly high SNR ($>$19).}
\label{tab:new-signal-quality}
\begin{tabular}{lrrr}
\toprule
\textbf{Benchmark} & \textbf{Mean Abs Change} & \textbf{SNR} & \textbf{Spearman} \\
\midrule
ARC Challenge  & 0.012 & 11.777 & 0.931 \\
ASSIN2 ENT     & 0.035 & 18.043 & 0.180 \\
ASSIN2 PAR     & 0.025 & 28.970 & 0.273 \\
BELEBELE       & 0.025 & 4.775  & 0.949 \\
CALAME         & 0.009 & 41.838 & 0.621 \\
Global PIQA    & 0.026 & 23.081 & 0.578 \\
HellaSwag      & 0.004 & 26.174 & 0.937 \\
LAMBADA        & 0.024 & 19.579 & 0.576 \\
MMLU           & 0.015 & 6.923  & 0.923 \\
\bottomrule
\end{tabular}
\end{table}

\subsubsection{Per-Benchmark Results}

\begin{figure}[!htbp]
\centering
\includegraphics[width=0.75\linewidth]{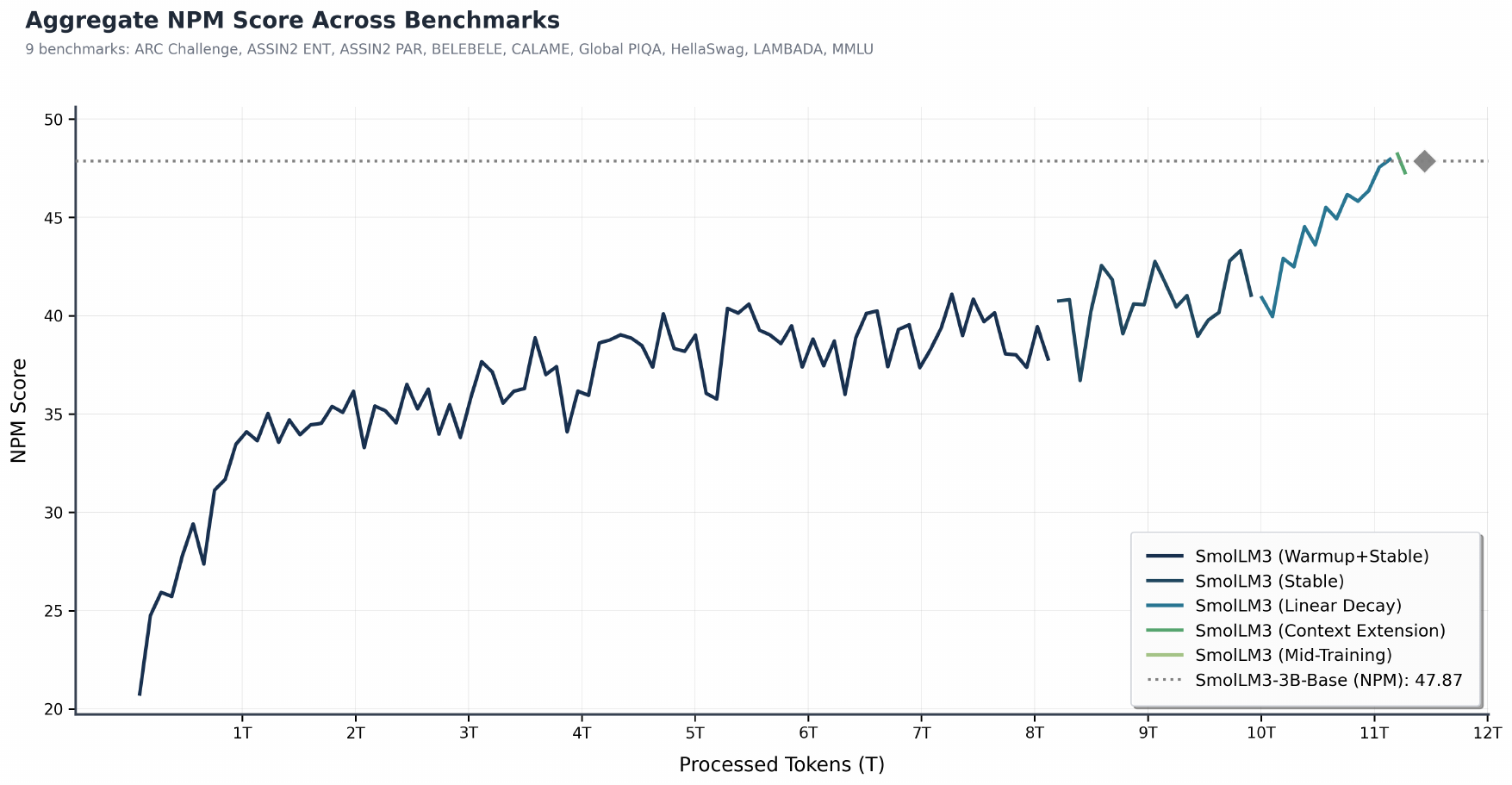}
\caption{NPM aggregated score for SmolLM3-3b on our new Portuguese evaluation suite. The score shows a clearer, monotonic upward trend with lower volatility than Garcia et al.'s evaluation suite.}
\label{fig:new-npm}
\end{figure}

\begin{figure}[!htbp]
\centering
\begin{tabular}{ccc}
\includegraphics[width=0.30\linewidth]{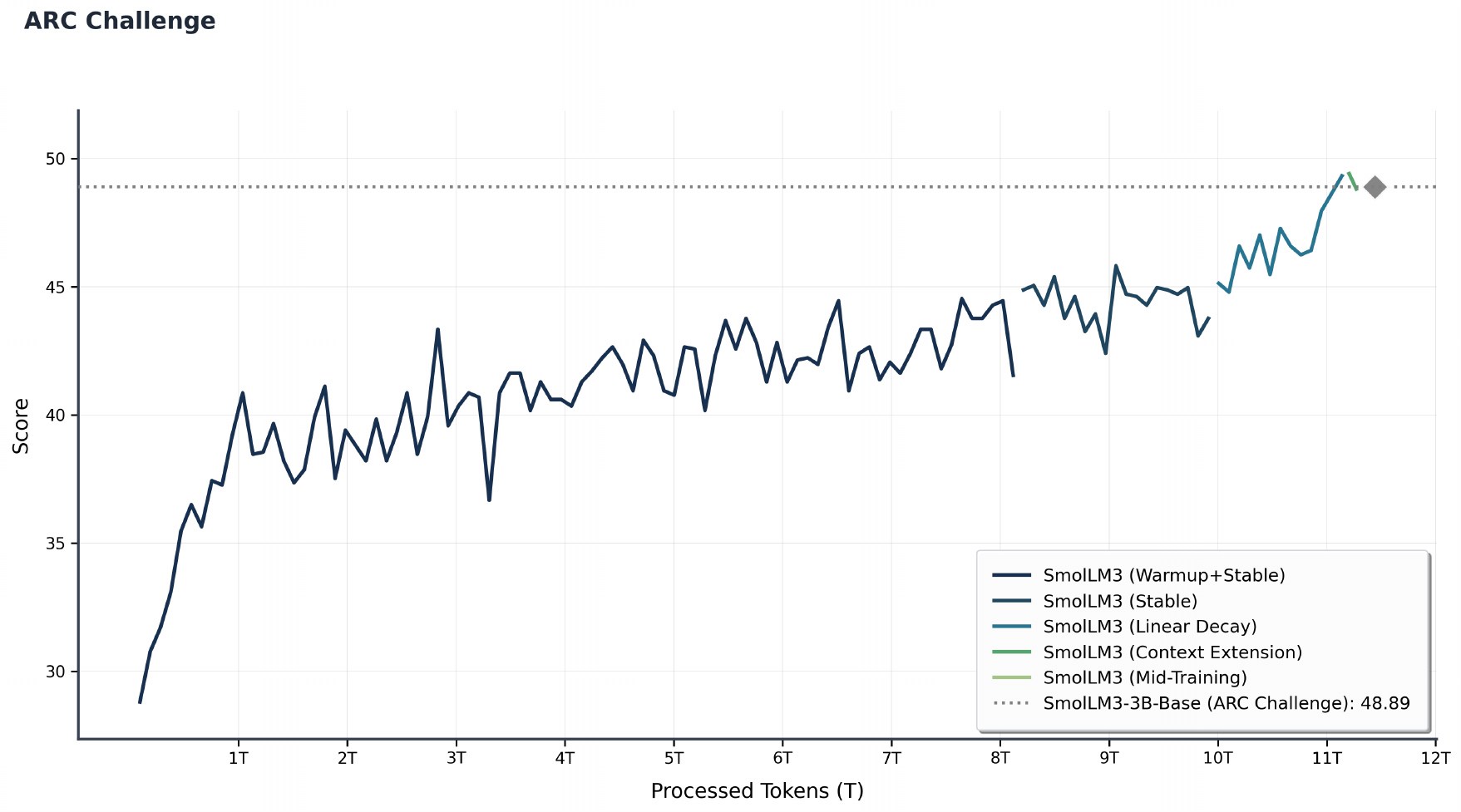} &
\includegraphics[width=0.30\linewidth]{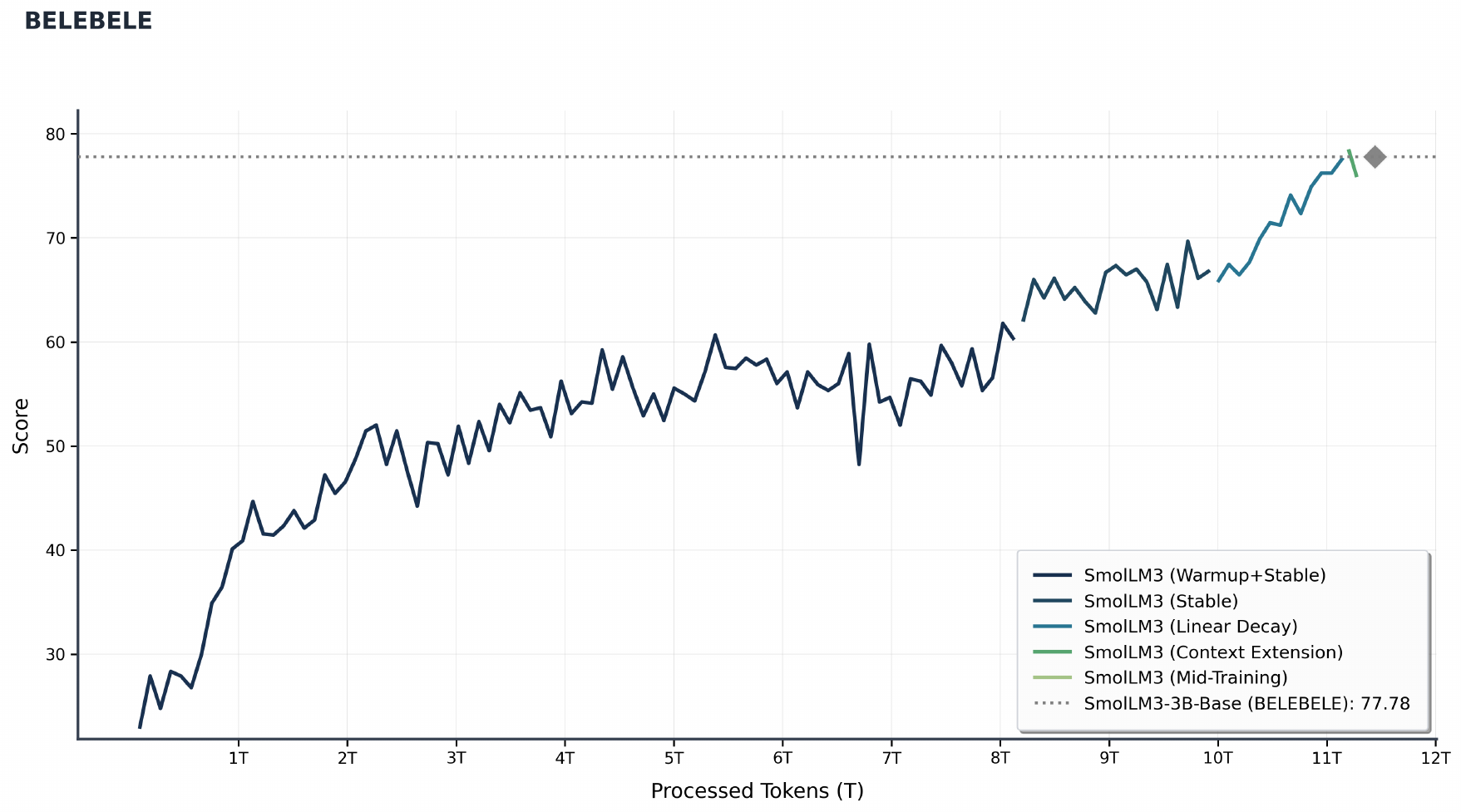} &
\includegraphics[width=0.30\linewidth]{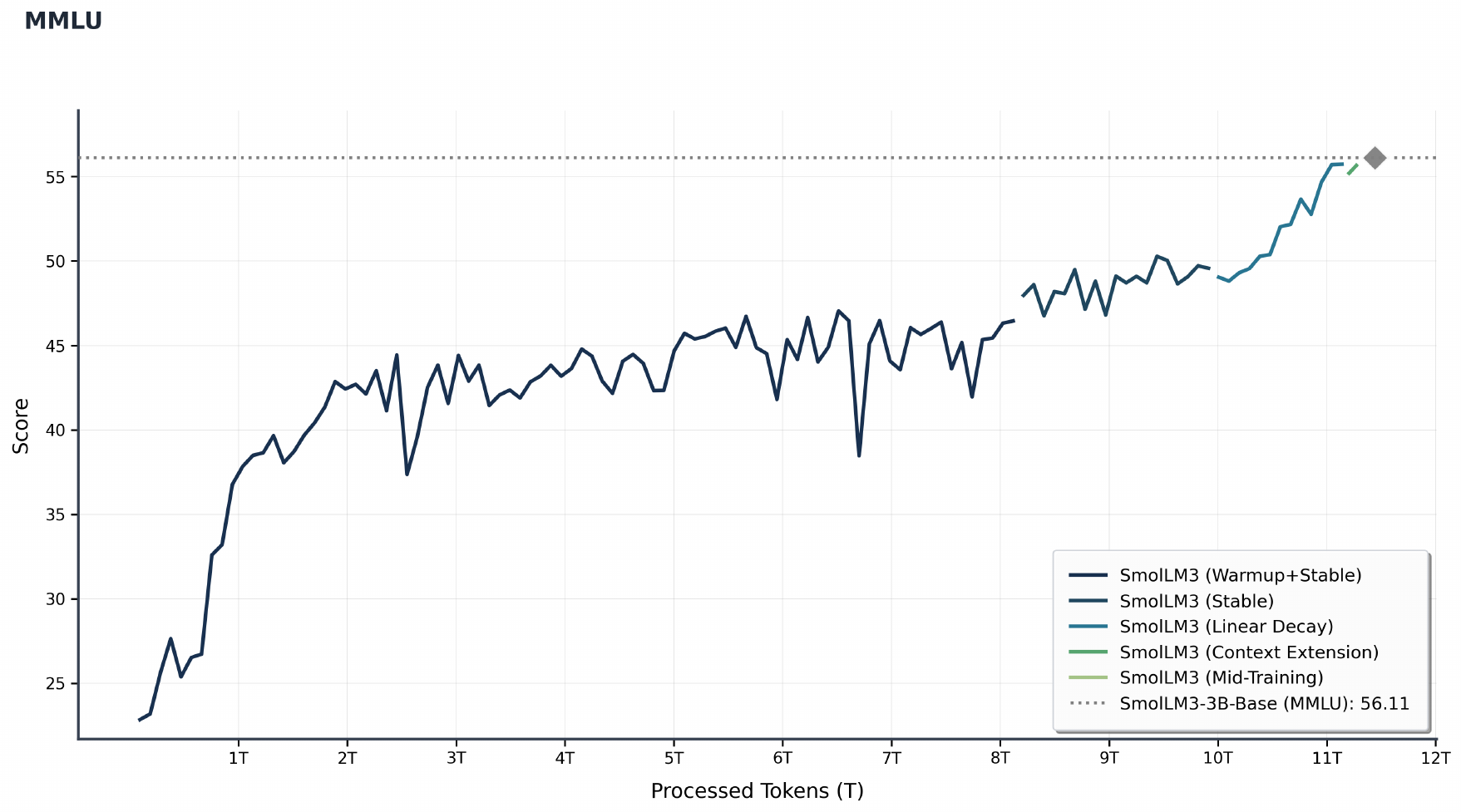} \\
\includegraphics[width=0.30\linewidth]{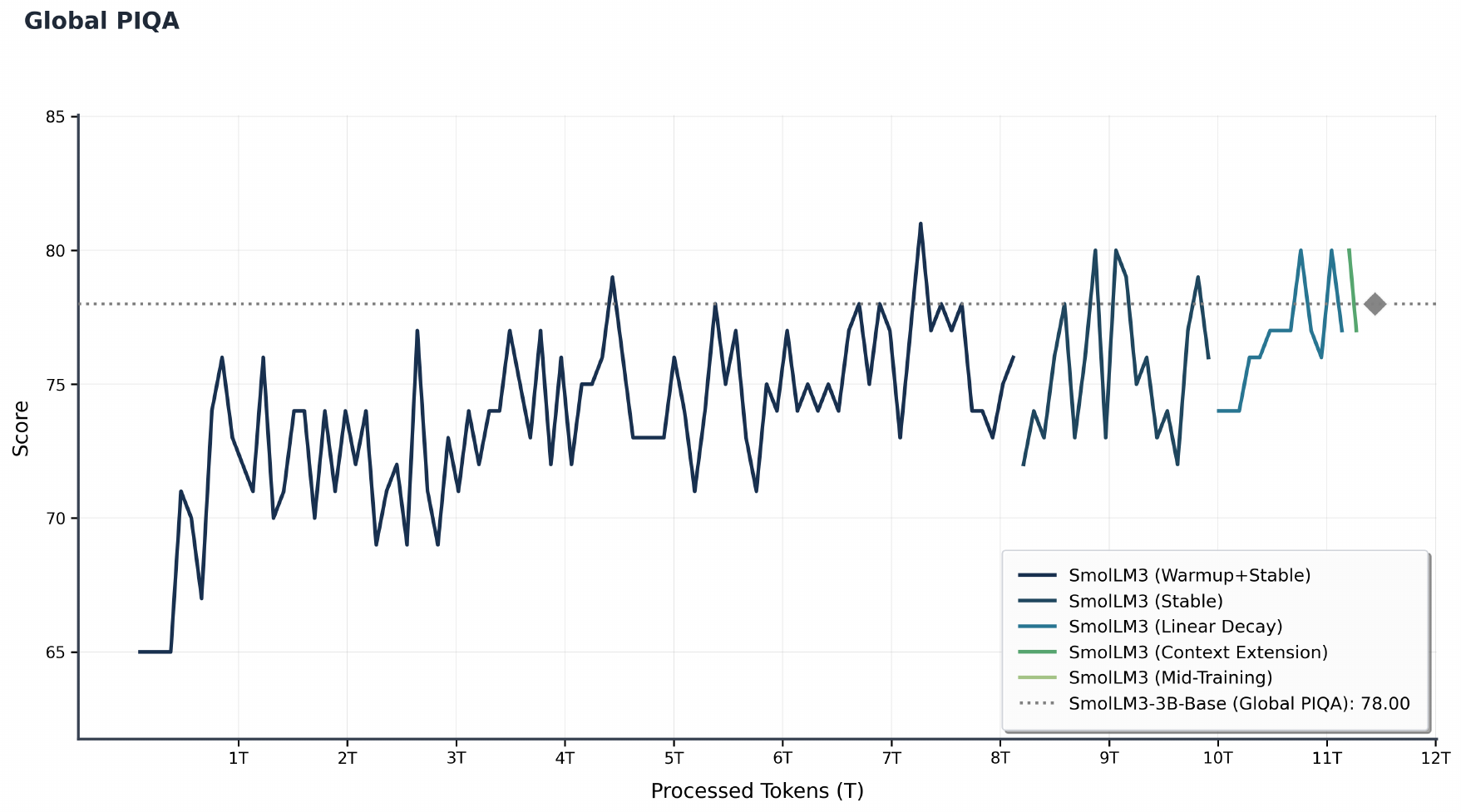} &
\includegraphics[width=0.30\linewidth]{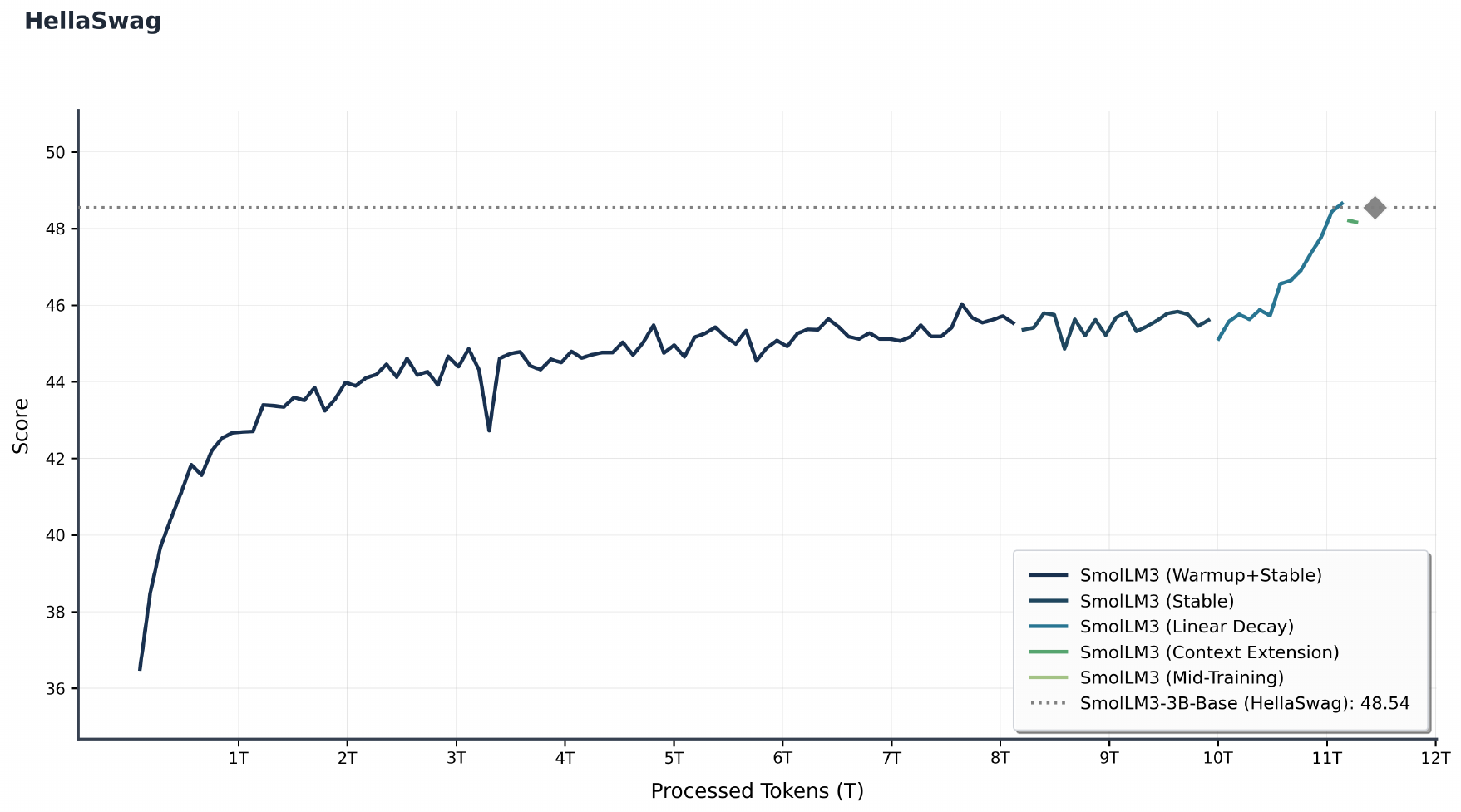} &
\includegraphics[width=0.30\linewidth]{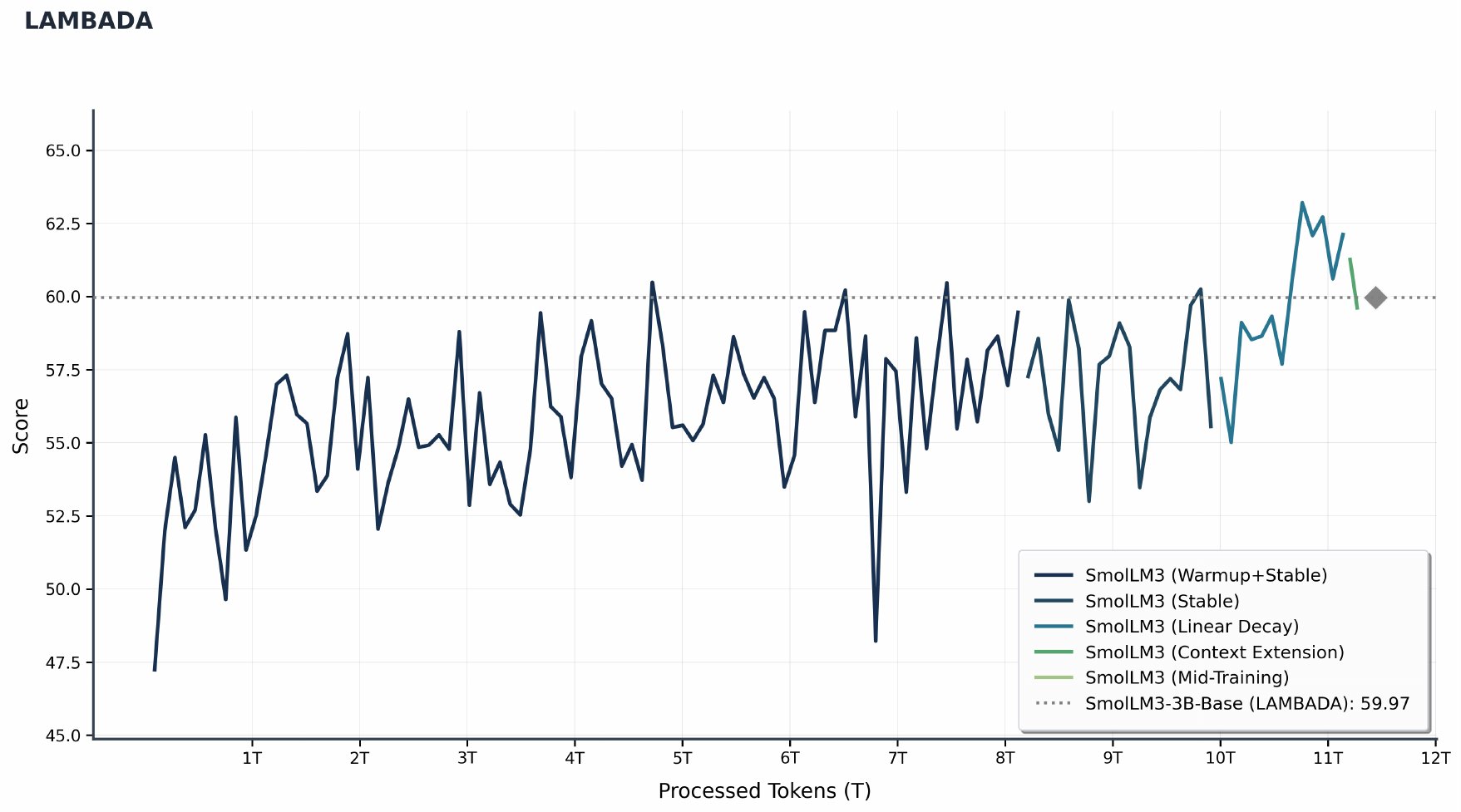} \\
\includegraphics[width=0.30\linewidth]{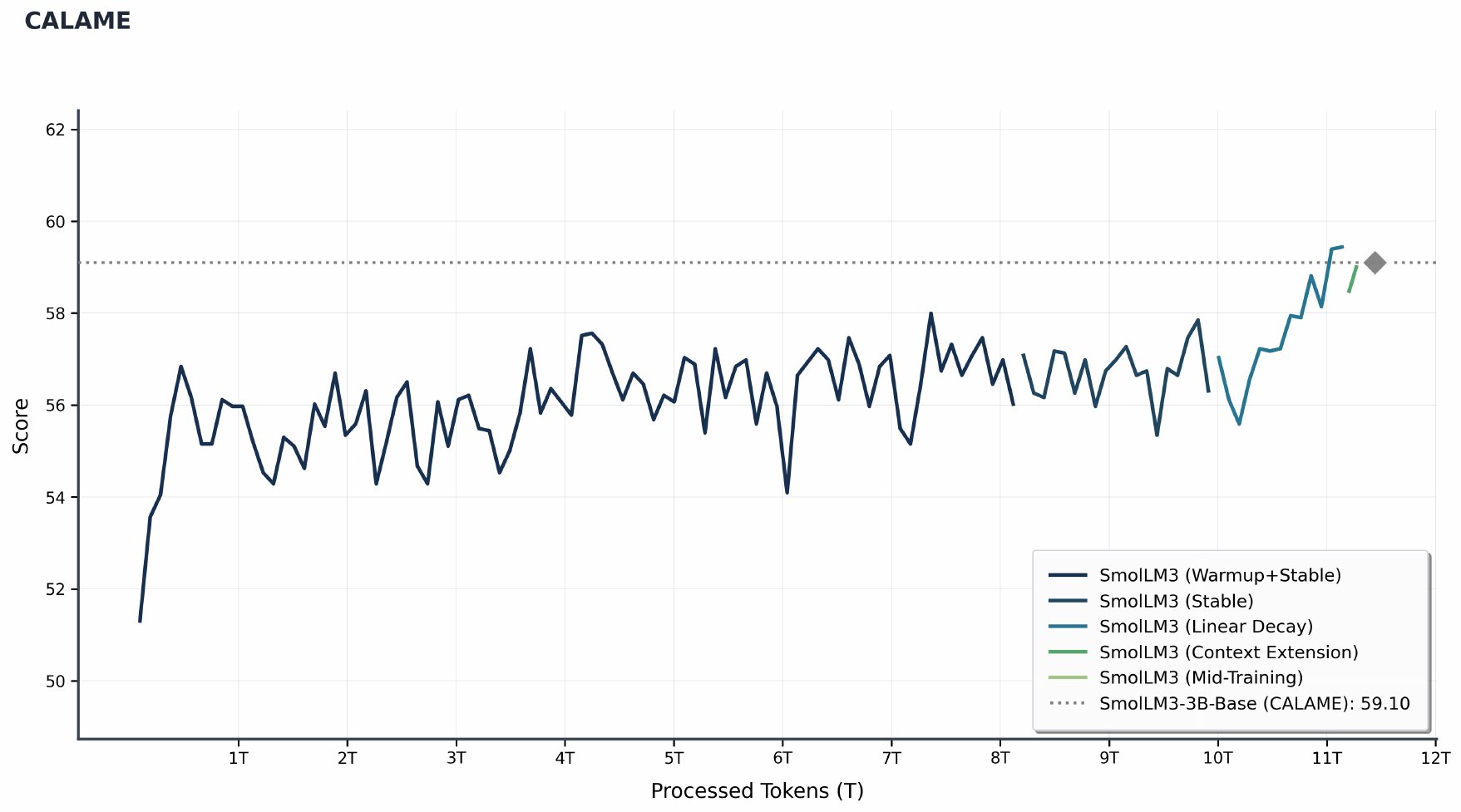} &
\includegraphics[width=0.30\linewidth]{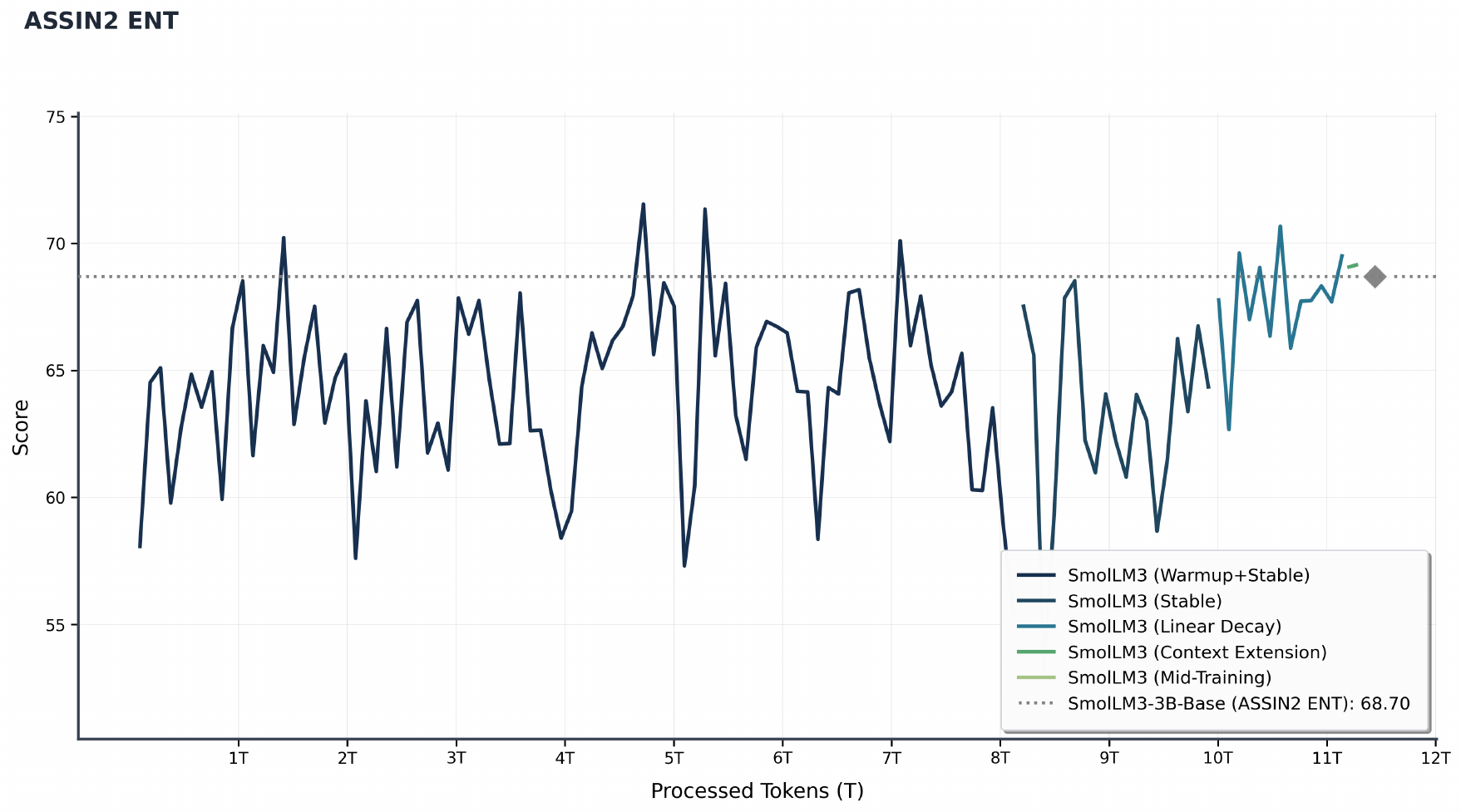} &
\includegraphics[width=0.30\linewidth]{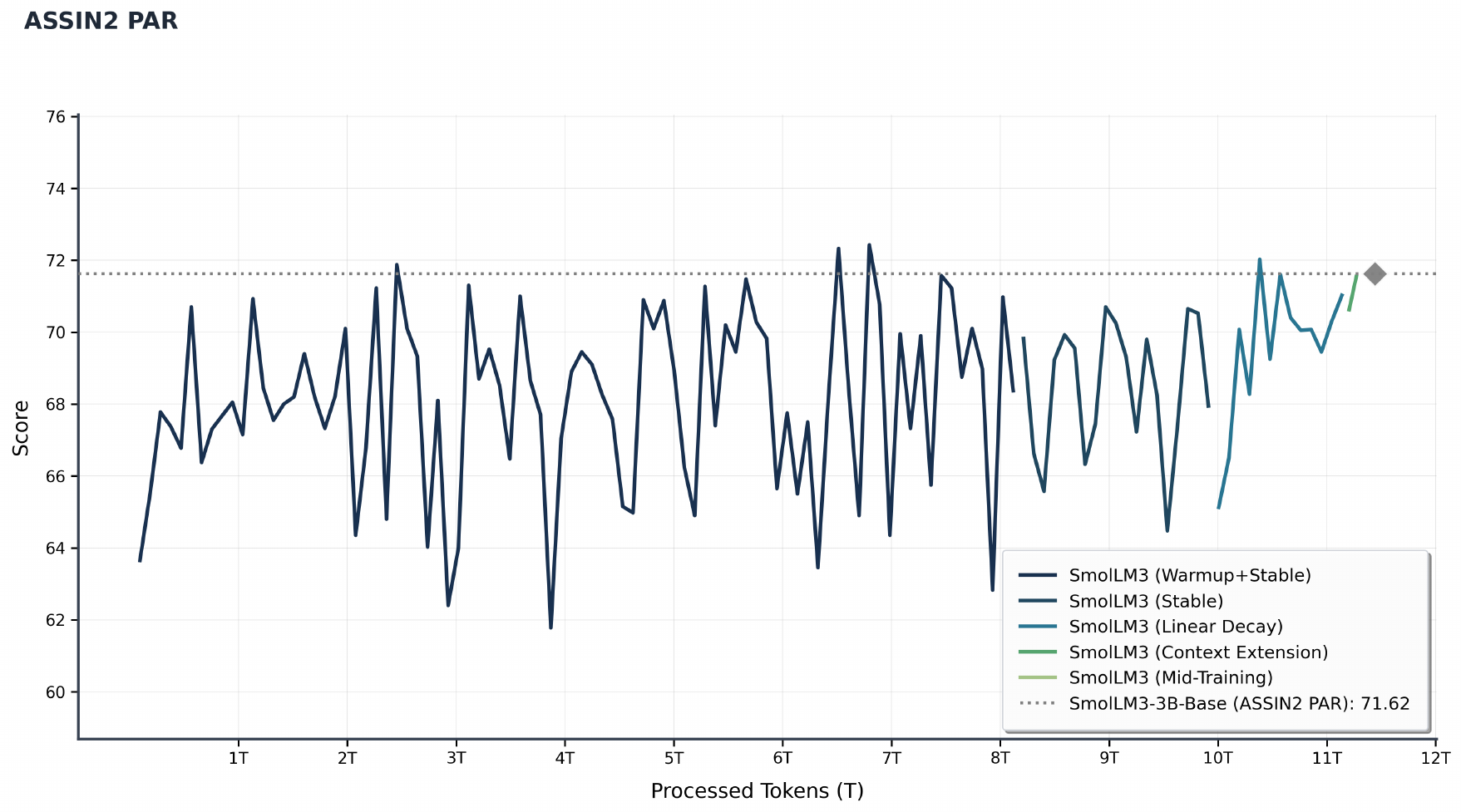} \\
\end{tabular}
\caption{Individual benchmark performance on our new Portuguese evaluation suite.}
\label{fig:new-benchmarks}
\end{figure}

Figures~\ref{fig:new-npm} and \ref{fig:new-benchmarks} show the NPM aggregated score and individual benchmark performance for SmolLM3-3b on our new evaluation suite.

\newpage

\begin{figure}[!htbp]
\centering
\begin{tabular}{cc}
\includegraphics[width=0.45\linewidth]{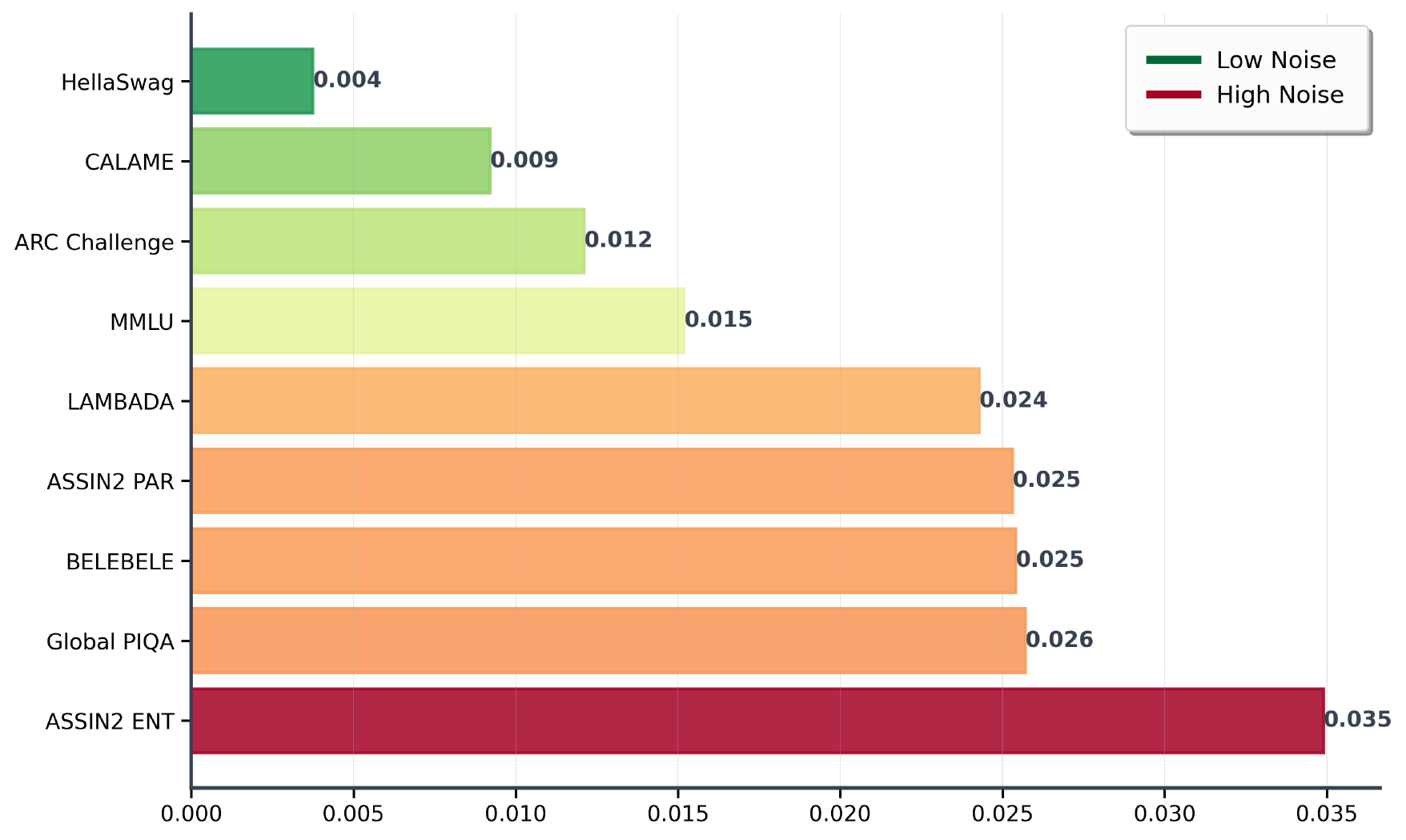} &
\includegraphics[width=0.45\linewidth]{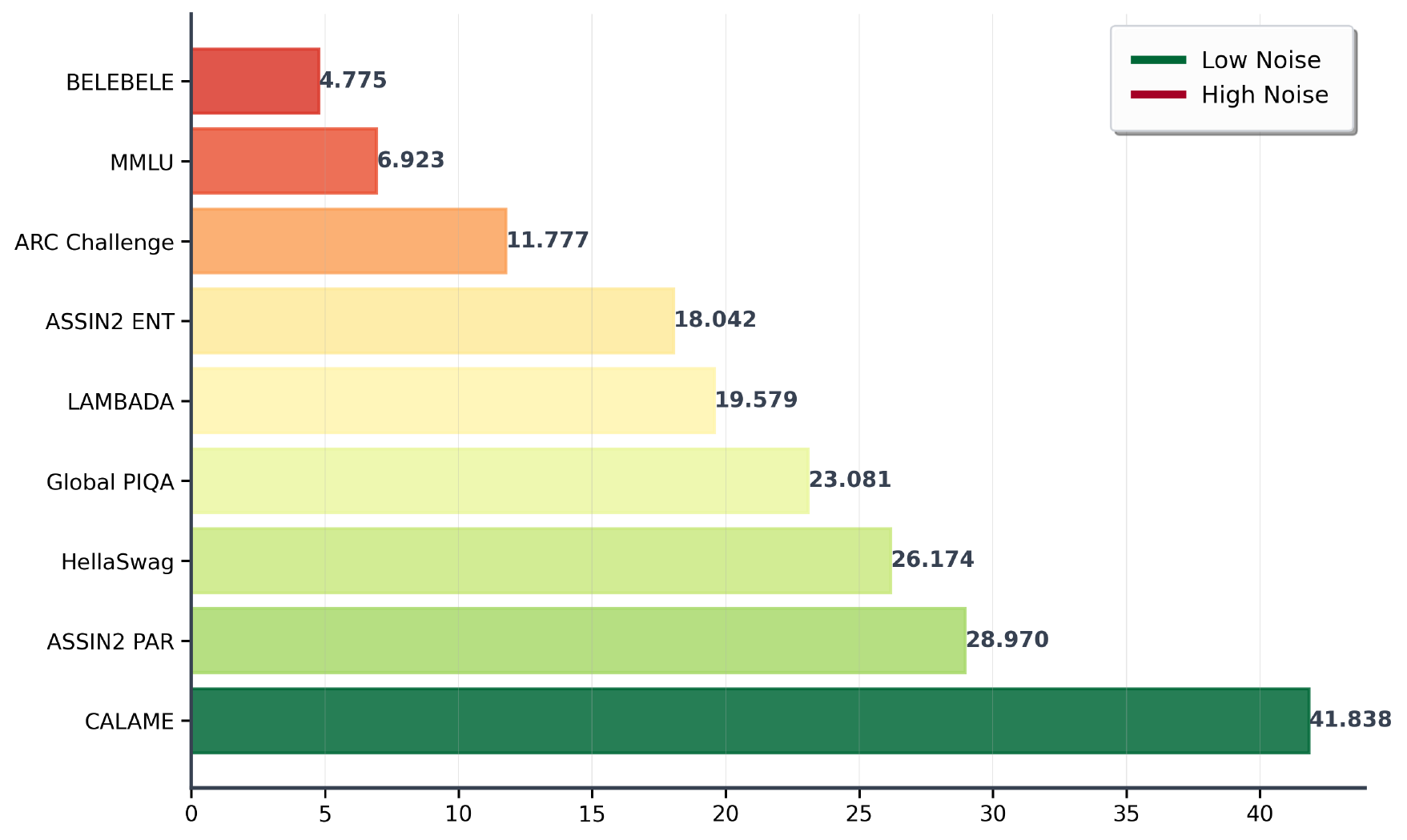} \\
(a) Mean Absolute Change & (b) Signal-to-Noise Ratio \\
\multicolumn{2}{c}{
\includegraphics[width=0.45\linewidth]{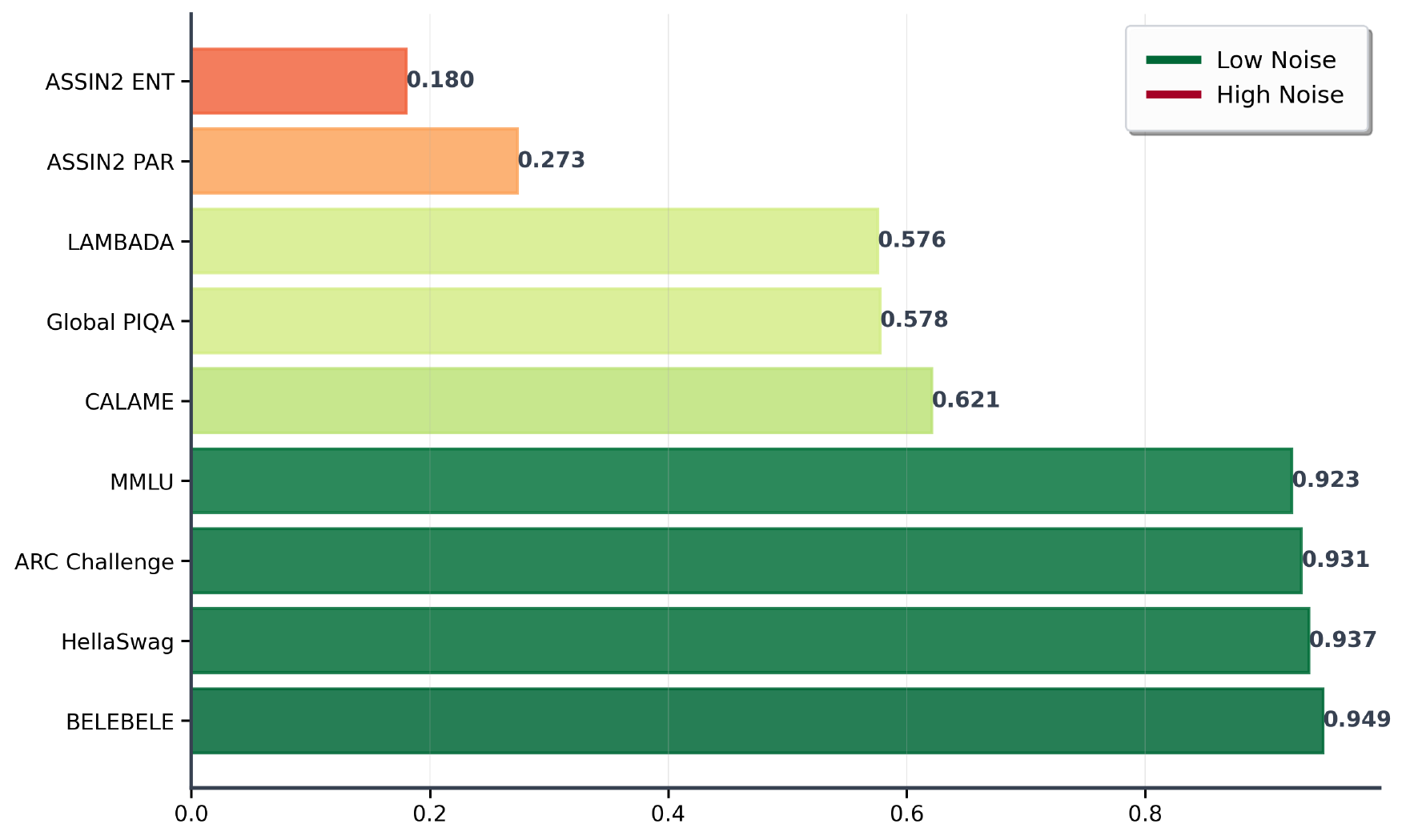}
} \\
\multicolumn{2}{c}{(c) Spearman Correlation}
\end{tabular}
\caption{Signal-quality metrics for our new Portuguese evaluation suite.}
\label{fig:new-noise}
\end{figure}

Figure~\ref{fig:new-noise} presents the three signal-quality metrics across all benchmarks in our new harness.

\clearpage
\subsection{Unified Suite: Easy Set and Hard Set}
\label{appendix:unified-harness}

To create a unified evaluation suite, we combine the most reliable benchmarks from both evaluation suites into two tiers based on their signal quality characteristics.

\subsubsection{Evaluation Configuration}

Table~\ref{tab:unified-config} presents the complete configuration of our unified Portuguese evaluation suite.

\begin{table}[!htbp]
\centering
\small
\caption{Configuration of our unified Portuguese evaluation suite. The Easy Set provides reliable signals at $<$200B tokens, while the Hard Set requires $>$660B tokens but offers complementary advanced evaluation.}
\label{tab:unified-config}
\begin{tabular}{lllrr}
\toprule
\textbf{Benchmark} & \textbf{n-shot} & \textbf{Type} & \textbf{Baseline (\%)} & \textbf{Metric} \\
\midrule
\multicolumn{5}{c}{\textbf{Easy Set}} \\
\midrule
CALAME          & 5-shot & Completion         & 0    & \texttt{acc}      \\
GlobalPIQA      & 5-shot & Completion         & 50   & \texttt{acc\_norm} \\
LAMBADA         & 5-shot & Completion         & 0    & \texttt{acc}      \\
ARC-Challenge   & 5-shot & MC-Q\&A            & 25   & \texttt{acc\_norm} \\
HellaSwag       & 5-shot & MC-Q\&A            & 25   & \texttt{acc\_norm} \\
\midrule
\multicolumn{5}{c}{\textbf{Hard Set}} \\
\midrule
ENEM            & 3-shot & MC-Q\&A            & 20   & \texttt{acc}      \\
BLUEX           & 3-shot & MC-Q\&A            & 22.5 & \texttt{acc}      \\
OAB Exams       & 3-shot & MC-Q\&A            & 25   & \texttt{acc}      \\
BELEBELE        & 5-shot & MC-Q\&A            & 25   & \texttt{acc\_norm} \\
MMLU            & 5-shot & MC-Q\&A            & 25   & \texttt{acc}      \\
\bottomrule
\end{tabular}
\end{table}

\subsubsection{Comparison of Easy Set vs. Hard Set}

Table~\ref{tab:easy-hard-comparison} compares the signal-quality characteristics of the Easy Set and Hard Set evaluations.

\begin{table}[!htbp]
\centering
\small
\caption{Comparison of signal-quality characteristics between Easy Set and Hard Set evaluations.}
\label{tab:easy-hard-comparison}
\begin{tabular}{lrrrr}
\toprule
\textbf{Benchmark} & \textbf{Tokens to Surpass (B)} & \textbf{SNR} & \textbf{Spearman} \\
\midrule
\multicolumn{4}{c}{\textbf{Easy Set}} \\
\midrule
ARC Challenge & 188  & 11.78 & 0.931 \\
CALAME        & $<$40 & 41.84 & 0.621 \\
Global PIQA   & 94   & 23.08 & 0.578 \\
HellaSwag     & 94   & 26.17 & 0.937 \\
LAMBADA       & $<$40 & 19.58 & 0.576 \\
\midrule
\textbf{Average (Easy)} & \textbf{103} & \textbf{24.49} & \textbf{0.729} \\
\midrule
\multicolumn{4}{c}{\textbf{Hard Set}} \\
\midrule
ENEM          & 755   & 4.49  & 0.878 \\
BLUEX         & 755   & 5.99  & 0.859 \\
OAB           & 1,038 & 8.24  & 0.810 \\
BELEBELE      & 660   & 4.78  & 0.949 \\
MMLU          & 755   & 6.92  & 0.923 \\
\midrule
\textbf{Average (Hard)} & \textbf{793} & \textbf{6.08} & \textbf{0.884} \\
\bottomrule
\end{tabular}
\end{table}

\clearpage

\begin{figure}[!htbp]
\centering
\begin{tabular}{cc}
\includegraphics[width=0.48\linewidth]{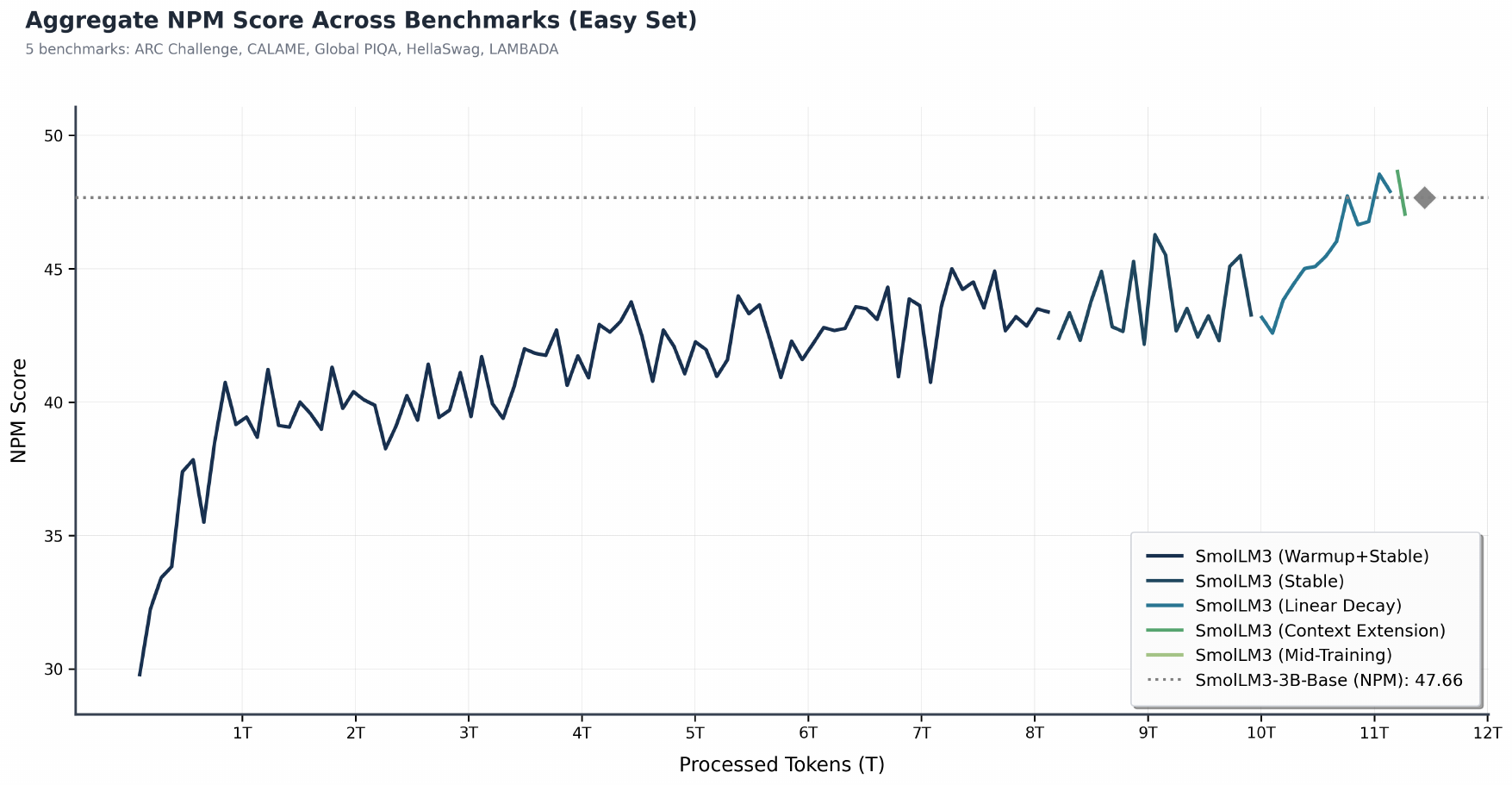} &
\includegraphics[width=0.48\linewidth]{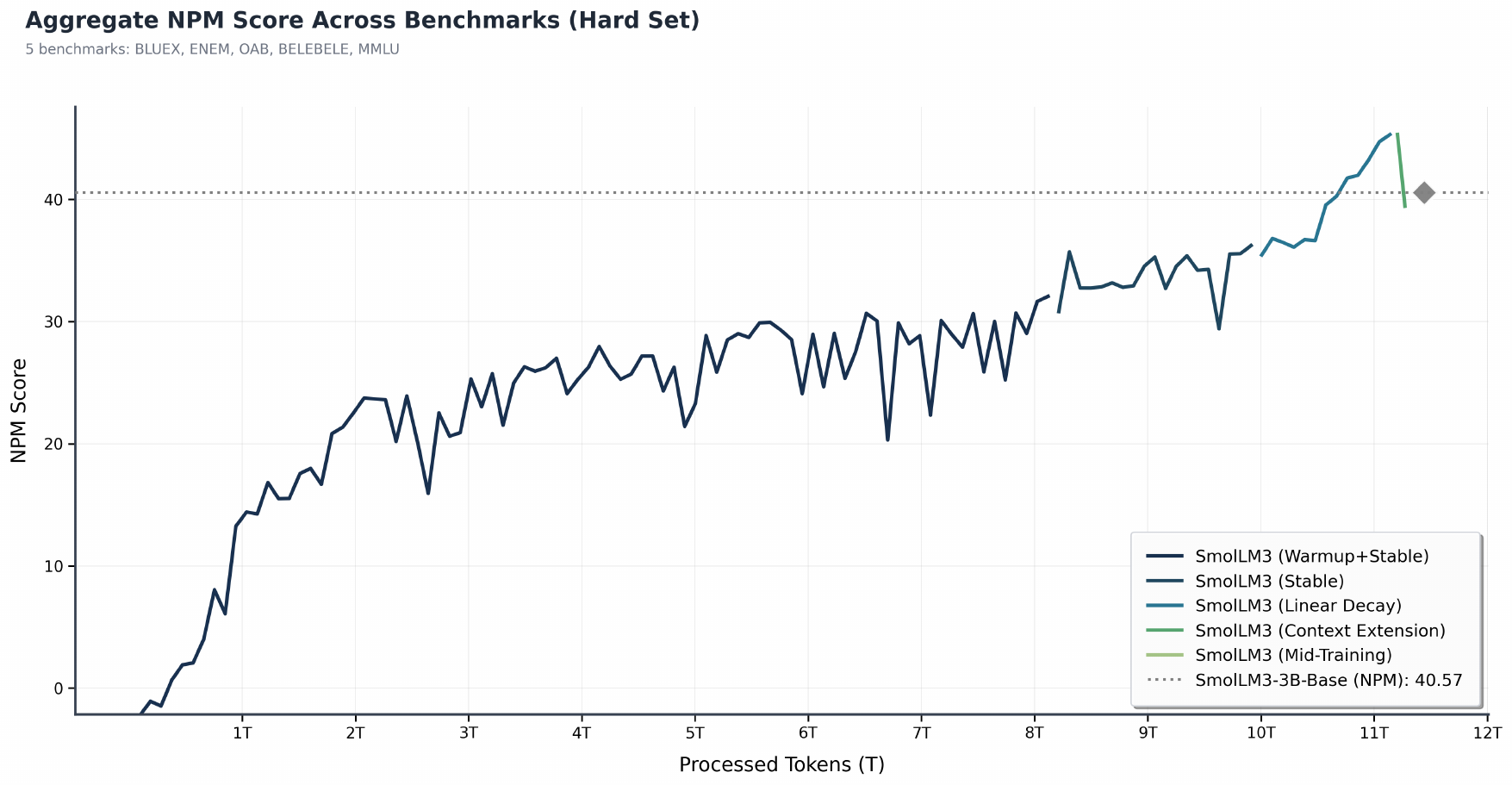} \\
(a) Easy Set & (b) Hard Set \\
\end{tabular}
\caption{NPM aggregated scores for the Easy Set and Hard Set. The Easy Set provides clear signals in early training stages ($<$200B tokens), whereas the Hard Set requires more training data but shows substantial improvements at later stages ($>$660B tokens).}
\label{fig:unified-npm}
\end{figure}

Figure~\ref{fig:unified-npm} shows the NPM aggregated scores for the Easy Set and Hard Set separately.

\subsection{Post-Training Evaluation Details}
\label{appendix:post-training-eval}

\subsubsection{IFEval-PT}

We manually translated and adapted 300 samples from the original IFEval dataset~\citep{zhou2023instructionfollowingevaluationlargelanguage} into Portuguese.

Table~\ref{tab:ifeval-metrics} describes the four evaluation metrics provided by IFEval-PT.

\begin{table}[!htbp]
\centering
\small
\caption{IFEval-PT evaluation metrics. Each metric is computed at either the prompt level (all instructions must be followed) or instruction level (per-instruction check).}
\label{tab:ifeval-metrics}
\begin{tabular}{p{4.5cm}p{9cm}}
\toprule
\textbf{Metric} & \textbf{Description} \\
\midrule
\texttt{prompt\_level\_strict\_acc} & Strict prompt accuracy: True only if \textit{all} instructions are followed exactly with no formatting flexibility. \\
\texttt{inst\_level\_strict\_acc} & Strict instruction accuracy: Checks each instruction separately with exact matching. \\
\texttt{prompt\_level\_loose\_acc} & Loose prompt accuracy: Like strict prompt accuracy, but allows minor formatting variations. \\
\texttt{inst\_level\_loose\_acc} & Loose instruction accuracy: Per-instruction check using loose (format-tolerant) matching. \\
\bottomrule
\end{tabular}
\end{table}

\subsubsection{GSM8K-PT}

We translated the GSM8K mathematical reasoning benchmark~\citep{cobbe2021trainingverifierssolvemath} using Qwen2.5-32B-Instruct, followed by manual review and correction, yielding 1,295 Portuguese math problems. The evaluation is performed via exact-match scoring with two extraction methods:

\begin{itemize}
    \item \textbf{strict-match:} Extracts numbers only from a required format tag (e.g., ``\#\#\#\# 42'').
    \item \textbf{flexible-extract:} Uses any detected number in the output as the answer, providing more lenient matching.
\end{itemize}

\subsubsection{RULER-PT}

We adapted the RULER long-context benchmark~\citep{hsieh2024rulerwhatsrealcontext, kim2025rulermeasureallbenchmarking} to Portuguese, including needle-in-a-haystack (NIAH) retrieval tasks, variable tracking, and common word extraction across multiple context lengths (1K--128K tokens). Table~\ref{tab:ruler-tasks} describes each task.

\begin{table}[!htbp]
\centering
\small
\caption{RULER-PT task descriptions. Tasks span retrieval, multi-hop reasoning, and aggregation across varying context lengths.}
\label{tab:ruler-tasks}
\begin{tabular}{p{3.5cm}p{2.5cm}p{7.5cm}}
\toprule
\textbf{Task} & \textbf{Category} & \textbf{Description} \\
\midrule
\texttt{niah\_pt\_single\_1}   & Retrieval (NIAH)  & Find a hidden key/value pair embedded in random tokens. \\
\texttt{niah\_pt\_single\_2}   & Retrieval (NIAH)  & Recover a key/value pair embedded in natural Portuguese text. \\
\texttt{niah\_pt\_single\_3}   & Retrieval (NIAH)  & Complex retrieval: values are long/random; exact match required. \\
\texttt{niah\_pt\_multikey\_1} & Retrieval (NIAH)  & Retrieve the correct value for a queried key (simple). \\
\texttt{niah\_pt\_multikey\_2} & Retrieval (NIAH)  & Match key to value in structured KV pairs. \\
\texttt{niah\_pt\_multikey\_3} & Retrieval (NIAH)  & Retrieve correct values amid noise and distractors (advanced). \\
\texttt{niah\_pt\_multivalue}  & Retrieval (NIAH)  & Output all values associated with a single key. \\
\texttt{niah\_pt\_multiquery}  & Retrieval (NIAH)  & Handle multiple queries; score each independently. \\
\texttt{ruler\_pt\_vt}         & Multi-Hop Tracing & Follow chained variable assignments to compute final value. \\
\texttt{ruler\_pt\_cwe}        & Aggregation       & Identify true top-k frequent words (uniform + noise distribution). \\
\texttt{ruler\_pt\_fwe}        & Aggregation       & Rank frequent words (heavy-tailed distribution). \\
\bottomrule
\end{tabular}
\end{table}

The RULER Score is computed as a weighted average of all task scores, reflecting the model's overall capability in handling long-context retrieval and reasoning tasks.

\subsubsection{HumanEval}

We evaluate code generation using the HumanEval benchmark~\citep{chen2021codex}, originally introduced alongside OpenAI Codex. HumanEval comprises 164 Python programming problems, each defined by a function signature, a natural-language specification, and hidden unit tests that verify functional correctness.

To ensure linguistic consistency with our Portuguese evaluation suite, we simply adapted the instruction prompts from the original HumanEval implementation in the LM-Evaluation-Harness to Portuguese, keeping the function signatures, canonical solutions, and test cases unchanged. No modifications were made to the underlying logic, test harness, or execution protocol. This preserves full comparability with prior work while isolating the effect of prompt language.

\begin{tcolorbox}[colback=gray!5!white, colframe=gray!75!black, title=HumanEval Portuguese Prompt]
\small

Escreva uma solução para o seguinte problema e certifique-se de que ela passe nos testes:

\begin{lstlisting}[language=Python]
```python
{{ prompt }}
```
\end{lstlisting}

Aqui está a função completa:

\begin{lstlisting}[language=Python]
```python
{{ prompt }}
[Model Completion ...]
\end{lstlisting}

\end{tcolorbox}

All models are evaluated in a zero-shot setting. We report pass@1, which measures the proportion of problems solved correctly by the model’s first generated solution. A solution is considered correct only if it passes all hidden unit tests.

\newpage
\section{Infrastructure: Details and Scalability}
\label{appendix:infrastructure-software-stack}

This appendix provides technical details about our computational infrastructure, software stack implementation, and scalability analysis. We present hardware specifications for the Marvin HPC cluster, detailed descriptions of our distributed training setup, and scaling experiments that characterize the performance of our training system across different configurations.

\subsection{Marvin HPC Cluster: Technical Specifications}
\label{appendix:infrastructure-marvin-cluster}

Marvin is a state-of-the-art tier 3 HPC cluster hosted at the University of Bonn. Below, we provide detailed specifications for all node types used in this work.

\subsubsection{Compute Node Configurations}

\paragraph{192 MPP Nodes (IntelSR Partition).}
\begin{itemize}
    \item CPU: 2$\times$ Intel Xeon ``Sapphire Rapids'' 48-core/96-thread at 2.10GHz
    \item RAM: 1024GB DDR5 4800MHz
    \item Local Storage: 1$\times$ 1.92TB SSD U.3 NVMe
    \item Total Aggregate: 18,432 cores, 196.6TB RAM
\end{itemize}

\paragraph{24 Large Memory Nodes (LM Partition).}
\begin{itemize}
    \item CPU: 2$\times$ Intel Xeon ``Sapphire Rapids'' 48-core/96-thread at 2.10GHz
    \item RAM: 2048GB DDR5 4800MHz
    \item Local Storage: 1$\times$ 1.92TB SSD U.3 NVMe
    \item Total Aggregate: 2,304 cores, 49.2TB RAM
\end{itemize}

\paragraph{5 Very Large Memory Nodes (VLM Partition).}
\begin{itemize}
    \item CPU: 2$\times$ Intel Xeon ``Sapphire Rapids'' 48-core/96-thread at 2.10GHz
    \item RAM: 4096GB DDR5 4800MHz
    \item Local Storage: 1$\times$ 3.84TB SSD U.3 NVMe
    \item Total Aggregate: 480 cores, 20.5TB RAM
\end{itemize}

\paragraph{32 A100 GPU Nodes (SGPU Partition).}
\begin{itemize}
    \item CPU: 2$\times$ AMD EPYC ``Milan'' 64-core/128-thread at 2.00GHz
    \item RAM: 1024GB DDR4 3200MHz
    \item GPU: 4$\times$ NVIDIA A100 80GB (NVLink-connected within node)
    \item Total Aggregate: 4,096 cores, 32.8TB RAM, 128 A100 GPUs
\end{itemize}

\paragraph{24 A40 GPU Nodes (MLGPU Partition).}
\begin{itemize}
    \item CPU: 2$\times$ AMD EPYC ``Milan'' 64-core/128-thread at 2.00GHz
    \item RAM: 512GB DDR4 3200MHz
    \item GPU: 8$\times$ NVIDIA A40 48GB
    \item Total Aggregate: 3,072 cores, 12.3TB RAM, 192 A40 GPUs
\end{itemize}

\subsubsection{Storage and Network Infrastructure}

\paragraph{Storage System.}
\begin{itemize}
    \item File System: Lustre
    \item Capacity: 5.6PB for user data
    \item Configuration: High-throughput parallel file system optimized for large-scale I/O
\end{itemize}

\paragraph{Network Interconnect.}
\begin{itemize}
    \item Technology: Mellanox InfiniBand NDR
    \item Bandwidth: 200Gb/s
    \item Topology: Non-blocking fat-tree for optimal collective communication
\end{itemize}

Different components of our pipeline leveraged different partitions based on their computational requirements:

\begin{itemize}
    \item \textbf{Data Processing:} Quality filtering, deduplication, and tokenization were conducted primarily on IntelSR, LM, and VLM partitions, utilizing high core counts and large memory capacities for parallel processing of our datasets.
    \item \textbf{Annotation and Filtering:} Learned filter training and synthetic data generation used the MLGPU partition (A40 nodes), balancing GPU capacity with cost efficiency for inference workloads.
    \item \textbf{Model Evaluation:} Benchmark evaluations were conducted on both MLGPU and SGPU partitions, depending on availability and model size.
    \item \textbf{Pretraining:} Large-scale distributed pretraining runs were executed exclusively on the SGPU partition (A100 nodes), which is optimized for highly scalable multi-node GPU applications.
    \item \textbf{Post-Training:} Supervised fine-tuning and preference alignment were conducted on both SGPU and MLGPU partitions, depending on model size and memory requirements.
\end{itemize}

\subsection{Software Stack: the Foundry}
\label{appendix:infrastructure-software-stack-foundry}

Our codebase, \href{https://github.com/Polygl0t/llm-foundry}{Polygl0t/llm-foundry}, is a custom open-source stack built on several foundational libraries that supports the entire lifecycle of large language model development, from data processing to model training and evaluation. It integrates several foundational libraries to provide a complete pipeline from data processing to model evaluation:

\begin{itemize}
    \item \textbf{PyTorch~\citep{paszke2019pytorch}:} Provides a flexible and performant foundation for distributed training, with native support for Distributed Data Parallelism (DDP), Fully Sharded Data Parallelism (FSDP), and mixed-precision training.
    \item \textbf{Hugging Face Transformers~\citep{wolf-etal-2020-transformers}:} Ensures compatibility with the broader ecosystem of pre-trained models, serving also as a standardized way to define and port known architectures.
    \item \textbf{vLLM~\citep{kwon2023efficient}:} Provides high-throughput, memory-efficient inference for large language models. We use vLLM for all inference-intensive tasks, including synthetic data generation and LLM-based annotation, achieving up to 4× higher throughput than standard Hugging Face inference via techniques such as PagedAttention and continuous batching.
    \item \textbf{Datatrove~\citep{penedo2024datatrove}:} Offers a comprehensive toolkit for large-scale text data processing, including extraction, filtering, and deduplication. Datatrove's modular pipeline architecture enabled efficient processing of our datasets across distributed CPU nodes.
    \item \textbf{Datasets~\citep{lhoest-etal-2021-datasets}:} Provides efficient data loading and processing for large-scale corpora, with support for streaming, memory-mapped access, and distributed caching. We use Datasets extensively for managing training data and evaluation benchmarks.
    \item \textbf{SentencePiece~\citep{kudo2018sentencepiece}:} Implements unsupervised text tokenization with efficient training on large corpora.
    \item \textbf{LM-Evaluation-Harness~\citep{eval-harness}:} Provides a unified framework for evaluating language models across diverse benchmarks. We extended the harness with Portuguese-specific tasks and used it for all pre- and post-training evaluations, ensuring standardized, reproducible benchmark results.
    \item \textbf{TRL (Transformer Reinforcement Learning)~\citep{vonwerra2020trl}:} Offers implementations of supervised fine-tuning (SFT) and preference optimization algorithms (DPO, APO). We use TRL for all post-training stages, including instruction fine-tuning and alignment with preference data.
    \item \textbf{Mergekit~\citep{goddard2025arceesmergekittoolkitmerging}:} A toolkit for merging pre-trained language models. It uses an out-of-core approach to perform several merging methods, including the tokenizer transplantation we employed for continual pretraining.
\end{itemize}

\subsubsection{Distributed Training}

We employ two complementary parallelism strategies to train our models:

\paragraph{Distributed Data Parallelism (DDP).}
DDP replicates the full model and optimizer state across GPUs and distributes training data across devices. During the backward pass, gradients are averaged across all GPUs using NCCL's AllReduce collective operation. DDP achieves near-linear scaling efficiency for models that fit comfortably within a single GPU's memory budget. We use DDP for our smaller models (0.5B--1.5B parameters) when training with modest batch sizes on a small number of nodes.

\paragraph{Fully Sharded Data Parallelism (FSDP).}
FSDP~\citep{zhao2023pytorchfsdpexperiencesscaling} implements the ZeRO optimization strategy~\citep{rajbhandari2020zeromemoryoptimizationstraining}, sharding optimizer states, gradients, and model parameters across the data-parallel dimension. This enables training models that exceed a single GPU's memory capacity. During the forward pass, FSDP materializes only the parameter shards required for the current layer (via AllGather), while during the backward pass, gradients are reduced and scattered (via ReduceScatter). We primarily use FSDP (v2) with no re-shard after the forward pass (equivalent to ZeRO Stage 2) for our larger model (3.7B parameters).

\subsubsection{Optimizations}

We implement several optimization techniques to maximize training efficiency:

\paragraph{Mixed-Precision Training with BF16 and TF32.}
All models are trained using BF16 (Brain Float 16) mixed precision, which reduces memory consumption and accelerates matrix operations while maintaining numerical stability better than FP16. BF16 uses the same exponent range as FP32 (8 bits) but reduces the mantissa to 7 bits, eliminating the need for loss scaling while preserving dynamic range. Additionally, we enable TF32 (TensorFloat-32) tensor cores on NVIDIA Ampere GPUs~\citep{valero2023mixed}.

\paragraph{Grouped-Query Attention (GQA).}
We implement grouped-query attention~\citep{ainslie2023gqatraininggeneralizedmultiquery} across all Tucano2 models, reducing the memory footprint of the key-value cache during inference. GQA groups multiple query heads into a single key-value head, substantially reducing the KV cache size while maintaining model quality.

\paragraph{Activation Recomputation (Re-materialization).}
Activation Recomputation~\citep{chen2016trainingdeepnetssublinear} reduces memory consumption by discarding intermediate activations during the forward pass and recomputing them during the backward pass. This trades computation for memory, enabling training with longer sequences, larger batch sizes, or larger models on fixed hardware.

\paragraph{FlashAttention2.}
We integrate FlashAttention2~\citep{dao2023flashattention2fasterattentionbetter}, a highly optimized attention kernel that fuses attention operations and exploits GPU memory hierarchy to achieve IO-optimal complexity. Unlike standard attention implementations that scale quadratically in memory ($O(N^2)$ for sequence length $N$), FlashAttention-2 achieves linear memory complexity ($O(N)$) through block-wise computation and on-chip SRAM tiling. On NVIDIA A100 GPUs, FlashAttention-2 reaches up to 73\% of theoretical peak FLOPS for attention operations.

\paragraph{Fused Triton Kernels via Liger.}
We use the Liger Kernel library~\citep{hsu2025ligerkernel}, which provides highly optimized Triton kernels for common language-model operations. Liger fuses multiple operations into a single GPU kernel, thereby reducing memory bandwidth requirements and kernel-launch overhead. Specifically, we use Liger's implementations of RMSNorm, RoPE (Rotary Position Embedding, SwiGLU, and Cross-Entropy Loss. Liger's optimizations yield approximately 20\% higher multi-GPU training throughput and up to 60\% memory savings compared to standard PyTorch implementations. The memory savings are particularly impactful for cross-entropy computation, as materializing logits for large vocabularies requires substantial GPU memory.

\subsection{Scaling Analysis}
\label{appendix:infrastructure-scaling-analysis}

To characterize the scalability of our training infrastructure, we conducted a scaling study using a 7B-parameter reference language model with a standard Llama-style architecture. The study was performed on the JUWELS Booster module,\footnote{A full description of JUWELS Booster is available in \href{https://apps.fz-juelich.de/jsc/hps/juwels/booster-overview.html}{apps.fz-juelich.de/jsc/hps/juwels/booster-overview.html}.} which provides a similar hardware environment in terms of accelerators (NVIDIA A100 GPUs) and interconnect (InfiniBand NDR 200Gb/s).

\subsubsection{Experimental Setup}

Table~\ref{tab:scaling-setup} summarizes the model architecture and training configuration used in our scaling experiments.

\begin{table}[h]
\centering
\caption{Experimental setup for scaling experiments on a 7B-parameter reference model.}
\label{tab:scaling-setup}
\begin{tabular}{ll}
\toprule
\multicolumn{2}{c}{\textbf{Model Configuration}} \\
\midrule
Parameters & 7 billion \\
Architecture & Llama (32 layers, 4096 hidden dim, 32 attn heads) \\
Sequence Length & 4096 tokens \\
Vocabulary Size & 49,152 tokens \\
Activation Function & SwiGLU \\
Position Embedding & RoPE (Rotary Position Embedding) \\
Normalization & RMSNorm \\
\midrule
\multicolumn{2}{c}{\textbf{Training Configuration}} \\
\midrule
Parallelism Strategy & FSDP with ZeRO Stage 3 (full parameter sharding) \\
Precision & BF16 mixed precision with TF32 matrix multiplications \\
Micro-Batch Size & 2 samples per GPU (8192 tokens per GPU) \\
Global Batch Size & 1024 samples ($\sim$4M tokens) \\
\bottomrule
\end{tabular}
\end{table}

\paragraph{Hardware Configurations.}
We evaluated five configurations ranging from 4 nodes (16 GPUs) to 64 nodes (256 GPUs), each node equipped with 4$\times$ NVIDIA A100 40GB GPUs:

\begin{itemize}
    \item 4 nodes: 16 GPUs, gradient accumulation steps = 32
    \item 8 nodes: 32 GPUs, gradient accumulation steps = 16
    \item 16 nodes: 64 GPUs, gradient accumulation steps = 8
    \item 32 nodes: 128 GPUs, gradient accumulation steps = 4
    \item 64 nodes: 256 GPUs, gradient accumulation steps = 2
\end{itemize}

\subsubsection{Scaling Results}

Table~\ref{tab:scaling-7b} presents scaling results across all configurations. We report the time per training step, the achieved tokens per second, the compute utilization (FLOPS per GPU), and the speedup relative to the 4-node baseline.

\begin{table}[h]
\centering
\caption{Scaling results for 7B-parameter model. Global batch size fixed at 1024 samples ($\sim$4M tokens).}
\label{tab:scaling-7b}
\begin{tabular}{ccccccc}
\toprule
\textbf{Nodes} & \textbf{GPUs} & \textbf{Time/Step (s)} & \textbf{Tokens/s} & \textbf{TFLOPS/GPU} & \textbf{Speedup} & \textbf{Efficiency (\%)} \\
\midrule
4  & 16  & 75.55 & 55,513  & 165.3 & 1.00$\times$  & 100.0 \\
8  & 32  & 38.24 & 109,731 & 163.5 & 1.98$\times$  & 98.8  \\
16 & 64  & 19.72 & 212,676 & 158.4 & 3.83$\times$  & 95.8  \\
32 & 128 & 10.11 & 414,968 & 154.5 & 7.47$\times$  & 93.4  \\
64 & 256 & 5.41  & 774,753 & 144.3 & 13.96$\times$ & 87.3  \\
\bottomrule
\end{tabular}
\end{table}

\paragraph{Key Observations.}

\begin{itemize}
    \item \textbf{Near-Linear Speedup:} Scaling from 4 nodes to 64 nodes (16$\times$ increase in hardware) yields a 13.96$\times$ speedup, corresponding to 87.3\% parallel efficiency.
    
    \item \textbf{Throughput Scaling:} Tokens per second increase from 55K (4 nodes) to 775K (64 nodes), enabling efficient training of large models on trillion-token datasets. At the 64-node configuration, the system processes 774K tokens/second, sufficient to complete a 1-trillion-token training run in approximately 15 days of continuous training.
    
    \item \textbf{Model FLOP Utilization:} At 4 nodes, we achieve 165 TFLOPS per A100 GPU, corresponding to approximately 53\% of the theoretical peak (312 TFLOPS for BF16 operations). As we scale to 64 nodes, FLOPS per GPU decreases slightly to 144 (46\%), reflecting increased communication overhead relative to computation. This is expected behavior.
    
    \item \textbf{Communication Overhead:} The gradual decrease in parallel efficiency from 98.8\% (8 nodes) to 87.3\% (64 nodes) indicates increasing communication costs as we scale. FSDP's AllGather and ReduceScatter collectives dominate communication volume, and their relative cost increases as per-GPU computation decreases.
    
\end{itemize}

\begin{figure}[h]
\centering
\includegraphics[width=0.48\textwidth]{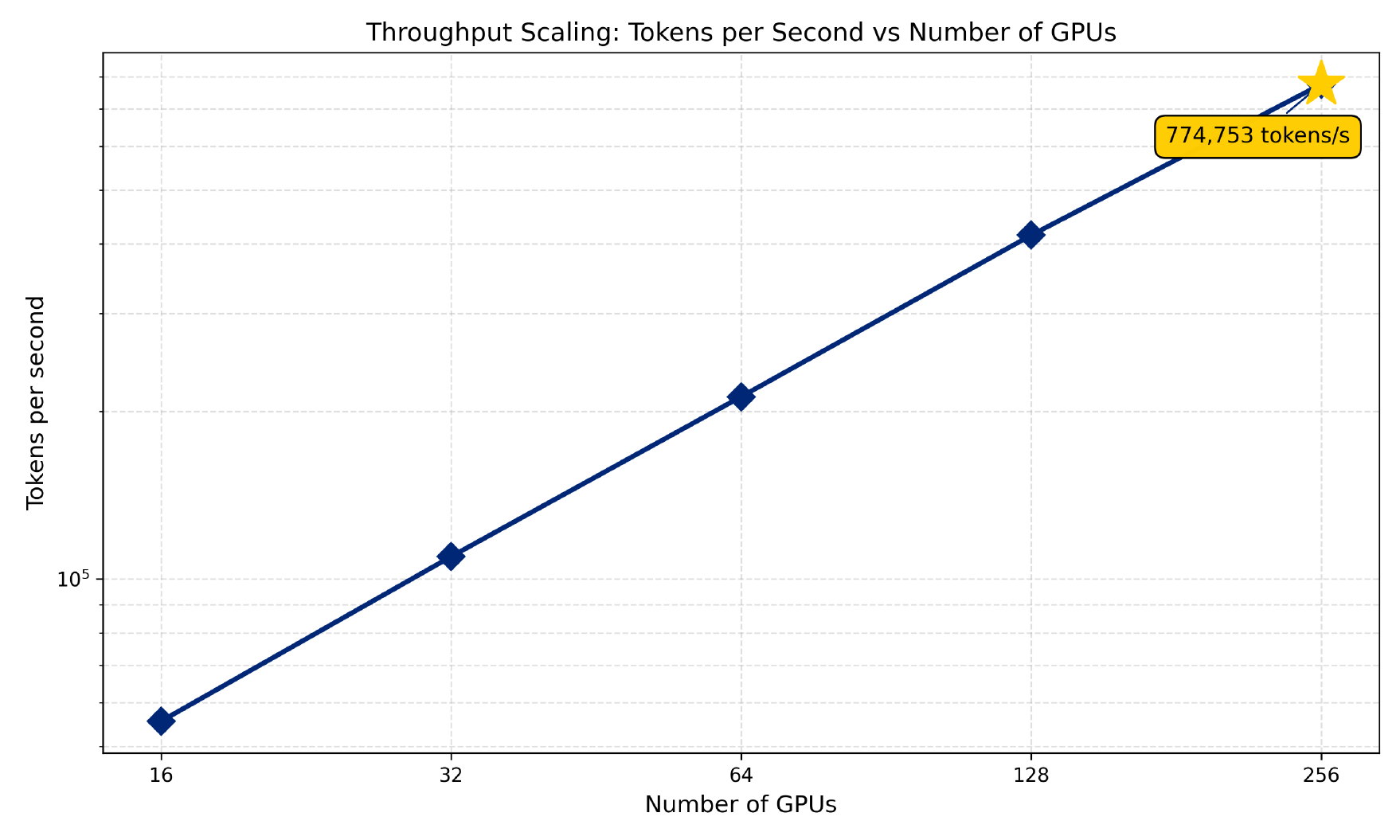}
\hfill
\includegraphics[width=0.48\textwidth]{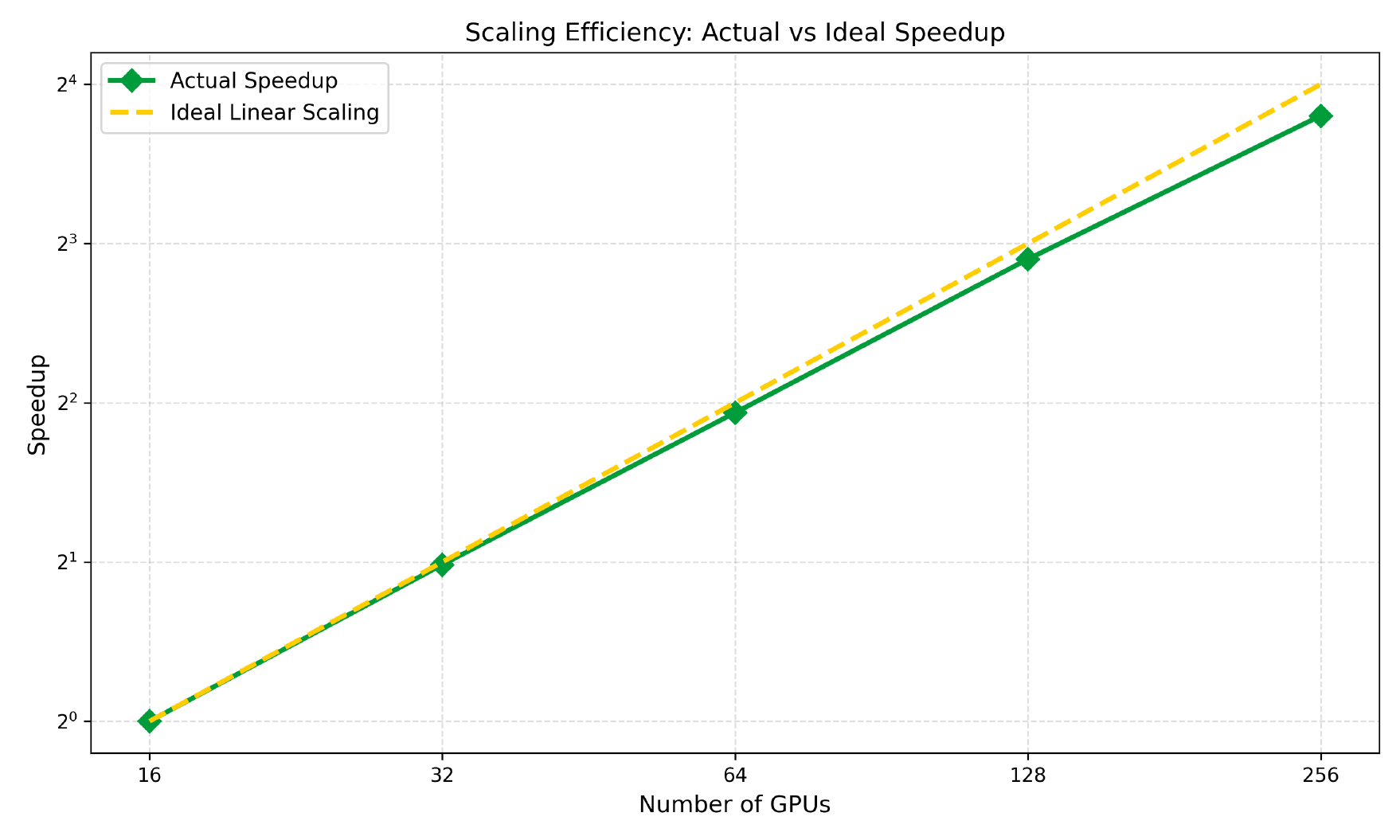}
\caption{Scaling analysis for 7B-parameter model. \textbf{Left:} Throughput (tokens/second) and achieved FLOPS per GPU as a function of node count. \textbf{Right:} Speedup relative to 4-node baseline, compared against ideal linear scaling. Both plots demonstrate near-linear scalability up to 64 nodes.}
\label{fig:scaling-analysis}
\end{figure}

Figure~\ref{fig:scaling-analysis} presents two complementary views of our scaling results. The left panel shows absolute throughput (tokens/second) and achieved FLOPS as a function of the number of nodes, demonstrating a near-linear increase in processing capacity. The right panel shows speedup relative to the 4-node baseline, along with a reference line indicating perfect linear scaling. The close alignment between observed speedup and the ideal scaling line up to 32 nodes, with only modest deviation at 64 nodes, confirms the scalability of our implementation.

\newpage
For completeness, we document key system configurations used during training in Table~\ref{tab:system-configurations}.

\begin{table}[h]
\centering
\caption{Training system software and configuration details.}
\label{tab:system-configurations}
\begin{tabular}{ll}
\toprule
\textbf{Component} & \textbf{Version/Configuration} \\
\midrule
\multicolumn{2}{c}{\textbf{Software Versions}} \\
\midrule
PyTorch & 2.8.0 \\
CUDA & 12.6.0 \\
NCCL & 2.18.3 \\
Transformers & 4.53.2 \\
Flash-Attention & 2.8.2 \\
Liger Kernel & 0.6.1 \\
\midrule
\multicolumn{2}{c}{\textbf{FSDP Configuration}} \\
\midrule
Sharding Strategy & \texttt{FULL\_SHARD} (ZeRO Stage 3) \\
Mixed Precision Policy & BF16+FP32/TF32 \\
CPU Offload & Disabled (all states kept on GPU) \\
Activation Checkpointing & Enabled \\
\midrule
\multicolumn{2}{c}{\textbf{NCCL Configuration}} \\
\midrule
Transport & InfiniBand (multi-node) / NVLink (intra-node) \\
IB Device & mlx5 (Mellanox ConnectX) \\
\texttt{NCCL\_IB\_DISABLE} & 0 (InfiniBand enabled) \\
\texttt{NCCL\_NET\_GDR\_LEVEL} & 5 (GPUDirect RDMA enabled) \\
\texttt{NCCL\_ALGO} & Tree (for large scale, $>$32 nodes) \\
\texttt{NCCL\_TIMEOUT} & 3600 (seconds) \\
\texttt{NCCL\_IB\_TIMEOUT} & 20 \\
\texttt{NCCL\_IB\_RETRY\_CNT} & 7 \\
\texttt{NCCL\_P2P\_DISABLE} & 0 (peer-to-peer enabled) \\
\texttt{NCCL\_SHM\_DISABLE} & 0 (shared memory enabled) \\
\bottomrule
\end{tabular}
\end{table}

\newpage
\section{Pretraining: Details}
\label{appendix:pretraining}

This appendix provides technical details for the pretraining experiments described in Section~\ref{sec:pretraining}.

\subsection{Architecture}
\label{appendix:pretraining-architecture}

Table~\ref{tab:appendix-architecture-specs} provides the complete architectural configuration for \tucanoptsmall. The model follows the Llama architecture~\citep{touvron2023llama} as implemented in Hugging Face Transformers~\citep{wolf-etal-2020-transformers}.

\begin{table}[h]
\centering
\caption{Architectural specifications.}
\label{tab:appendix-architecture-specs}
\begin{tabular}{lp{6cm}}
\toprule
\textbf{Parameter} & \textbf{Value} \\
\midrule
Total Parameters & 670,127,616 \\
Architecture & Llama \\
Hidden Activation & SwiGLU \\
Normalization & RMSNorm \\
Positional Embeddings & RoPE \\
Hidden Layer Size ($d_{\text{model}}$) & 1,536 \\
Feed-Forward (Intermediate) Size & 3,072 \\
Maximum Context Length & 4,096 tokens \\
Number of Attention Heads & 16 \\
Number of Layers ($n_{\text{layer}}$) & 28 \\
Attention Head Dimension & 96 \\
Number of Key/Value Heads (GQA) & 8 \\
Tied Input/Output Embeddings & True \\
Vocabulary Size & 49,152 \\
\bottomrule
\end{tabular}
\end{table}

\subsection{Optimizer Experiments: AdamW vs. AdamW+Muon}
\label{appendix:pretraining-optimizer-ablation}

To assess the potential benefits of the Muon optimizer~\citep{jordan2024muon} for our training setup, we conducted a controlled pilot study comparing the standard AdamW optimizer against a hybrid AdamW+Muon configuration.

\paragraph{Experimental Setup.} We trained two 0.6B-parameter models (identical architecture to \tucanoptsmall) for 12,000 optimization steps, corresponding to approximately 25 billion tokens (global batch size: 2,097,152 tokens). Both runs used the warmup-stable-decay learning rate schedule~\citep{hagele2024scalinglawscomputeoptimaltraining} with 2,000 warmup steps, 9,000 stable steps, and approximately 1,000 cooldown steps (1-sqrt Decay). The AdamW configuration used a maximum learning rate of $7 \times 10^{-4}$, while the Muon configuration used $7 \times 10^{-3}$. In the hybrid configuration, Muon was applied to all matrix-shaped hidden weights (including attention and feed-forward projections), while AdamW was applied to embeddings, normalization layers, bias terms, and all scalar parameters. The same weight decay coefficient was used for all decayed parameter groups, but Muon and AdamW used different learning rates. AdamW used $\beta_1$, $\beta_2$, and $\epsilon$, while Muon used $\beta_2$ as its momentum parameter.

\paragraph{Evaluation Metrics.} To measure the stability of the training process under both optimizers, we computed gradient-based metrics at each training step by recording the L2 norm of gradients across all model parameters. We define:

\begin{itemize}
    \item \textbf{Variability ($\sigma/\mu$):} The coefficient of variation, computed as the standard deviation divided by the mean gradient norm. This is a relative measure of dispersion, normalized by the average gradient scale.
    
    \item \textbf{Volatility ($\sigma(\Delta g_t)$):} The standard deviation of the changes in gradient norms between consecutive steps, where $\Delta g_t = \|g_t\| - \|g_{t-1}\|$. This metric captures the extent to which the gradient norm fluctuates from step to step.
\end{itemize}

\paragraph{Results.} Table~\ref{tab:appendix-optimizer-metrics} summarizes the gradient stability metrics for both configurations. The AdamW+Muon configuration exhibits substantially lower volatility (0.0769 vs. 0.1909) and reduced variability (1.6139 vs. 1.7906), suggesting smoother optimization dynamics. Final perplexity scores also favored the AdamW+Muon setup.

\begin{table}[h]
\centering
\caption{Gradient stability metrics for AdamW vs. AdamW+Muon over 12,000 steps ($\sim$25B tokens).}
\label{tab:appendix-optimizer-metrics}
\begin{tabular}{lcc}
\toprule
\textbf{Metric} & \textbf{AdamW} & \textbf{AdamW+Muon} \\
\midrule
Variability ($\sigma/\mu$) & 1.7906 & \textbf{1.6139} \\
Volatility ($\sigma(\Delta g_t)$) & 0.1909 & \textbf{0.0769} \\
\bottomrule
\end{tabular}
\end{table}

\begin{figure}[h]
\centering
\begin{subfigure}[b]{0.48\textwidth}
\centering
\includegraphics[width=\textwidth]{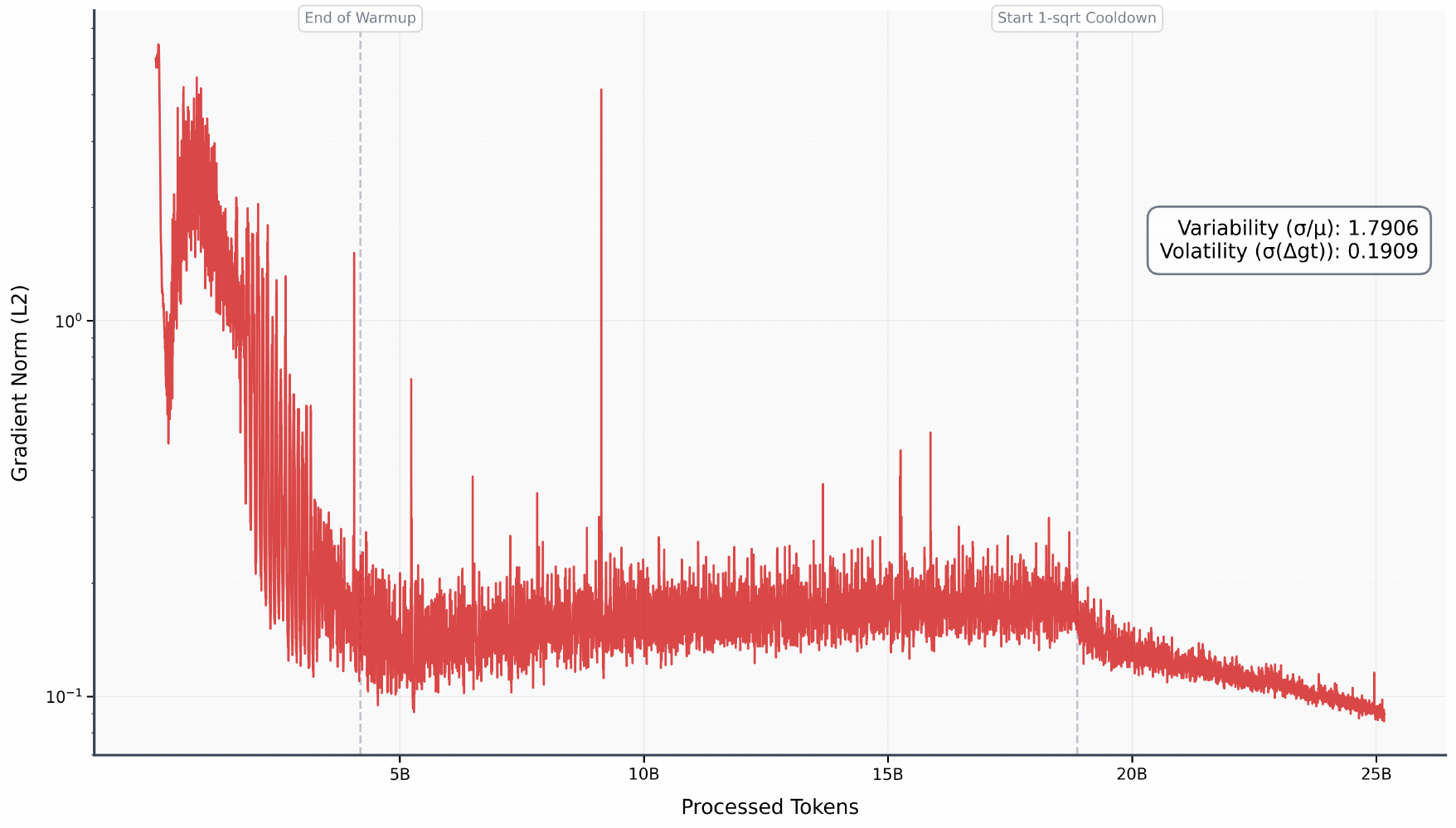}
\caption{AdamW}
\end{subfigure}
\hfill
\begin{subfigure}[b]{0.48\textwidth}
\centering
\includegraphics[width=\textwidth]{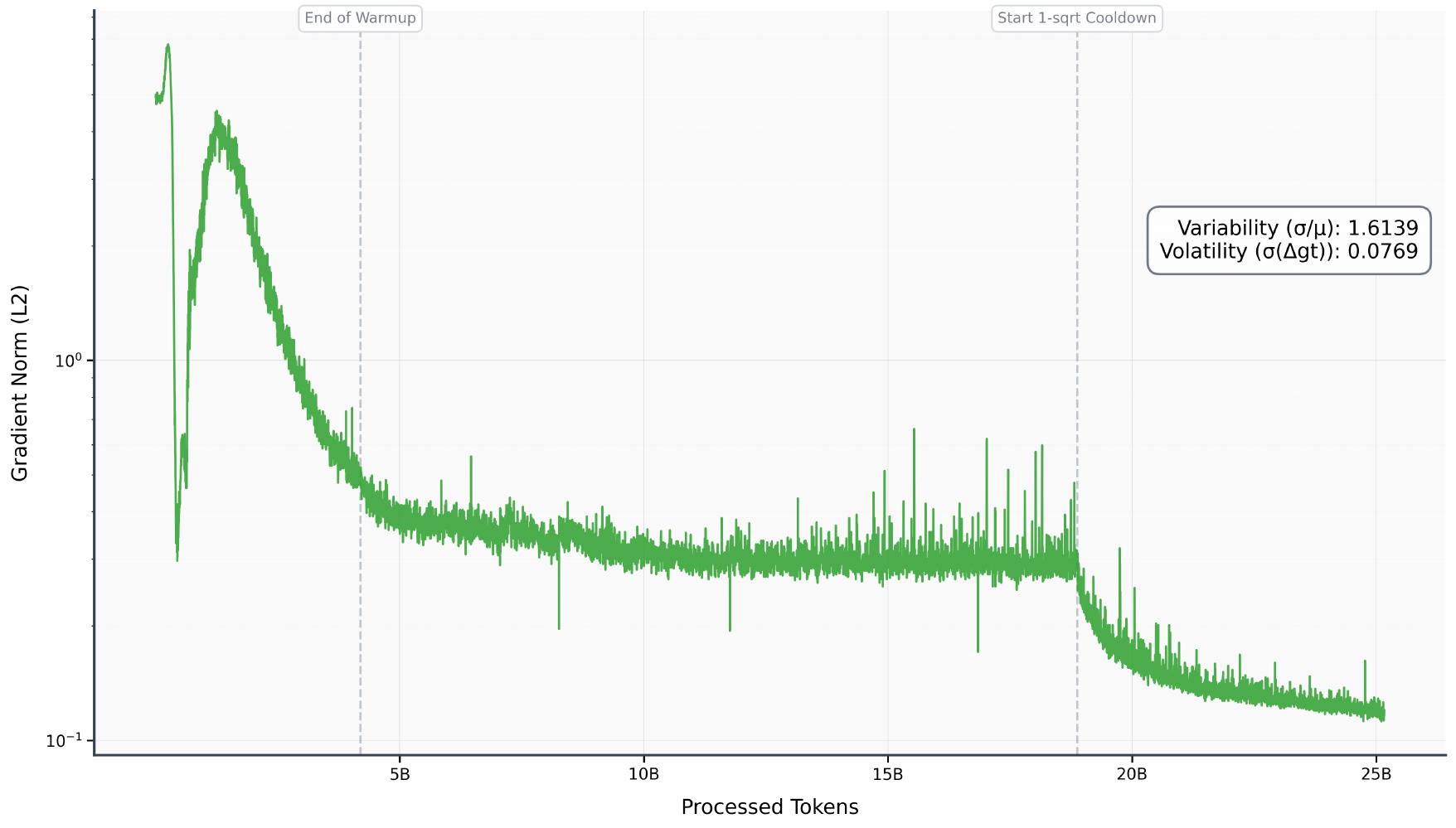}
\caption{AdamW+Muon}
\end{subfigure}
\caption{Gradient norm trajectories for AdamW (left) and AdamW+Muon (right) over 12,000 steps.}
\label{fig:appendix-optimizer-gradient-norms}
\end{figure}

Figure~\ref{fig:appendix-optimizer-gradient-norms} presents the gradient norm trajectories for both configurations throughout the training run.

\paragraph{Key Takeaways and Caveats.} The preliminary results suggest that Muon may significantly reduce gradient volatility, potentially contributing to more stable convergence and better generalization. However, these findings should be interpreted with caution, as they are based on a single model size and a relatively short training run. More comprehensive studies across different model scales, training durations, and data mixtures would be needed to fully characterize the benefits and potential trade-offs of using Muon in large-scale LLM training. Nevertheless, based on these encouraging initial results and the growing body of anecdotal evidence from large-scale deployments~\citep{nanochat, kimiteam2026kimik2openagentic, liu2025muonscalablellmtraining, chang2025convergencemuon}, we adopted AdamW+Muon as part of our full pretraining recipe.

\subsection{Hyperparameter Settings}
\label{appendix:pretraining-hyperparameters}

Table~\ref{tab:appendix-hyperparameters} provides the complete hyperparameter configuration used for training \tucanoptsmall. These values were derived using the empirical scaling laws from DeepSeek LLM~\citep{deepseekai2024deepseekllmscalingopensource}.

\begin{table}[h]
\centering
\caption{Hyperparameter settings for pretraining.}
\label{tab:appendix-hyperparameters}
\begin{tabular}{lp{6cm}}
\toprule
\textbf{Training Configuration} & \textbf{Value} \\
\midrule
Parallelism Strategy & DDP \\
Activation Checkpointing & False \\
Total Batch Size & 2,097,152 tokens ($2^{21}$) \\
Micro Batch Size (per GPU) & 16 samples \\
Gradient Accumulation Steps & 4 \\
Maximum Context Length & 4,096 tokens \\
Total Training Steps & 195,000 \\
Checkpointing Frequency & Every 5,000 steps \\
\midrule
\multicolumn{2}{c}{\textit{\textbf{Learning Rate Schedule}}} \\
\midrule
Schedule Type & Warmup-Stable-Decay (WSD) \\
Warmup Steps (Stage 1 only) & 2,000 \\
AdamW Maximum Learning Rate & $7 \times 10^{-4}$ \\
Muon Maximum Learning Rate & $7 \times 10^{-3}$ \\
Minimum Learning Rate & 0.0 \\
Stage 1 Decay Type & None (stable) \\
Stage 2 Decay Type & None (stable) \\
Stage 3 Decay Type & 1-sqrt Decay \\
\midrule
\multicolumn{2}{c}{\textit{\textbf{Optimizer Configuration}}} \\
\midrule
Optimizer & AdamW + Muon  \\
Weight Decay & 0.1 (excluding embeddings and norms) \\
Adam Beta 1 & 0.9 \\
Adam Beta 2 & 0.95 \\
Adam Epsilon & $1 \times 10^{-8}$ \\
Maximum Gradient Norm (Clipping) & 1.0 \\
\midrule
\multicolumn{2}{c}{\textit{\textbf{Resource Consumption}}} \\
\midrule
Total Training Tokens & $\sim$408 billion \\
Estimated FLOPs & $1.47 \times 10^{22}$ \\
Energy Consumption & 872.61 kWh \\
Carbon Footprint (NRW, Germany) & 332.42 kg CO$_2$e \\
\bottomrule
\end{tabular}
\end{table}

\paragraph{Derivation from Scaling Laws.} We applied the DeepSeek LLM scaling heuristics~\citep{deepseekai2024deepseekllmscalingopensource}, which relate the compute budget ($C$) to optimal hyperparameters via power-law relationships. The compute budget is estimated using the adjusted DeepSeek formulation:

\begin{equation}
C = \left(72 \, n_{\text{layer}} \, d_{\text{model}}^{2} + 12 \, n_{\text{layer}} \, d_{\text{model}} \, \ell_{\text{seq}}\right) D
\end{equation}

where $n_{\text{layer}} = 28$, $d_{\text{model}} = 1{,}536$, $\ell_{\text{seq}} = 4{,}096$, and $D \approx 408 \times 10^9$ tokens. Plugging in these values yields $C \approx 1.47 \times 10^{22}$ FLOPs. Using the DeepSeek scaling heuristics:

\begin{align}
\text{Max Learning Rate} &= 0.3118 \cdot C^{-0.125} \approx 7 \times 10^{-4} \\
\text{Batch Size} &= 0.2920 \cdot C^{0.3271} \approx 2.1 \times 10^{6} \text{ tokens}
\end{align}

We rounded the batch size to the nearest power of two ($2^{21} = 2{,}097{,}152$) for hardware efficiency.

\paragraph{Weight Decay Strategy.} Following recent best practices from OLMo 2~\citep{olmo20252olmo2furious} and SmolLM 3~\citep{bakouch2025smollm3}, we apply weight decay selectively, excluding embeddings and normalization layers.

\subsection{Data Mixture}
\label{appendix:pretraining-data-mixtures}

Tables~\ref{tab:appendix-stage1-mixture}, \ref{tab:appendix-stage2-mixture}, and \ref{tab:appendix-stage3-mixture} provide specifications for the data mixtures used in each training stage. We iteratively refined all proportions and repetition factors during early training checkpoints by monitoring the aggregate NPM score on the Easy Set evaluations.

\subsubsection{Language Proportions per Stage}

Table~\ref{tab:appendix-language-proportions} summarizes the language distribution across all three training stages. Throughout the training run, we maintain Portuguese as the majority language (63\% overall), with the proportion increasing in the final stage to emphasize high-quality native content.

\begin{table}[h]
\centering
\caption{Language proportions and token count per training stage.}
\label{tab:appendix-language-proportions}
\begin{tabular}{lcccc}
\toprule
\textbf{Stage} & \textbf{Portuguese (\%)} & \textbf{English (\%)} & \textbf{Steps} & \textbf{Total Tokens} \\
\midrule
Warmup+Stable & 61\% ($\sim$180B) & 39\% ($\sim$112B) & 100,000 & $\sim$292B \\
Stable & 54\% ($\sim$76B) & 46\% ($\sim$64B) & 60,000 & $\sim$140B \\
1-sqrt-Decay & 75\% ($\sim$58B) & 25\% ($\sim$20B) & 35,000 & $\sim$74B \\
\midrule
\textbf{Total} & \textbf{63\%} ($\sim$\textbf{314B}) & \textbf{37\%} ($\sim$\textbf{196B}) & \textbf{195,000} & $\sim$\textbf{408B} \\
\bottomrule
\end{tabular}
\end{table}

\subsubsection{Stage 1 (Warmup+Stable) Data Mixture}

Stage 1 spans 100,000 optimization steps, corresponding to approximately 292 billion tokens. The learning rate follows a linear warmup for the first 2,000 steps, reaching a peak of $7 \times 10^{-4}$ for AdamW and $7 \times 10^{-3}$ for Muon, then remains stable at this peak for the next 98,000 steps. The data mixture emphasizes high-quality educational content from both Portuguese and English sources.

\begin{table}[h]
\centering
\caption{Stage 1 (Warmup+Stable) data mixture: 100,000 steps, $\sim$292B tokens.}
\label{tab:appendix-stage1-mixture}
\begin{tabular}{lp{5cm}cc}
\toprule
\textbf{Dataset} & \textbf{Subset} & \textbf{Size (Tokens)} & \textbf{Repetition} \\
\midrule
\gigaverbo & Education Score $\geq$ 3 & 90B & 2$\times$ \\
FineWeb-Edu & Education Score $\geq$ 3 & 88B & 1$\times$ \\
FineMath & Education Score $\geq$ 3 & 24B & 1$\times$ \\
\midrule
\multicolumn{2}{l}{\textbf{Portuguese Subtotal}} & \textbf{180B (61\%)} & \\
\multicolumn{2}{l}{\textbf{English Subtotal}} & \textbf{112B (39\%)} & \\
\multicolumn{2}{l}{\textbf{Stage 1 Total}} & \textbf{292B} & \\
\bottomrule
\end{tabular}
\end{table}

\subsubsection{Stage 2 (Stable) Data Mixture}

Stage 2 spans 60,000 optimization steps, corresponding to approximately 140 billion tokens. The learning rate remains constant at its peak throughout this stage. The data mixture becomes more diverse, incorporating synthetic data and reasoning-focused datasets.

\begin{table}[h]
\centering
\caption{Stage 2 (Stable) data mixture: 60,000 steps, $\sim$140B tokens.}
\label{tab:appendix-stage2-mixture}
\begin{tabular}{lp{5cm}cc}
\toprule
\textbf{Dataset} & \textbf{Subset} & \textbf{Size (Tokens)} & \textbf{Repetition} \\
\midrule
\gigaverbo & Education Score $\geq$ 4 & 28B & 2$\times$ \\
\gigaverbosynth & All & 10B & 2$\times$ \\
FineWeb-Edu & Education Score $\geq$ 4 & 14B & 1$\times$ \\
Cosmopedia v2 & All & 30B & 1$\times$ \\
FineMath & Education Score $\geq$ 4 & 8B & 1$\times$ \\
Big Reasoning Traces & All & 2B & 1$\times$ \\
Math Meta Reasoning & All & 1B & 1$\times$ \\
OpenScience & All & 9B & 1$\times$ \\
\midrule
\multicolumn{2}{l}{\textbf{Portuguese Subtotal}} & \textbf{76B (54\%)} & \\
\multicolumn{2}{l}{\textbf{English Subtotal}} & \textbf{64B (46\%)} & \\
\multicolumn{2}{l}{\textbf{Stage 2 Total}} & \textbf{140B} & \\
\bottomrule
\end{tabular}
\end{table}

\subsubsection{Stage 3 (1-sqrt Decay) Data Mixture}

Stage 3 spans 35,000 optimization steps, corresponding to approximately 74 billion tokens. The learning rate is annealed from its peak value using a 1/sqrt decay schedule, as recommended by~\citet{hagele2024scalinglawscomputeoptimaltraining}. The data mixture further increases the proportion of Portuguese content to 75\%.

\begin{table}[h]
\centering
\caption{Stage 3 (1-sqrt Decay) data mixture: 35,000 steps, $\sim$74B tokens.}
\label{tab:appendix-stage3-mixture}
\begin{tabular}{lp{5cm}cc}
\toprule
\textbf{Dataset} & \textbf{Subset} & \textbf{Size (Tokens)} & \textbf{Repetition} \\
\midrule
\gigaverbo & Education Score $\geq$ 4 & 28B & 1$\times$ \\
\gigaverbosynth & All & 10B & 3$\times$ \\
FineMath & Education Score $\geq$ 4 & 8B & 1$\times$ \\
Big Reasoning Traces & All & 2B & 1$\times$ \\
Math Meta Reasoning & All & 1B & 1$\times$ \\
OpenScience & All & 9B & 1$\times$ \\
\midrule
\multicolumn{2}{l}{\textbf{Portuguese Subtotal}} & \textbf{58B (75\%)} & \\
\multicolumn{2}{l}{\textbf{English Subtotal}} & \textbf{20B (25\%)} & \\
\multicolumn{2}{l}{\textbf{Stage 3 Total}} & \textbf{78B} & \\
\bottomrule
\end{tabular}
\end{table}

\subsection{Training Dynamics}
\label{appendix:pretraining-gradient-norms}

\begin{figure}[h]
\centering
\includegraphics[width=0.95\linewidth]{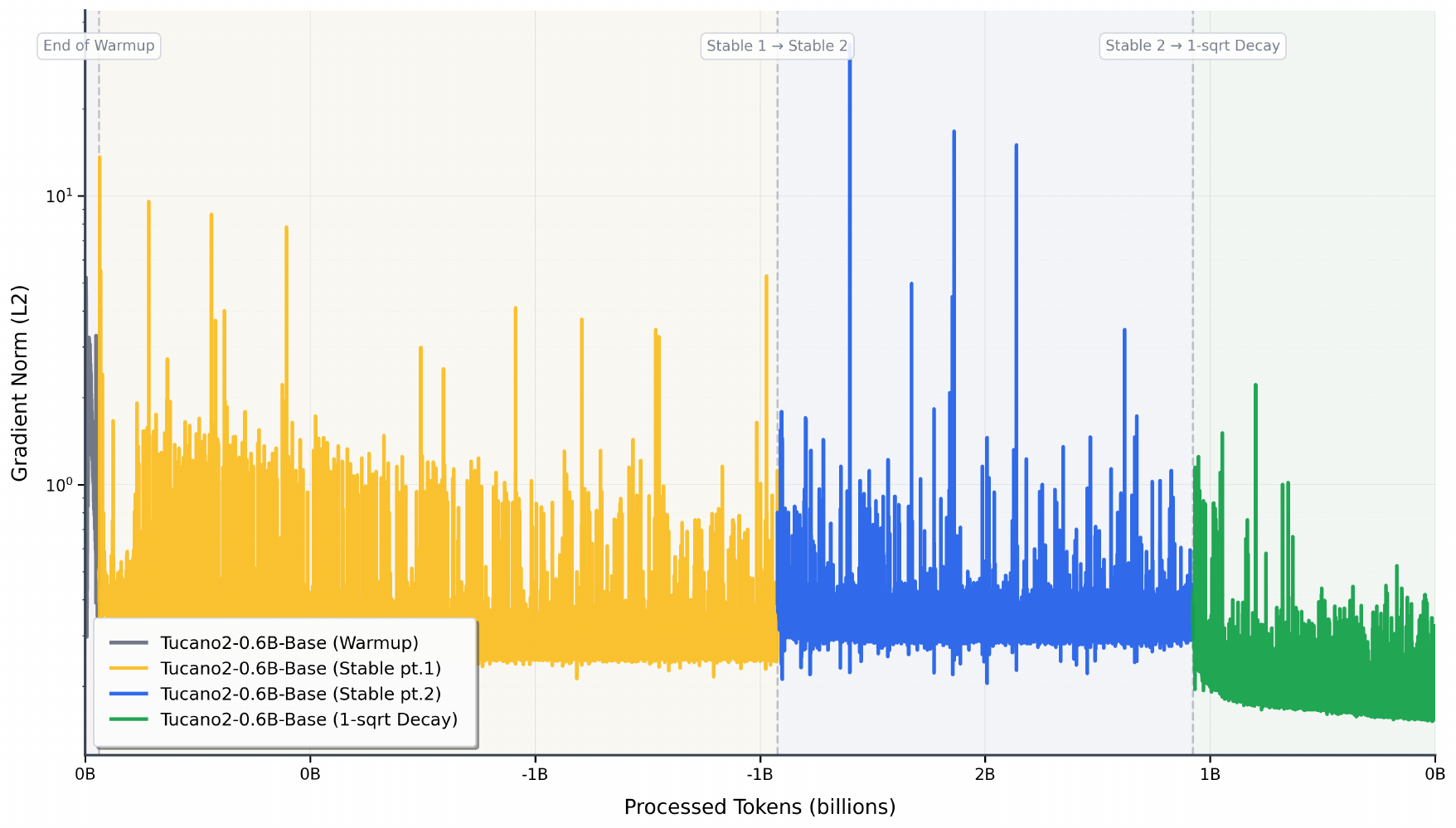}
\caption{Gradient norms throughout the complete pretraining run (195,000 steps).}
\label{fig:appendix-gradient-norms-full}
\end{figure}

Figure~\ref{fig:appendix-gradient-norms-full} presents the complete gradient norm trajectory throughout the entire training run across all three stages. The gradient norms exhibit expected patterns: a gradual increase during the warmup, stable behavior during the constant-learning-rate phases, and a gradual decay during the final annealing phase. No signs of gradient explosion or vanishing gradients are observed during training.

\paragraph{Stage-Specific Behaviors.} During Stage 1 (Warmup+Stable), gradient norms increase smoothly during the warmup phase (steps 0--2,000) and then stabilize with moderate fluctuations. Stage 2 (Stable) exhibits slightly higher variability in gradient norms due to a more diverse data distribution, but remains well within the stable range. Stage 3 (1-sqrt Decay) exhibits a gradually decreasing gradient norm as the learning rate is annealed, consistent with the model converging to a refined final state. These behaviors align with expectations for multi-stage training with WSD schedules~\citep{hagele2024scalinglawscomputeoptimaltraining, allal2025smollm2smolgoesbig, bakouch2025smollm3}.

\subsection{Per-Benchmark Results}
\label{appendix:pretraining-detailed-results}

\begin{figure}[h]
\centering
\begin{subfigure}[b]{0.48\textwidth}
    \centering
    \includegraphics[width=\textwidth]{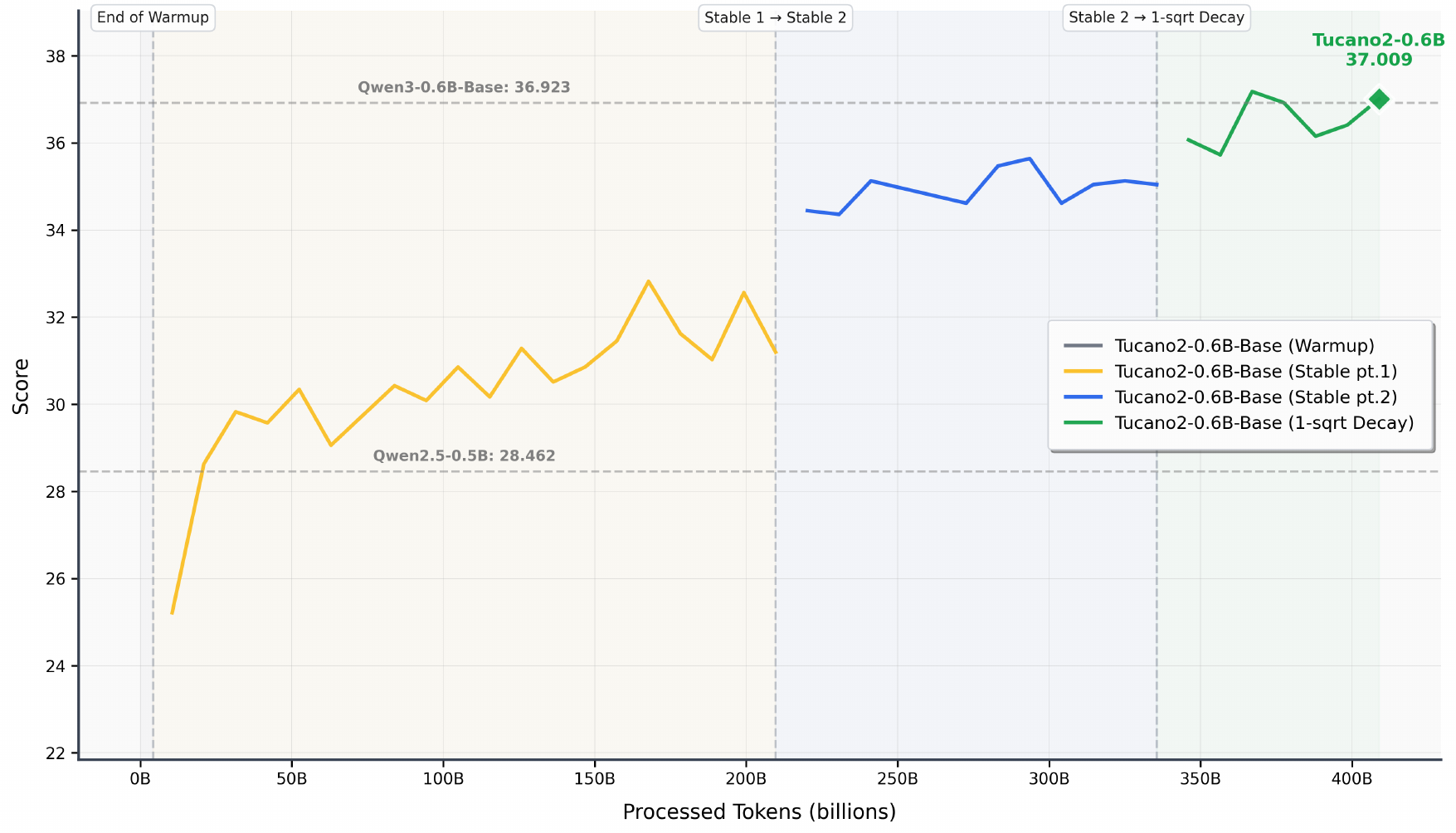}
    \caption{ARC Challenge}
    \label{fig:appendix-easy-arc}
\end{subfigure}
\hfill
\begin{subfigure}[b]{0.48\textwidth}
    \centering
    \includegraphics[width=\textwidth]{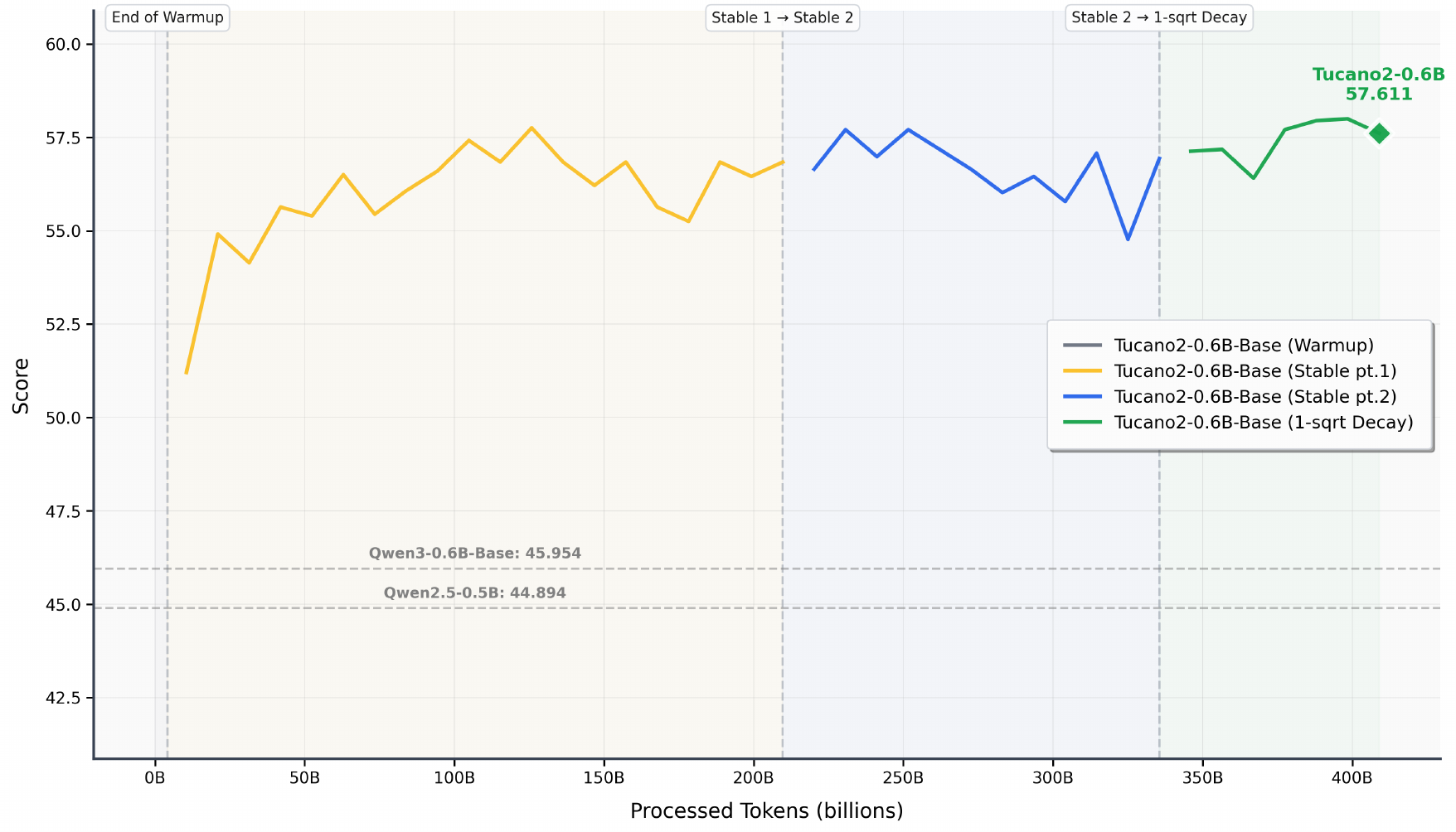}
    \caption{Calame}
    \label{fig:appendix-easy-calame}
\end{subfigure}

\vspace{0.5cm}

\begin{subfigure}[b]{0.48\textwidth}
    \centering
    \includegraphics[width=\textwidth]{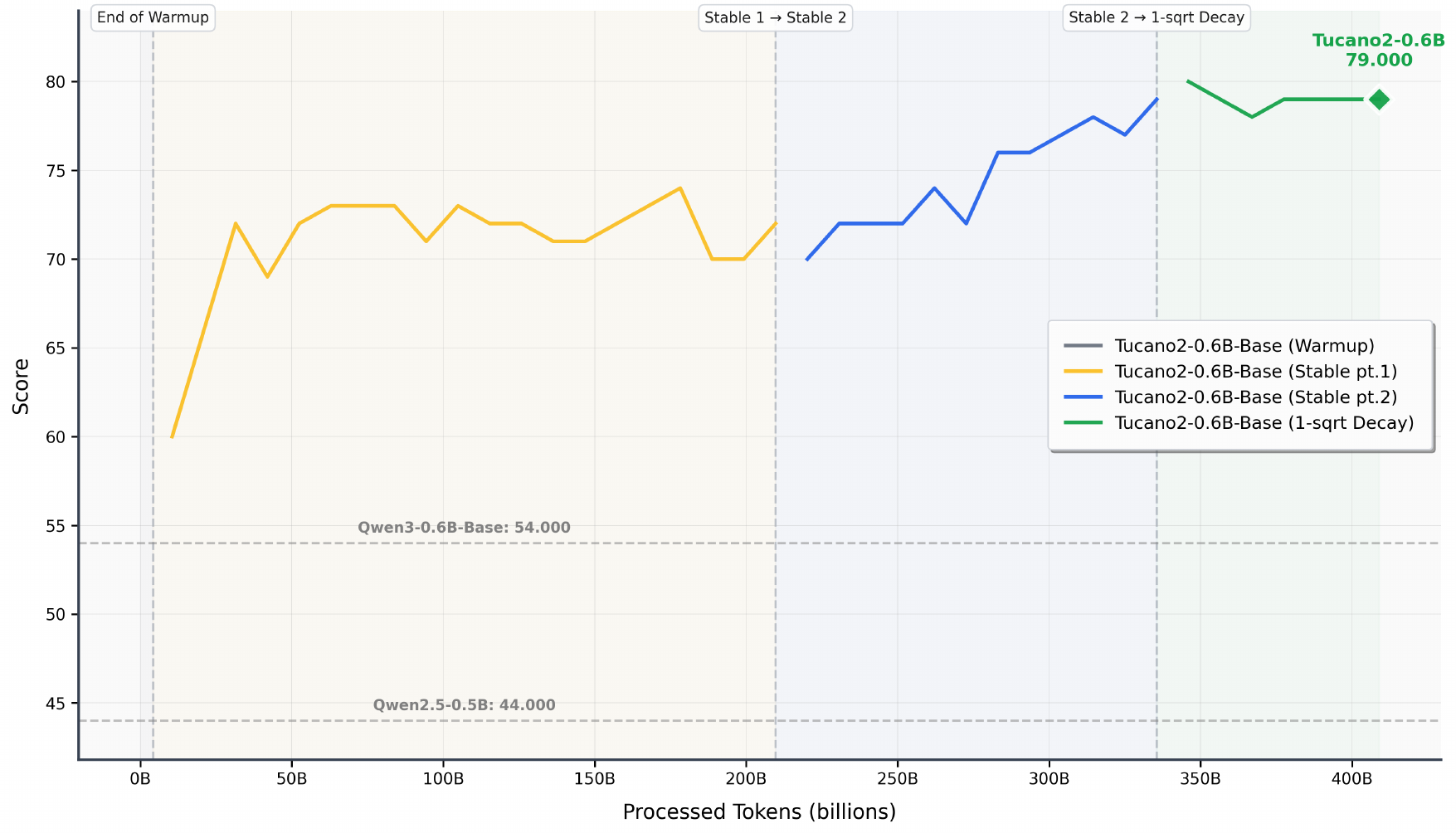}
    \caption{Global PIQA}
    \label{fig:appendix-easy-piqa}
\end{subfigure}
\hfill
\begin{subfigure}[b]{0.48\textwidth}
    \centering
    \includegraphics[width=\textwidth]{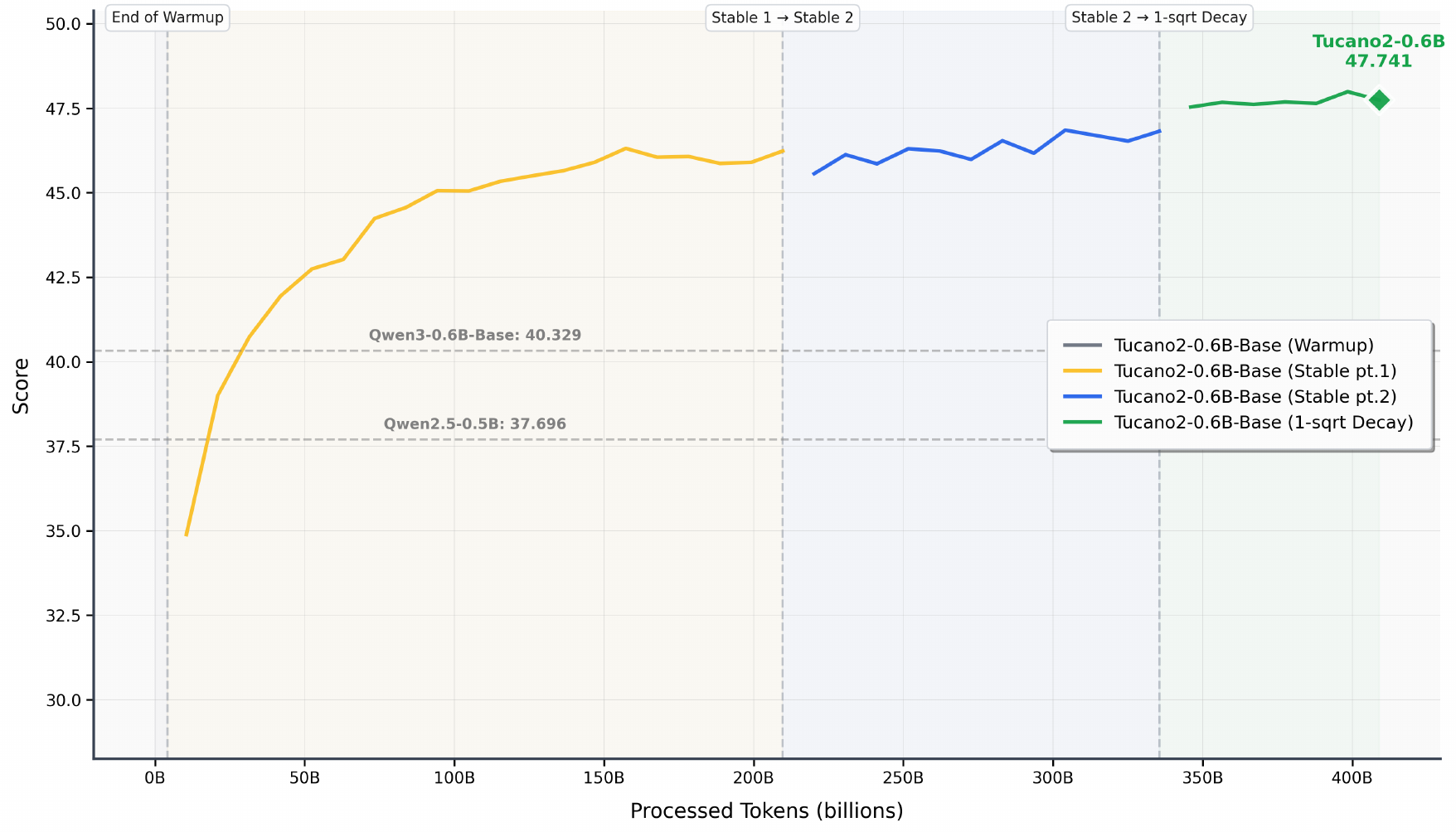}
    \caption{HellaSwag}
    \label{fig:appendix-easy-hellaswag}
\end{subfigure}

\vspace{0.5cm}

\begin{subfigure}[b]{0.48\textwidth}
    \centering
    \includegraphics[width=\textwidth]{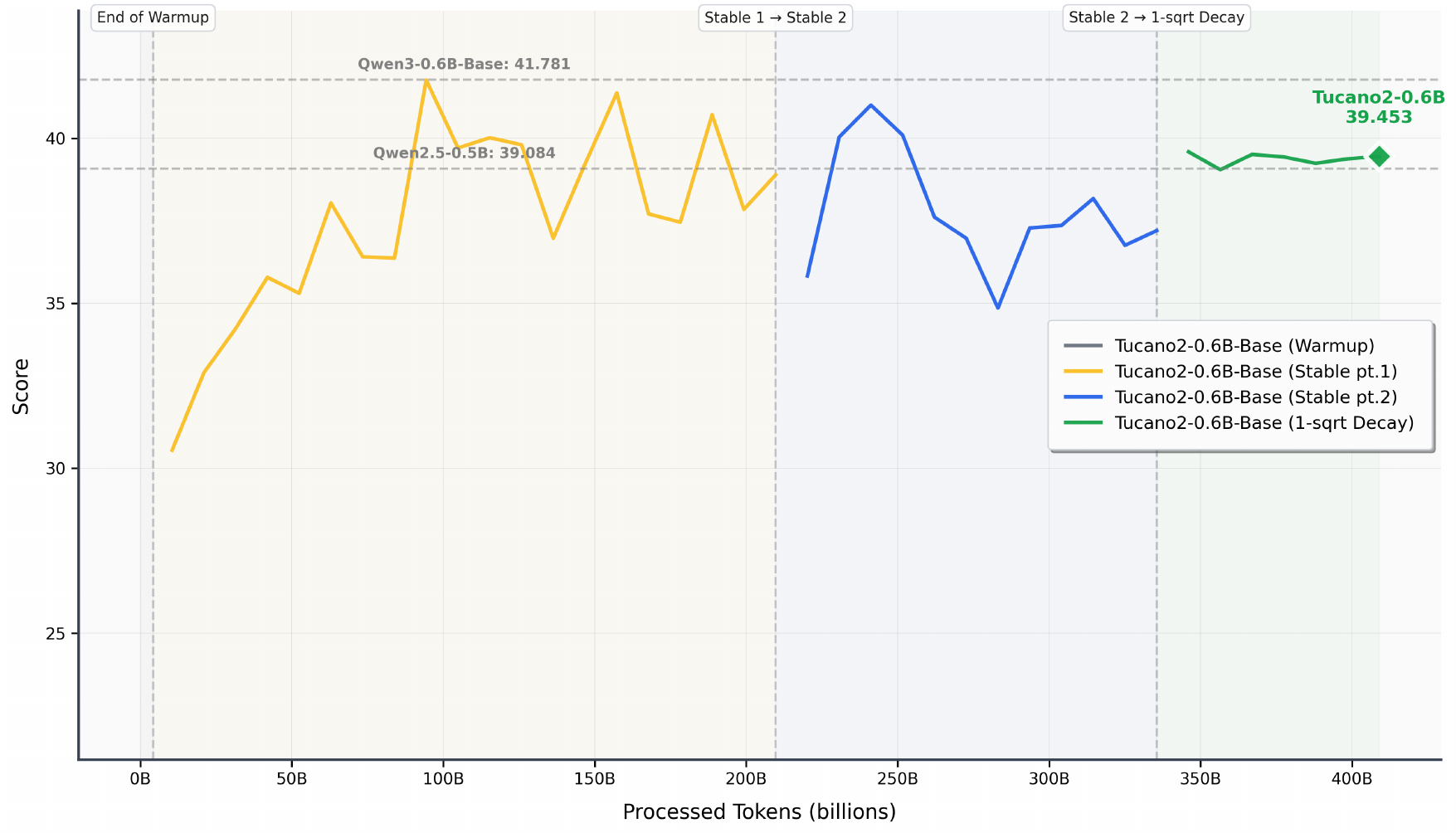}
    \caption{Lambada}
    \label{fig:appendix-easy-lambada}
\end{subfigure}

\caption{Easy Set evaluations across training checkpoints.}
\label{fig:appendix-easy-set}
\end{figure}

\begin{figure}[h]
\centering
\begin{subfigure}[b]{0.48\textwidth}
    \centering
    \includegraphics[width=\textwidth]{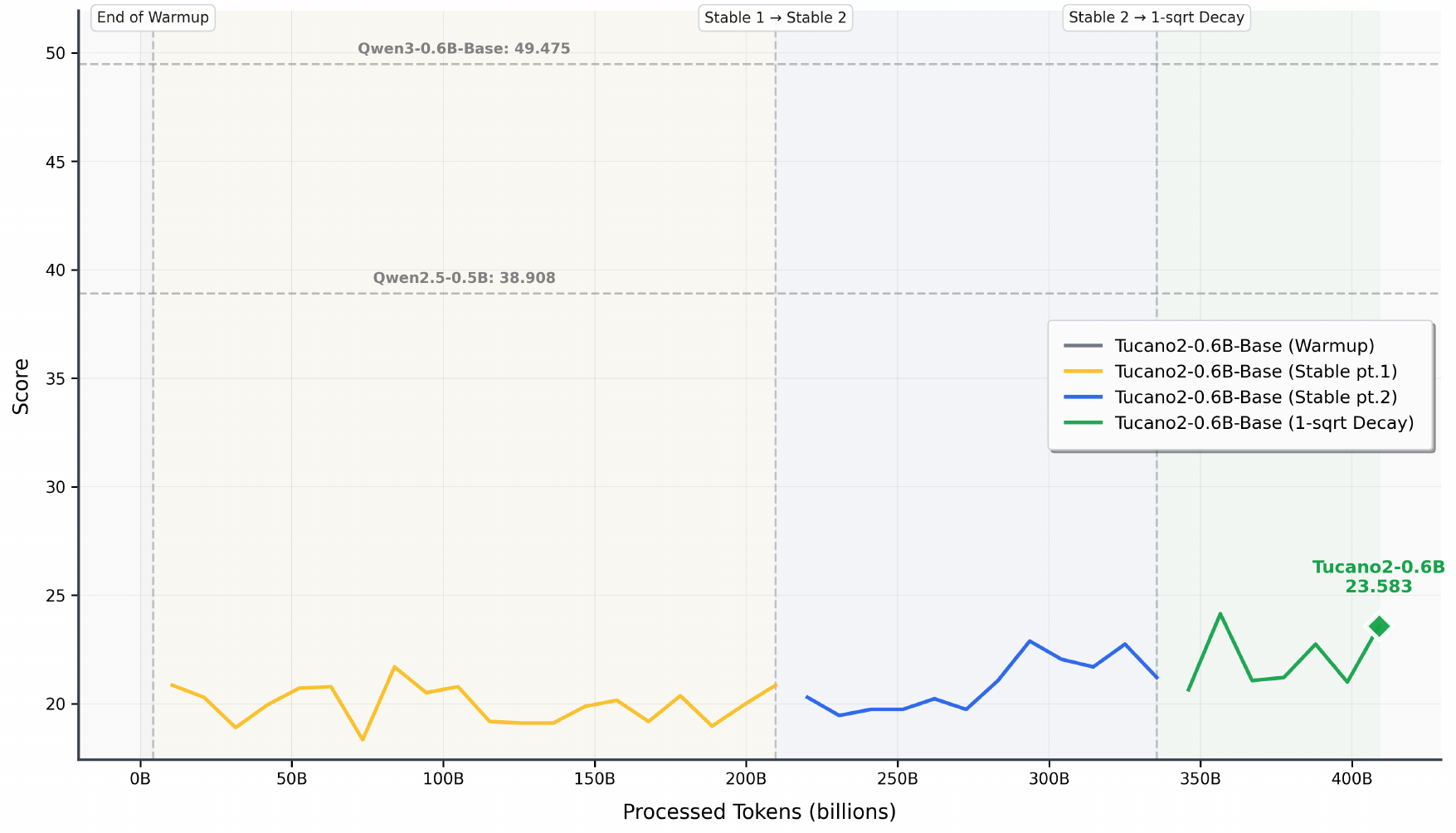}
    \caption{ENEM}
    \label{fig:appendix-hard-enem}
\end{subfigure}
\hfill
\begin{subfigure}[b]{0.48\textwidth}
    \centering
    \includegraphics[width=\textwidth]{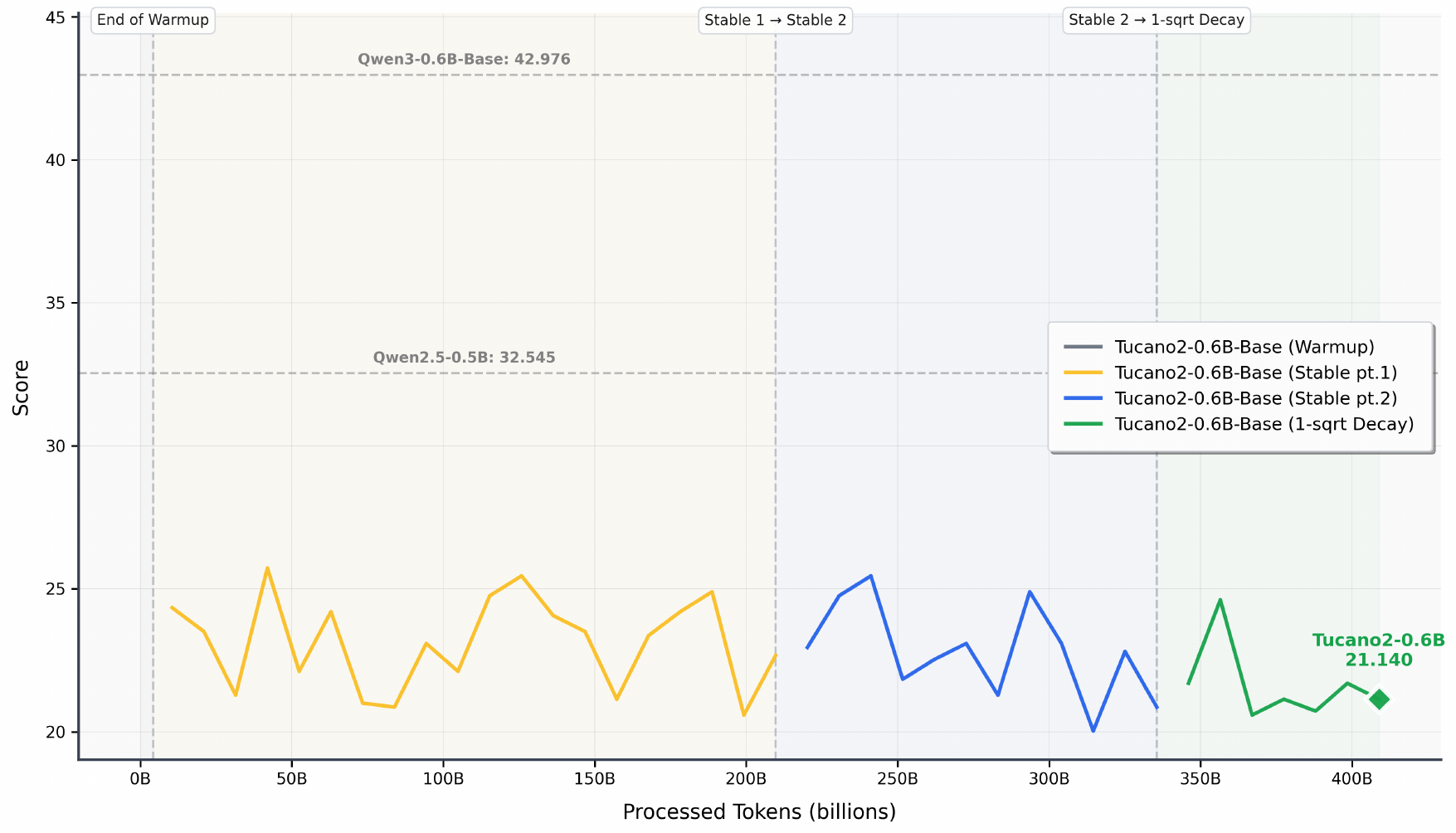}
    \caption{BLUEX}
    \label{fig:appendix-hard-bluex}
\end{subfigure}

\vspace{0.5cm}

\begin{subfigure}[b]{0.48\textwidth}
    \centering
    \includegraphics[width=\textwidth]{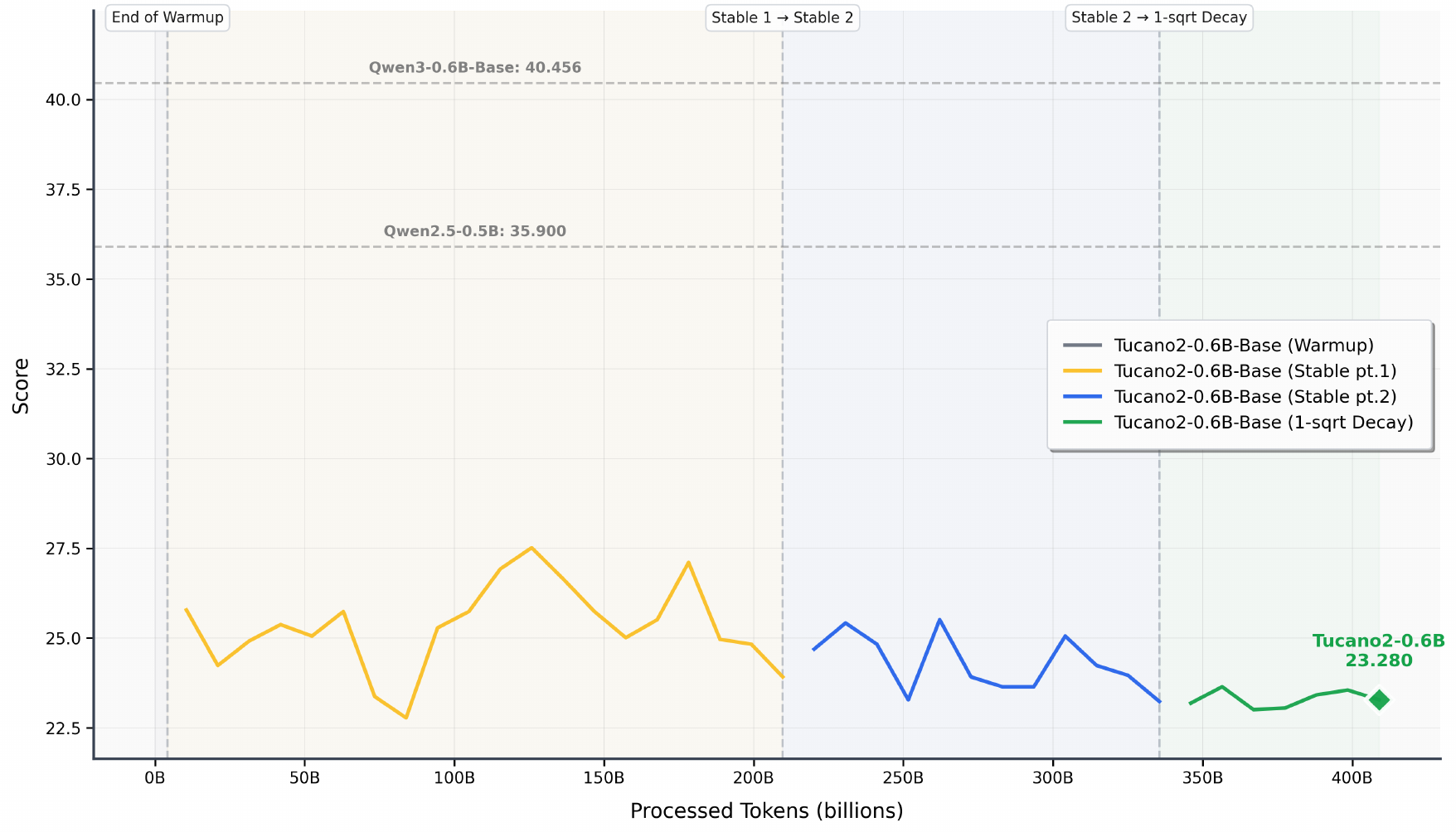}
    \caption{OAB}
    \label{fig:appendix-hard-oab}
\end{subfigure}
\hfill
\begin{subfigure}[b]{0.48\textwidth}
    \centering
    \includegraphics[width=\textwidth]{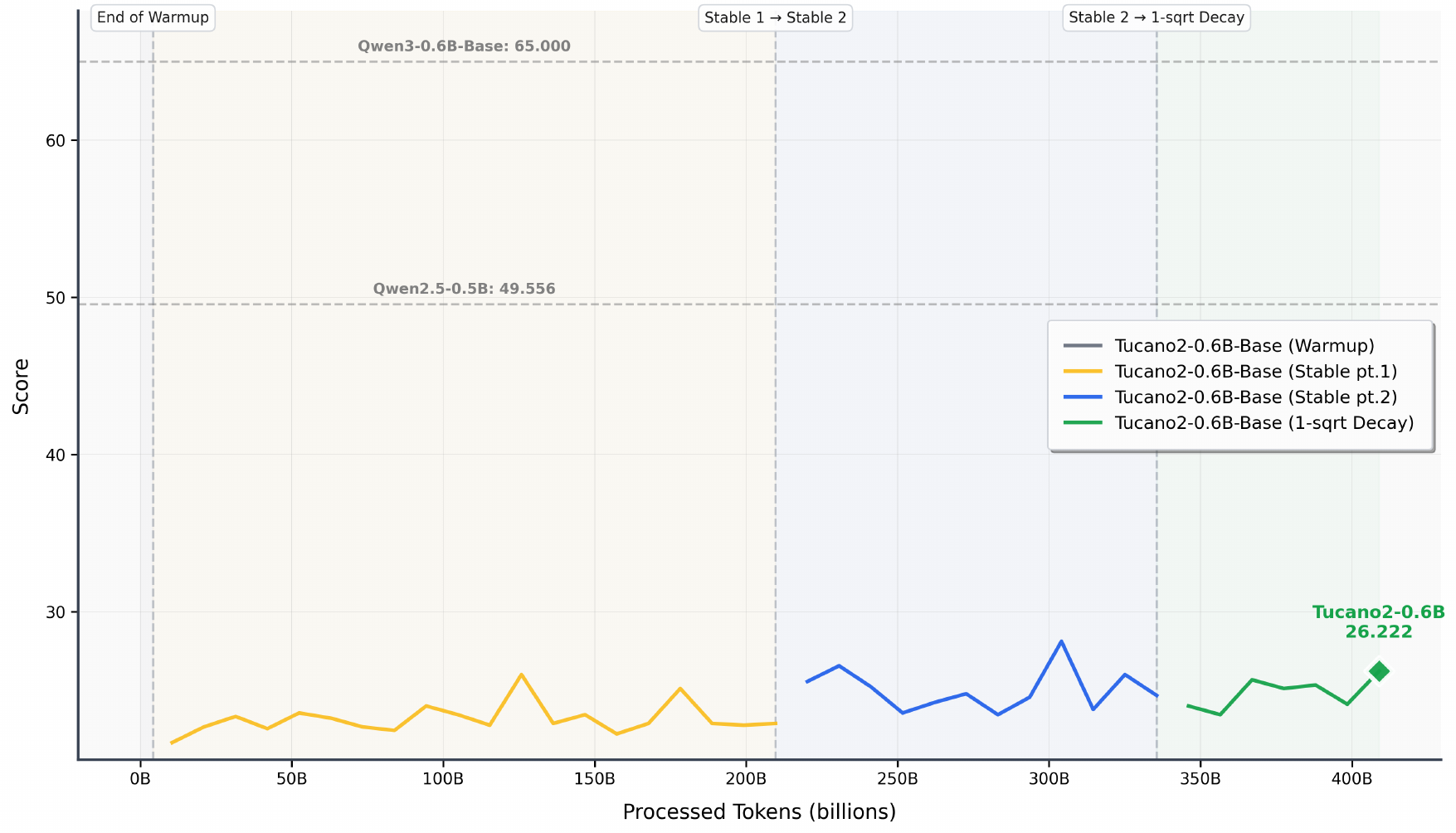}
    \caption{Belebele}
    \label{fig:appendix-hard-belebele}
\end{subfigure}

\vspace{0.5cm}

\begin{subfigure}[b]{0.48\textwidth}
    \centering
    \includegraphics[width=\textwidth]{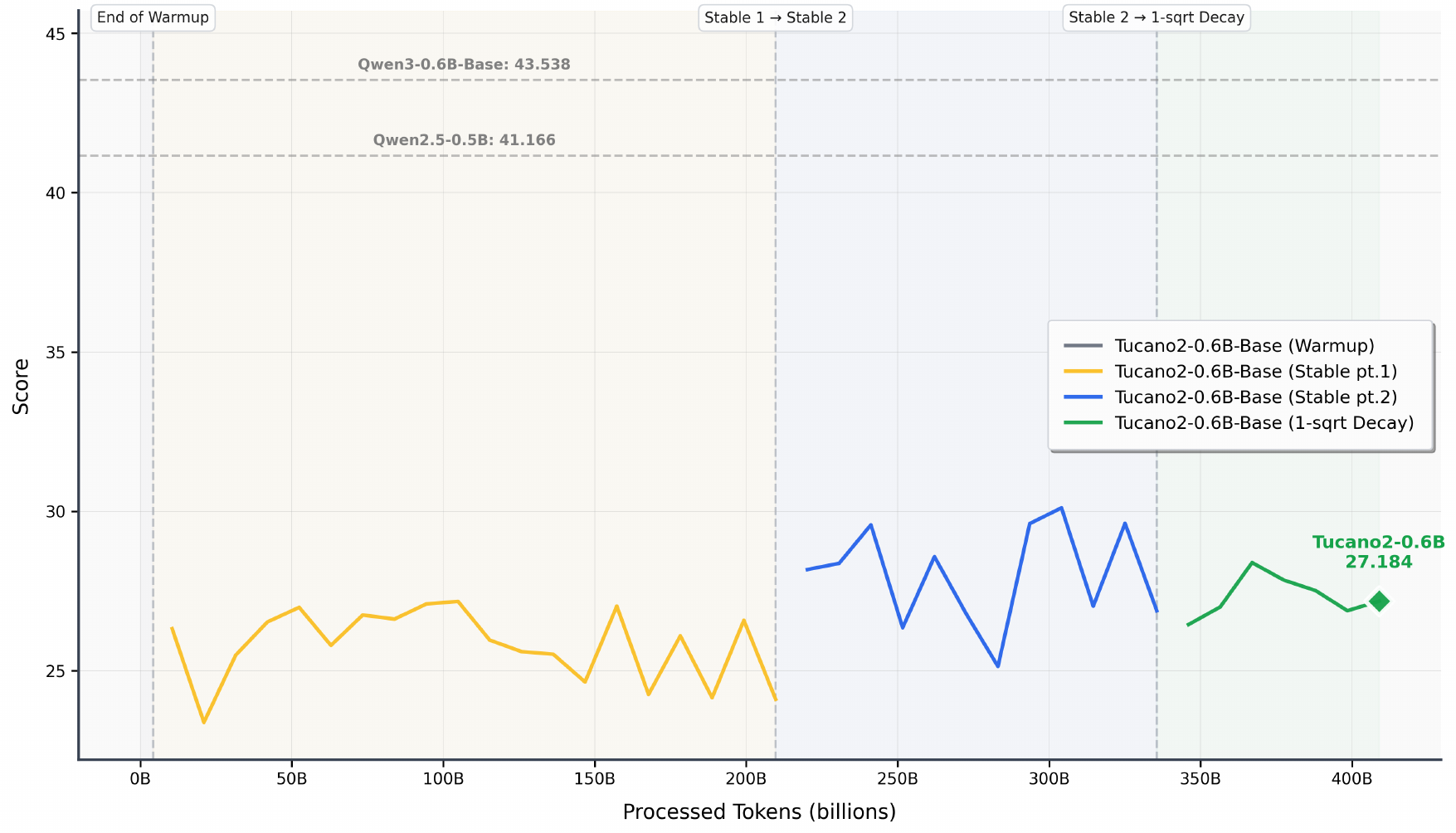}
    \caption{MMLU}
    \label{fig:appendix-hard-mmlu}
\end{subfigure}

\caption{Hard Set evaluations across training checkpoints.}
\label{fig:appendix-hard-set}
\end{figure}

This section provides per-benchmark performance curves for all tasks in both the Easy Set (Figure~\ref{fig:appendix-easy-set}) and Hard Set (Figure~\ref{fig:appendix-hard-set}).

\subsubsection{Easy Set}

\newpage
\subsubsection{Hard Set}

\subsubsection{Comparisons}

Tables~\ref{tab:appendix-easy-set-comparison} and~\ref{tab:appendix-hard-set-comparison} present results for the full evaluation suite, separated into Easy Set and Hard Set tasks.

\begin{table}[h]
\centering
\small
\caption{Easy Set evaluations.}
\label{tab:appendix-easy-set-comparison}
\begin{tabular}{lccccc}
\toprule
 & \textbf{ARC-C} & \textbf{CALAME} & \textbf{Global-PIQA} & \textbf{HellaSwag} & \textbf{Lambada} \\
\midrule
Qwen3-0.6B & 36.92 & 45.95 & 54 & 40.33 & 41.78 \\
\textbf{Tucano2-0.6B} & \textbf{37.01} & \textbf{57.61} & \textbf{79} & 47.74 & 39.45 \\
Qwen2.5-0.5B & 28.46 & 44.89 & 44 & 37.7 & 39.08 \\
Curi{\'o}-1.1B & 30.43 & 59.25 & 75 & \textbf{49.45} & \textbf{46.69} \\
Tucano-2b4 & 30.43 & 50.34 & 73 & 48.85 & 32.39 \\
Curi{\'o}-Edu-1B & 32.22 & 54.91 & 69 & 46.3 & 42.93 \\
Carvalho-1.3B & 27.01 & 53.42 & 63 & 38.53 & 33.59 \\
GlorIA-1.3B & 26.41 & 54.67 & 64 & 36.35 & 36.68 \\
\bottomrule
\end{tabular}
\end{table}

\newpage

\begin{table}[h]
\centering
\small
\caption{Hard Set evaluations.}
\label{tab:appendix-hard-set-comparison}
\begin{tabular}{lccccc}
\toprule
 & \textbf{BLUEX} & \textbf{ENEM} & \textbf{OAB} & \textbf{Belebele} & \textbf{MMLU} \\
\midrule
Qwen3-0.6B & 42.98 & 49.48 & 40.46 & 65 & 43.54 \\
\textbf{Tucano2-0.6B} & 21.14 & 23.58 & 23.28 & 26.22 & 27.18 \\
Qwen2.5-0.5B & 32.55 & 38.91 & 35.9 & 49.56 & 41.17 \\
Curi{\'o}-1.1B & 21.56 & 21.06 & 23.1 & 22.89 & 26.35 \\
Tucano-2b4 & 25.45 & 21.62 & 26.74 & 25.89 & 26.24 \\
Curi{\'o}-Edu-1B & 23.5 & 19.87 & 25.01 & 26.22 & 25.43 \\
Carvalho-1.3B & 19.33 & 18.12 & 22.32 & 26.44 & 24.82 \\
GlorIA-1.3B & 4.31 & 2.52 & 4.69 & 22.78 & 23.69 \\
\bottomrule
\end{tabular}
\end{table}

\paragraph{Key Observations.}
\begin{itemize}
    \item On individual Easy Set evaluations (Table~\ref{tab:appendix-easy-set-comparison}), \tucanoptsmall achieves best-in-class performance on ARC Challenge, Calame, and Global PIQA, highlighting strengths in science reasoning, Portuguese language modeling, and physical commonsense.
    \item Hard Set performance (Table~\ref{tab:appendix-hard-set-comparison}) remains close to random baselines for all models trained on $<$1T tokens, confirming our analysis in Section~\ref{sec:building-evaluation-harness} that these benchmarks require substantially larger training budgets.
    \item Despite using significantly less compute than Curi{\'o}-1.1B (which trained on 1 trillion tokens of pretraining and 150 billion tokens of continual pretraining), \tucanoptsmall achieves comparable overall performance.
\end{itemize}

\newpage
\section{Continual Pretraining: Details}
\label{appendix:continual_pretraining}

This appendix provides supplementary material for the continual pretraining experiments described in Section~\ref{sec:continual_pretraining}, including OMP-based tokenizer transplantation results, hyperparameter configurations, data mixtures, per-benchmark evaluation results, and the full performance vs. compute breakdown.

\subsection{Tokenizer Transplantation: Initial Results}
\label{appendix:cpt-omp-results}

Table~\ref{tab:cpt-omp-transplantation} reports the per-benchmark performance of Qwen3 base models before and after OMP-based tokenizer transplantation. The transplanted models retain a substantial portion of the original performance, especially on Hard Set benchmarks. For instance, the transplanted Qwen3-0.6B already outperforms the from-scratch Tucano2-0.6B-Base on all Hard Set benchmarks despite receiving no additional training. Easy Set degradation is concentrated in token-prediction-sensitive tasks (CALAME, LAMBADA), consistent with the expected impact of vocabulary replacement on next-token prediction distributions.

\begin{table}[h]
\centering
\small
\caption{Per-benchmark performance before and after OMP-based tokenizer transplantation. \textbf{Easy Set:} ARC Challenge (ARC), CALAME (CAL), Global PIQA (PIQA), HellaSwag (HS), LAMBADA (LAM). \textbf{Hard Set:} BLUEX (BLX), ENEM, OAB, BELEBELE (BEL), MMLU.}
\label{tab:cpt-omp-transplantation}
\resizebox{\textwidth}{!}{%
\begin{tabular}{lcccccccccc}
\toprule
 & \textbf{ARC} & \textbf{CAL} & \textbf{PIQA} & \textbf{HS} & \textbf{LAM} & \textbf{BLX} & \textbf{ENEM} & \textbf{OAB} & \textbf{BEL} & \textbf{MMLU} \\
\midrule
Qwen3-0.6B-Base             & 36.92 & 45.95 & 54 & 40.32 & 41.78 & 42.97 & 49.47 & 40.45 & 65.00 & 43.53 \\
\quad + transplantation     & 28.11 & 20.52 & 54 & 34.80 & 25.16 & 39.77 & 42.12 & 37.08 & 59.11 & 40.03 \\
\midrule
Qwen3-1.7B-Base             & 47.17 & 53.56 & 67 & 52.55 & 50.80 & 57.16 & 65.22 & 45.78 & 77.88 & 55.48 \\
\quad + transplantation     & 36.83 & 26.39 & 63 & 41.43 & 26.12 & 42.00 & 58.78 & 44.28 & 74.22 & 50.75 \\
\midrule
Qwen3-4B-Base               & 54.52 & 57.94 & 77 & 63.19 & 60.37 & 69.95 & 77.60 & 55.58 & 87.88 & 68.59 \\
\quad + transplantation     & 42.64 & 23.84 & 61 & 46.92 & 33.08 & 51.87 & 35.96 & 48.29 & 81.66 & 58.93 \\
\bottomrule
\end{tabular}}
\end{table}

\subsection{Hyperparameter Settings}
\label{appendix:cpt-hyperparameters}

Table~\ref{tab:cpt-hyperparameters} provides the complete training configuration for all three continual pretraining runs. Learning rates were selected via systematic sweeps of 10,000 steps each, targeting strong Easy Set performance while preserving the Hard Set capabilities inherited from the Qwen3 base models. The AdamW+Muon optimizer variant used for \tucanoptsmall was also evaluated in small-scale experiments but consistently underperformed AdamW, likely due to a mismatch between the optimizer state and the Qwen3 pretrained weights.

\newpage
\begin{table}[h]
\centering
\small
\caption{Hyperparameter settings for continual pretraining.}
\label{tab:cpt-hyperparameters}
\begin{tabular}{lccc}
\toprule
\textbf{Training Configuration} & \textbf{0.5B} & \textbf{1.5B} & \textbf{3.7B} \\
\midrule
Parallelism Strategy        & DDP              & DDP              & FSDP2 (ZeRO-2)  \\
Activation Checkpointing    & False            & False            & True \\
Total Batch Size            & 1,048,576 tokens & 1,048,576 tokens & 1,048,576 tokens \\
Micro Batch Size  & 8 samples        & 4 samples        & 16 samples        \\
Accumulation Steps & 4                & 8                & 2                \\
Context Length      & 4,096 tokens     & 4,096 tokens     & 4,096 tokens     \\
Training Steps        & 50,000           & 100,000          & 50,000           \\
Checkpointing     & Every 2,500 steps & Every 2,500 steps & Every 2,500 steps \\
\midrule
\multicolumn{4}{c}{\textit{\textbf{Learning Rate Schedule}}} \\
\midrule
Schedule Type      & Warmup--Cosine  & Warmup--Cosine  & Warmup--Cosine  \\
Warmup Steps       & 100                   & 200                   & 100                   \\
Peak Learning Rate & $2.5\times10^{-4}$    & $1\times10^{-4}$      & $7.5\times10^{-5}$    \\
Minimum LR         & 0.0                   & 0.0                   & 0.0                   \\
\midrule
\multicolumn{4}{c}{\textit{\textbf{Optimizer Configuration}}} \\
\midrule
Optimizer          & AdamW            & AdamW            & AdamW            \\
Weight Decay       & 0.1              & 0.1              & 0.1              \\
$\beta_1$          & 0.9              & 0.9              & 0.9              \\
$\beta_2$          & 0.95             & 0.95             & 0.95             \\
$\epsilon$         & $1\times10^{-8}$ & $1\times10^{-8}$ & $1\times10^{-8}$ \\
Gradient Clip Norm & 1.0              & 1.0              & 1.0              \\
\midrule
\multicolumn{4}{c}{\textit{\textbf{Resource Consumption}}} \\
\midrule
Total Training Tokens & $\sim$50B          & $\sim$100B         & $\sim$50B           \\
Estimated FLOPs       & $1.5\times10^{20}$ & $9\times10^{20}$   & $1.11\times10^{21}$ \\
Energy Consumption    & 225 kWh            & 878 kWh            & 1,223 kWh           \\
Carbon Footprint      & 86 kg CO$_2$e      & 334 kg CO$_2$e     & 466 kg CO$_2$e      \\
\bottomrule
\end{tabular}
\end{table}

\subsection{Data Mixture}
\label{appendix:cpt-data-mixtures}

Table~\ref{tab:cpt-data-mixtures} details the data composition for each continual pretraining run. All mixtures are purely Portuguese, combining curated web text from \gigaverbo (filtered by education score) with synthetic augmentations from \gigaverbosynth. This deliberate shift from the multilingual mixtures used during from-scratch pretraining is intended to maximize Portuguese-specific adaptation within a constrained compute budget. The 1.5B model benefits from a larger training budget (100B tokens), enabling inclusion of lower-threshold web data (education score $\geq$ 3) and additional repetition of high-quality subsets. For the 0.5B and 3.7B models, the data budget was limited to 50B tokens due to project timeline and compute constraints.

\begin{table}[h]
\centering
\small
\caption{Data mixture composition for continual pretraining. Subset numbers denote education score thresholds from \gigaverbo.}
\label{tab:cpt-data-mixtures}
\begin{tabular}{lccc}
\toprule
\textbf{Dataset} & \textbf{0.5B} & \textbf{1.5B} & \textbf{3.7B} \\
\midrule
\gigaverbo~(subset 3) & ---            & 12B ($\times$1) & ---            \\
\gigaverbo~(subset 4) & 28B ($\times$1) & 28B ($\times$2) & 28B ($\times$1) \\
\gigaverbo~(subset 5) & 0.1B ($\times$1) & 0.1B ($\times$2) & 0.1B ($\times$2) \\
\textbf{Total Web Text} & \textbf{28.1B} & \textbf{68.2B} & \textbf{28.2B} \\
\midrule
\gigaverbosynth        & 10B ($\times$2) & 10B ($\times$3) & 10B ($\times$2) \\
\textbf{Total Synthetic} & \textbf{20B} & \textbf{30B} & \textbf{20B} \\
\midrule
\textbf{Web : Synthetic} & \textbf{60\,:\,40} & \textbf{70\,:\,30} & \textbf{60\,:\,40} \\
\bottomrule
\end{tabular}
\end{table}

\subsection{Per-Benchmark Results}
\label{appendix:cpt-detailed-results}

\begin{figure}[h]
\centering

\includegraphics[width=0.8\linewidth]{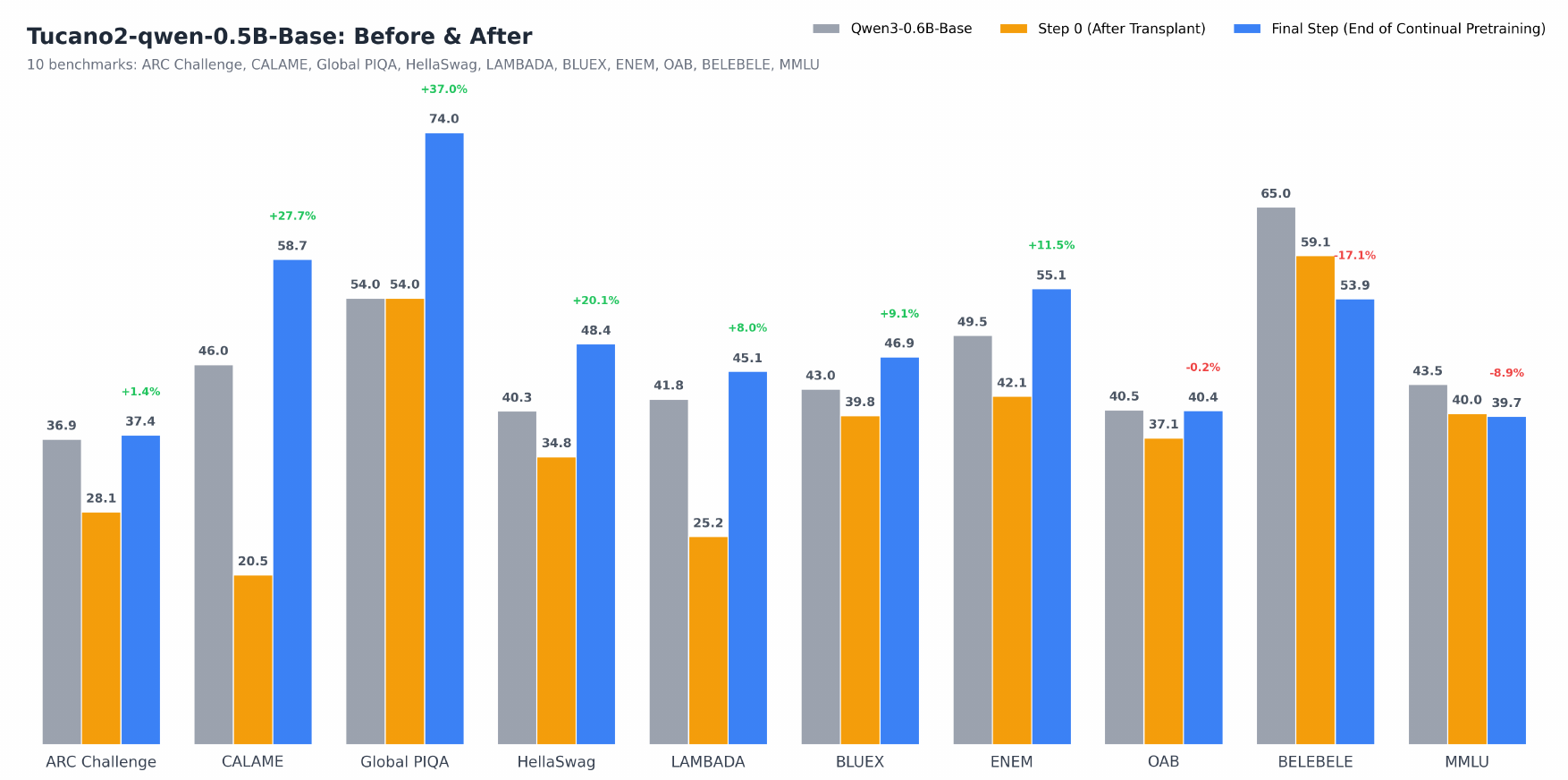}
\caption{Per-benchmark comparison: Qwen3-0.6B-Base vs.\ \tucanocptsmall.}
\label{fig:cpt-before-after-0.5B}

\vspace{1em}

\includegraphics[width=0.8\linewidth]{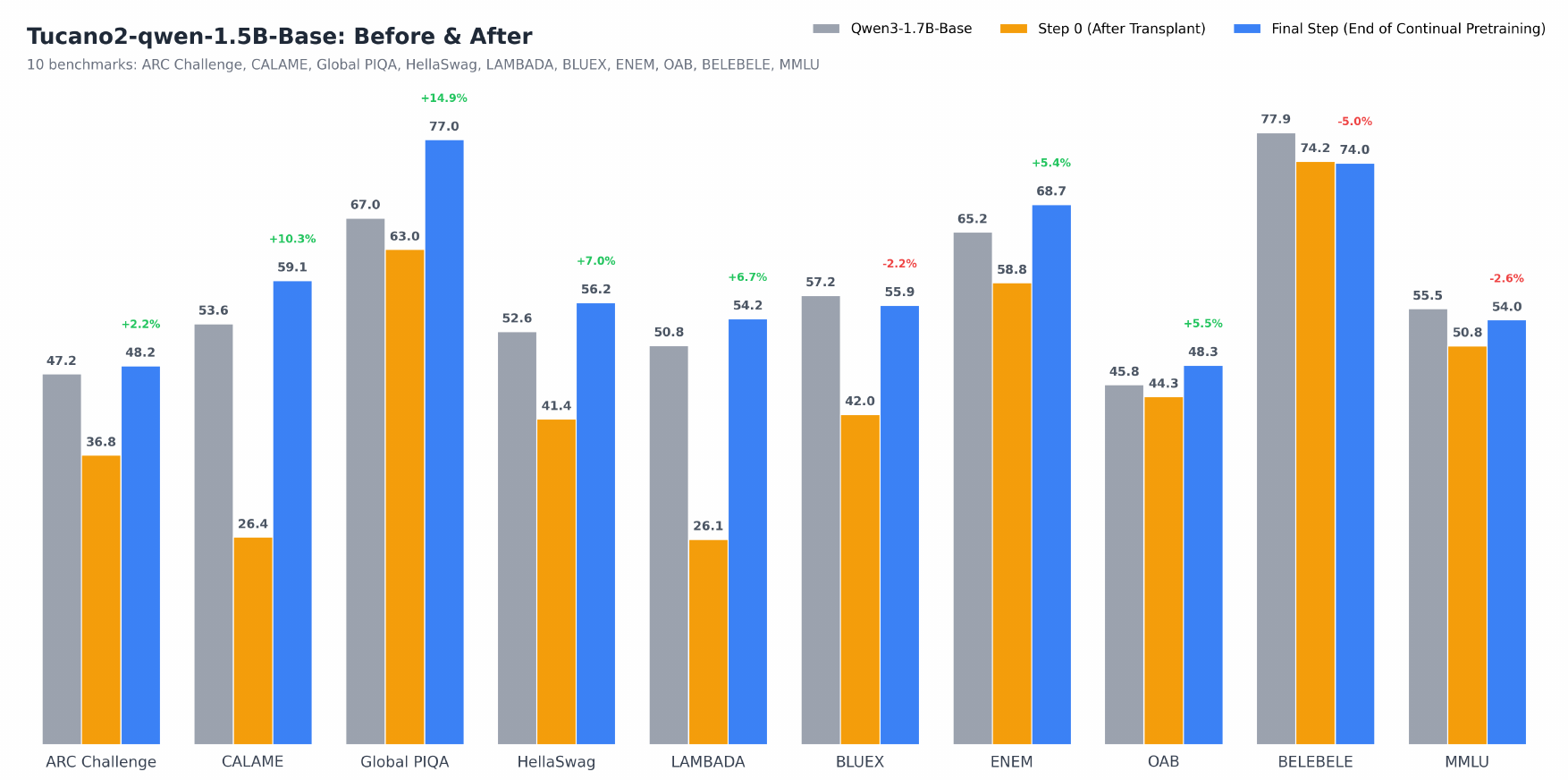}
\caption{Per-benchmark comparison: Qwen3-1.7B-Base vs.\ \tucanocptmedium.}
\label{fig:cpt-before-after-1.5B}

\end{figure}

Figures~\ref{fig:cpt-before-after-0.5B} and~\ref{fig:cpt-before-after-1.5B} show per-benchmark performance shifts for the 0.5B and 1.5B models, respectively.

Visualizations of individual benchmark performance across training steps, as well as the training dynamics (loss curves, gradient norms) for all continual pretraining runs, are available in the following repositories:

\begin{itemize}
    \item \tucanocptsmall: \href{https://huggingface.co/Polygl0t/Tucano2-qwen-0.5B-Base/tree/main/.plots}{Polygl0t/Tucano2-qwen-0.5B-Base}
    \item \tucanocptmedium: \href{https://huggingface.co/Polygl0t/Tucano2-qwen-1.5B-Base/tree/main/.plots}{Polygl0t/Tucano2-qwen-1.5B-Base}
    \item \tucanocptlarge: \href{https://huggingface.co/Polygl0t/Tucano2-qwen-3.7B-Base/tree/main/.plots}{Polygl0t/Tucano2-qwen-3.7B-Base}
\end{itemize}

\newpage
\subsubsection{Easy Set}

Table~\ref{tab:cpt-easy-set} reports per-benchmark results on the Easy Set evaluations (ARC-Challenge, CALAME, Global PIQA, HellaSwag, LAMBADA) across all evaluated models.

\begin{table}[h]
\centering
\small
\caption{Easy Set evaluations.}
\label{tab:cpt-easy-set}
\begin{tabular}{lccccc}
\toprule
 & \textbf{ARC-C} & \textbf{CALAME} & \textbf{G.\ PIQA} & \textbf{HellaSwag} & \textbf{LAMBADA} \\
\midrule
\textbf{\tucanocptlarge} & 57.78 & 61.08 & 83 & 65.32 & 62.53 \\
Qwen2.5-7B                      & 54.19 & 58.96 & 78 & 67.92 & 59.52 \\
Qwen3-4B-Base                   & 54.53 & 57.95 & 77 & 63.19 & 60.37 \\
SmolLM3-3B-Base                  & 51.37 & 59.15 & 81 & 65.57 & 59.89 \\
Qwen2.5-3B                      & 45.21 & 58.38 & 75 & 59.44 & 57.17 \\
\textbf{\tucanocptmedium} & 48.21 & 59.06 & 77 & 56.25 & 54.20 \\
Curi\'{o}-Edu-7B                 & 50.94 & 60.79 & 86 & 66.48 & 64.62 \\
Qwen3-1.7B-Base                 & 47.18 & 53.56 & 67 & 52.55 & 50.81 \\
Curi\'{o}-7B                    & 48.03 & 63.44 & 89 & 67.58 & 65.94 \\
Llama-3.2-3B                    & 41.11 & 54.48 & 69 & 59.14 & 59.48 \\
Granite-3.3-2B                  & 41.37 & 58.77 & 70 & 60.81 & 58.22 \\
\textbf{\tucanocptsmall} & 37.44 & 58.67 & 74 & 48.43 & 45.14 \\
Qwen3-0.6B-Base                 & 36.92 & 45.95 & 54 & 40.33 & 41.78 \\
Llama-2-7B                      & 42.14 & 54.53 & 67 & 56.76 & 59.73 \\
\textbf{\tucanoptsmall}              & 37.01 & 57.61 & 79 & 47.74 & 39.45 \\
Qwen2.5-0.5B                    & 28.46 & 44.89 & 44 & 37.70 & 39.08 \\
Curi\'{o}-1.1B                  & 30.43 & 59.25 & 75 & 49.45 & 46.69 \\
Tucano-2b4                      & 30.43 & 50.34 & 73 & 48.85 & 32.39 \\
Curi\'{o}-Edu-1B                & 32.22 & 54.91 & 69 & 46.30 & 42.93 \\
Llama-3.2-1B                    & 31.71 & 50.00 & 55 & 45.27 & 45.60 \\
Tucano-1b1                      & 30.09 & 48.94 & 68 & 44.10 & 28.43 \\
Tucano-630m                     & 28.72 & 47.30 & 68 & 40.37 & 26.20 \\
Carvalho-pt-gl-1.3B             & 27.01 & 53.42 & 63 & 38.53 & 33.59 \\
TeenyTinyLlama-460m             & 27.35 & 42.49 & 59 & 34.81 & 21.56 \\
Tucano-160m                     & 25.56 & 43.59 & 59 & 33.73 & 21.64 \\
TeenyTinyLlama-160m             & 24.02 & 39.79 & 58 & 29.89 & 17.74 \\
GlorIA-1.3B                     & 26.41 & 54.67 & 64 & 36.35 & 36.68 \\
\bottomrule
\end{tabular}
\end{table}

\newpage
\subsubsection{Hard Set}

Table~\ref{tab:cpt-hard-set} reports per-benchmark results on the Hard Set evaluations (BLUEX, ENEM, OAB Exams, BELEBELE, MMLU) across all evaluated models.

\begin{table}[h]
\centering
\small
\caption{Hard Set evaluations.}
\label{tab:cpt-hard-set}
\begin{tabular}{lccccc}
\toprule
 & \textbf{BLUEX} & \textbf{ENEM} & \textbf{OAB} & \textbf{BELEBELE} & \textbf{MMLU} \\
\midrule
\textbf{\tucanocptlarge} & 66.20 & 77.54 & 58.45 & 83.67 & 65.40 \\
Qwen2.5-7B                      & 65.92 & 75.02 & 55.03 & 89.67 & 68.55 \\
Qwen3-4B-Base                   & 69.96 & 77.61 & 55.58 & 87.89 & 68.59 \\
SmolLM3-3B-Base                  & 54.52 & 61.37 & 45.51 & 77.67 & 56.19 \\
Qwen2.5-3B                      & 58.28 & 67.32 & 50.34 & 83.22 & 59.79 \\
\textbf{\tucanocptmedium} & 55.91 & 68.72 & 48.29 & 74.00 & 54.04 \\
Curi\'{o}-Edu-7B                 & 47.15 & 58.64 & 43.78 & 53.00 & 45.14 \\
Qwen3-1.7B-Base                 & 57.16 & 65.22 & 45.79 & 77.89 & 55.49 \\
Curi\'{o}-7B                    & 43.39 & 50.59 & 39.68 & 45.33 & 40.83 \\
Llama-3.2-3B                    & 50.35 & 53.04 & 39.45 & 68.89 & 48.28 \\
Granite-3.3-2B                  & 45.34 & 54.02 & 39.54 & 65.67 & 45.63 \\
\textbf{\tucanocptsmall} & 46.87 & 55.14 & 40.36 & 53.89 & 39.68 \\
Qwen3-0.6B-Base                 & 42.98 & 49.48 & 40.46 & 65.00 & 43.54 \\
Llama-2-7B                      & 31.29 & 31.77 & 35.49 & 41.44 & 38.64 \\
\textbf{\tucanoptsmall}               & 21.14 & 23.58 & 23.28 & 26.22 & 27.18 \\
Qwen2.5-0.5B                    & 32.55 & 38.91 & 35.90 & 49.56 & 41.17 \\
Curi\'{o}-1.1B                  & 21.56 & 21.06 & 23.10 & 22.89 & 26.35 \\
Tucano-2b4                      & 25.45 & 21.62 & 26.74 & 25.89 & 26.24 \\
Curi\'{o}-Edu-1B                & 23.50 & 19.87 & 25.01 & 26.22 & 25.43 \\
Llama-3.2-1B                    & 24.06 & 23.93 & 26.06 & 33.33 & 28.51 \\
Tucano-1b1                      & 25.45 & 21.55 & 26.38 & 25.67 & 25.26 \\
Tucano-630m                     & 26.70 & 21.69 & 26.92 & 27.33 & 25.60 \\
Carvalho-pt-gl-1.3B             & 19.33 & 18.12 & 22.32 & 26.44 & 24.82 \\
TeenyTinyLlama-460m             & 25.87 & 20.15 & 27.02 & 28.11 & 26.65 \\
Tucano-160m                     & 24.76 & 20.57 & 17.22 & 23.44 & 25.77 \\
TeenyTinyLlama-160m             & 22.53 & 18.89 & 22.32 & 26.78 & 25.74 \\
GlorIA-1.3B                     & 4.31  & 2.52  & 4.69  & 22.78 & 23.69 \\
\bottomrule
\end{tabular}
\end{table}

\newpage
\subsection{Performance vs.\ Compute}
\label{appendix:cpt-performance-vs-compute}

Table~\ref{tab:cpt-performance-vs-compute} provides a detailed breakdown of compute costs for all evaluated models, including separate accounting of pretraining (PT) and continual pretraining (CPT) FLOPs. Total compute is estimated as $C = 6ND$.

Even with a limited compute budget for continual pretraining, the Tucano2-qwen-3.7B-Base model outperforms all similarly sized models in our evaluation suite. The continual pretraining FLOPs represent a negligible fraction of the total compute: $<$0.13\% for the 3.7B model, $<$0.25\% for the 1.5B model, and $<$0.12\% for the 0.5B model.

\begin{table}[h]
\centering
\small
\caption{Performance vs.\ compute breakdown for all evaluated models. PT = Pretraining, CPT = Continual Pretraining.}
\label{tab:cpt-performance-vs-compute}
\resizebox{\textwidth}{!}{%
\begin{tabular}{lcccccccc}
\toprule
 & \textbf{Params (B)} & \textbf{PT Tokens (B)} & \textbf{CPT Tokens (B)} & \textbf{Total Tokens (B)} & \textbf{PT FLOPs} & \textbf{CPT FLOPs} & \textbf{Total FLOPs} & \textbf{NPM} \\
\midrule
\textbf{\tucanocptlarge} & 3.7  & 36,000 & 50  & 36,050 & 8.64e+23 & 1.11e+21 & 8.65e+23 & 59.21 \\
Qwen2.5-7B                  & 7.0  & 18,000 & --- & 18,000 & 7.56e+23 & ---      & 7.56e+23 & 57.97 \\
Qwen3-4B-Base               & 4.0  & 36,000 & --- & 36,000 & 8.64e+23 & ---      & 8.64e+23 & 57.86 \\
SmolLM3-3B-Base              & 3.0  & 11,200 & --- & 11,200 & 2.02e+23 & ---      & 2.02e+23 & 50.25 \\
Qwen2.5-3B                  & 3.0  & 18,000 & --- & 18,000 & 3.24e+23 & ---      & 3.24e+23 & 50.15 \\
\textbf{\tucanocptmedium} & 1.5  & 36,000 & 100 & 36,100 & 3.67e+23 & 9.0e+20  & 3.68e+23 & 47.89 \\
Curi\'{o}-Edu-7B             & 7.0  & 2,000  & 20  & 2,020  & 8.40e+22 & 8.40e+20 & 8.48e+22 & 45.66 \\
Qwen3-1.7B-Base             & 1.7  & 36,000 & --- & 36,000 & 3.67e+23 & ---      & 3.67e+23 & 44.48 \\
Curi\'{o}-7B                & 7.0  & 2,000  & 150 & 2,150  & 8.40e+22 & 6.30e+21 & 9.03e+22 & 42.78 \\
Llama-3.2-3B                & 3.0  & 9,000  & --- & 9,000  & 1.62e+23 & ---      & 1.62e+23 & 40.50 \\
Granite-3.3-2B              & 2.0  & 12,000 & --- & 12,000 & 1.44e+23 & ---      & 1.44e+23 & 39.96 \\
\textbf{\tucanocptsmall} & 0.5  & 36,000 & 50  & 36,050 & 1.30e+23 & 1.50e+20 & 1.30e+23 & 35.35 \\
Qwen3-0.6B-Base             & 0.6  & 36,000 & --- & 36,000 & 1.30e+23 & ---      & 1.30e+23 & 29.39 \\
Llama-2-7B                  & 7.0  & 2,000  & --- & 2,000  & 8.40e+22 & ---      & 8.40e+22 & 29.36 \\
\textbf{\tucanoptsmall}           & 0.6  & 408    & --- & 408    & 1.47e+21 & ---      & 1.47e+21 & 20.63 \\
Qwen2.5-0.5B                & 0.5  & 18,000 & --- & 18,000 & 5.40e+22 & ---      & 5.40e+22 & 19.89 \\
Curi\'{o}-1.1B              & 1.1  & 1,000  & 150 & 1,150  & 6.60e+21 & 9.90e+20 & 7.59e+21 & 19.23 \\
Tucano-2b4                  & 2.4  & 515    & --- & 515    & 7.42e+21 & ---      & 7.42e+21 & 17.87 \\
Curi\'{o}-Edu-1B            & 1.1  & 1,000  & 20  & 1,020  & 6.60e+21 & 1.32e+20 & 6.73e+21 & 17.72 \\
Llama-3.2-1B                & 1.0  & 9,000  & --- & 9,000  & 5.40e+22 & ---      & 5.40e+22 & 16.57 \\
Tucano-1b1                  & 1.1  & 250    & --- & 250    & 1.65e+21 & ---      & 1.65e+21 & 15.44 \\
Tucano-630m                 & 0.63 & 211    & --- & 211    & 7.98e+20 & ---      & 7.98e+20 & 14.89 \\
Carvalho-pt-gl-1.3B         & 1.3  & 26     & 5   & 31     & 2.03e+20 & 3.90e+19 & 2.42e+20 & 12.54 \\
TeenyTinyLlama-460m         & 0.46 & 6.2    & --- & 6.2    & 1.71e+19 & ---      & 1.71e+19 & 11.18 \\
Tucano-160m                 & 0.16 & 169    & --- & 169    & 1.62e+20 & ---      & 1.62e+20 & 8.78  \\
TeenyTinyLlama-160m         & 0.16 & 6.2    & --- & 6.2    & 5.95e+18 & ---      & 5.95e+18 & 7.71  \\
GlorIA-1.3B                 & 1.3  & 35     & --- & 35     & 2.73e+20 & ---      & 2.73e+20 & 5.92  \\
\bottomrule
\end{tabular}}
\end{table}
\newpage
\section{Instruct-Completion Quality Annotators: Details}
\label{appendix:instruct-quality-annotators}

To filter \gigaverbosft at scale, we fine-tuned Qwen3-4B into two quality-assessment models from 500K LLM-annotated interactions.

\paragraph{Model Variants.} We developed the following quality-assessment models:
\begin{itemize}
    \item Classification model (quality scoring): Regression-based continuous score prediction (1--5).
    \item Conditional generation model (quality explanation): Generates a JSON object (\texttt{\{score, reason\}}) with a custom chat template.
\end{itemize}

\subsection{Training Configurations}

For both models, we use a custom chat template that converts data in ChatML format into a single string input, and the output is either a scalar score (classification model) or a JSON object (generation model).

\paragraph{Training Hyperparameters.} Both models were fine-tuned from Qwen3-4B with the configurations shown in Table~\ref{tab:quality-hyperparameters}.

\begin{table}[h]
\centering
\caption{Training configurations for Instruct-Completion Quality Annotators.}
\label{tab:quality-hyperparameters}
\small
\begin{tabular}{lll}
\toprule
\textbf{Training Configuration} & \textbf{Classification Model} & \textbf{Generation Model} \\
\midrule
Base model & Qwen3-4B & Qwen3-4B \\
Epochs & 2 & 3 \\
Batch size & 64 & 64 \\
Maximum context length & 6{,}032 tokens & 6{,}032 tokens \\
Optimizer & AdamW (cosine schedule) & AdamW (cosine schedule) \\
Warmup steps & 100 & 100 \\
Peak learning rate & $5 \times 10^{-5}$ & $5 \times 10^{-5}$ \\
Embedding layer & Frozen & Frozen \\
Chat template & Custom & Custom \\
\bottomrule
\end{tabular}
\end{table}

\subsection{Evaluation Results}

Table~\ref{tab:quality-results} presents the validation performance of both Instruct-Completion Quality Annotators.

\begin{table}[h]
\centering
\caption{Performance metrics for Instruct-Completion Quality Annotators.}
\label{tab:quality-results}
\small
\begin{tabular}{llcc}
\toprule
& \textbf{Task} & \textbf{F1 Macro} & \textbf{F1 ($\geq$3 threshold)} \\
\midrule
Classification model & Quality scoring & \textbf{0.80} & \textbf{0.98} \\
Conditional generation model & Quality explanation & 0.72 & 0.88 \\
\bottomrule
\end{tabular}
\end{table}

\newpage
\section{Post-training: Details}
\label{appendix:post_training}

This appendix provides additional details on the post-training pipeline described in Section~\ref{sec:post_training}, including dataset construction details, quality-assessment model training, full training configurations, per-benchmark results across all model scales, and breakdowns of long-context evaluation.

\subsection{Task Taxonomy: Details}
\label{appendix:sft_task_taxonomy}

Table~\ref{tab:sft-task-taxonomy} enumerates the 12 task types in \gigaverbosft together with brief descriptions.

\begin{table}[h]
\centering
\small
\caption{Task taxonomy.}
\label{tab:sft-task-taxonomy}
\begin{tabular}{ll}
\toprule
\textbf{Task Type} & \textbf{Description} \\
\midrule
Code Generation             & Instruction-guided code writing and synthesis \\
Function Calling            & Tool calling and structured command invocation \\
General Instruction         & Diverse tasks across multiple domains \\
Mathematical Reasoning      & Problem-solving with step-by-step solutions \\
Reasoning                   & Explicit reasoning traces (\texttt{<think>}$\ldots$\texttt{</think>}) \\
Retrieval-Augmented Gen.    & Context-aware question answering \\
Rewriting                   & Paraphrasing and stylistic transformations \\
Structured Output Gen.      & Producing formatted data (JSON) \\
Summarization               & Extractive and abstractive summary generation \\
System Prompts              & Role-based instruction adherence \\
Translation                 & Bilingual transfer (Portuguese $\leftrightarrow$ English) \\
Math / Math CoT                    & Mathematics with (and without) chain-of-thought traces \\
\bottomrule
\end{tabular}
\end{table}

\newpage
\subsection{GigaVerbo-v2 SFT: Statistics}
\label{appendix:sft_visualizations}

\begin{figure}[h]
\centering
\begin{minipage}[t]{0.48\linewidth}
    \centering
    \includegraphics[width=\linewidth]{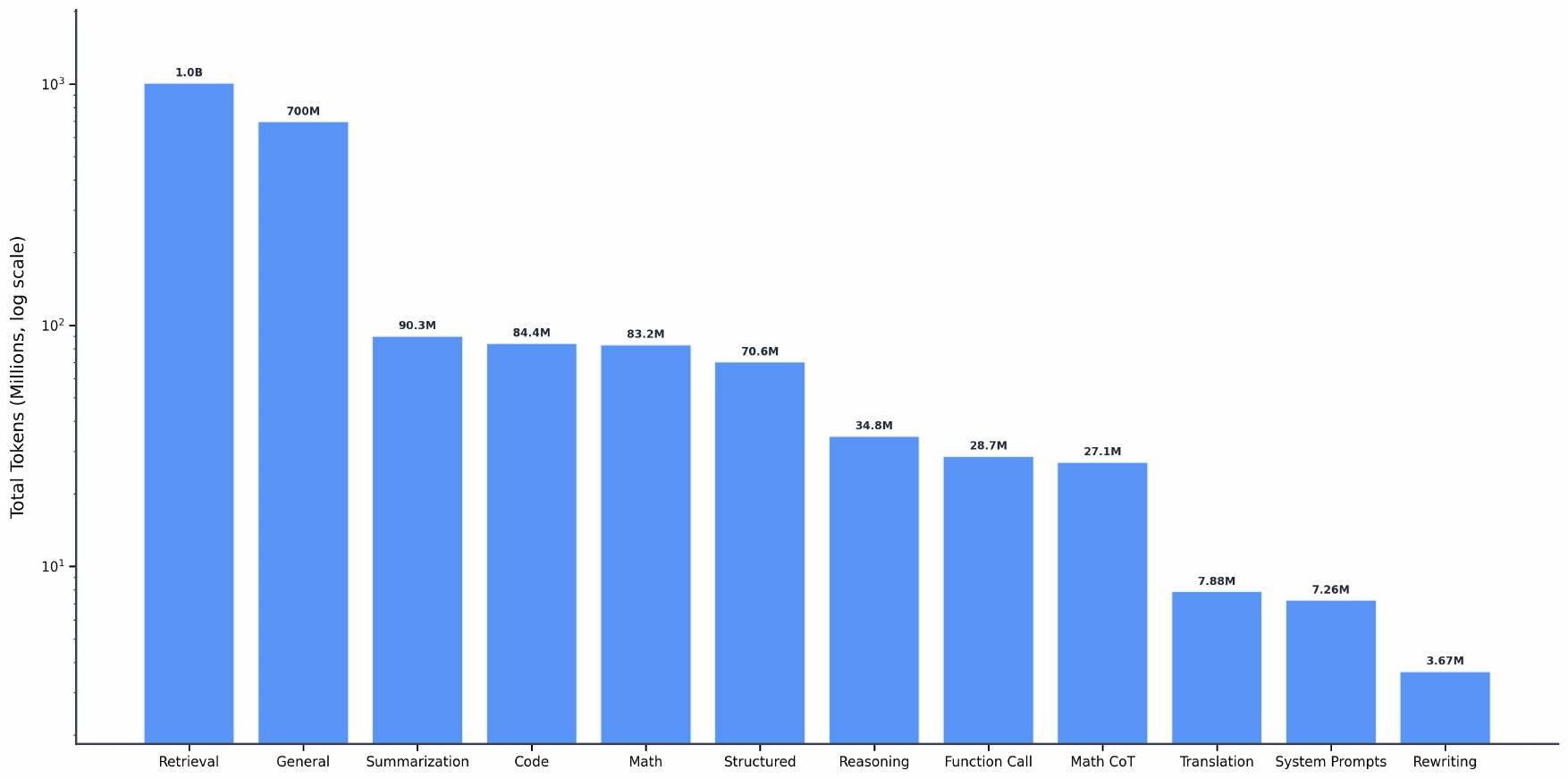}
    \caption*{(a) Token distribution per subset.}
\end{minipage}
\hfill
\begin{minipage}[t]{0.48\linewidth}
    \centering
    \includegraphics[width=\linewidth]{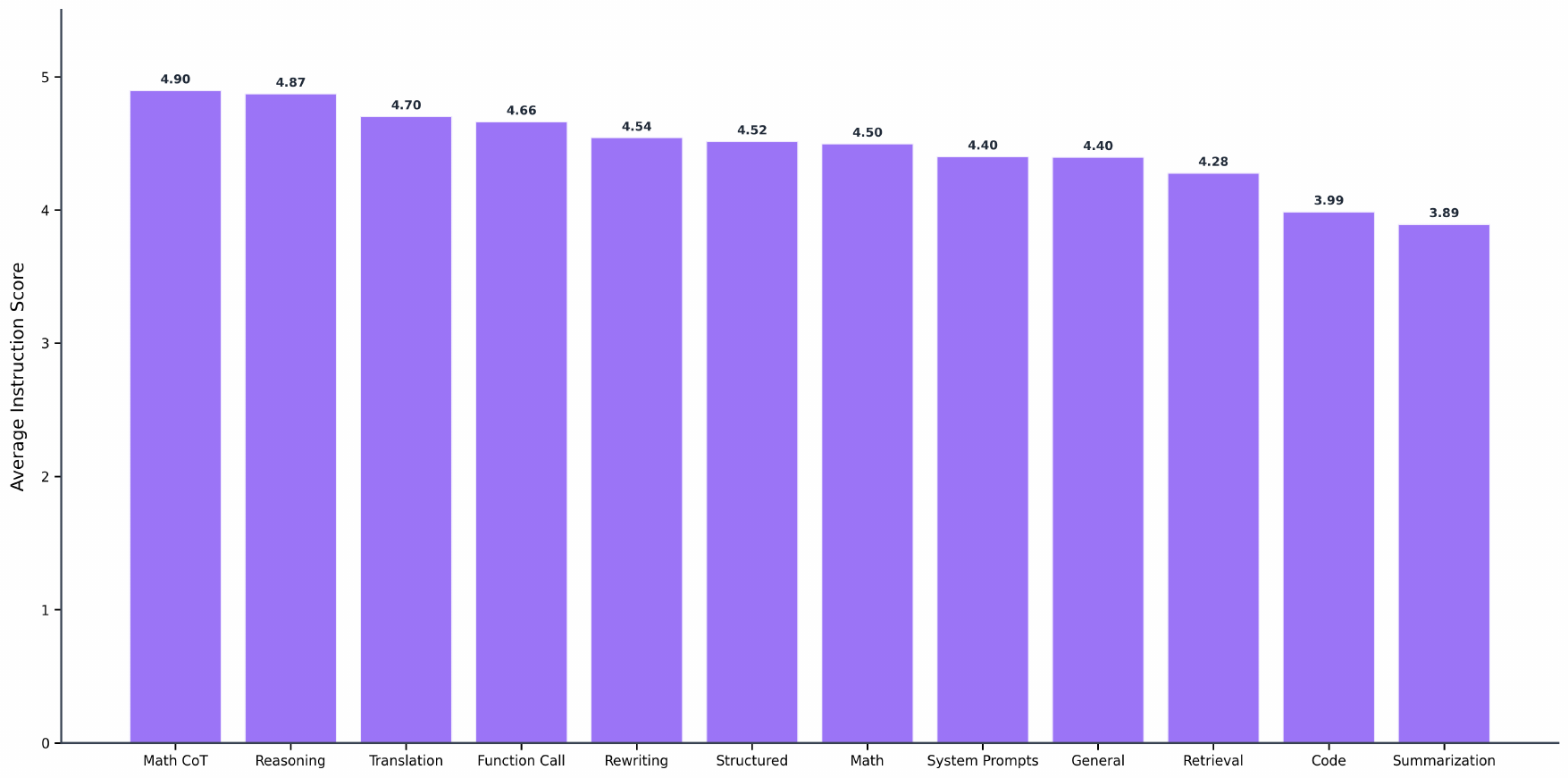}
    \caption*{(b) Quality score distribution.}
\end{minipage}
\caption{SFT dataset statistics after filtering.}
\label{fig:sft-visualizations}
\end{figure}

Figure~\ref{fig:sft-visualizations} displays the distribution of tokens and quality scores in \gigaverbosft.

\subsection{GigaVerbo-v2 Preferences: Statistics}
\label{appendix:preferences_details}

Table~\ref{tab:preferences-details} provides the full composition of the \gigaverbopreferences dataset.

\begin{table}[h]
\centering
\small
\caption{Preference dataset composition.}
\label{tab:preferences-details}
\begin{tabular}{lrrrrr}
\toprule
\textbf{Subset} & \textbf{Examples} & \textbf{Chosen Tok.} & \textbf{Rejected Tok.} & \textbf{Total Tok.} \\
\midrule
Harmless (reasoning)       &  9,641 & 6,105,673  & 4,571,353  & 10,677,026 \\
Harmless (no reasoning)    & 10,521 & 4,711,020  & 4,057,440  &  8,768,460 \\
Harmful (reasoning)        &  4,008 & 1,462,284  & 3,242,908  &  4,705,192 \\
Harmful (no reasoning)     &  4,267 &   809,240  & 2,795,035  &  3,604,275 \\
\midrule
\textbf{Total}             & \textbf{28,437} & \textbf{14,278,217} & \textbf{13,666,736} & \textbf{27,945,053} \\
\bottomrule
\end{tabular}
\end{table}

For the \textit{harmless} subset, chosen responses were generated by Qwen2.5-32B-Instruct with chain-of-thought reasoning, while rejected responses came from Qwen2.5-7B-Instruct. For the \textit{harmful} subset, chosen responses used safety-oriented constitutions applied to Qwen2.5-32B-Instruct, while rejected (compliant) responses were produced by an abliterated variant of Qwen2.5-32B-Instruct

\begin{figure}[h]
\centering
\begin{minipage}[t]{0.48\linewidth}
    \centering
    \includegraphics[width=\linewidth]{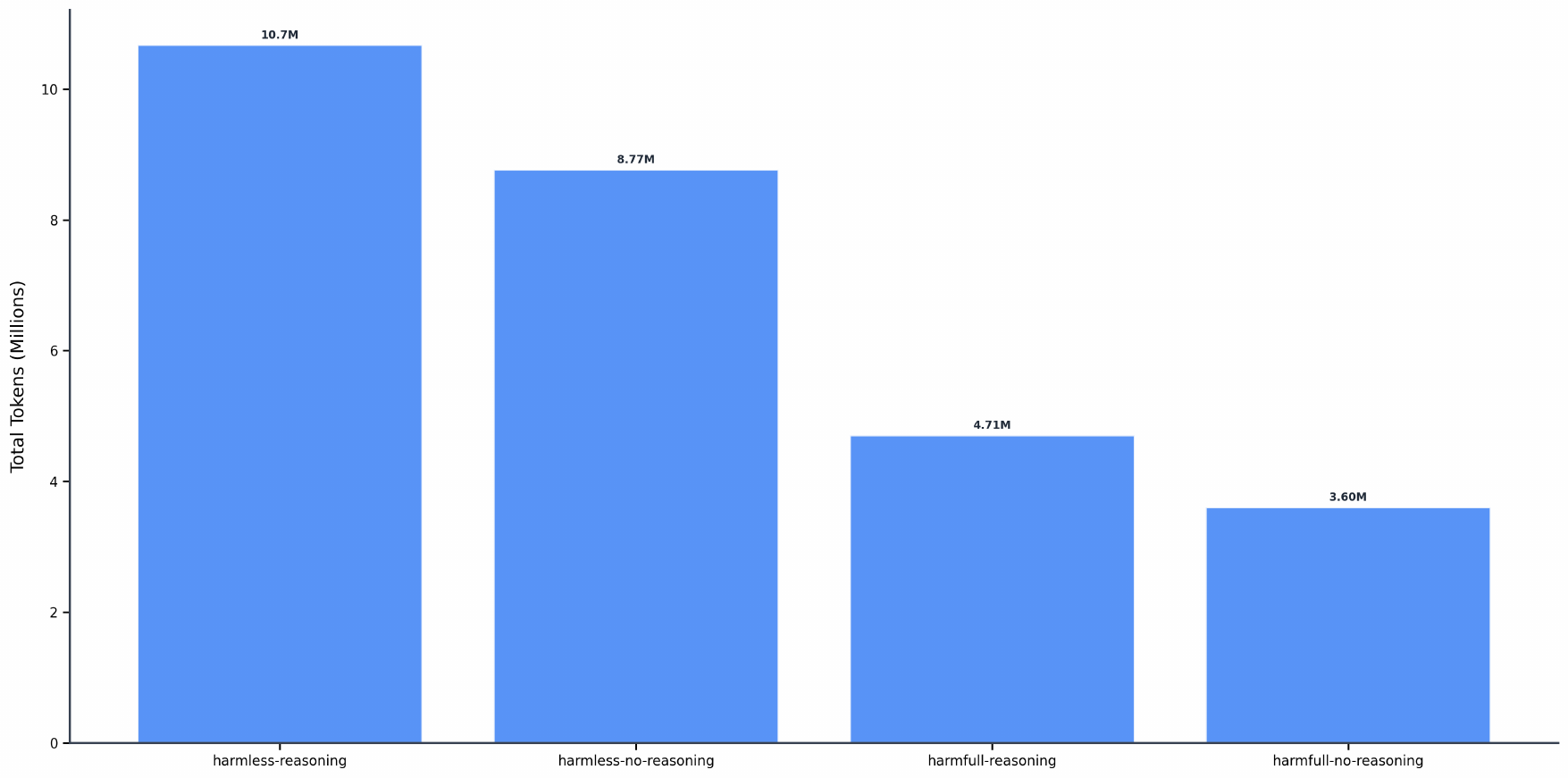}
    \caption*{(a) Token distribution per subset.}
\end{minipage}
\hfill
\begin{minipage}[t]{0.48\linewidth}
    \centering
    \includegraphics[width=\linewidth]{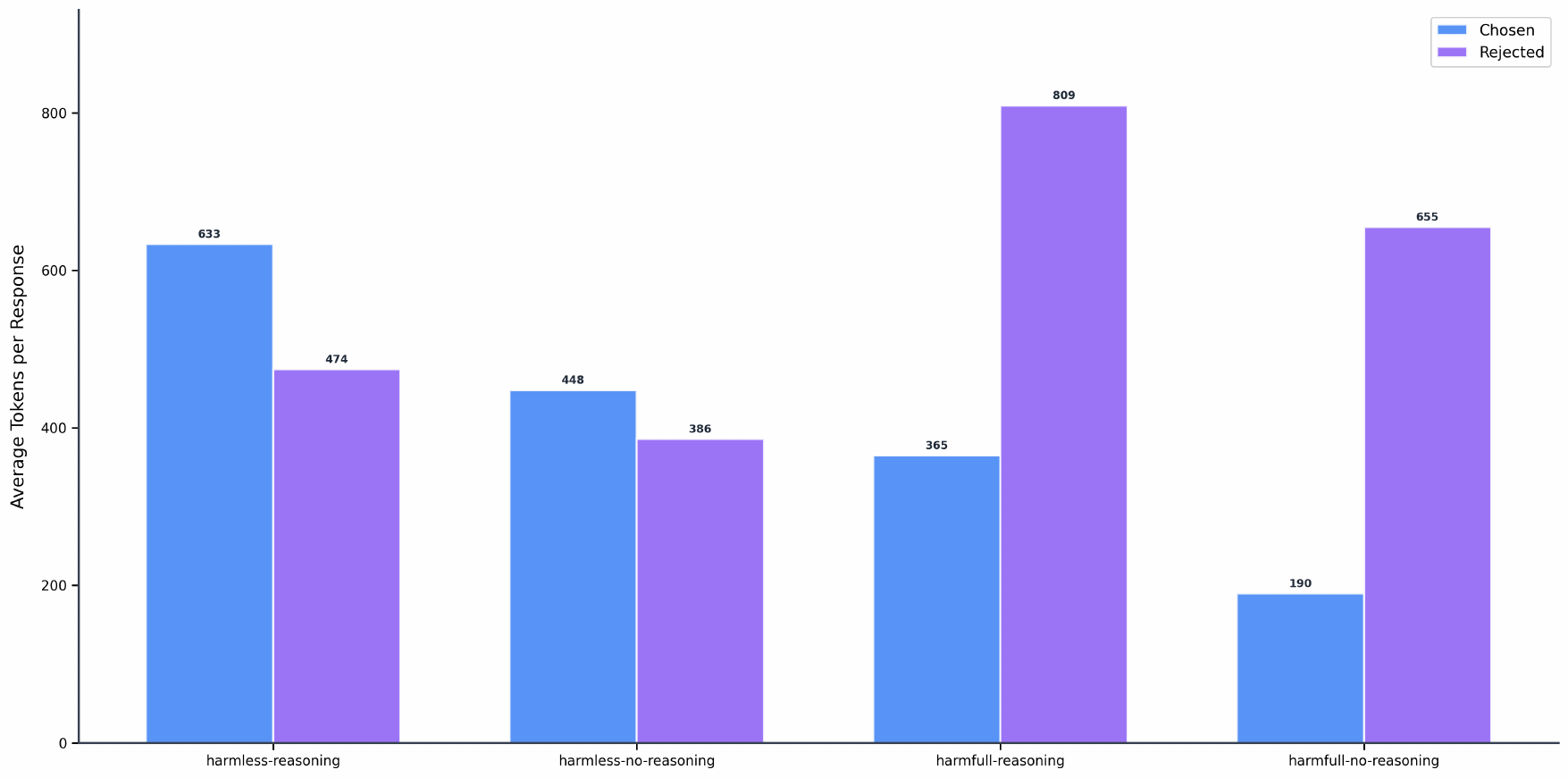}
    \caption*{(b) Average chosen vs.\ rejected length.}
\end{minipage}
\caption{Preference dataset statistics.}
\label{fig:preferences-visualizations}
\end{figure}

\subsection{Constitutional Prompts}
\label{appendix:constitutional_ai}

All constitutions used in our pipeline are available in \href{https://huggingface.co/datasets/Polygl0t/gigaverbo-v2-preferences/tree/main}{Polygl0t/gigaverbo-v2-preferences}.

\begin{itemize}
    \item \href{https://huggingface.co/datasets/Polygl0t/gigaverbo-v2-preferences/blob/main/HARMFULL_CONSTITUTION.md}{Harmful Constitution}.
    \item \href{https://huggingface.co/datasets/Polygl0t/gigaverbo-v2-preferences/blob/main/HARMLESS_CONSTITUTION.md}{Harmless Constitution}.
    \item \href{https://huggingface.co/datasets/Polygl0t/gigaverbo-v2-preferences/blob/main/REASONING_CONSTITUTION.md}{Reasoning Constitution}.
\end{itemize}

\subsection{Training Configurations}
\label{appendix:post_training_configs}

Tables~\ref{tab:instruct-config} and~\ref{tab:think-config} provide the full hyperparameter configurations for the Instruct and Think variants, respectively. Values separated by ``/'' denote SFT / APO stages.

\begin{table}[h]
\centering
\small
\caption{Hyperparameter settings for Instruct variants (SFT / APO).}
\label{tab:instruct-config}
\begin{tabular}{lccc}
\toprule
\textbf{Training Configuration} & \textbf{0.5B} & \textbf{1.5B} & \textbf{3.7B} \\
\midrule
Parallelism Strategy        & DDP           & DDP           & FSDP2 (ZeRO-2)           \\
Activation Checkpointing    & False         & True          & True          \\
Total Batch Size            & 524,288 tokens & 524,288 tokens & 524,288 tokens \\
Micro Batch Size            & 4 / 4 samples & 4 / 4 samples & 4 / 1 samples \\
Accumulation Steps          & 4 / 4         & 4 / 4         & 4 / 32        \\
Context Length              & 4,096 tokens  & 4,096 tokens  & 4,096 tokens  \\
Training Steps              & 68,635 / 1,115 & 68,635 / 1,115 & 68,635 / 1,115 \\
\midrule
\multicolumn{4}{c}{\textit{\textbf{Learning Rate Schedule}}} \\
\midrule
Schedule Type               & Warmup--Cosine & Warmup--Cosine & Warmup--Cosine \\
Warmup Ratio                & 0.1           & 0.1           & 0.1           \\
Peak Learning Rate          & $8.5e^{-5}$ / $5e^{-6}$  & $7.5e^{-5}$ / $5e^{-6}$  & $5e^{-5}$ / $5e^{-6}$ \\
Minimum LR                  & 0.0           & 0.0           & 0.0           \\
\midrule
\multicolumn{4}{c}{\textit{\textbf{Optimizer Configuration}}} \\
\midrule
Optimizer                   & AdamW         & AdamW         & AdamW         \\
Weight Decay                & 0.0           & 0.0           & 0.0           \\
$\beta_1$                   & 0.9           & 0.9           & 0.9           \\
$\beta_2$                   & 0.95          & 0.95          & 0.95          \\
$\epsilon$                  & $1e^{-8}$ & $1e^{-8}$ & $1e^{-8}$ \\
Gradient Clip Norm          & 1.0           & 1.0           & 1.0           \\
\midrule
\multicolumn{4}{c}{\textit{\textbf{SFT}}} \\
\midrule
Epochs                      & 5             & 5             & 5             \\
Packing                     & False         & False         & False         \\
Assistant-Only Loss         & True          & True          & True          \\
\midrule
\multicolumn{4}{c}{\textit{\textbf{APO}}} \\
\midrule
Epochs                      & 5             & 5             & 5             \\
Loss Type                   & \texttt{apo\_zero}     & \texttt{apo\_zero}     & \texttt{apo\_zero}     \\
$\beta$                     & 0.5           & 0.5           & 0.5           \\
\midrule
\multicolumn{4}{c}{\textit{\textbf{Resource Consumption}}} \\
\midrule
Energy Consumption          & 33 kWh        & 170 kWh       & 290 kWh       \\
Carbon Footprint            & 12.7 kg CO$_2$e & 65 kg CO$_2$e & 110 kg CO$_2$e \\
\bottomrule
\end{tabular}
\end{table}

\newpage
\begin{table}[h]
\centering
\small
\caption{Hyperparameter settings for Think variants (SFT / APO).}
\label{tab:think-config}
\begin{tabular}{lccc}
\toprule
\textbf{Training Configuration} & \textbf{0.5B} & \textbf{1.5B} & \textbf{3.7B} \\
\midrule
Parallelism Strategy        & DDP           & DDP           & FSDP2 (ZeRO-2)           \\
Activation Checkpointing    & False         & True          & True          \\
Total Batch Size            & 524,288 tokens & 524,288 tokens & 524,288 tokens \\
Micro Batch Size            & 4 / 4 samples & 4 / 4 samples & 4 / 1 samples \\
Accumulation Steps          & 4 / 4         & 4 / 4         & 4 / 32        \\
Context Length              & 4,096 tokens  & 4,096 tokens  & 4,096 tokens  \\
Training Steps              & 3,060 / 535   & 3,060 / 535   & 3,060 / 535   \\
\midrule
\multicolumn{4}{c}{\textit{\textbf{Learning Rate Schedule}}} \\
\midrule
Schedule Type               & Warmup--Cosine & Warmup--Cosine & Warmup--Cosine \\
Warmup Ratio                & 0.1           & 0.1           & 0.1           \\
Peak Learning Rate          & $8.5e^{-5}$ / $5e^{-6}$  & $7.5e^{-5}$ / $5e^{-6}$  & $5e^{-5}$ / $5e^{-6}$ \\
Minimum LR                  & 0.0           & 0.0           & 0.0           \\
\midrule
\multicolumn{4}{c}{\textit{\textbf{Optimizer Configuration}}} \\
\midrule
Optimizer                   & AdamW         & AdamW         & AdamW         \\
Weight Decay                & 0.0           & 0.0           & 0.0           \\
$\beta_1$                   & 0.9           & 0.9           & 0.9           \\
$\beta_2$                   & 0.95          & 0.95          & 0.95          \\
$\epsilon$                  & $1e^{-8}$ & $1e^{-8}$ & $1e^{-8}$ \\
Gradient Clip Norm          & 1.0           & 1.0           & 1.0           \\
\midrule
\multicolumn{4}{c}{\textit{\textbf{SFT}}} \\
\midrule
Epochs                      & 5             & 5             & 5             \\
Packing                     & False         & False         & False         \\
Assistant-Only Loss         & True          & True          & True          \\
\midrule
\multicolumn{4}{c}{\textit{\textbf{APO}}} \\
\midrule
Epochs                      & 5             & 5             & 5             \\
Loss Type                   & \texttt{apo\_zero}     & \texttt{apo\_zero}     & \texttt{apo\_zero}     \\
$\beta$                     & 0.5           & 0.5           & 0.5           \\
\midrule
\multicolumn{4}{c}{\textit{\textbf{Resource Consumption}}} \\
\midrule
Energy Consumption          & 2.66 kWh      & 5.5 kWh       & 29 kWh        \\
Carbon Footprint            & 1.23 kg CO$_2$e & 2.55 kg CO$_2$e & 13.3 kg CO$_2$e \\
\bottomrule
\end{tabular}
\end{table}

\subsection{Data Mixture}
\label{appendix:data_mixtures}

\subsubsection{Instruct Models}

For the Instruct SFT stage, we selected samples with \texttt{instruct\_score} $\geq 4.5$ across all subsets, except General Instruction Following (where all samples were retained).

\begin{table}[h]
\centering
\small
\caption{Instruct SFT data mixture.}
\label{tab:instruct-sft-mixture}
\begin{tabular}{lrr}
\toprule
\textbf{Task Category} & \textbf{Tokens (approx.)} & \textbf{\%} \\
\midrule
General Instruction Following   & 700M   & 80.1 \\
Structured Output Generation    &  35M   &  4.0 \\
Math and CoT                    &  27M   &  3.1 \\
Function Calling                & 17.5M  &  2.0 \\
Preference Data (Chosen)        &  14M   &  1.6 \\
Translation                     & 5.7M   &  0.7 \\
Coding                          & 2.3M   &  0.3 \\
Retrieval Augmented Generation  & 2.2M   &  0.3 \\
Summarization                   & 290K   & 0.03 \\
\midrule
\textbf{Total}                  & \textbf{${\sim}$874M} &  \\
\bottomrule
\end{tabular}
\end{table}

For Instruct APO, reasoning traces were stripped from preference pairs containing \texttt{<think>}$\ldots$\texttt{</think>} tags, allowing all 28,437 pairs to be used:

\begin{table}[h]
\centering
\small
\caption{Instruct APO data mixture.}
\label{tab:instruct-apo-mixture}
\begin{tabular}{lrr}
\toprule
\textbf{Subset} & \textbf{Samples} & \textbf{\%} \\
\midrule
Harmless (no reasoning)         & 10,521 & 37.0 \\
Harmless (reasoning stripped)   &  9,641 & 33.9 \\
Harmful (no reasoning)          &  4,267 & 15.0 \\
Harmful (reasoning stripped)    &  4,008 & 14.1 \\
\midrule
\textbf{Total}                  & \textbf{28,437} & \\
\bottomrule
\end{tabular}
\end{table}

\subsubsection{Think Models}

Think SFT used exclusively reasoning data:

\begin{table}[h]
\centering
\small
\caption{Think SFT data mixture.}
\label{tab:think-sft-mixture}
\begin{tabular}{lrr}
\toprule
\textbf{Task Category} & \textbf{Tokens (approx.)} & \textbf{\%} \\
\midrule
Reasoning & 34M & 100 \\
\bottomrule
\end{tabular}
\end{table}

Think APO retained only pairs with explicit reasoning traces:

\begin{table}[h]
\centering
\small
\caption{Think APO data mixture.}
\label{tab:think-apo-mixture}
\begin{tabular}{lrr}
\toprule
\textbf{Subset} & \textbf{Samples} & \textbf{\%} \\
\midrule
Harmless (with reasoning)       &  9,641 & 70.6 \\
Harmful (with reasoning)        &  4,008 & 29.4 \\
\midrule
\textbf{Total}                  & \textbf{13,649} & \\
\bottomrule
\end{tabular}
\end{table}

Key differences: (1)~Instruct SFT contains ${\sim}$25$\times$ more tokens than Think SFT; (2)~Instruct strips reasoning traces while Think retains them; (3)~Instruct APO uses all 28K pairs while Think APO uses 13.6K pairs with reasoning traces.

\newpage
\subsection{Per-Benchmark Results}
\label{appendix:nstruct-think-results}

\subsubsection{Instruct Variants}

Table~\ref{tab:instruct-individual} provides individual benchmark scores for all Instruct variants and baselines.

\begin{table}[h]
\centering
\scriptsize
\caption{Individual benchmark scores for Instruct variants and baselines.}
\label{tab:instruct-individual}
\setlength{\tabcolsep}{3pt}
\begin{tabular}{lccccccccc}
\toprule
 & \textbf{BLUEX} & \textbf{ENEM} & \textbf{OAB} & \textbf{ARC} & \textbf{BELEBELE} & \textbf{MMLU} & \textbf{IFEval} & \textbf{GSM8K} & \textbf{HumanEval} \\
\midrule
\textbf{\tucanocptlargeinstruct}    & 64.53 & 72.92 & 54.31 & 60.34 & 85.22 & 64.64 & 41.67 & 53.81 & 47.56 \\
Jurema-7B                  & 63.42 & 70.96 & 64.97 & 52.56 & 88.44 & 49.91 & 47.00 & 30.29 & 75.61 \\
Qwen2.5-3B-Inst.           & 56.88 & 68.65 & 46.79 & 41.71 & 84.00 & 58.22 & 63.33 & 51.90 & 70.73 \\
Qwen3-4B                   & 63.28 & 72.15 & 50.30 & 43.08 & 83.67 & 26.93 & 79.33 & 39.88 & 86.59 \\
Gemma-3-Gaia-4b-it         & 50.90 & 64.52 & 43.46 & 54.70 & 78.89 & 51.49 & 70.33 & 51.29 & 64.02 \\
SmolLM3-3B                 & 53.55 & 64.73 & 41.00 & 52.74 & 78.67 & 53.23 & 69.67 & 44.44 & 68.29 \\
Llama-3.2-3B-Inst.         & 52.02 & 59.13 & 44.97 & 43.93 & 78.56 & 52.14 & 62.67 & 55.10 & 48.17 \\
Qwen2.5-1.5B-Inst.         & 52.02 & 61.79 & 44.28 & 39.74 & 76.00 & 51.91 & 42.00 & 42.16 & 48.78 \\
\textbf{\tucanocptmediuminstruct}    & 52.85 & 62.70 & 43.42 & 50.26 & 77.56 & 52.54 & 34.33 & 19.71 & 26.22 \\
Qwen3-1.7B                 & 50.49 & 58.64 & 32.62 & 36.67 & 64.89 & 30.56 & 65.00 & 26.94 & 64.02 \\
\textbf{\tucanocptsmallinstruct}    & 40.33 & 53.60 & 40.73 & 38.63 & 62.33 & 41.46 & 30.00 & 18.49 & 10.37 \\
Qwen3-0.6B                 & 34.91 & 41.15 & 29.75 & 30.51 & 37.11 & 26.48 & 55.00 & 31.66 & 39.02 \\
Llama-3.2-1B-Inst.         & 30.04 & 34.01 & 30.84 & 32.82 & 41.56 & 35.15 & 44.33 & 26.56 & 29.27 \\
Qwen2.5-0.5B-Inst.         & 30.18 & 34.08 & 29.34 & 27.44 & 50.67 & 39.54 & 31.00 & 14.69 & 24.39 \\
Tucano-2b4-Inst.           & 25.87 & 20.01 & 26.74 & 31.97 & 24.00 & 26.72 & 15.00 &  2.05 &  0.00 \\
Tucano-1b1-Inst.           & 22.95 & 19.94 & 25.33 & 30.00 & 24.89 & 25.59 & 13.33 &  1.67 &  0.00 \\
\bottomrule
\end{tabular}
\end{table}

\begin{figure}[h]
\centering
\includegraphics[width=0.85\linewidth]{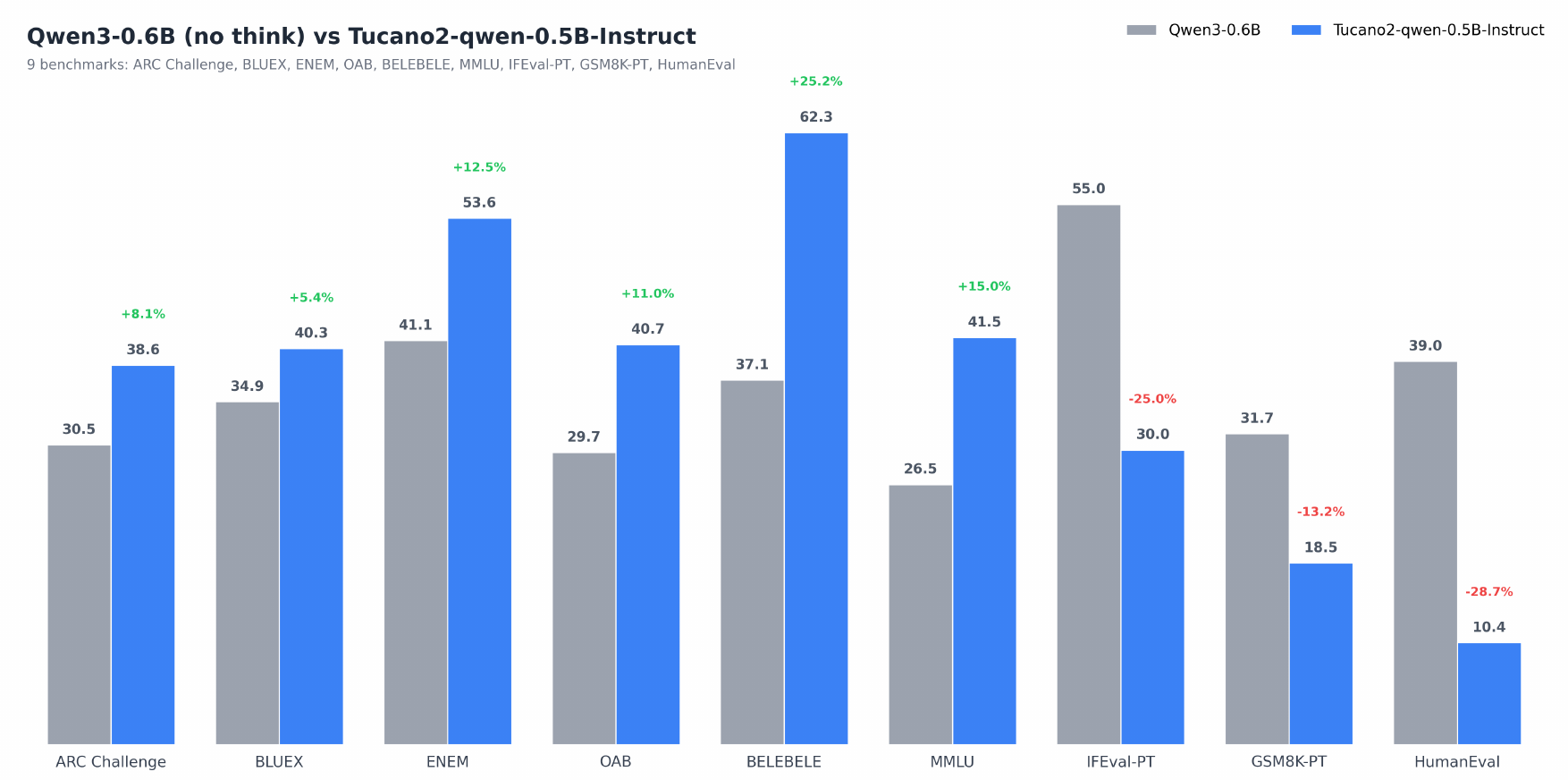}
\caption{Per-benchmark comparison for 0.5B-Instruct.}
\label{fig:instruct-0.5B-comparison}
\end{figure}

\begin{figure}[h]
\centering
\includegraphics[width=0.85\linewidth]{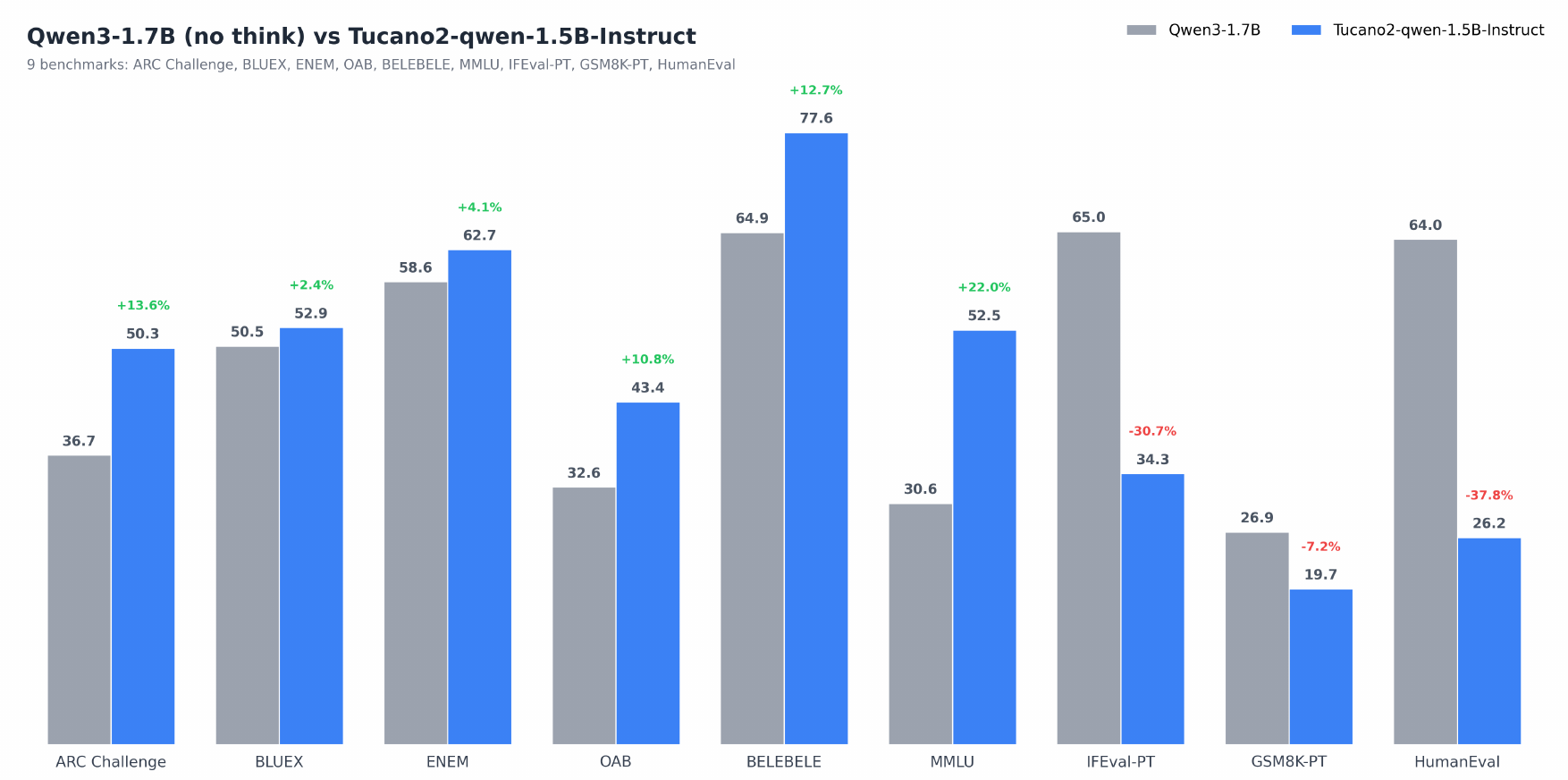}
\caption{Per-benchmark comparison for 1.5B-Instruct.}
\label{fig:instruct-1.5B-comparison}
\end{figure}

Figures~\ref{fig:instruct-0.5B-comparison} and~\ref{fig:instruct-1.5B-comparison} show per-benchmark comparisons at the 0.5B and 1.5B scales.

\subsubsection{Think Variants}

Table~\ref{tab:think-individual} provides individual benchmark scores for all Think variants and baselines.

\begin{table}[h]
\centering
\scriptsize
\caption{Individual benchmark scores for Think variants and baselines.}
\label{tab:think-individual}
\setlength{\tabcolsep}{3pt}
\begin{tabular}{lcccccccc}
\toprule
 & \textbf{BLUEX} & \textbf{ENEM} & \textbf{OAB} & \textbf{ARC} & \textbf{BELEBELE} & \textbf{MMLU} & \textbf{IFEval} & \textbf{GSM8K} \\
\midrule
\textbf{\tucanocptlargethink}    & 63.00 & 71.52 & 53.76 & 55.38 & 83.56 & 61.18 & 31.67 & 56.70 \\
SmolLM3-3B                 & 48.82 & 60.60 & 42.19 & 52.56 & 77.78 & 52.82 & 64.67 & 72.15 \\
Qwen3-4B                   & 78.30 & 85.23 & 47.97 & 39.66 & 23.00 & 31.53 & 84.00 & 77.02 \\
Qwen3-1.7B                 & 68.29 & 72.50 & 39.32 & 35.64 & 22.89 & 22.85 & 68.33 & 71.69 \\
\textbf{\tucanocptmediumthink}    & 39.22 & 39.89 & 34.26 & 42.82 & 67.67 & 43.30 & 33.67 & 22.83 \\
Qwen3-0.6B                 & 43.53 & 48.71 & 34.21 & 29.91 & 22.89 & 22.85 & 58.33 & 58.37 \\
\textbf{\tucanocptsmallthink}    & 34.49 & 31.98 & 27.02 & 32.74 & 36.11 & 36.08 & 27.67 & 14.61 \\
\bottomrule
\end{tabular}
\end{table}

\begin{figure}[h]
\centering
\includegraphics[width=0.85\linewidth]{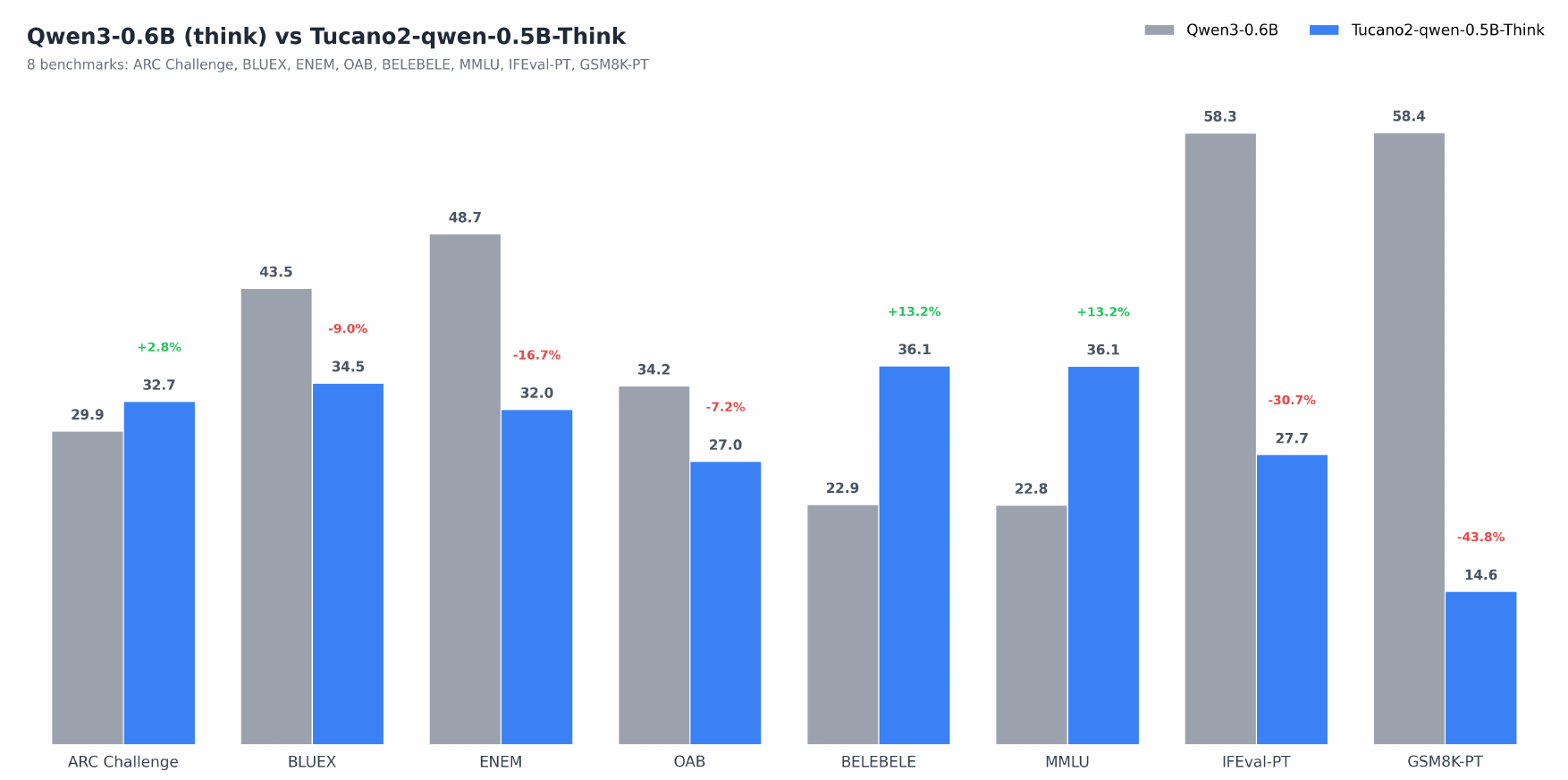}
\caption{Per-benchmark comparison for 0.5B-Think.}
\label{fig:think-0.5B-comparison}
\end{figure}

\begin{figure}[h]
\centering
\includegraphics[width=0.85\linewidth]{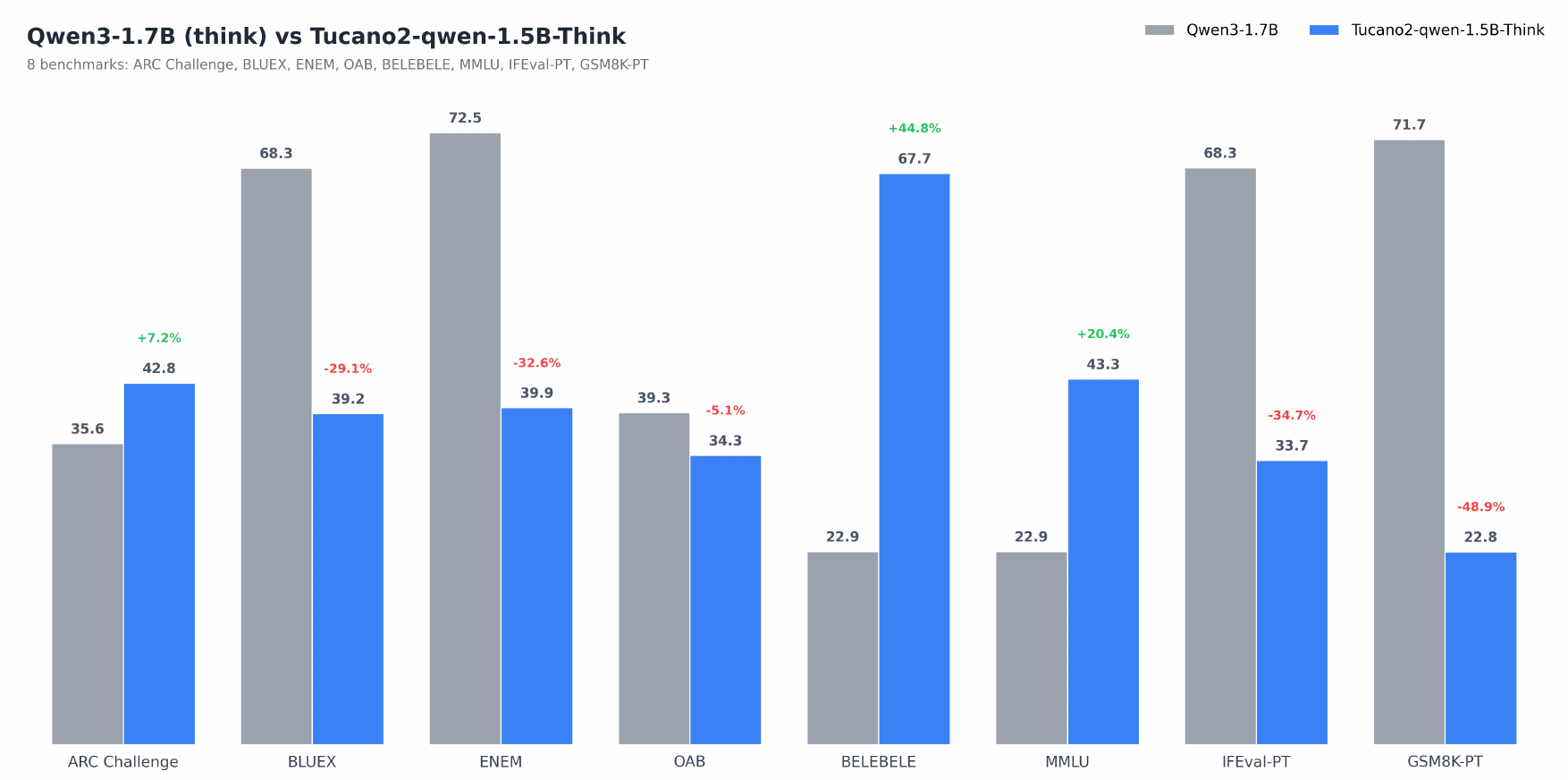}
\caption{Per-benchmark comparison for 1.5B-Think.}
\label{fig:think-1.5B-comparison}
\end{figure}

Figures~\ref{fig:think-0.5B-comparison} and~\ref{fig:think-1.5B-comparison} show per-benchmark comparisons at the 0.5B and 1.5B scales.

\subsubsection{Instruct and Think Highlights}
\label{appendix:instruct-think-highlights}

\paragraph{Instruct variants.}
Across all parameter scales, the Tucano2 Instruct variants demonstrate consistent advantages on knowledge-intensive and domain-specific benchmarks. \tucanocptlargeinstruct achieves the highest Knowledge \& Reasoning score among all 3--4B models, with gains distributed across BLUEX, ENEM, OAB, ARC-Challenge, BELEBELE, and MMLU. Mathematical reasoning is a particular strength: on GSM8K-PT, the 3.7B Instruct model scores 53.81, outperforming both Qwen3-4B (39.88) and Gemma-3-Gaia-PT-BR-4b-it (51.29).

\paragraph{Think variants.}
The comparatively lower aggregate performance of Think variants relative to some baselines can be attributed to: (1)~a substantial shortage of high-quality Portuguese reasoning data during training; (2)~significantly shorter reasoning traces compared to Qwen3 and SmolLM3; and (3)~a constrained 4,096-token context window (roughly half the inference budget available to competing models). Despite these constraints, Tucano2-qwen-3.7B-Think surpasses both Qwen3-4B and SmolLM3-3B on knowledge-intensive benchmarks while reasoning entirely in Portuguese---demonstrating that Portuguese-native chain-of-thought reasoning is viable and competitive even under constrained budgets. Think variants underperform on instruction-following (IFEval-PT) and coding (HumanEval) benchmarks.

\paragraph{Think vs.\ Instruct: HumanEval Degradation.}
When reasoning mode is enabled, coding performance (HumanEval) degrades substantially across the hybrid models we evaluated, as shown in Table~\ref{tab:think-humaneval-drop}. Average HumanEval performance drops by ${\sim}$29 percentage points when switching from Instruct to Think mode (inference token budget = 8,192).

\begin{table}[h]
\centering
\small
\caption{HumanEval scores: Instruct vs.\ Think mode.}
\label{tab:think-humaneval-drop}
\begin{tabular}{lccc}
\toprule
\textbf{Model} & \textbf{Instruct} & \textbf{Think} & \textbf{$\Delta$} \\
\midrule
Qwen3-4B    & 86.59 & 50.61 & $-$35\% \\
SmolLM3-3B  & 68.29 & 33.54 & $-$34\% \\
Qwen3-1.7B  & 64.02 & 38.41 & $-$25\% \\
Qwen3-0.6B  & 39.02 & 17.68 & $-$21\% \\
\midrule
\textit{Average} &  &  & $-$29\% \\
\bottomrule
\end{tabular}
\end{table}

\subsubsection{RULER-PT: Results}
\label{appendix:ruler-details}

Tables~\ref{tab:ruler-1024},~\ref{tab:ruler-2048}, and~\ref{tab:ruler-4096} present per-task RULER-PT scores at context lengths of 1,024, 2,048, and 4,096 tokens for the 3.7B-scale models.

\begin{table}[h]
\centering
\small
\caption{RULER-PT detailed results at 1,024 tokens.}
\label{tab:ruler-1024}
\begin{tabular}{lccc}
\toprule
\textbf{Metric} & \textbf{Qwen3-4B} & \textbf{3.7B-Inst.} & \textbf{3.7B-Think} \\
\midrule
RULER Score (aggregate)                      & 0.966 & 0.795 & 0.817 \\
\midrule
NIAH multi-key (line retrieval)              & 0.886 & 0.814 & 0.834 \\
NIAH multi-key (KV retrieval)                & 1.000 & 0.800 & 0.744 \\
NIAH multi-key                               & 0.998 & 0.844 & 0.870 \\
NIAH multi-query                             & 0.886 & 0.768 & 0.797 \\
NIAH multi-value                             & 1.000 & 0.769 & 0.778 \\
NIAH single needle (passkey)                 & 1.000 & 0.830 & 0.846 \\
NIAH single needle (vanilla)                 & 1.000 & 0.808 & 0.808 \\
NIAH single needle (essay UUID)              & 0.998 & 0.778 & 0.878 \\
Common word extraction                       & 0.881 & 0.490 & 0.601 \\
Frequent word extraction                     & 0.979 & 0.860 & 0.879 \\
Variation tracing                            & 1.000 & 0.989 & 0.953 \\
\bottomrule
\end{tabular}
\end{table}

\newpage
\begin{table}[h]
\centering
\small
\caption{RULER-PT detailed results at 2,048 tokens.}
\label{tab:ruler-2048}
\begin{tabular}{lccc}
\toprule
\textbf{Metric} & \textbf{Qwen3-4B} & \textbf{3.7B-Inst.} & \textbf{3.7B-Think} \\
\midrule
RULER Score (aggregate)                      & 0.984 & 0.710 & 0.765 \\
\midrule
NIAH multi-key (line retrieval)              & 1.000 & 0.720 & 0.820 \\
NIAH multi-key (KV retrieval)                & 0.998 & 0.696 & 0.590 \\
NIAH multi-key                               & 0.998 & 0.860 & 0.868 \\
NIAH multi-query                             & 1.000 & 0.712 & 0.741 \\
NIAH multi-value                             & 1.000 & 0.595 & 0.759 \\
NIAH single needle (passkey)                 & 1.000 & 0.792 & 0.824 \\
NIAH single needle (vanilla)                 & 1.000 & 0.610 & 0.818 \\
NIAH single needle (essay UUID)              & 1.000 & 0.712 & 0.882 \\
Common word extraction                       & 0.906 & 0.316 & 0.362 \\
Frequent word extraction                     & 0.927 & 0.839 & 0.897 \\
Variation tracing                            & 1.000 & 0.963 & 0.857 \\
\bottomrule
\end{tabular}
\end{table}

\begin{table}[h]
\centering
\small
\caption{RULER-PT detailed results at 4,096 tokens.}
\label{tab:ruler-4096}
\begin{tabular}{lccc}
\toprule
\textbf{Metric} & \textbf{Qwen3-4B} & \textbf{3.7B-Inst.} & \textbf{3.7B-Think} \\
\midrule
RULER Score (aggregate)                      & 0.979 & 0.686 & 0.707 \\
\midrule
NIAH multi-key (line retrieval)              & 1.000 & 0.684 & 0.770 \\
NIAH multi-key (KV retrieval)                & 0.996 & 0.610 & 0.472 \\
NIAH multi-key                               & 0.996 & 0.782 & 0.782 \\
NIAH multi-query                             & 0.999 & 0.677 & 0.695 \\
NIAH multi-value                             & 0.999 & 0.522 & 0.653 \\
NIAH single needle (passkey)                 & 1.000 & 0.786 & 0.798 \\
NIAH single needle (vanilla)                 & 1.000 & 0.574 & 0.824 \\
NIAH single needle (essay UUID)              & 0.996 & 0.696 & 0.806 \\
Common word extraction                       & 0.897 & 0.562 & 0.571 \\
Frequent word extraction                     & 0.883 & 0.713 & 0.667 \\
Variation tracing                            & 1.000 & 0.940 & 0.741 \\
\bottomrule
\end{tabular}
\end{table}

The largest performance gaps appear on multi-key KV retrieval and common word extraction tasks, which require maintaining and integrating information across extended passages.

\newpage
\subsection{Inference Samples}
\label{appendix:inference_samples}

Below, we provide inference samples from both the Instruct and Think variants of our 3.7B models. While the models still exhibit some signs of hallucinations and verbosity, even in zero-shot settings, they demonstrate substantially improved instruction-following capability compared to the first Tucanos.

\subsubsection*{Instruct Variant}

\begin{tcolorbox}[colback=gray!5!white, colframe=gray!75!black, title=Coding]
\small
\textbf{User:} Como eu posso abrir um arquivo CSV em Python usando a biblioteca pandas?

\textbf{Assistant:} Para ler dados de arquivos \texttt{CSV} utilizando a poderosa ferramenta chamada \textbf{Pandas} na linguagem de programa\c{c}\~{a}o Python, voc\^{e} pode usar uma fun\c{c}\~{a}o simples como esta:

\begin{verbatim}
import pandas as pd

def carregar_csv(caminho):
    """Funcao para carregamento dos dados do formato csv
    atraves da lib Pandas."""
    try:
        df = pd.read_csv(caminho)
        print("Arquivo carregado!")
        return df
    except Exception as erro:
        raise ValueError('Ocorreu algum problema durante leitura:',
                         str(erro))

if __name__ == '__main__':
    nome_do_arquivo = 'seuarquivocsv.csv'
    resultado = carregar_csv(nome_do_arquivo)
\end{verbatim}

Este script define duas partes principais: primeiro, a defini\c{c}\~{a}o funcional (\texttt{carregar\_csv}) que recebe um argumento string com o caminho do arquivo \texttt{.CSV}; depois, tratamento b\'{a}sico de excess\~{o}es caso haja problemas relacionados \`{a} conex\~{a}o ou formata\c{c}\~{a}o incorreta do ficheiro.
\end{tcolorbox}

\begin{tcolorbox}[colback=gray!5!white, colframe=gray!75!black, title=Math]
\small
\textbf{User:} Como eu posso resolver o seguinte problema: 2x + 3 = 11?

\textbf{Assistant:} Para encontrar a solu\c{c}\~{a}o para x, voc\^{e} precisa isolar x em um lado da equa\c{c}\~{a}o. Aqui est\~{a}o as etapas:
\begin{enumerate}
    \item Subtraia 3 de ambos os lados: $2x + 3 - 3 = 11 - 3$, resultando em $2x = 8$.
    \item Divida cada termo por 2: $x = 4$.
\end{enumerate}
Portanto, $x = 4$.
\end{tcolorbox}

\begin{tcolorbox}[colback=gray!5!white, colframe=gray!75!black, title=Rewriting]
\small
\textbf{User:} Remova todas as informa\c{c}\~{o}es irrelevantes da frase a seguir.

\textit{``O novo restaurante que abriu no centro da cidade, de propriedade do primo de Jorge, que costumava ser chef de um restaurante em Ipanema, serve uma variedade de cozinhas de todo o mundo.''}

\textbf{Assistant:} Um novo restaurante aberto no centro da cidade oferece diversas culin\'{a}rias internacionais.
\end{tcolorbox}

\begin{tcolorbox}[colback=gray!5!white, colframe=gray!75!black, title=Tool Use]
\small
\textbf{System:} Utilize as ferramentas para resolver as tarefas que o usu\'{a}rio lhe solicitar. Se as ferramentas forem insuficientes, explique ao usu\'{a}rio que n\~{a}o \'{e} poss\'{i}vel completar a tarefa. A ferramenta dispon\'{i}vel \'{e} \texttt{create\_todo}, que cria uma nova tarefa com campos \texttt{task\_description} e \texttt{due\_date} (formato \texttt{YYYY-MM-DD}).

\textbf{User:} Eu preciso criar uma nova tarefa para o meu projeto.

\textbf{Assistant:} Claro, eu posso ajudar com isso. Voc\^{e} poderia, por favor, me fornecer a descri\c{c}\~{a}o da tarefa e a data de entrega?

\textbf{User:} A tarefa \'{e} finalizar o relat\'{o}rio do projeto e a data de entrega \'{e} 2022-03-15.

\textbf{Assistant:}
\begin{verbatim}
<tool_call>
{"name": "create_todo", "arguments": {"task_description":
    "Finalizar o relatorio do projeto", "due_date": "2022-03-15"}}
</tool_call>
\end{verbatim}
\end{tcolorbox}

\begin{tcolorbox}[colback=gray!5!white, colframe=gray!75!black, title=Similarity Scoring]
\small
\textbf{User:} Frase 1: O artista pintou um quadro inspirado nos campos floridos da primavera. Frase 2: O criador produziu uma obra que representava paisagens urbanas cinzentas. Pergunta: Qu\~{a}o similares s\~{a}o as duas frases? D\^{e} uma pontua\c{c}\~{a}o entre 1,0 a 5,0.

\textbf{Assistant:} 2.5. As duas frases tratam sobre obras de arte, mas diferem significativamente quanto aos temas retratados --- natureza versus cidade ---, resultando numa pontua\c{c}\~{a}o intermedi\'{a}ria de 2.5.
\end{tcolorbox}

\subsubsection*{Think Variant}

\begin{tcolorbox}[colback=gray!5!white, colframe=gray!75!black, title=Math]
\small
\textbf{User:} Como eu posso resolver o seguinte problema: 2x + 3 = 11?

\textbf{$\langle$think$\rangle$} O usu\'{a}rio est\'{a} me perguntando como resolver a equa\c{c}\~{a}o $2x + 3 = 11$. A ideia principal \'{e} isolar $x$ no lado esquerdo. Subtraindo 3 de ambos os lados: $2x = 8$. Dividindo por 2: $x = 4$. Verificando: $2(4) + 3 = 11$ \checkmark \textbf{$\langle$/think$\rangle$}

\textbf{Assistant:} Vamos resolver passo a passo:
\begin{enumerate}
    \item Isolar o termo com $x$: $2x + 3 - 3 = 11 - 3 \Rightarrow 2x = 8$
    \item Resolver para $x$: $\frac{2x}{2} = \frac{8}{2} \Rightarrow x = 4$
\end{enumerate}
Verifica\c{c}\~{a}o: $2(4) + 3 = 8 + 3 = 11$ \checkmark
\end{tcolorbox}

\begin{tcolorbox}[colback=gray!5!white, colframe=gray!75!black, title=Rewriting]
\small
\textbf{User:} Remova todas as informa\c{c}\~{o}es irrelevantes da frase a seguir.

\textit{``O novo restaurante que abriu no centro da cidade, de propriedade do primo de Jorge, que costumava ser chef de um restaurante em Ipanema, serve uma variedade de cozinhas de todo o mundo.''}

\textbf{$\langle$think$\rangle$} O usu\'{a}rio est\'{a} me pedindo para remover informa\c{c}\~{o}es irrelevantes. A informa\c{c}\~{a}o relevante gira em torno do estabelecimento: sua localiza\c{c}\~{a}o atual (centro da cidade) e suas caracter\'{i}sticas principais (variedade de cozinhas). A rela\c{c}\~{a}o de propriedade com o primo de Jorge adiciona contexto social, mas a hist\'{o}ria anterior como chef \'{e} desnecess\'{a}ria. Elimino essa informa\c{c}\~{a}o e simplifico a express\~{a}o final. \textbf{$\langle$/think$\rangle$}

\textbf{Assistant:} O novo restaurante que abriu no centro da cidade, de propriedade do primo de Jorge, serve uma diversidade de cozinhas internacionais.
\end{tcolorbox}

\begin{tcolorbox}[colback=gray!5!white, colframe=gray!75!black, title=History]
\small
\textbf{User:} Liste os principais eventos envolvendo a Revolu\c{c}\~{a}o Farroupilha.

\textbf{$\langle$think$\rangle$} O usu\'{a}rio est\'{a} me perguntando para listar os principais eventos da Revolu\c{c}\~{a}o Farroupilha. Vou organizar esses eventos cronologicamente, mencionando datas exatas, batalhas significativas e acordos finais, al\'{e}m de figuras-chave associadas ao conflito entre o Rio Grande do Sul e o governo imperial brasileiro durante 1835--1845. \textbf{$\langle$/think$\rangle$}

\textbf{Assistant:} Claro! A Revolu\c{c}\~{a}o Farroupilha, tamb\'{e}m conhecida como Guerra dos Farrapos, foi um conflito armado ocorrido no estado atual do Rio Grande do Sul, Brasil, durante os anos de 1835 a 1845. Aqui est\~{a}o alguns dos principais eventos:
\begin{enumerate}
    \item \textbf{Declara\c{c}\~{a}o de Independ\^{e}ncia (1835):} No dia 20 de setembro de 1835, liderados por Bento Gon\c{c}alves da Silva, as tropas ga\'{u}chas declararam independ\^{e}ncia do Imp\'{e}rio Brasileiro, estabelecendo a Rep\'{u}blica Rio-Grandense.
    \item \textbf{Batalha de Seival (1836):} Esta batalha marcou uma vit\'{o}ria decisiva das for\c{c}as farroupilhas contra as tropas imperiais. Foi aqui que Giuseppe Garibaldi refor\c{c}ou significativamente suas fileiras.
    \item \textbf{Funda\c{c}\~{a}o da Capital Provis\'{o}ria --- Pelotas (1837):} Ap\'{o}s vencerem algumas batalhas iniciais, os rebeldes escolheram Pelotas como capital provis\'{o}ria da nova rep\'{u}blica.
    \item \textbf{Tratado de Ponche Verde (1845):} Este tratado p\^{o}s fim oficialmente \`{a} guerra ap\'{o}s negocia\c{c}\~{o}es mediadas. Os termos garantiram indeniza\c{c}\~{o}es financeiras \`{a}s fam\'{i}lias combatentes farroupilhas.
\end{enumerate}
\end{tcolorbox}

\end{document}